\documentclass[10pt,journal,compsoc]{IEEEtran}
\ifCLASSOPTIONcompsoc
\usepackage[nocompress]{cite}
\else
\usepackage{cite}
\fi

\usepackage{amsmath,amssymb,amsfonts}
\usepackage{algorithmic}
\usepackage{algorithm}
\usepackage{array}
\usepackage[caption=false,font=normalsize,labelfont=sf,textfont=sf]{subfig}
\usepackage{textcomp}
\usepackage{stfloats}
\usepackage{url}
\usepackage{verbatim}
\usepackage{multirow}
\usepackage{cite}
\usepackage{graphicx}
\usepackage{picinpar}
\usepackage{url}
\usepackage{flushend}
\usepackage[latin1]{inputenc}
\usepackage{colortbl}
\usepackage{soul}
\usepackage{pifont}
\usepackage{color}
\usepackage{alltt}
\usepackage[colorlinks]{hyperref}
\usepackage{enumerate}
\usepackage{siunitx}
\usepackage{breakurl}
\usepackage{epstopdf}
\usepackage{pbox}
\usepackage{amssymb}
\usepackage{threeparttable}
\usepackage{booktabs}
\usepackage{array}
\usepackage{tabularx}
\usepackage[dvipsnames]{xcolor}
\usepackage{ragged2e}  % For justifying abstract
\usepackage{booktabs}
\usepackage{times}
\ifCLASSINFOpdf

\else

\fi

\definecolor{my_purple}{HTML}{9903F0}
\definecolor{my_green}{HTML}{156C09}
\definecolor{my_orange}{HTML}{FF3300}

\begin{document}

% \title{What is the Neural Brain for Embodied Agents? Insights from Neuroscience}

%\title{Neural Brain: A Novel Neuroscience-inspired Framework for Embodied Agents}

\title{Neural Brain: A Neuroscience-inspired Framework for Embodied Agents}

% The quastions (added by jian) need to be solved.
% 1. The literature reviewed needs to be added to GitHub
% 2. Each section from 3 to 6 needs an explanation diagram and a table of representative methods
% 3. When writing the future direction in section 3 to 6, the reviewed current research needs to be combined with the Neural Brain in section2, which means the future directions are written to narrow this gap

% The quastions (added by xiongtao) need to be solved.
% 1.Shouldn't the first letter of “Neural Brain” in the full text be capitalized? And also “human brain”?
% 2.Check if the abbreviation appears for the first time in the full text and use it
% 3.The spacing between the caption and the text needs to be adjusted and unified
% 4.Figure and Table reference should be unified. 
% 5.The first letter of each word should be capitalized consistently

% \author{Jian Liu$^{*}$, Xiongtao Shi$^{*}$, Thai Duy Nguyen, Haitian Zhang, Tianxiang Zhang, \\Jae Won Cho, Erik Cambria,~\IEEEmembership{Fellow,~IEEE}, and Lin Wang}

\author{Jian Liu$^{*}$, Xiongtao Shi$^{*}$, Thai Duy Nguyen$^{\ddag}$, Haitian Zhang$^{\ddag}$, Tianxiang Zhang$^{\ddag}$, Wei Sun, Yanjie Li, Athanasios V. Vasilakos,~\IEEEmembership{Senior Member,~IEEE}, Giovanni Iacca,~\IEEEmembership{Senior Member,~IEEE}, Arshad Ali Khan, Arvind Kumar, Jae Won Cho,~\IEEEmembership{Member,~IEEE}, Ajmal Mian,~\IEEEmembership{Senior Member,~IEEE}, Lihua Xie,~\IEEEmembership{Fellow,~IEEE}, Erik Cambria,~\IEEEmembership{Fellow,~IEEE}, and Lin Wang$^{\dag}$,~\IEEEmembership{Member,~IEEE}
	\IEEEcompsocitemizethanks{
      \IEEEcompsocthanksitem $^{*}$ Equal Contribution, $^{\ddag}$ Co-second Author, $^{\dag}$ Corresponding Author (linwang@ntu.edu.sg).
	\IEEEcompsocthanksitem Jian Liu, Xiongtao Shi, Thai Duy Nguyen, Haitian Zhang, Tianxiang Zhang, Lihua Xie, and Lin Wang are with the School of Electrical and Electronic Engineering, Nanyang Technological University, Singapore.
        \IEEEcompsocthanksitem Jian Liu and Wei Sun are with the School of Artificial Intelligence and Robotics, Hunan University, China.
        \IEEEcompsocthanksitem Xiongtao Shi and Yanjie Li are with the School of Intelligence Science and Engineering, Harbin Institute of Technology (Shenzhen), China.
       \IEEEcompsocthanksitem  Athanasios Vasilakos is with the Department of Information and Communication Technology, University of Agder, Norway. 
        \IEEEcompsocthanksitem Giovanni Iacca is with the Department of Information Engineering and Computer Science, University of Trento, Italy.
        \IEEEcompsocthanksitem  Arshad Ali Khan is with Elm Company, London, UK.
        \IEEEcompsocthanksitem Arvind Kumar is with the Division of Computational Science and Technology, KTH Royal Institute of Technology, Sweden, 
	\IEEEcompsocthanksitem Jae Won Cho is with the School of Artificial Intelligence and Data Science, Sejong University, Korea.
	\IEEEcompsocthanksitem Ajmal Mian is with the Department of Computer Science of the University of Western Australia, Australia.
        \IEEEcompsocthanksitem Erik Cambria is with the College of Computing and Data Science, Nanyang Technological University, Singapore.

	}
}

\IEEEtitleabstractindextext{%
\justifying
	\begin{abstract}
  The rapid evolution of artificial intelligence (AI) has shifted from static, data-driven models to dynamic systems capable of perceiving and interacting with real-world environments. Despite advancements in pattern recognition and symbolic reasoning, current AI systems, such as large language models, remain disembodied, unable to physically engage with the world. This limitation has driven the rise of embodied AI, where autonomous agents, such as humanoid robots, must navigate and manipulate unstructured environments with human-like adaptability.
  At the core of this challenge lies the concept of Neural Brain, a central intelligence system designed to drive embodied agents with human-like adaptability. A Neural Brain must seamlessly integrate multimodal sensing and perception with cognitive capabilities. Achieving this also requires an adaptive memory system and energy-efficient hardware-software co-design, enabling real-time action in dynamic environments.
  This paper introduces a unified framework for the Neural Brain of embodied agents, addressing two fundamental challenges: (1) defining the core components of Neural Brain and (2) bridging the gap between static AI models and the dynamic adaptability required for real-world deployment. To this end, we propose a biologically inspired architecture that integrates multimodal active sensing, perception-cognition-action function, neuroplasticity-based memory storage and updating, and neuromorphic hardware/software optimization. Furthermore, we also review the latest research on embodied agents across these four aspects and analyze the gap between current AI systems and human intelligence.
  By synthesizing insights from neuroscience, we 
  %believe this paper can 
  outline a roadmap towards the development of generalizable, autonomous agents capable of human-level intelligence in real-world scenarios. Our project page is at \href{https://github.com/CNJianLiu/Neural-Brain-for-Embodied-Agents}{Neural-Brain-for-Embodied-Agents}.
	% The rapid evolution of artificial intelligence (AI) has shifted from static, data-driven models to dynamic systems capable of perceiving and interacting with real-world environments. Despite advancements in pattern recognition and symbolic reasoning, current AI systems, such as large language models, remain disembodied, unable to physically engage with the world. This limitation has driven the rise of embodied AI, where autonomous agents, like humanoid robots, must navigate and manipulate unstructured environments with human-like adaptability. At the core of this challenge is the ``Neural Brain", a biologically inspired system that integrates sensing, function, memory storage and updating, and hardware/software. The Key idea of this paper is to introduce a unified framework for the Neural Brain of embodied agents, emphasizing hierarchical neural architectures, predictive coding, multimodal sensory integration, etc. Embodied agent requires not only multimodal sensing (e.g., vision, language, tactile, spatial awareness) but also cognitive capabilities like learning, reasoning, and decision-making, supported by adaptive memory systems and energy-efficient neuromorphic hardware-software co-design. The key challenges are defining the Neural Brain and bridging the gap between static AI models and the dynamic adaptability needed for real-world deployment. By synthesizing advancements in neuroscience, robotics, and machine learning, this work can outline a roadmap for developing generalizable, autonomous agents capable of real-world intelligence.
	\end{abstract}
	
	% Note that keywords are not normally used for peerreview papers.
	}

\maketitle

\IEEEdisplaynontitleabstractindextext

\IEEEpeerreviewmaketitle

\IEEEraisesectionheading{\section{Introduction}\label{Introduction}}

\begin{figure*}[t!]
	\centering
	\includegraphics[width=\textwidth]{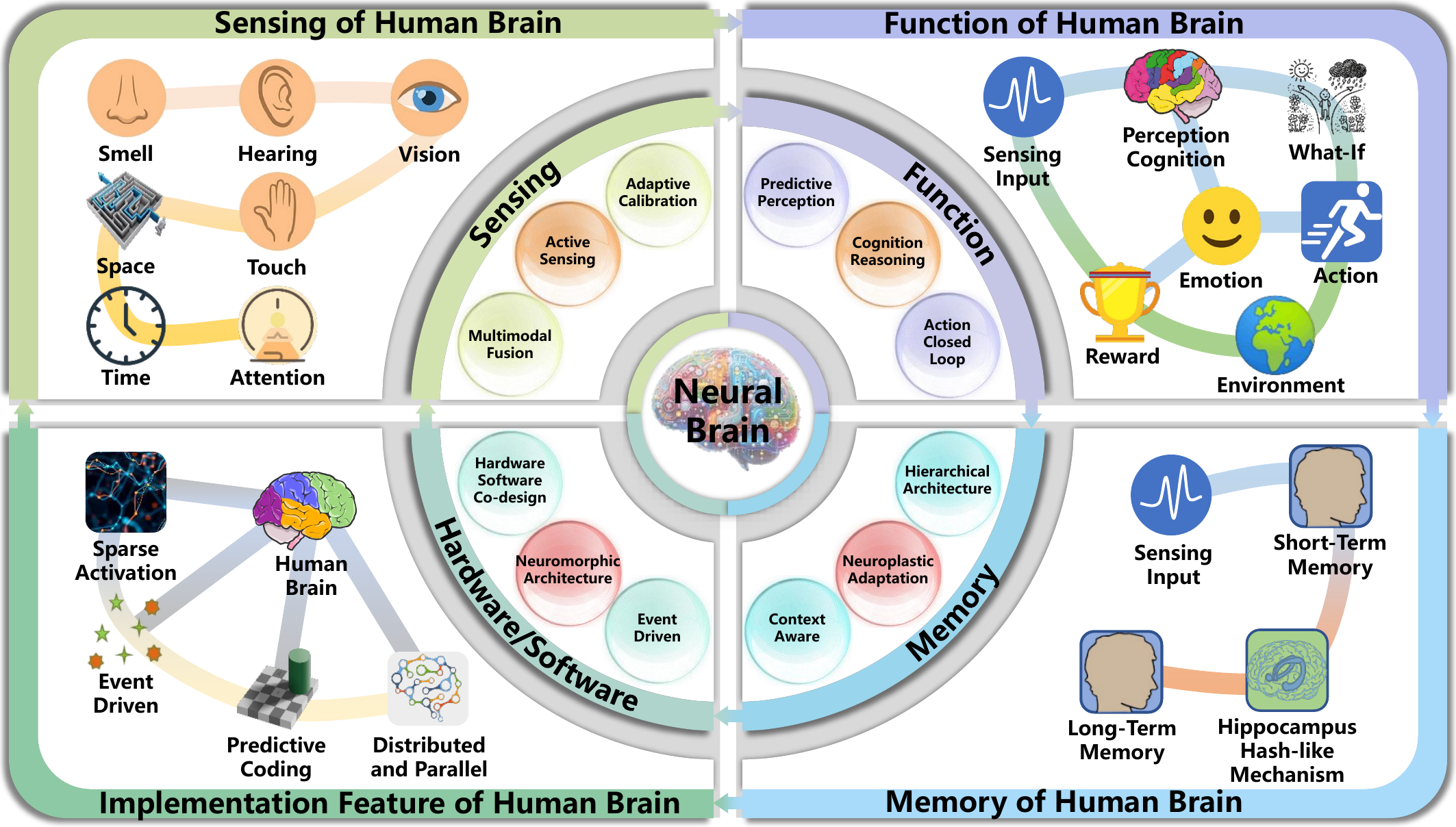}
	\vspace{-1em}
	\caption{%The inner Neural Brain is inspired by the outer Human Brain.
    Human brain-inspired Neural Brain. The human brain comprises four key components: sensing, function (perception, cognition, action), memory (short-term and long-term), and implementation features, such as sparse activation, event-driven processing, predictive coding, and distributed and parallel mechanisms. Inspired by insights from neuroscience, we propose the concept of a Neural Brain for Embodied Agents, which integrates these principles into four distinct modules. The sensing module incorporates multimodal fusion, active sensing, and adaptive calibration to enhance perceptual capabilities. The function module encompasses predictive perception, cognitive reasoning, and action, including an action-closed loop to ensure continuous interaction with the environment. The memory module features a hierarchical architecture, neuroplastic adaptation, and context awareness, enabling agents to store and retrieve information dynamically and efficiently. Finally, the hardware/software module is characterized by event-driven processing, neuromorphic architecture, and hardware-software co-design, ensuring robust and flexible operation. These four core ideas, derived from the structure and functionality of the human brain, aim to empower embodied agents to adapt, learn, and perform effectively in real-world, embodied environments.}
	\label{Fig_teaser_figure}
	\vspace{-1em}
\end{figure*}

\IEEEPARstart{A}{rtificial} intelligence (AI) has undergone a fundamental shift over the past decade, evolving from static, data-driven models to dynamic, interactive systems capable of perceiving and acting within physical environments~\cite{survey1, survey2, liu2024surveywm, zhang2024survey, 7pillars}. While contemporary AI, exemplified by large language models (LLMs)~\cite{brown2020languagemodelsfewshotlearners, achiam2023gpt, chowdhery2023palm, raffel2023exploringlimitstransferlearning, meta2024llama3}, has achieved remarkable success in pattern recognition and symbolic reasoning, these models remain fundamentally disembodied, lacking the capacity to interact with the world in a physically grounded manner~\cite{NMIschulze2025visual, liu2024surveywm}. This has driven the emergence of embodied AI, where autonomous agents actively navigate and manipulate their surroundings like the Atlas Robot of Boston Dynamics~\cite{bostonDynamics2023atlas}, Optimus Robot of Tesla~\cite{tesla2023optimus}, and Unitree G1~\cite{unitreeG1}. However, operating in unstructured real-world environments poses significant challenges, necessitating interdisciplinary advancements at the intersection of robotics, neuroscience, and machine learning~\cite{NMIachterberg2023spatially, NMIhan2024lifelike, NMImeng2025preserving, zaadnoordijk2022lessons, NNSmiconi2025neural}.

\par Given the transition above, a key question emerges: How should an embodied agent be designed to achieve intelligent behavior? Insights can be drawn from biological intelligence, particularly human cognition, which represents one of the most advanced forms of embodied intelligence~\cite{NMIschulze2025visual, NMICOGvon2024cognitive, NMICOGgornet2024automated}. The human brain seamlessly integrates sensory processing, perception, cognition, action, and memory, enabling adaptive behavior in complex and dynamic environments. Neuroscientific research~\cite{jia2023neuroscience} has revealed that this capability arises from hierarchical and distributed neural architectures. For example, the hippocampus encodes spatial representations for navigation, the prefrontal cortex facilitates goal-directed planning, and the cerebellum refines motor execution through predictive error correction. These subsystems interact through predictive coding mechanisms, allowing the brain to continuously update internal models of the world and minimize sensory prediction errors.

Crucially, intelligence is not solely a function of computation but is deeply rooted in embodiment (i.e., the bidirectional interaction between an agent's physical structure, its environment, and its neural processing)~\cite{survey2, liu2024surveywm}. However, the research of embodied intelligence inspired by the human brain from a neuroscience perspective remains \textbf{largely unexplored}, leaving a significant gap in the field.
Even though research on embodied agents has developed rapidly in the past two years~\cite{xisurvey, wangsurvey, xiesurvey, durantesurvey, ma2024survey}, there is currently still no well-established architecture or theoretical foundation for constructing a Neural Brain that can comprehensively drive an embodied agent. %We analyze that 
This is mainly due to two challenges. The first challenge is \textbf{Definitional}: There is no concrete definition of the Neural Brain for embodied agents. % is still lacking. 
Specifically, what functions should the Neural Brain encompass, and what should its architecture look like? The academic and industrial community has yet to reach a clear consensus on these questions. The second challenge is \textbf{Implementational}: Most contemporary AI models are designed with static models. However, a Neural Brain supporting embodied agents should be more human-like in its adaptability and flexibility. Contemporary AI models still face the above significant challenges in achieving real-time, flexible control of embodied agents in dynamic environments. Existing approaches, such as modular perception-cognition-action pipelines~\cite{kim2024openvla,huanginner,sarch2025vlm} and end-to-end reinforcement learning~\cite{he2025asap,ji2024exbody2,cheng2024expressive,peng2021amp,gu2025humanoid}, have demonstrated promising results but remain limited in adaptability, integration, and energy efficiency. Industrial robotic systems often rely on manually engineered state machines, while academic research predominantly focuses on task-specific neural architectures. Both approaches struggle with generalization across novel contexts. For example, LLMs such as GPT-4~\cite{achiam2023gpt} and DeepSeek-R1~\cite{guo2025deepseek}, despite their sophisticated linguistic capabilities, lack direct physical grounding, while robotic controllers trained through simulation-to-reality transfer often exhibit brittleness when deployed in unstructured environments. These limitations underscore the absence of a unified framework that enables embodied agents to learn, predict, and act in a closed-loop fashion, akin to biological intelligence.

% 处理挑战，给出定义（四方面）
\par To tackle the above challenges, we draw inspiration from the human brain, adopting a \textbf{neuroscience perspective} to define the Neural Brain and its components. Specifically, the Neural Brain for embodied agents is a biologically inspired computational framework that synthesizes principles from neuroscience, robotics, and machine learning to facilitate autonomous and adaptive interaction within unstructured environments. Designed to emulate the hierarchical and distributed architecture of the human brain, it integrates four parts: multimodal active sensing (Sensing), closed-loop perception-cognition-action cycles (Function), neuroplasticity-driven memory systems (Memory), and energy-efficient neuromorphic hardware-software co-design (Hardware/Software), as shown in Fig.~\ref{Fig_teaser_figure}.

% 从四个方面展开调研
% Since achieving adaptability and autonomy in embodied agents requires a system analogous to the human brain~\cite{zador2023catalyzing,bartolozzi2022embodied,jia2023neuroscience}
\par Several existing research studies highlight the necessity of multiple sensing modalities, integrated cognitive processes, robust memory mechanisms, and specialized hardware/software architectures. For example, some models~\cite{zhao2023learning, zhao2024aloha, chi2023diffusion} employ a vision-to-action architecture, emphasizing the critical role of visual sensing and perception in embodied agents. Expanding on this, several advanced frameworks~\cite{team2024octo,wang2025scaling,kim2024openvla,liu2024rdt,black2024pi_0,zhanghirt} incorporate both vision and language into action architectures.

Beyond vision and language, other sensory modalities also play a crucial role in embodied intelligence. Specifically, some methods~\cite{chen2020soundspaces,liu2020embodied,severino2024evolution} emphasize the importance of auditory, tactile, and olfactory sensing and perception. Furthermore, some models~\cite{zhu2024spa,yang2024thinking} underscore the critical role of 3D spatial awareness for embodied agents. These insights collectively reinforce the need for multi modal senses and perception~\cite{zaadnoordijk2022lessons}, where vision, language, auditory, tactile, olfactory, and spatial awareness contribute to a comprehensive Neural Brain for embodied agents. Crucially, multiple perception streams alone are not sufficient. Higher-level cognitive abilities, such as learning~\cite{gupta2021embodied}, thinking~\cite{zawalski2024robotic}, reasoning~\cite{huanginner}, problem-solving~\cite{li2025embodied}, decision-making~\cite{chen2023towards}, are crucial for embodied agents to function effectively. Furthermore, to perform tasks over extended time horizons in unseen environments, embodied agents must have the ability to store and update memory~\cite{fang2019scene,sarch2025vlm,zhang2024survey}. Finally, the development of embodied agents necessitates a hardware and software architecture that enables efficient sensing, functions (perception, cognition, and action), and the storage and updating of memory. For example, neuromorphic chips~\cite{davies2021advancing,haessig2018spiking,bhattasali2022neural} support spiking neural networks (SNNs), enabling low-power, event-driven computation. Additionally, deep learning frameworks, such as PyTorch~\cite{paszke2019pytorch} and TensorFlow~\cite{abadi2016tensorflow} provide software platforms for training and deploying large-scale AI models. Overall, these existing works have not systematically combined the four parts of the Neural Brain, and few studies have been conducted from the perspective of neuroscience.

% 与现有survey本质不同及主要贡献highlight
\par With the rapid development of AI agent research, several surveys have emerged~\cite{xisurvey, wangsurvey, xiesurvey, durantesurvey, ma2024survey, liu2025advanceschallengesfoundationagents}. For example, Xi \emph{et al.}~\cite{xisurvey} provided a comprehensive survey on LLM-based agents, proposing a unified framework consisting of brain, perception, and action, and exploring their applications in single-agent, multi-agent, and human-agent cooperation scenarios, as well as their social dynamics in agent societies. Durante \emph{et al.}~\cite{durantesurvey} defined Agent AI as interactive systems integrating multimodal perception and embodied action, leveraging foundation models, external knowledge, and human feedback to enhance intelligence. Their model explores how grounded environments mitigate model hallucinations and enable agents to operate in both physical and virtual worlds.
%Ma \emph{et al.}~\cite{ma2024survey} presented a survey on vision-language-action models (VLAs) for embodied AI, categorizing research into component-level models, low-level action policies, and high-level task planners. It also summarizes key resources, discusses current challenges, and outlines future directions for advancing VLAs in language-conditioned robotic tasks.
Liu \emph{et al.}~\cite{liu2025advanceschallengesfoundationagents} presented a brain-inspired modular framework that systematically explores the core components, self-evolution, multi-agent collaboration, and safety mechanisms of LLM-driven intelligent agents, integrating insights from cognitive science to chart their development and challenges. Overall, these surveys provide a comprehensive overview of current AI agent research from different perspectives.

%\textbf{Distinguished from these research works}, 
\textbf{Unlike prior studies}, 
this paper is the first to introduce an innovative perspective by defining ``\textbf{Neural Brain}'' of embodied agents through the lens of neuroscience. We \emph{not only propose this pioneering definition but also provide a comprehensive design framework for the purpose.} Subsequently, we revisit the existing literature in alignment with this novel framework, highlighting gaps and challenges, and outlining promising directions for future research. Specifically, we synthesize advances in neuroscience, robotics, and machine learning, as shown in Fig.~\ref{Fig_section_1_illustration}, to establish a foundation for Neural Brains in embodied agents. The proposed framework seeks to replicate key principles of biological cognition, including active sensing, a tightly coupled perception-cognition-action loop, etc. By integrating theoretical insights with practical engineering considerations, we aim to advance AI beyond task-specific optimization, laying the groundwork for achieving generalizable embodied intelligence. Our main contributions and highlights are as follows:

\begin{figure}[t!]
\centering
	\includegraphics[scale=0.4]{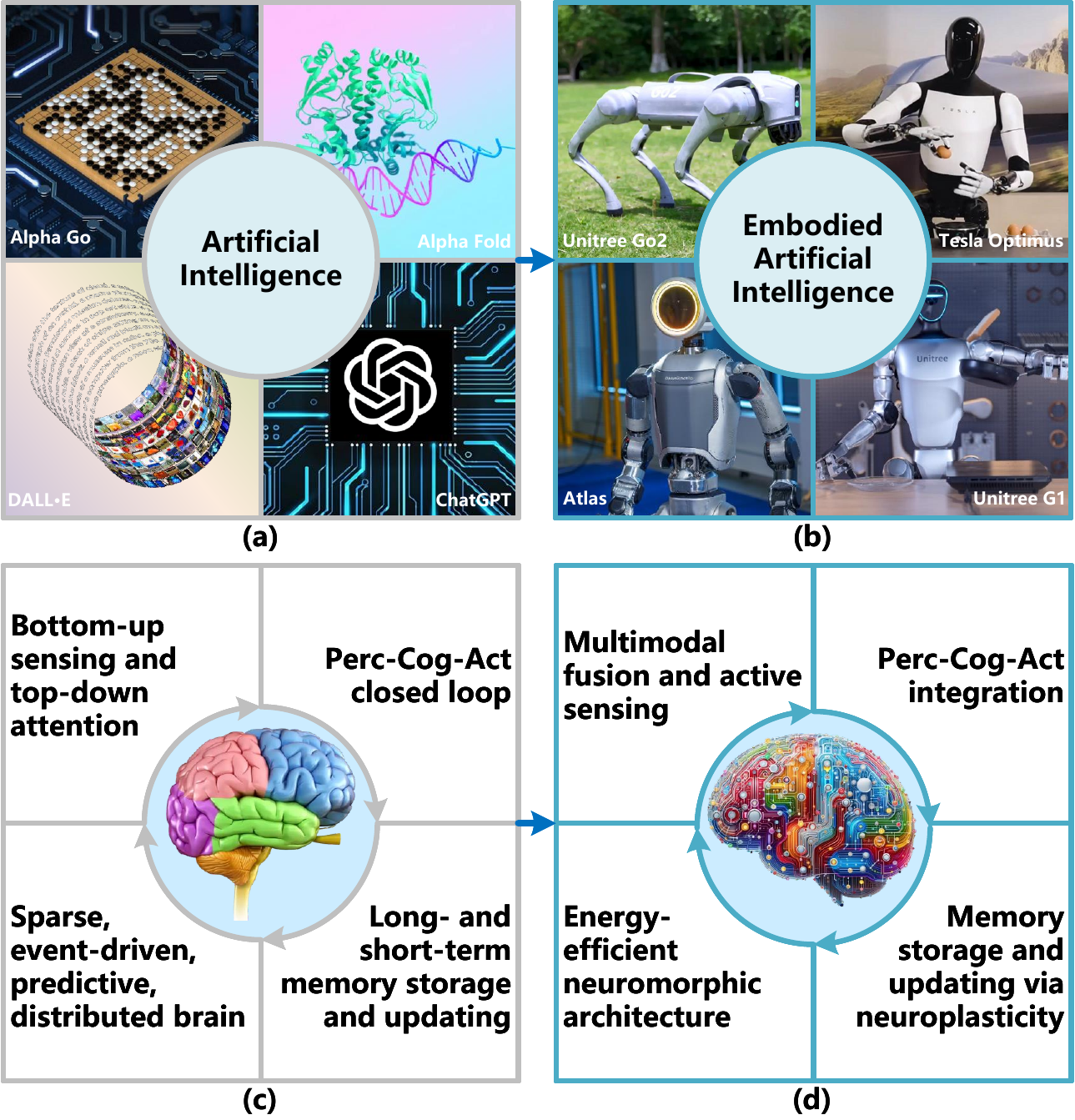}
  \vspace{-0.5em}
	\caption{The evolution from AI to embodied AI. (a) AI excels in pattern recognition but lacks physical interaction with the real world. (b) Embodied AI enables robots like Atlas of Boston Dynamics and Unitree G1 to perceive and act in their environment. (c) Inspired by the human brain, intelligence arises from neural processes that integrate sensing, perception, cognition, action, and memory. (d) Furthermore, this work proposes a concept of ``Neural Brain" for embodied agents, combining neuroscience to achieve generalizable embodied AI.}
	\label{Fig_section_1_illustration}
  \vspace{-1em}
\end{figure}

\vspace{-0.5em}
\begin{itemize} 
\item{To the best of our knowledge, we are the first to explore the design of a brain for embodied agents from a neuroscience perspective. This offers a biologically inspired foundation to enhance the adaptability, efficiency, and cognitive capabilities of embodied agents.}

\item{We identify the fundamental components of a Neural Brain, guided by the core functionality of the human brain and insights from neuroscience. These components encompass sensing, function, memory, and hardware/software. Establishing these capabilities lays the foundation for supporting embodied agents operating in unstructured environments.}

\item{We propose a biologically inspired Neural Brain architecture that integrates multimodal active sensing, integrated perception-cognition-action function, neuroplasticity-driven memory storage and updating, and neuromorphic hardware/software optimization. By synthesizing insights from neuroscience, we outline a strategic roadmap for developing generalizable embodied agents to achieve human-level intelligence in real-world scenarios.}
  
\item{We conduct a comprehensive review of the representative and latest research on AI agents across the defined four dimensions and critically analyze the existing gap between current AI agents and human intelligence. This enables readers to gain a clear understanding of the disparity between current research and the ideal, offering valuable insights for future investigations.}
\end{itemize}
\vspace{-0.5em}
\par The rest of this paper is organized as follows: Sec.~\ref{section_Human_Brain_Neural_Brain} introduces the Neural Brain concept in relation to human neuroscience, offering a formal definition of the Neural Brain; Sec.~\ref{section_Sensing_for_Neural_Brain},~\ref{section_Neural_Brain_Perception_Cognition_Action},~\ref{section_Neural_Brain_Knowledge_Memory_Storage_and_Update},~\ref{section_Neural_Brain_Hardware_and_Software} elaborate on the key components, beginning with sensing in Sec.~\ref{section_Sensing_for_Neural_Brain}, followed by perception-cognition-action in Sec.~\ref{section_Neural_Brain_Perception_Cognition_Action}, memory storage and updating in Sec.~\ref{section_Neural_Brain_Knowledge_Memory_Storage_and_Update}, and hardware and software architectures in Sec.~\ref{section_Neural_Brain_Hardware_and_Software}; Sec.~\ref{section_Dissuasion_and_Further_Directions} discusses current challenges, outlines promising directions for future research, and summarizes this paper. By integrating perspectives from neuroscience, AI, and robotics, this work aspires to bridge the gap between theoretical neuroscience and practical implementation, ultimately contributing to the development of a more robust Neural Brain capable of guiding intelligent embodied agents in dynamic real-world scenarios.

\section{From Human Brain to Neural Brain}\label{section_Human_Brain_Neural_Brain}

\par In this section, we examine the human brain from a neuroscience perspective, then introduce the concept of the Neural Brain by drawing parallels between human neuroscience and computational models. The key idea is to extract fundamental principles of human brain function, and use them to inform the design of a Neural Brain that can operate in real-world, embodied environments.

%%%%%%%%%%%%%%%%%%%%%%%%%%%% Xiongtao %%%%%%%%%%%%%%%%%%%%%%%%%%%%%%%%%%

% 总的思路
% 1.Bottom-Up Sensing and Top-Down Attention
% 1.1.从书中找到所有的传感器，并从支撑这些传感器是Bottom-Up Sensing的角度写
% 1.2.从书中找到所有的传感器，并从支撑这些传感器是Top-Down Attention的角度写
% 2.The Perception-Cognition-Action Closed Loop
% 2.1.在Perception部分，主要是基于GAN的Perception,主要有三点，GAN，memory,和不断学习
% 2.2.在Cognition部分，主要有四个点，强化学习的reward的概念很重要，memory很重要，路径图，反事实思考
% 2.3.在Action部分，反射性动作，反应性动作，基于目的的动作，动作通过练习熟练，误差可以不断修正
% 3.Efficient Memory Storage and Update，现在暂时只有长期记忆，感觉应该假如短期记忆
% 3.1.Storage and Update，存储的硬件和hashing
% 3.2.Optimization，竞争性存储，遗忘，和存储大概而不是细节
% 4.Sparse, Event-Driven, Predictive, Distributed Brain

\subsection{Human Brain: Insights from Neuroscience}

\par Neuroscientific studies offer valuable inspiration for constructing a Neural Brain for embodied agents. Drawing from the architecture and functioning of human brain, we can uncover principles for building embodied agents that mimic the human brain.

% 对应第二章和7.1到7.8

\subsubsection{Bottom-Up Sensing and Top-Down Attention}\label{2.1.1}

\par
\begin{figure}[t!]
\centering
	\includegraphics[scale=0.75]{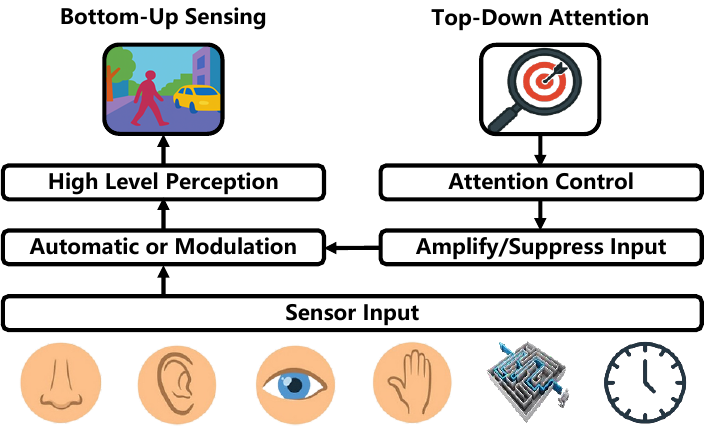}
  \vspace{-1.5em}
	\caption{A schematic overview illustrating how bottom-up sensing and top-down attention contribute to perception, with bottom-up processes driven by sensory input and top-down mechanisms modulating attention--either by amplifying or suppressing specific stimuli--based on task goals.}
	\label{Fig_Bottom_Up_Sensing_and_Top_Down_Attention}
  \vspace{-1em}
\end{figure}

 The human brain processes sensory information through two key mechanisms: Bottom-Up Sensing, which passively encodes raw stimuli, and Top-Down Attention, which actively adjusts perception based on cognitive goals, past experiences, and task relevance. Working together, these mechanisms ensure both accurate and flexible perception, allowing efficient processing of complex inputs (see Fig.~\ref{Fig_Bottom_Up_Sensing_and_Top_Down_Attention}).

\par \textbf{(1) Bottom-Up Sensing:} Bottom-Up Sensing is an automatic, data-driven process that extracts environmental information without conscious effort. It forms the foundation of sensory perception by encoding raw stimuli, which are later refined through higher-order perception and cognitive processing.

\par In olfaction, odors are detected all the time without effort. The olfactory system sorts odor signals early, as seen in fruit flies, where smell receptor neurons (ORNs) send signals to projection neurons (PNs) and then to Kenyon cells (KCs). Only the top 5\% of KCs keep firing for the odor, enabling sparse encoding for efficient processing. Similarly, animals find their way by comparing smells from each nostril, using a process like the gradient descent method. Similarly, hearing mainly works through Bottom-Up processing, where sound waves are analyzed continuously. The cochlea processes sound details using a logarithmic scale, while special connections help smooth hearing signals. Humans figure out where sounds come from by noticing small differences in time, loudness, and phase between our ears, making it easy to hear direction. Likewise, vision also relies on Bottom-Up processing, where light stimuli are automatically structured into neural representations. Signals from the retina reach the primary visual cortex (V1), encoding edges, motion, and color before forming higher-order perceptions. Specialized neurons further distinguish faces, emotions, and textures, ensuring rapid and unconscious recognition of essential stimuli. Tactile perception follows a similar Bottom-Up model, as mechanoreceptors transduce pressure, vibration, and texture into neural signals, which ascend to the somatosensory cortex. This hierarchical system allows real-time environmental responsiveness without cognitive control. In addition, spatial localization integrates sensory inputs automatically. The hippocampus and medial entorhinal cortex (MEC) generate internal spatial maps, with place cells encoding locations and grid cells tracking distances. Time perception also operates passively, as hippocampal neurons encode temporal intervals for sequence recognition and motor coordination. Moreover, speed and acceleration perception are continuously processed across brain regions. Speed cells in the MEC monitor velocity, while hippocampal acceleration encoding fine-tunes motor adjustments. The vestibular system maintains balance and motion perception without active attention. Vector-based encoding further enables unconscious spatial mapping, with boundary and object vector cells constructing spatial reference frames for real-time navigation.

\par \textbf{(2) Top-Down Attention}: In contrast to Bottom-Up Sensing, Top-Down Attention exerts cognitive control over sensory perception, selectively enhancing, suppressing, or filtering stimuli based on context, goals, and prior experiences. In olfaction, past experiences help recognize odors, making familiar scents stand out more. The limbic system links odors with emotions and memories, exemplifying Top-Down modulation. Similarly, in hearing, focusing on certain sounds helps understand speech in noisy places (e.g., the cocktail party effect). Cross-modal effects, like the McGurk effect, show how vision can change what we hear, proving the brain's control over attention. Vision is also shaped by Top-Down processes, where attention highlights important details and reduces distractions. The brain predicts missing parts based on past patterns. In tactile, attention sharpens detail when feeling textures, and experience improves sensitivity, as seen in blind individuals reading Braille. The brain adjusts spatial mapping based on needs, changing grid cell activity for flexible navigation. Time perception also shifts, with brain rhythms in the hippocampus lowering time awareness during deep focus, causing time to feel distorted. Predictions help sense speed and acceleration, allowing quick movement adjustments. The balance system uses Top-Down control to ignore self-caused motion, keeping perception steady. The brain adjusts spatial encoding based on the task, switching between self-centered (egocentric) and world-centered (allocentric) perspectives as needed.

\par Bottom-Up sensing quickly processes environmental information automatically, while Top-Down attention fine-tunes and prioritizes it based on goals. These two systems work together to ensures not only efficient sensory encoding but also adaptive filtering and prioritization of information, enabling organisms to swiftly respond to environmental demands and make context-aware decisions.

\subsubsection{The Perception-Cognition-Action Closed Loop}\label{2.1.2}

\par
\begin{figure}[t!]
\centering
	\includegraphics[scale=0.75]{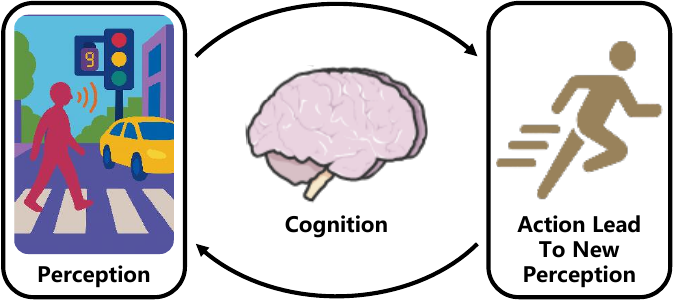}
  \vspace{-0.5em}
	\caption{Schematic illustration of the closed-loop framework of multimodal perception, cognition, and action. The process begins with sensing-driven perception--integrating multimodal sensory inputs such as vision, audition, space, and time. These perceptions are processed through cognitive mechanisms to infer context and determine appropriate responses, and resulting actions dynamically influence subsequent perceptual input.}
	\label{Fig_The_Perception_Cognition_Action_Closed_Loop}
  \vspace{-1em}
\end{figure}

 The Perception-Cognition-Action Loop is a key brain framework that continuously processes information, adapts to new experiences, and improves behavior. Instead of separate steps, perception, cognition, and action work together in a dynamic cycle. Perception interprets sensory input using predictions, cognition turns it into understanding, and action creates feedback that updates both perception and cognition. As illustrated in Fig.~\ref{Fig_The_Perception_Cognition_Action_Closed_Loop}, this ongoing loop supports learning, navigation, and smart decision-making in complex situations.

% 在Perception中的GAN, 对应7.8, A More Versatile Generative Adversarial Network (GAN) in the Brain?

\par \textbf{(1) Perception:} Perception is an active, predictive process where the brain anticipates, corrects mistakes, and uses memory to improve sensory interpretation. It works like a Generative Adversarial Network (GAN), constantly comparing predictions with real input to refine perception and optimize accuracy.

\par Rather than simply processing sensory input, the brain generates expectations based on past experiences and refines them through real-time feedback. This predictive ability extends across sensory systems. Since predictions are not always accurate, the brain corrects errors dynamically. The predictive model acts as a ``generator'', while an ``error detector'' compares expectations with actual input, making necessary adjustments. Memory also plays a crucial role, filling in missing details and influencing recognition. The hippocampus aids this process through sharp-wave ripple replay. To adapt to new environments, the perceptual system undergoes continuous learning. Synaptic weight changes optimize perception over time, as seen in driving: beginners rely on sensory input and thinking, but with experience, prediction improves.

\par Perception is a continuous, adaptive process shaped by predictive modeling, memory integration, and neural plasticity. It allows individuals to anticipate, compare, and refine their understanding of the world.

\par \textbf{(2) Cognition:} Cognition is an active, predictive, and adaptive process, not just passive information storage~\cite{maocom}. The brain builds internal models, predicts future events, and improves understanding through experience. This predictive ability supports individuals in adapting to change. Causal reasoning, including counterfactual thinking, enables individuals to mentally simulate ``what-if'' scenarios, optimize actions, and enhance adaptability before execution.

\par Reinforcement learning is an essential mechanism in cognition, strengthening beneficial behaviors over time. Dopamine-driven learning and reward feedback help refine decision-making. The brain uses replay and preplay mechanisms, guided by sharp-wave ripples, to compare past experiences with new situations, reinforcing knowledge and adapting strategies. Reverse replay aids in evaluating rewards by backtracking from goals, optimizing goal-directed actions. Together, these processes form a modular, feedforward network that enhances cognitive flexibility and prediction. In parallel, memory is fundamental to cognition, allowing the brain to complete missing details, recognize patterns, and anticipate information. The hippocampus and entorhinal cortex integrate episodic memory with real-time perception, forming a continuous feedback loop between past experiences and present input. Replay mechanisms refine stored patterns, preventing information overload and improving retrieval efficiency. This predictive memory processing enables proactive rather than reactive cognition. Additionally, thinking is closely linked to spatial navigation, using similar systems to organize ideas and make decisions. These systems were first developed to help find resources, but now also support abstract thinking and problem-solving. Just as mental maps help arrange knowledge, guide planning, and encourage new ideas, logical thinking follows structured paths, like route maps. Events are stored as points along a path, similar to how certain brain cells track locations. By creating shorter, more general paths, the brain improves speed and flexibility, making knowledge easier to use in different situations. Just as some brain cells predict movement, thought pathways imagine possible outcomes, improving decision-making and problem-solving. Counterfactual reasoning ability to mentally explore different possibilities is further strengthened by ``what-if'' thinking, which helps the brain test other choices before acting. When facing uncertainty, it temporarily blocks a thought path to see if another route leads to the same goal. This works like scientific testing and probability-based learning, where past experience helps guide decisions. If another option works well, the brain adjusts its thinking, making learning and adaptation better. Ultimately, cognition is a self-optimizing system, constantly refining itself through prediction, learning, and adaptation. Adults rarely encounter entirely ``new'' situations; instead, the brain matches new experiences with existing knowledge or recombines stored information, fostering creative thinking and problem-solving.

\par Emotion is also an essential element of reasoning because it provides the cognitive system with the ability to evaluate, prioritize, and assign significance to information in a way that purely logical processes cannot achieve alone~\cite{hourglass}. In the human brain, the amygdala plays a central role in processing emotion, and its influence extends deeply into our ability to make decisions, evaluate risks, assign value, and respond adaptively to changing circumstances. Neuroscience research, particularly the work of Antonio Damasio~\cite{damdes}, has shown that individuals with damage to emotional centers such as the amygdala or ventromedial prefrontal cortex may retain their logical reasoning abilities yet become severely impaired in real-world decision-making. This paradox highlights that emotion is not the opposite of reason but rather a vital contributor to it. The amygdala, for example, helps identify emotionally salient stimuli, encode emotional memories, and modulate attention and valuation processes in ways that support efficient reasoning. When evaluating multiple options, emotional signals processed by the amygdala and integrated with the prefrontal cortex help prioritize what matters most and why~\cite{liukno}. This insight has profound implications for artificial intelligence. Traditional AI systems are built around logical reasoning and pattern recognition, but often lack mechanisms to assign emotional or contextual value to information. As a result, these systems may struggle with ambiguous, uncertain, or socially complex situations that humans navigate easily by leveraging emotion. By modeling some of the functions of the amygdala, AI can begin to approach human-like reasoning. For example, affective computing allows machines to detect, interpret, and respond to human emotions, while emotion-aware reinforcement learning enables agents to simulate internal states that guide decision-making under constraints. Incorporating emotion also improves memory prioritization, attention modulation, and goal-directed behavior, helping AI agents decide not just what they can do, but what they should do in a given situation. Frameworks like sentic computing~\cite{senticnet} and the Personalized Sentiment Analysis Pyramid~\cite{zhuneu} explicitly incorporate affective signals into reasoning processes. These systems model not just subjectivity and polarity, but also deeper emotional and intentional layers, mirroring how the amygdala contributes to emotional salience and behavioral relevance in humans. Ultimately, emotion provides a grounding mechanism for reasoning, enabling both biological and artificial systems to make decisions that are not only rational but also context-sensitive, value-aligned, and socially intelligent. By drawing inspiration from the brain's emotion circuits, particularly the amygdala, AI can move beyond cold calculation toward a form of reasoning that is more robust, adaptive, and human-compatible.

\par By integrating memory, predictive modeling, error correction, and emotional evaluation, cognition enables efficient navigation of an ever-changing world. Through its dynamic interaction with perception, memory, emotion, and reasoning, cognition forms the foundation of intelligence, enhancing learning, decision-making, and problem-solving.

% 从简单的动作到复杂的动作，并且，动作可以逐渐熟练，到偏差修正，到最后的总结

\par \textbf{(3) Action:} Action completes the Perception-Thinking-Action Loop, where sensory information and thought processes turn into movements. However, the loop stays dynamic-actions change the environment, creating new sensory feedback that updates future perception. This continuous cycle supports adaptive learning and real-time improvement of behavior.

\par Reflexive actions are rapid, automatic responses requiring minimal cognitive processing, ensuring survival by enabling quick reactions to environmental changes. For example, the spinal cord controls reflexes like the withdrawal reflex, where pain signals cause an instant movement response. Building upon this, reactive actions extend beyond simple reflexes by incorporating immediate sensory feedback. The motor system continuously adjusts based on proprioceptive and exteroceptive inputs, ensuring stability and coordination. For example, balance corrections while walking on uneven terrain involve rapid sensorimotor integration, with the cerebellum fine-tuning movements in real time. Progressing further, goal-directed behaviors involve prediction, planning, and cognitive control. The brain builds internal models to anticipate action outcomes, optimizing execution. Predictive control is evident in tasks like catching a ball, where trajectory and speed estimations guide movement. In addition, motor learning improves movement through practice and experience, making actions more automatic. Procedural memory helps gradually reduce conscious effort. Activities like riding a bike or playing an instrument show how repeated practice makes movements more automatic. Error correction is also crucial for adaptive motor control. The brain monitors differences between intended and actual outcomes, making real-time adjustments. In tasks that involve vision and movement, sensory signals improve hand-eye coordination. The cerebellum fine-tunes movements by correcting prediction mistakes, helping actions stay accurate and efficient.

\par Action is not merely a response to stimuli but a dynamic, predictive process driven by cognition, learning, and adaptation. Reflexive and reactive actions ensure immediate responses, while goal-directed behaviors enable strategic decision-making. Motor learning refines execution, and reinforcement learning optimizes action selection. Every action reshapes the environment, generating new sensory inputs that restart the loop, fostering adaptive intelligence.

\par From the human neuroscience perspective, integrating cognition into the perception-action loop can be conceptualized in four primary levels from shallow to deep: memory-driven Observe-Orient-Act (OOA), which relies on past experiences to guide actions; knowledge/reasoning-driven Observe-Orient-Decision-Act (OODA), based on rational analysis and reasoning; association/creation-driven Observe-Orient-Create-Act (OOCA), which involves nonlinear thinking and creativity; and hypothesis/discovery-driven Observe-Orient-Hypothesis-Act (OOHA), focusing on exploration through hypothesis generation and discovery. These modes reflect distinct cognitive processes in how the brain integrates perception and action.

\par The Perception-Cognition-Action Loop allows constant learning and improvement. Perception fine-tunes sensory input by predicting and fixing mistakes, cognition uses experience to make better choices, and actions change the environment, keeping the cycle going. This self-adjusting system supports intelligence, problem-solving, and goal-driven behavior, helping people adapt and interact with a changing world.

% 参考第四章和第五章
% 1.记忆不是由单一神经元（如“祖母细胞”）存储，而是涉及多个神经元的网络。这些神经元共同参与，使得单个细胞丢失不会导致记忆消失
% 2.记忆不仅依赖单个神经元，还涉及更大规模的神经网络,快速信号传递（纳秒级）,传统突触传递（毫秒级）,慢速体积传输（数秒至数分钟）
% 3.新经验的存储方式：写入已有的相关神经元（与旧经验重叠）。分配到新的神经元（独立存储）。
% 4.神经元的竞争机制,较活跃的神经元 更可能参与记忆存储
% 5.记忆巩固：哈希（Hashing）视角
% 6.海马系统与不同皮层区域之间的连接变化 快速但短暂。皮层区域之间的连接变化 缓慢但持久。
% 7.反馈哈希模型（Feedback Hashing Model）,记忆存储不是简单覆盖，而是通过哈希和匹配机制进行优化。
%
% 1.在亚细胞水平上，每个突触的权重 可能取决于树突棘的形态
% 2.

% 总的思路，1.记忆存储的硬件条件，2.Memory Replay and Hashing, 3.资源竞争，4.遗忘，5.存储效率

\subsubsection{Efficient Memory Storage and Update}\label{2.1.3}

\par
\begin{figure}[t!]
\centering
	\includegraphics[scale=0.75]{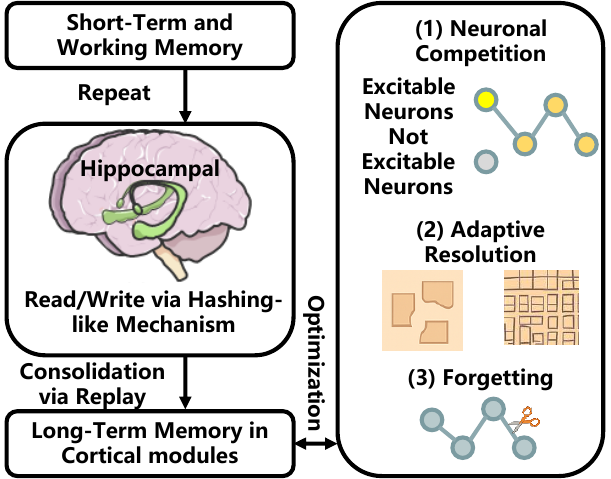}
  \vspace{-0.5em}
	\caption{Schematic representation of efficient memory storage and update, where information enters through short-term and working memory and is transferred to the hippocampus via repetition and rehearsal. The hippocampus encodes and retrieves information through a hashing-like mechanism and consolidates it via memory replay. Long-term memory is refined through neuronal competition, task-dependent adaptive resolution, and forgetting via synaptic weakening.}
	\label{Fig_Efficient_Memory_Storage_and_Update}
  \vspace{-1em}
\end{figure}

 The brain relies on both short-term and long-term memory, each serving distinct yet interconnected roles. Short-term and working memory temporarily store and manipulate information before it is either discarded or transferred to long-term storage, preserving only essential details. For long-term retention, the brain employs efficient strategies to balance stability and adaptability, involving synaptic modifications, neural competition, and memory replay. A schematic representation of these processes is provided in Fig.~\ref{Fig_Efficient_Memory_Storage_and_Update}.

\par \textbf{(1) Short-term and working memory:} Short-term memory (STM) allows the brain to temporarily store information for a few seconds to minutes, enabling immediate processing but with limited capacity, typically around 7$\pm$2 items (Miller's Law). Unlike long-term memory, which relies on synaptic modifications, STM is sustained through persistent neural activity in the prefrontal cortex (PFC) rather than structural changes. Closely related to STM, working memory (WM) extends this function by actively manipulating information for reasoning, decision-making, and problem-solving, integrating inputs from the PFC, parietal cortex, and hippocampus. While WM enhances cognitive flexibility, its reliance on sustained neural firing makes it more susceptible to interference. Despite its fleeting nature, some STM content is consolidated into long-term memory through repetition, rehearsal, and hippocampal-cortical interactions, embedding select information into stable neural networks for future retrieval.

\par \textbf{(2) Storage and Update:} Long-term Memory storage and changes depend on small adjustments to synapse strength, which control how neurons connect. Learning makes useful connections stronger, while weaker ones are removed through competition. The hippocampus not only stores memory but also acts as a dynamic hashing system that efficiently encodes and retrieves information by linking experiences to pre-existing memory structures. Memory consolidation involves structured replay of past experiences, where forward replay replicates the original sequence to strengthen memory.

\par \textbf{(3) Optimization:} Memory encoding is competitive, with highly excitable neurons at the time of learning more likely to be recruited into memory networks. Such a competitive encoding process naturally leads to fewer cells and synapses that represent distant events, while some events occupy many more cells than other events to begin with. Memory networks also optimize resolution based on storage demands. For instance, anterior hippocampal place cells have larger receptive fields than posterior ones, reflecting a spatial encoding gradient where high-resolution details are preserved only when necessary. Forgetting also plays a crucial role in efficient memory management by eliminating outdated or irrelevant information. Memory decay occurs through synaptic weakening, where underused connections undergo long-term depression (LTD), allowing stronger memories to overwrite weaker ones.

\par Memory storage and updating rely on short-term and long-term memory, integrating synaptic plasticity, neural competition, and predictive encoding. Short-term and working memory enable rapid adaptation, while long-term memory ensures lasting retention through structural changes. By balancing stability and flexibility, the brain efficiently manages memory traces, reinforcing important details while discarding irrelevant ones, continuously refining its knowledge for lifelong learning.

% The brain prioritizes new information while avoiding redundant storage through neuronal competition. Memory encoding is competitive, with highly excitable neurons at the time of learning more likely to be recruited into memory networks. If two related experiences occur close in time, their engrams may be linked, enhancing associative memory. In contrast, experiences outside this excitability window recruit separate neuronal populations, ensuring memory distinction. New neurons enhance memory flexibility by increasing storage capacity and representational diversity. In the dentate gyrus, adult neurogenesis supports cognitive adaptability, as newly integrated neurons strengthen storage while enabling efficient updates. Forgetting plays a key role in efficient memory management by eliminating outdated or irrelevant information. Memory decay occurs through synaptic weakening, where underused connections undergo long-term depression (LTD), reducing their neural influence. Stronger new memories can overwrite weaker ones due to resource competition. Moreover, the brain optimizes memory storage by encoding essential patterns rather than redundant details, using predictive coding and generalization. Rather than storing every experience in full detail, the memory system extracts core patterns and rules. And, memory networks adjust detail level based on storage needs. For example, place cells in the front part of the hippocampus cover larger areas than those in the back. This shows a pattern where fine details are kept only when needed.

\subsubsection{Sparse, Event-Driven, Predictive, Distributed Brain}\label{2.1.4}

% 1.节能和稀疏激活
% 2.事件驱动处理以实现实时适应
% 3.记忆引导和预测控制
% 4.运动控制中的分布式和并行处理,需要添加五种处理器，参考8.1

\par The brain achieves efficient low-power operation through a sparse, event-driven, and distributed parallel processing system, as illustrated in Fig.~\ref{Fig_Sparse_Event_Driven_Predictive, Distributed_Brain}, unlike conventional computing models that rely on dense, continuous, centralized, and high-power operations.

\begin{figure}[t!]
\centering
	\includegraphics[scale=0.75]{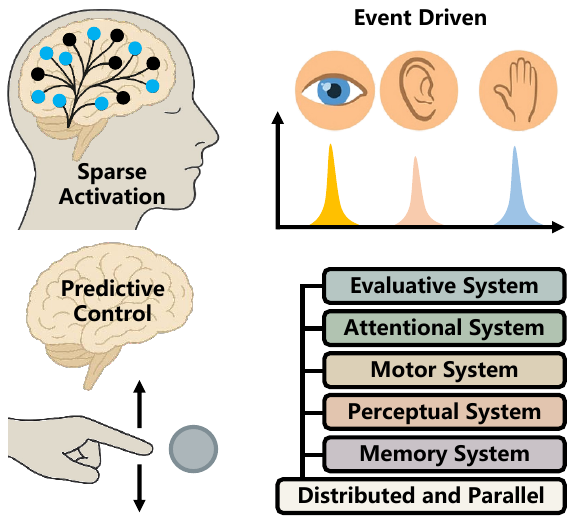}
  \vspace{-0.5em}
	\caption{Visual representation of the brain's sparse, event-driven, predictive, and distributed architecture for efficient motor coordination. The diagram illustrates how energy-efficient control is achieved through selective neuronal activation, real-time responses to sensory events, internal predictive models for smooth motion, and parallel processing distributed across five functional systems.}
\label{Fig_Sparse_Event_Driven_Predictive, Distributed_Brain}
  \vspace{-1em}
\end{figure}

\par \textbf{(1) Sparse Activation:} Even though the brain manages billions of neurons and trillions of connections, it uses only about 20W of power. This efficiency comes from activating only the necessary neurons, avoiding extra activity and saving energy. The same rule applies to movement--rather than using all motor units, the brain only activates the ones needed for the task.

\par \textbf{(2) Event-Driven Processing:} The brain operates on an event-driven model, responding dynamically to external stimuli rather than maintaining constant computational activity. For example, balance and body position feedback adjust movements only when needed, helping maintain stability without extra processing. Basic movements like walking and posture are managed by the spinal cord and brainstem, reducing the workload on higher brain areas and saving mental energy.

\par \textbf{(3) Predictive Control:} Predictive processing allows the brain to optimize movement without constant recalibration. The cerebellum constructs internal models that anticipate sensory consequences, enabling smooth and precise motion. Through motor learning and synaptic plasticity, movements become increasingly automated, reducing cognitive load and energy demand over time.

\par \textbf{(4) Distributed and Parallel Processing:} Rather than relying on a central unit, the brain integrates five key systems: Memory System, Perceptual System, Motor System, Attentional System, and Evaluative System. Like Swarm Learning, motor control is decentralized and event-driven, minimizing energy use by activating only the necessary circuits and adapting quickly to sensory feedback, ensuring precise and efficient coordination.

\par In summary, the brain ensures efficient body coordination through sparse activation, event-driven processing, distributed parallel computation, and predictive modeling. This low-power, adaptive system enables precise motor control while conserving resources, highlighting the brain's superiority over conventional computational models in balancing efficiency and complexity.

\subsubsection{Summary of Human Brain}

\par The human brain is an adaptive, predictive, and efficient system that integrates perception, cognition, memory, and action. It balances Bottom-Up Sensing for raw data processing with Top-Down Attention for selective focus. Cognition relies on predictive learning, reinforcement, and spatial reasoning. Action completes the Perception-Cognition-Action Loop, guided by predictive control, event-driven processing, and distributed motor coordination to enhance efficiency. Memory optimizes storage through neural plasticity, competition, and selective forgetting, using hierarchical encoding, replay mechanisms, and predictive hashing for rapid retrieval and adaptability. Through sparse activation and dynamic learning, the brain continuously refines its understanding, enabling intelligent adaptation to complex environments.

%%%%%%%%%%%%%%%%%%%%%%%%%%%% Jian %%%%%%%%%%%%%%%%%%%%%%%%%%%%%%%%%%
\subsection{Definition of Neural Brain from Neuroscience}

\par
\begin{figure*}[t!]
\centering
\includegraphics[width=\textwidth]{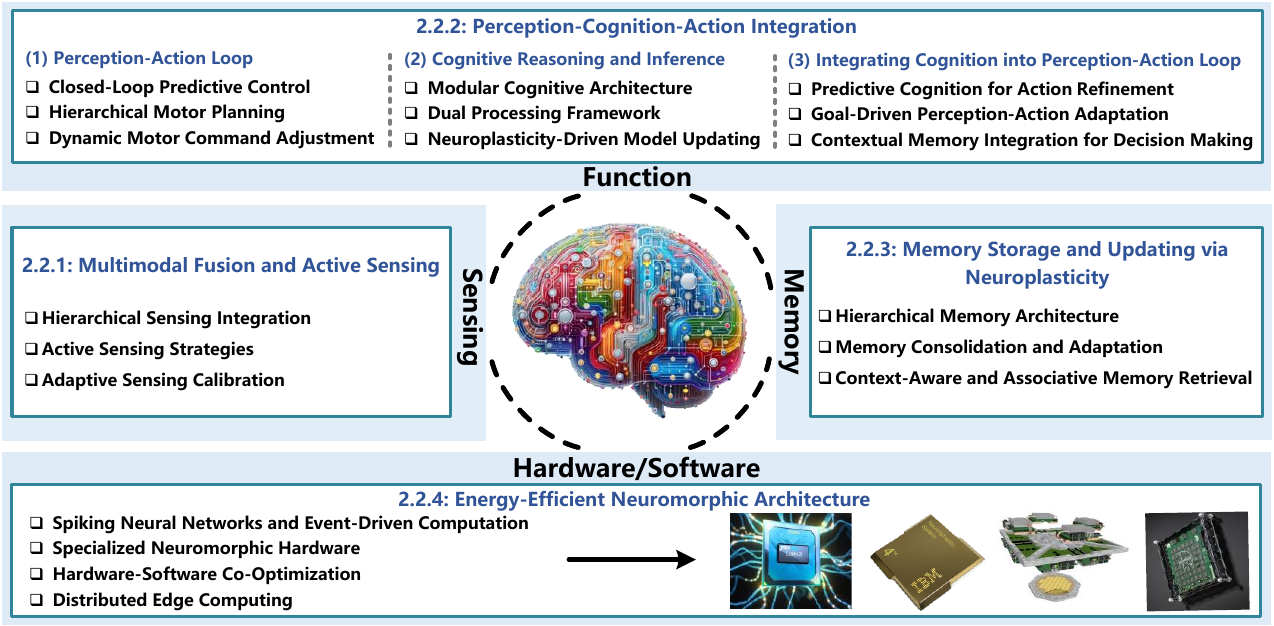}
\vspace{-1.5em}
\caption{Illustration of our proposed Neural Brain, which is inspired by neuroscience and consists of four parts: multimodal fusion and
active \textbf{sensing}, perception-cognition-action integration \textbf{function}, \textbf{memory} storage and updating via neuroplasticity, and energy-efficient neuromorphic \textbf{hardware/software} architecture.}
\label{Fig2.2}
\vspace{-1em}
\end{figure*}

 The Neural Brain for embodied agents is a biologically inspired computational framework that synthesizes principles from neuroscience, robotics, and machine learning to facilitate autonomous and adaptive interaction within unstructured environments. Designed to emulate the hierarchical and distributed architecture of the human brain, it integrates multimodal and active sensing (Sensing), closed-loop perception-cognition-action cycles (Function), neuroplasticity-driven memory systems (Memory), and energy-efficient neuromorphic hardware-software co-design (Hardware/Software), as shown in Fig.~\ref{Fig2.2}. The following sections formalize its core components and their neuroscientific underpinnings.

\subsubsection{Multimodal Fusion and Active Sensing}
\par \textbf{Neuroscientific Basis:}
The human brain processes multimodal sensory information, including vision, audition, tactile, olfaction, and taste, through a hierarchical integration mechanism. Primary sensory cortices extract low-level features, while association cortices synthesize these inputs into a unified perceptual representation~\cite{affectivebci}. In addition, active perception mechanisms, such as saccadic eye movements and selective attention, optimize information acquisition by prioritizing salient stimuli and dynamically adjusting sensory focus.

\par
\begin{figure}[t!]
\centering
	\includegraphics[scale=0.82]{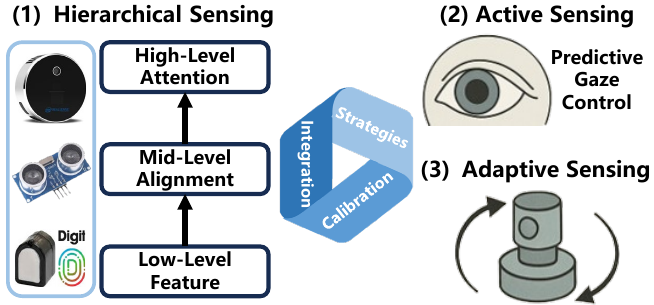}
  \vspace{-1.5em}
	\caption{Diagram of neuroscience-inspired multimodal fusion and active sensing, which includes 3 parts: hierarchical sensing \textbf{integration}, active sensing \textbf{strategies}, and adaptive sensing \textbf{calibration}.}
	\label{Fig2.2.1}
  \vspace{-1em}
\end{figure}

 \textbf{Design Principles:}
\textbf{(1) Hierarchical Sensing Integration:} The Neural Brain should adopt a multi-tiered sensory processing framework that mirrors the functional organization of the human sensory cortex. At the primary level, dedicated sensor modules need to extract spatial-temporal features such as edges, textures, spectral properties, and motion cues. Intermediate processing layers should perform cross-modal alignment to ensure consistency across different sensory streams. Higher-order layers, akin to the association cortices in the brain, need to integrate multimodal information through attention-based fusion mechanisms, generating a coherent and semantically rich environmental representation. This hierarchical structure facilitates robust perception and enhances downstream cognitive and motor processes. \textbf{(2) Active Sensing Strategies:} Inspired by the human brain's ability to actively modulate sensory acquisition, the Neural Brain should implement an adaptive sensing mechanism that dynamically adjusts sensor configurations in response to environmental uncertainties. Through reinforcement learning and self-supervised optimization, the system can optimize sensor orientation, resolution, and sampling frequency autonomously to prioritize high-value information. Just as biological systems employ saccadic eye movements and selective attention to enhance perception, the Neural Brain should incorporate predictive gaze control and attention-guided sensing to minimize redundancy, reduce computational overhead, and improve perceptual efficiency in real-time decision-making. \textbf{(3) Adaptive Sensing Calibration:} To maintain perceptual stability in dynamic and uncertain environments, the Neural Brain should integrate continuous self-calibration mechanisms that compensate for sensor noise, drift, and varying external conditions. Inspired by the adaptive gain control observed in biological sensory systems, the framework needs to incorporate real-time recalibration strategies that adjust sensor parameters based on environmental feedback. These mechanisms should leverage statistical modeling, uncertainty estimation, and self-correcting feedback loops to ensure consistent and high-fidelity data acquisition. By dynamically adapting to fluctuations in sensory inputs, the Neural Brain can sustain reliable perception and enhance the robustness of downstream cognitive and motor processes. The overall multimodal fusion and active sensing diagram is shown in Fig.~\ref{Fig2.2.1}.

\par \textbf{Example:} In autonomous driving, Hierarchical Sensing Integration enables the fusion of multimodal data from cameras, LiDAR, and radar, leveraging the complementary strengths of each sensor for a more comprehensive environmental perception. Active Sensing Strategies dynamically adjust sensor focus based on real-time road conditions, optimizing perception by prioritizing critical features. Adaptive Sensing Calibration ensures robust performance across varying weather and lighting conditions by compensating for sensor drifts and environmental distortions, thereby enhancing perception reliability and decision-making accuracy.

\subsubsection{Perception-Cognition-Action Integration}
\begin{figure}[t!]
\centering
	\includegraphics[scale=0.73]{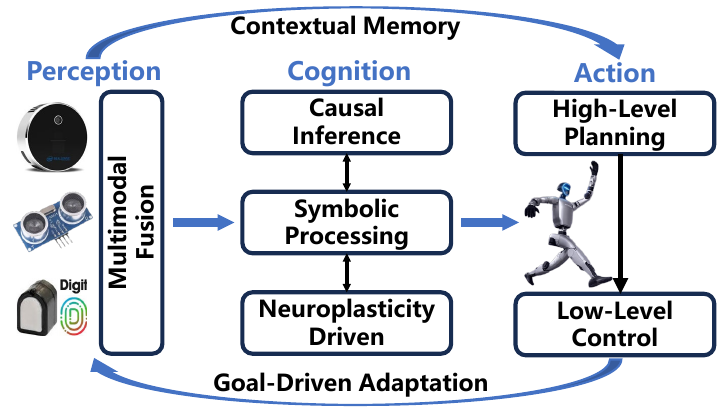}
  \vspace{-1.5em}
	\caption{Diagram of neuroscience-inspired perception-cognition-action integration, which aims to integrate cognitive ability into the perception-action process and close the loop through contextual memory and goal-driven adaptation.}
	\label{Fig2.2.2}
  \vspace{-1em}
\end{figure}

The Neural Brain needs to implement a closed-loop perception-action cycle while also enabling high-level cognitive reasoning and inference, functions largely absent in conventional deep learning models. The overall perception-cognition-action integration diagram is shown in Fig.~\ref{Fig2.2.2}.

\noindent \emph{(a) Perception-Action Loop}
\par \textbf{Neuroscientific Basis:} Human motor control relies on continuous feedback loops between sensory input and motor execution. The cerebellum refines movements through predictive coding, while the prefrontal cortex plans and coordinates actions across multiple timescales.

\par \textbf{Design Principles:}
\textbf{(1) Closed-Loop Predictive Control:} The Neural Brain should be equipped with an internal predictive model that continuously updates and refines motion trajectories. It needs to achieve this by comparing the expected sensory feedback with the observed outcomes, facilitating the refinement of the model for more accurate predictions and enhanced control. \textbf{(2) Hierarchical Motor Planning:} Drawing inspiration from the interaction between the prefrontal cortex and subcortical regions, the Neural Brain should support a clear separation between high-level motor planning (focused on goal-oriented behavior) and low-level motor control (which involves real-time trajectory correction). This separation allows for more efficient management of cognitive resources, enabling the Neural Brain to address complex tasks without compromising real-time performance. \textbf{(3) Dynamic Motor Command Adjustment:} The Neural Brain needs to facilitate real-time adaptation by supporting high-frequency sensory updates and ensuring minimal latency in responses. In this way, it should be capable of dynamically adjusting motor commands in reaction to environmental changes and fluctuations, allowing the system to maintain flexibility and precision in diverse situations.

% \par \textbf{Example:} In robotic manipulation, the perception-action loop enables real-time trajectory correction. Visual and tactile feedback refine grasp strategies, ensuring robust object handling even in unstructured settings.

\noindent \emph{(b) Cognitive Reasoning and Inference}
\par \textbf{Neuroscientific Basis:} The ability of the human brain to generalize, reason, and infer causal relationships emerges from the complex interactions among the prefrontal cortex, hippocampus, and parietal lobe. These regions operate synergistically to facilitate symbolic reasoning, abstract cognition, and long-term memory consolidation. However, these higher-order cognitive functions remain largely underdeveloped in contemporary deep learning models such as LLMs.

\par \textbf{Design Principles:}
\textbf{(1) Modular Cognitive Architecture:} The Neural Brain should integrate three core cognitive subsystems to enable comprehensive cognition. First, a \textit{Causal Inference Module} is needed to develop and continuously refine internal models of the world, facilitating the understanding of cause-and-effect relationships. Second, a \textit{Symbolic Processing Module} should handle logic-based reasoning, providing compensatory mechanisms for the inherent limitations of sub-symbolic neural representations. Finally, \textit{Contextual Memory Integration} should allow the incorporation of background knowledge, supporting robust decision-making by grounding it in prior experiences and contextual information. \textbf{(2) Dual Processing Framework:} To enhance decision-making, the system should combine neural network-based perception with explicit symbolic reasoning. It can leverage advanced techniques such as graph neural networks and adversarial learning to enable effective causal inference, supporting reasoning under uncertainty and ensuring more robust decision-making processes in complex, dynamic environments. \textbf{(3) Neuroplasticity-Driven Model Updating:} Inspired by hippocampal-prefrontal interactions, the system should continuously refine its internal models based on new sensory inputs. This process is critical for ensuring long-term adaptability and enabling the system to make robust inferences in novel scenarios. By incorporating neuroplasticity, the system can adjust to changes in the environment and optimize its decision-making capabilities over time.

% \par \textbf{Example:} In disaster response, the Neural Brain processes multimodal data (video, audio, sensor feeds) to assess situational risks, infer causal dependencies, and generate optimal rescue strategies using prior knowledge and real-time observations.

\noindent \emph{(c) Integrating Cognition into the Perception-Action Loop:} 
%Traditional perception-action loops primarily rely on reactive control, limiting their ability to generalize across tasks and anticipate future states. Integrating cognition into this loop enables goal-directed, predictive, and context-aware decision-making, allowing the system to adapt dynamically to complex and uncertain environments. In biological systems, cognition enhances sensorimotor processing by incorporating causal reasoning, predictive modeling, and memory-driven inference, facilitating anticipatory action selection and long-term behavioral optimization. The interaction between the prefrontal cortex, basal ganglia, and cerebellum enables human motor control to be not merely reflexive but strategic and adaptive. 

\textbf{Neuroscientific Basis:} The human brain refines its perception-action cycle through predictive coding and reinforcement learning, using past experiences to guide future actions. The prefrontal cortex and associative cortical areas form internal models that govern attention, expectation, and decision-making. Meanwhile, the hippocampus and entorhinal cortex store episodic memories, supporting experience-based learning, and the dopaminergic system assigns reward values to behavioral outcomes, influencing action selection. Integrating these principles allows the Neural Brain to move beyond reactive control, enabling adaptive, experience-driven perception-action strategies.

\textbf{Design Principles:} \textbf{(1) Predictive Cognition for Action Refinement:} The Neural Brain should integrate predictive modeling mechanisms that continuously estimate future sensory states based on current motor commands and environmental dynamics. By incorporating Bayesian inference and reinforcement learning, the system can anticipate external perturbations and adjust control strategies preemptively. This predictive capability reduces reaction delays, enhances motion smoothness, and minimizes unnecessary corrective actions, enabling proactive rather than purely reactive behavior. \textbf{(2) Goal-Driven Perception-Action Adaptation:} To ensure that perception remains task-relevant and intention-driven, the Neural Brain needs to incorporate cognitive modulation of sensorimotor processing. High-level task objectives should dynamically influence sensory attention and motor planning, ensuring that perception is not merely passive but actively shaped by behavioral goals. This approach enables the system to prioritize critical environmental cues while filtering out irrelevant data, optimizing computational efficiency, and decision accuracy. \textbf{(3) Contextual Memory Integration for Decision Making:} Effective decision-making in dynamic environments requires context-dependent memory retrieval, allowing the system to leverage past experiences when evaluating action strategies. The Neural Brain should establish hierarchical memory representations, where low-level sensorimotor interactions are associated with high-level cognitive abstractions. By incorporating mechanisms inspired by the hippocampus-prefrontal cortex interactions, the system can retrieve relevant episodic experiences to inform ongoing decision-making, improving adaptability in previously encountered scenarios. 

\par \textbf{Example:} A cognitively enhanced robotic system performing complex manipulation tasks in unstructured environments demonstrates the seamless integration of perception, cognition, and action. While conventional robotic manipulation relies on reactive feedback loops, the Neural Brain needs to incorporate predictive decision-making, goal-directed adaptation, and memory-driven learning to improve efficiency and adaptability. \emph{(a) Perception-Action Loop} enables real-time adaptation by continuously refining grasp strategies based on visual and tactile feedback. When handling objects with varying physical properties, the Neural Brain should dynamically adjust force and contact points in response to unexpected slippage, ensuring stability while minimizing potential damage. Rather than relying on pre-programmed responses, the system needs to interpret sensory inputs and modulate motor actions in real time, enabling flexible interaction in dynamic environments. \emph{(b) Cognitive Reasoning and Inference} allows the system to anticipate object behavior rather than merely reacting to sensory input. For example, when lifting a container with an unknown liquid, the Neural Brain needs to estimate potential shifts in the center of mass and proactively adjust its grasp strategy. By incorporating reinforcement learning and internal simulations, the system should refine motor execution before instability occurs, improving precision and robustness. This predictive capability minimizes reaction delays and ensures smoother manipulation under uncertainty. \emph{(c) Integrating Cognition into the Perception-Action Loop} enables the Neural Brain to optimize decision-making through experience-based adaptation. When encountering an unstable load, the system can retrieve relevant grasping strategies and apply them preemptively. This episodic memory retrieval reduces reliance on trial-and-error learning, facilitating generalization across manipulation tasks while enhancing efficiency and resilience. By associating past sensorimotor interactions with high-level cognitive models, the system should refine response strategies and improve adaptability in novel scenarios. 

\par By integrating real-time sensory feedback, predictive cognition, and memory-driven adaptation, the Neural Brain enhances robotic manipulation, enabling autonomous assembly, human-robot collaboration, and tool-assisted operations in unstructured environments.

\subsubsection{Memory Storage and Updating via Neuroplasticity}
\par \textbf{Neuroscientific Basis:}
The human brain organizes memory hierarchically, with short-term working memory primarily managed by the prefrontal cortex and long-term memory consolidation governed by the hippocampus. Through neuroplasticity, the brain continuously adapts by strengthening neural connections associated with significant experiences while eliminating irrelevant information through synaptic pruning, ensuring efficient memory storage and retrieval.

\par
\begin{figure}[t!]
\centering
	\includegraphics[scale=0.88]{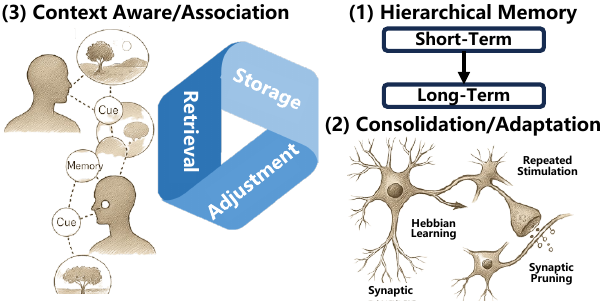}
  \vspace{-1.5em}
	\caption{Diagram of neuroscience-inspired memory storage and updating via neuroplasticity, which includes 3 parts: hierarchical memory \textbf{storage}, memory consolidation and adaptation \textbf{adjustment}, and context-aware and associative memory \textbf{retrieval}.}
	\label{Fig2.2.3}
  \vspace{-1em}
\end{figure}

 \textbf{Design Principles:}
\textbf{(1) Hierarchical Memory Architecture:} The Neural Brain should implement a structured memory system that mirrors the organization of human memory. Short-term memory should support transient information storage, enabling immediate decision-making and task execution, while long-term memory should function as a persistent knowledge repository, allowing for experience-driven learning, adaptation, and generalization across diverse contexts. The interaction between these two memory systems needs to be regulated to optimize information flow, ensuring that relevant experiences are efficiently transferred from short-term to long-term storage. \textbf{(2) Mechanisms for Memory Consolidation and Adaptation:} Inspired by Hebbian learning and long-term potentiation (LTP), the Neural Brain should employ biologically plausible mechanisms for consolidating critical information while filtering out redundant or outdated data. This process can involve synaptic strengthening for frequently reinforced patterns and selective forgetting through synaptic pruning, akin to the optimization observed in neuroplasticity. Adaptive memory mechanisms should dynamically reweigh stored information based on relevance and frequency of usage, preventing memory saturation and maintaining efficiency over time. \textbf{(3) Context-Aware and Associative Memory Retrieval:} To facilitate efficient knowledge utilization, the Neural Brain should implement an advanced retrieval system that activates relevant stored information based on real-time contextual cues. Drawing inspiration from human associative recall processes, this mechanism should integrate semantic understanding, episodic memory linkage, and probabilistic retrieval models to ensure that past experiences are leveraged in a manner that is contextually appropriate. This approach would enhance the agent's ability to make informed predictions, infer causal relationships, and adapt strategies dynamically based on prior interactions and environmental feedback. The overall memory storage and updating diagram is shown in Fig.~\ref{Fig2.2.3}.

\par \textbf{Example:} In human-AI dialogue systems, Hierarchical Memory Architecture organizes conversational knowledge into short-term and long-term memory, enabling immediate response generation while preserving essential context for sustained interactions. Mechanisms for Memory Consolidation and Adaptation facilitate the selective reinforcement of critical dialogue patterns and the dynamic refinement of knowledge representations, ensuring adaptability to evolving user interactions. Context-Aware and Associative Memory Retrieval enhances response coherence by leveraging semantic relationships and episodic recall, allowing the system to retrieve relevant conversational history based on contextual cues and user intent, thereby improving long-term contextual understanding.

\subsubsection{Energy-Efficient Neuromorphic Architecture}
\par \textbf{Neuroscientific Basis:}
Despite handling vast sensory and cognitive workloads, the human brain operates at ~20W, far outperforming conventional AI systems in efficiency. This is achieved through sparse activation, event-driven computation, and parallel processing.

\par \textbf{Design Principles:}
\textbf{(1) Spiking Neural Networks and Event-Driven Computation:} The Neural Brain should leverage SNNs, which operate asynchronously and selectively activate neurons in response to relevant stimuli. This event-driven computation paradigm reduces redundant processing, enhances energy efficiency, and more closely mimics the sparse and temporally precise firing patterns observed in biological neural systems. To fully exploit the advantages of SNNs, the framework needs to incorporate efficient spike-based encoding schemes, neuromorphic learning algorithms, and temporal processing strategies that support real-time perception and decision-making. \textbf{(2) Specialized Neuromorphic Hardware:} Custom neuromorphic chips, such as Intel Loihi and IBM TrueNorth, should be utilized to enable highly parallelized, low-power computation. These architectures are designed to emulate biological neural circuits by supporting massively distributed processing, asynchronous event-driven communication, and local memory storage at the neuron level. By integrating specialized hardware optimized for neuromorphic computing, the Neural Brain can significantly improve scalability, computational efficiency, and real-time processing capabilities, particularly in resource-constrained embedded systems and edge devices. \textbf{(3) Hardware-Software Co-Optimization:} To maximize efficiency, the Neural Brain needs to be developed with a co-optimization approach that tightly integrates hardware and software. Learning algorithms should be designed to exploit architectural properties such as sparsity, locality, and event-driven signal propagation~\cite{pansel}. This requires adapting training paradigms to operate efficiently within neuromorphic constraints, ensuring that model architectures, data flows, and memory hierarchies are optimized for real-time, power-efficient inference. Additionally, frameworks supporting neuromorphic computing, such as SpiNNaker and BrainScaleS, can be incorporated to bridge the gap between algorithmic efficiency and hardware capabilities. \textbf{(4) Distributed Edge Computing:} To minimize latency and reduce dependence on centralized computing resources, the framework should adopt a distributed computing architecture where time-sensitive tasks are processed at the edge while cloud-based updates refine global models. This hybrid approach allows real-time decision-making to be executed locally, reducing communication overhead and improving responsiveness in dynamic environments. Adaptive model compression, federated learning, and efficient on-device inference strategies need to be integrated to ensure that edge devices can operate autonomously while periodically synchronizing with cloud-based knowledge repositories for long-term adaptation and knowledge consolidation. Figure~\ref{Fig2.2} shows some existing energy-efficient neuromorphic architectures.

\par \textbf{Example:} In mobile robotics, Spiking Neural Networks and Event-Driven Computation enable energy-efficient real-time processing by mimicking biological neural dynamics, allowing asynchronous multimodal data fusion and adaptive decision-making. Specialized Neuromorphic Hardware, designed with low-power architectures and event-driven circuits, enhances computational efficiency under extreme power constraints. Hardware-Software Co-Optimization ensures seamless integration between neuromorphic processors and task-specific algorithms, optimizing both inference speed and energy consumption. Furthermore, Distributed Edge Computing leverages neuromorphic processing across multiple edge nodes, reducing reliance on centralized computation and extending operational autonomy in resource-constrained environments.

\subsubsection{Summary of Neural Brain}
\par By integrating key neuroscientific principles, including hierarchical perception, closed-loop sensorimotor control, modular cognitive reasoning, neuroplastic memory adaptation, and energy-efficient neuromorphic computation, the Neural Brain establishes a biologically inspired framework for embodied intelligence. This architecture not only overcomes the limitations of contemporary deep learning models in causal reasoning and adaptive generalization but also enables embodied agents to achieve greater efficiency and robustness in complex, dynamic environments. As a fundamental step toward scalable, resource-efficient, and autonomous embodied intelligence, this framework has the potential to serve as a solid foundation for the next generation of robotics and AI-driven systems.

\subsection{Overlaps between the Human and Neural Brain}

\par The Neural Brain incorporates key computational principles inspired by the human brain to enable embodied intelligence, as shown in Fig.~\ref{Fig_teaser_figure}. While not replicating biological structures directly, it draws from neuroscience to enhance adaptability, perception, and decision-making in real-world environments. The fundamental overlaps between the human brain and the Neural Brain can be categorized into four key domains: hierarchical sensing and attention, perception-cognition-action integration, memory storage and retrieval, and energy-efficient computation.

\noindent \textbf{Hierarchical Sensing and Attention:} Both the human brain and the Neural Brain process sensory information hierarchically, integrating bottom-up perception with top-down cognitive modulation. The human brain encodes raw sensory stimuli through bottom-up processing while simultaneously refining perception via top-down attention based on task relevance and prior experiences. Similarly, the Neural Brain employs multimodal sensory fusion, integrating vision, touch, language, and spatial awareness to construct a unified representation of the environment. Inspired by active perception mechanisms such as saccadic eye movements and selective attention, it dynamically adjusts sensor focus, resolution, and sampling strategies to prioritize salient information and improve computational efficiency. This adaptive approach enables robust perception in complex, unstructured environments.

\noindent \textbf{Perception-Cognition-Action Integration:} Both the human brain and the Neural Brain rely on closed-loop processing to refine actions based on sensory feedback and cognitive reasoning. The human brain continuously integrates bottom-up sensory inputs with top-down cognitive modulation, enabling task-driven perception and adaptive motor control. Similarly, the Neural Brain employs hierarchical perception, predictive modeling, and reinforcement learning to optimize decision-making and motor execution. By leveraging predictive coding and real-time error correction, both systems enhance efficiency in dynamic environments.

\noindent \textbf{Memory Storage and Retrieval:} The human brain efficiently balances short-term working memory with long-term consolidation through neuroplasticity-driven mechanisms. Episodic and semantic memories are dynamically retrieved and updated based on contextual relevance. The Neural Brain adopts similar principles by structuring memory hierarchically, enabling experience-driven adaptation. Hebbian learning-inspired memory consolidation, combined with episodic retrieval, allows the Neural Brain to refine behavior based on past interactions, facilitating continual learning and generalization across tasks.

\noindent \textbf{Energy-Efficient Computation:} Unlike traditional deep learning models that operate with high computational overhead, both the human brain and the Neural Brain utilize sparse, event-driven processing for efficiency. The human brain activates only task-relevant neurons, optimizing energy consumption while maintaining high-level cognitive performance. Inspired by this, the Neural Brain leverages SNNs, neuromorphic hardware, and distributed edge computing to reduce power consumption while sustaining real-time responsiveness. These architectural principles allow embodied agents to function autonomously in resource-constrained environments.

\subsection{Remarks and Discussions}
\par The development of the Neural Brain represents a shift from conventional AI paradigms toward biologically inspired embodied intelligence. Unlike static models trained on fixed datasets, the Neural Brain emphasizes continual adaptation, real-time interaction, and efficient decision-making in unstructured environments. However, several key challenges remain.

\textbf{(1) Bridging Neuroscience and Computation:} Despite significant progress in translating neuroscientific principles into AI and robotics, challenges persist in bridging the gap between biological intelligence and artificial models. While predictive coding, hierarchical memory, and neuromorphic processing provide promising directions, achieving the level of flexibility and generalization seen in the human brain remains an open research problem. Future work should explore biologically plausible architectures that integrate deep learning, symbolic reasoning, and neurophysiological constraints.

\textbf{(2) Scalability and Generalization:} Ensuring that the Neural Brain can generalize across diverse tasks and environments remains a fundamental challenge. Unlike human cognition, which adapts seamlessly to novel scenarios, current AI models struggle with out-of-distribution generalization. Addressing this requires further advancements in unsupervised representation learning, meta-learning, and multimodal perception-action integration to enhance the robustness of embodied intelligence.

\textbf{(3) Ethical and Safety Considerations:} As embodied agents with Neural Brains become more autonomous, ethical concerns surrounding decision-making, human-robot collaboration, and unintended biases must be addressed. Ensuring interpretability, safety, and robustness in real-world applications is critical. Future research should incorporate human-in-the-loop frameworks, uncertainty-aware decision models, and regulatory guidelines to ensure responsible deployment.

\textbf{Research Directions:} Moving forward, the Neural Brain should evolve towards more efficient and biologically plausible implementations. Key directions include:
\begin{itemize}
  \item Enhancing neuromorphic computing with more efficient software and hardware co-design.
  \item Developing self-supervised and few-shot learning techniques for real-world generalization.
  \item Investigating the role of curiosity-driven and intrinsic motivation models in embodied agents.
  \item Expanding human-robot interaction frameworks that integrate emotional and social cognition.
\end{itemize}

\par By integrating neuroscience, robotics, and AI, the Neural Brain lays the groundwork for the next generation of embodied intelligence. Its ability to merge sensory processing, cognition, and adaptive memory mechanisms represents a significant step toward achieving robust, autonomous, and efficient agents capable of operating in complex real-world environments.

\section{Sensing for Neural Brain}\label{section_Sensing_for_Neural_Brain}
 
In this section, we delve into the neuroscience-inspired design of sensing capabilities for an embodied agent's Neural Brain. We begin by examining how the human brain coordinates multimodal sensing through interconnected bottom-up and top-down pathways. We then review the state of the art across several sensory channels-vision, audition, tactile perception, olfaction, spatial awareness, and time, and analyze each through the lens of three unifying design principles:  (1) Hierarchical Sensing Integration, (2) Active Sensing Strategies, and (3) Adaptive Sensing Calibration. Finally, we discuss the remaining challenges, emphasizing these three unifying themes in the sensing domain as crucial embodied agent.

\noindent \textbf{Insights of Human Sensing from Neuroscience:}
The complete insights as illustrated in Sec. \ref{2.1.1}. In short, neuroscience research shows that human perception arises from intricate coordination between bottom-up sensing and top-down attention. Bottom-up pathways capture raw stimuli-such as light, sound waves, or chemical signals, and transform them into progressively more abstract neural representations. Alongside this automatic, data-driven flow, top-down attention exerts cognitive control over which signals are prioritized, amplified, or suppressed. This mechanism is essential for focusing on behaviorally relevant stimuli. This dynamic coordination of bottom-up sensing and top-down attention underpins efficient and robust perception in humans, offering rich inspiration for designing multimodal sensing in embodied agents.

\subsection{Embodied Agent Sensing}

\par Although a considerable body of research has explored sensing solutions for embodied agents, significant gaps remain between current methods and the robust, adaptable sensing observed in humans. Below, we provide an overview of sensing capabilities reported in the literature. For each sensing ability, we highlight key approaches and identify limitations related to hierarchical sensing integration, active sensing strategies, and adaptive sensing calibration. Some representative methods are shown in Table \ref{label3.1}.

\subsubsection{Multimodal and Hierarchical Sensing}\label{Multimodal and Hierarchical Sensing}

\begin{figure*}[t!]
\centering
	\includegraphics[width=0.98\textwidth]{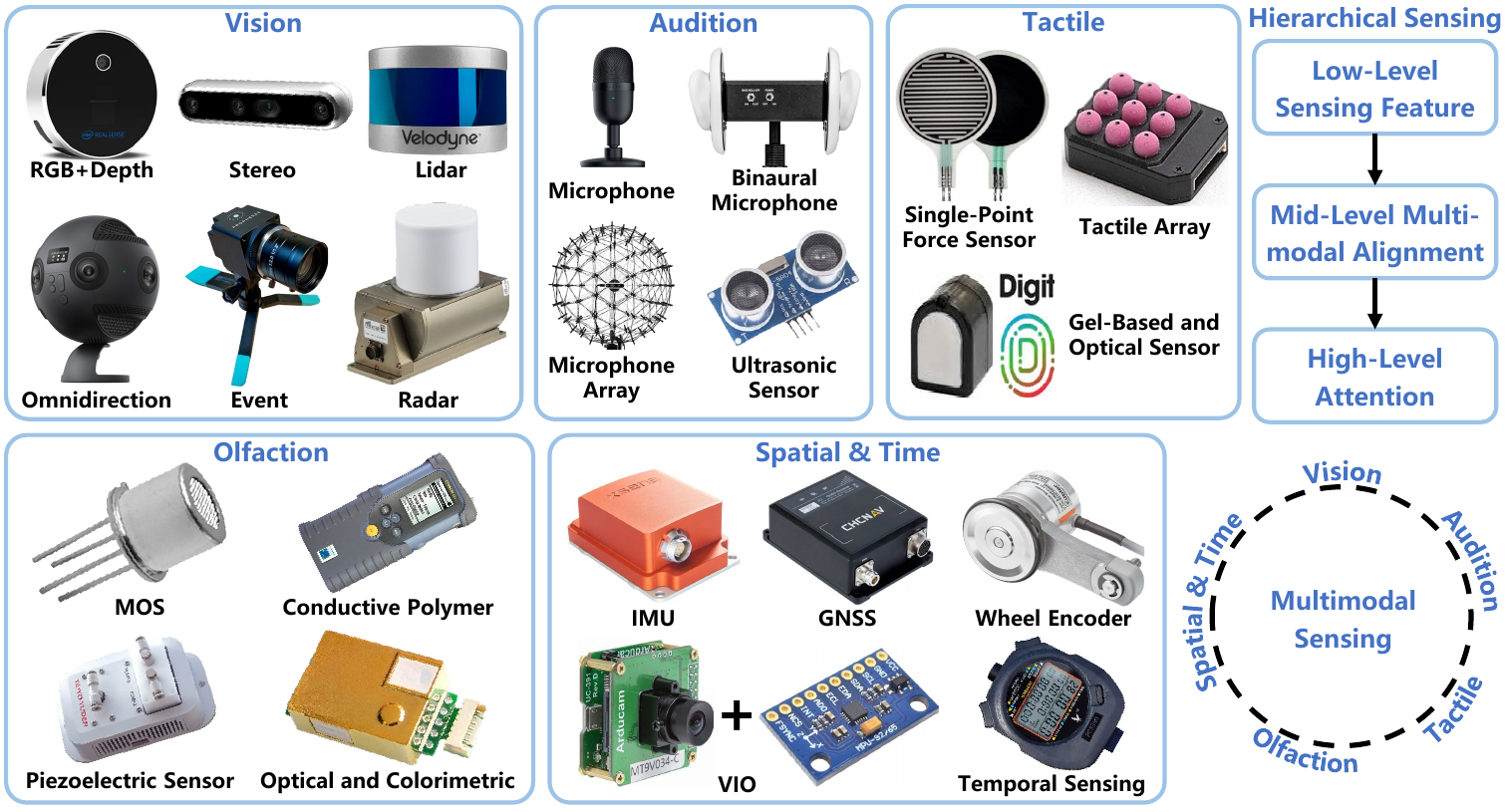}
    \vspace{-1em}
	\caption{The multimodal and hierarchical sensing of embodied agents mainly includes five types of sensors: vision, audition, tactile, olfaction, and spatial and time. Hierarchical sensing includes low-level sensing, mid-level multimodal alignment, and high-level attention.}
	\label{Fig3.1.1}
    \vspace{-1em}
\end{figure*}

A hallmark of biological perception is seamless multimodal fusion: vision modulates audition in the McGurk effect \cite{tiippana2014mcgurk}; touch sharpens visual object recognition; and olfactory cues bias taste perception. Visual cues can override heard phonemes; tactile feedback can disambiguate visual edges. These cross-modal influences illustrate how the human brain performs tightly coupled integration across sensory modalities---not only preserving low-level details but also achieving high-level semantic alignment through layered processing. In contrast, contemporary AI systems still largely treat each modality in isolation, lacking the brain's holistic and dynamic integration strategies. This highlights the need for hierarchical sensing integration, in which sensory signals are acquired, aligned, and fused in a structured, layered fashion to enable coherent and adaptive perception. The overall illustration and visualization of multimodal and hierarchical sensing of embodied agents as shown in Fig. \ref{Fig3.1.1}.

\begin{table*}[t!]
\caption{Representative multimodal \& hierarchical sensing (Sec. \ref{Multimodal and Hierarchical Sensing}) and active sensing \& adaptive calibration (Sec. \ref{Active Sensing and Adaptive Calibration}) methods for embodied agent.}
\vspace{-1em}
\renewcommand{\arraystretch}{1.5}
    \centering
    \resizebox{\linewidth}{!}{
    \begin{tabular}{ccccc|cc}
    \toprule[2pt]
    \multicolumn{5}{c|}{Multimodal \& Hierarchical Sensing}  & \multicolumn{2}{c}{Active Sensing \& Adaptive Calibration}\\
    \midrule
         Vision & Audition & Tactile & Olfaction & Spatial/Time & Active Sensing & Adaptive Calibration\\
    \midrule
         Hartley \emph{et al.} \cite{hartley1992stereo} & Eargle \emph{et al.} \cite{eargle2012microphone} &  Dahiya \emph{et al.} \cite{dahiya2009tactile} &  Yang \emph{et al.} \cite{yang20211d} &  Lechner \emph{et al.} \cite{lechner2000global} &  Zaraki \emph{et al.} \cite{zaraki2014designing}  &  Liao \emph{et al.} \cite{liao2023deep} \\
         
         Wasenm{\"u}ller \emph{et al.} \cite{wasenmuller2017comparison} &  Yang \emph{et al.} \cite{yang2022deepear} &  Heins \emph{et al.} \cite{heins2024force} &  Nambiar \emph{et al.} \cite{lewis2004comparisons} &  Lee \emph{et al.} \cite{lee2004robust} &  Ren \emph{et al.} \cite{ren2023robot} &  Chen \emph{et al.} \cite{chen2023polymer} \\
         
         Gallego \emph{et al.}  \cite{gallego2020event} &  Elko \emph{et al.} \cite{elko2008microphone} &   Tenzer \emph{et al.} \cite{tenzer2014feel} &  Gautschi \emph{et al.} \cite{gautschi2002piezoelectric}  &  Zhang \emph{et al.} \cite{zhang2018tutorial}  &  Kemna \emph{et al.} \cite{kemna2017multi}  &  Chen \emph{et al.} \cite{chen2024self} \\
         
         Behroozpour \emph{et al.}  \cite{behroozpour2017lidar} &  Shung \emph{et al.} \cite{shung1996ultrasonic} &  Donlon \emph{et al.} \cite{donlon2018gelslim} & Liu \emph{et al.} \cite{liu2020recent}  &  Li \emph{et al.} \cite{li2020spatio}  &  Ai \emph{et al.} \cite{ai2022deep}  &  Cao \emph{et al.} \cite{cao2022gvins} \\
    \bottomrule[2pt]
    \end{tabular}
    }
    \label{label3.1}
    \vspace{-1em}
\end{table*}

(1) Vision: The vision system in embodied agents leverages a broad range of sensors, each offering distinct perceptual capabilities tailored to different environmental conditions and task demands. These modalities collectively form the foundation for robust and adaptive visual understanding in physical, interactive settings.

\textbf{Monocular cameras} serve as the most fundamental vision sensors, capturing high-resolution 2D images with low cost and power consumption. Though inherently lacking depth information, they enable rich appearance-based perception, supporting tasks such as object detection \cite{zou2023object}, tracking \cite{yilmaz2006object}, and texture-based learning \cite{kiechle2018model}. \textbf{Stereo cameras} \cite{hartley1992stereo} exploit spatial disparity between two monocular lenses to estimate scene depth, mimicking the human visual system's binocular perception. They provide real-time, pixel-level depth and are widely adopted in visual SLAM \cite{engel2015large}, path planning \cite{hrabar20083d}, and navigation tasks \cite{Guo_navigation}. Some advanced systems allow baseline reconfiguration or focus tuning, enabling range-adaptive depth perception in dynamic scenes. \textbf{Depth cameras} \cite{li2019survey} capture dense 3D geometry through either structured light or time-of-flight (ToF). Structured light systems, such as Kinect v1 \cite{wasenmuller2017comparison}, project known infrared patterns and infer depth via deformation analysis. These are especially suited for indoor human-robot interaction. ToF cameras emit modulated IR signals and estimate per-pixel depth from return delay or phase shift, enabling fast, robust sensing under motion and varying lighting, which is essential for manipulation, hand-eye coordination, and contact-rich control. \textbf{Event cameras} \cite{gallego2020event} offer a biologically inspired sensing paradigm by asynchronously detecting changes in pixel-level brightness, producing sparse but temporally precise event streams. Their high dynamic range, microsecond latency, and temporal sparsity make them well-suited for agile robotic control, collision avoidance, and high-speed tracking. Unlike traditional frame-based cameras, event sensors prioritize motion and discontinuity, encoding only informative changes in the scene. \textbf{Omnidirectional cameras} \cite{nayar1997catadioptric} capture panoramic 360-degree visual data, enabling holistic situational awareness from a single viewpoint. These are particularly beneficial in navigation and social interaction analysis. When integrated with local sensors such as depth or stereo cameras, omni-vision acts as a global perceptual layer, guiding attention and augmenting local detail with broader context. \textbf{LiDAR} \cite{behroozpour2017lidar} systems actively probe the 3D environment by emitting laser pulses and measuring their time-of-flight to generate point clouds. They serve as high-resolution spatial scaffolds for 3D mapping and long-range localization. For embodied agents, LiDAR data often anchors other modalities, enabling terrain-aware navigation, object avoidance, and context-aware behavior in large-scale environments. \textbf{Radar} systems \cite{patole2017automotive} complement vision with long-range, weather-robust sensing using radio waves. Though their spatial resolution is lower than LiDAR, radar excels at detecting relative motion, velocity, and occluded targets-especially in fog, rain, or dust. This makes radar a vital component in safety-critical outdoor navigation \cite{ort2020autonomous}.

(2) Audition: The auditory system \cite{vanarse2016review, rubel2002auditory} in embodied agents enables sound-based perception \cite{bizley2013and}, supporting capabilities such as speech interaction \cite{guenther2015role}, spatial localization \cite{battaglia2003bayesian}, obstacle detection \cite{rodriguez2012assisting}, and multimodal fusion \cite{song2023multi}. Compared to vision, auditory sensors provide key advantages in environments with limited visibility or occlusion, and serve as lightweight complements to more data-intensive sensors. The core hardware components include microphones \cite{eargle2012microphone}, microphone arrays \cite{elko2008microphone}, and ultrasonic transducers \cite{shung1996ultrasonic}.

\textbf{Microphones} \cite{eargle2012microphone} are the most basic auditory sensors, transducing air pressure fluctuations into electrical signals. Single microphones enable general sound detection and amplitude-based activity classification but are limited in directional awareness. Despite their simplicity, they serve as essential input channels for human-robot communication, auditory cue detection, and sound-triggered responses. \textbf{Binaural microphones} \cite{yang2022deepear}, often placed at head-equivalent positions on robots, simulate interaural level and time differences, enabling front-back disambiguation and azimuth estimation. This biologically inspired layout supports immersive spatial audio processing and conversational interaction, crucial for assistive robots and service agents operating in close human proximity. \textbf{Microphone arrays} \cite{elko2008microphone}, composed of multiple spatially arranged microphones, extend sensing capabilities by enabling directional hearing through time-difference-of-arrival or phase analysis. These arrays support beamforming, source separation, and acoustic localization, analogous to the human auditory system's binaural processing. Linear, circular, or spherical array geometries are used depending on platform constraints and coverage requirements. Modern arrays can focus auditory attention toward specific directions, suppressing noise from irrelevant sectors-mirroring top-down auditory focus in human cognition. \textbf{Ultrasonic sensors} \cite{shung1996ultrasonic} transducers emit and receive high-frequency sound waves (typically 40 kHz) to estimate distances via time-of-flight. Due to their low cost and robustness, ultrasonic sensors are widely used for short-range obstacle detection, docking, and wall-following tasks. In swarms or dynamic environments, arrays of ultrasonic sensors offer 360-degree near-field awareness, acting as reactive sensing layers for navigation and collision avoidance.

(3) Tactile: Tactile sensing \cite{dahiya2009tactile} endows embodied agents with the ability to physically interact with their environment through direct contact, providing rich information about texture, force, pressure, vibration, compliance, and surface geometry. Unlike vision or audition, which are predominantly exteroceptive and remote, tactile sensing is inherently proximal and interactive, enabling nuanced manipulation \cite{tegin2005tactile}, contact-rich behaviors \cite{yu2023mimictouch}, and fine-grained environmental understanding \cite{huang20243d}.

\textbf{Single-point force sensors} \cite{heins2024force} represent the most fundamental form of tactile input. Typically based on piezoresistive, capacitive, or piezoelectric mechanisms, these sensors convert applied force into electrical signals, allowing for basic contact detection and force magnitude estimation. They are commonly integrated into grippers, fingertips, or joints, supporting essential functions such as grasp confirmation, contact detection, and overload protection. \textbf{Tactile arrays} \cite{tenzer2014feel} extend single-point sensors into high-resolution 2D matrices, often inspired by the spatial layout of mechanoreceptors in the human skin. These arrays can resolve fine-grained pressure patterns, allowing agents to infer object contours, surface textures, and distributed contact forces. Advanced tactile skins may incorporate multimodal sensing channels-such as temperature or vibration, in a layered structure, mimicking the hierarchical integration found in biological dermal systems. The spatially distributed nature of these arrays enables the system to localize contact, estimate surface curvature, and adjust force distribution in real-time. \textbf{Gel-based and optical tactile sensors} \cite{donlon2018gelslim} provide high-resolution tactile imaging by measuring the deformation of an elastomeric medium upon contact. Embedded cameras and light sources track surface displacement and indentation patterns, offering visual reconstructions of contact geometry at micrometer precision. These sensors bridge the gap between tactile and visual modalities, enabling high-resolution perception of slip, shear, and micro-geometry during manipulation. Their rich signal structure supports downstream learning-based contact models and slip prediction algorithms.

(4) Olfaction: Olfactory sensing \cite{sankaran2012biology} equips embodied agents with the ability to detect and interpret chemical cues in the environment, supporting a wide range of tasks such as gas leak detection, hazardous material identification, environmental monitoring, and even nuanced interaction through scent-based communication. Though less extensively explored than vision or audition, olfaction offers a fundamentally distinct perceptual modality rooted in molecular recognition. At the core of robotic olfactory systems lie sensor arrays composed of chemically diverse sensing elements. Each element within the array is fabricated with distinct materials. Upon exposure to specific chemicals, these elements undergo measurable changes in their electrical or physical properties.

\textbf{Metal-oxide-semiconductor (MOS) sensors} \cite{yang20211d} operate based on redox reactions that occur at a heated metal oxide surface. When gas molecules interact with this surface, they modulate its electrical conductivity, enabling detection of specific chemical species. These sensors are prized for their robustness, low cost, and broad-spectrum sensitivity, making them a practical choice for mobile robotic platforms engaged in environmental monitoring or hazardous material detection. However, their reliance on elevated operating temperatures and relatively slow recovery times may limit responsiveness in time-critical tasks such as real-time leak tracing or dynamic source localization. In contrast, \textbf{conductive polymer sensors} \cite{lewis2004comparisons} provide a lower-power alternative well-suited for energy-constrained platforms. These sensors respond to gas exposure by altering the conductivity of organic polymer films, allowing embodied agents to detect a broader range of volatile organic compounds at ambient temperatures. Despite their flexibility and compact form factors, they are more vulnerable to environmental drift and signal degradation, requiring periodic recalibration or adaptive compensation-particularly in long-term deployments or highly dynamic conditions. For tasks demanding ultra-sensitive chemical discrimination, such as identifying low-concentration toxins \cite{feng2010colorimetric} or distinguishing between structurally similar odorants \cite{laska1999olfactory}, \textbf{piezoelectric sensors} \cite{gautschi2002piezoelectric} devices offer exceptional precision. These sensors detect nanogram-scale mass changes on coated substrates by monitoring shifts in resonance frequency caused by molecular adsorption. Their ability to resolve minute chemical differences makes them ideal for applications in medical diagnostics \cite{kononenko2001machine}, food quality assessment \cite{gomiero2018food}, or micro-contaminant tracking \cite{mpatani2021adsorption} in confined environments. Another category gaining traction in robotic olfaction is \textbf{optical and colorimetric sensors} \cite{liu2020recent}, which transduce chemical interactions into detectable variations in light absorption, reflection, or emission spectra. These systems enable non-contact detection, making them particularly advantageous for embodied agents that must operate near sensitive or inaccessible targets-such as detecting explosive traces or identifying early fire hazards without physical contact. Their compact integration into transparent or disposable substrates further supports modular, application-specific sensing configurations.

(5) Spatial and Time: Spatial and temporal sensing enables embodied agents to maintain a coherent understanding of their own motion, orientation, and environmental dynamics over time-essential for navigation \cite{Guo_navigation}, coordination, memory formation, and interaction within complex, changing environments. Unlike traditional exteroceptive modalities, spatial and temporal perception often relies on proprioceptive and inertial measurements, providing critical support for continuous tracking, prediction, and control. \textbf{Inertial Measurement Units (IMUs)} \cite{ahmad2013reviews} form the cornerstone of spatiotemporal awareness in embodied systems. These compact modules typically integrate accelerometers, gyroscopes, and sometimes magnetometers to capture linear acceleration, angular velocity, and orientation relative to the Earth's magnetic field. IMUs provide high-frequency, low-latency measurements that enable real-time state estimation, dead-reckoning, and stabilization. Their outputs are essential for mobile platforms, legged locomotion, and aerial vehicles, supporting functions such as attitude control, gait timing, and trajectory prediction. Although IMUs offer short-term ego-motion estimation, they are prone to cumulative drift, requiring correction through fusion with exteroceptive signals such as vision or LiDAR. \textbf{Global Navigation Satellite Systems (GNSS)} \cite{lechner2000global}---including GPS \cite{xu2007gps}, GLONASS, and Galileo \cite{revnivykh2017glonass}---provide absolute positioning by triangulating signals from orbiting satellites. In open outdoor environments, GNSS ensures macro-scale localization with meter-level accuracy, making it indispensable for large-scale tasks such as autonomous driving \cite{joubert2020developments}, agricultural robotics \cite{vougioukas2019agricultural}, and outdoor exploration. However, GNSS signals often degrade or fail in indoor spaces, urban canyons, or under dense canopies, prompting the need for hybrid localization strategies that fuse GNSS with inertial or visual data. \textbf{Wheel encoders} \cite{lee2004robust} and \textbf{joint encoders} serve as fundamental proprioceptive components for wheeled and articulated robots. By measuring incremental rotations or joint angles, these sensors provide high-resolution motion tracking over short time spans. They enable embodied agents to reconstruct fine-grained postural changes, estimate local displacement, and execute closed-loop motor control. When combined with inertial and visual feedback, encoder signals contribute to robust odometry, calibration, and sensor fusion pipelines. \textbf{Visual-inertial odometry (VIO)} systems \cite{zhang2018tutorial} combine inertial data with visual cues (typically from monocular or stereo cameras) to estimate motion trajectories in real time. These systems closely emulate the spatiotemporal integration observed in humans, where visual and vestibular cues cooperate to maintain balance and spatial awareness. In such architectures, fast inertial updates handle rapid motion, while slower visual corrections mitigate drift-forming a hierarchical fusion scheme that parallels biological perception across multiple timescales. \textbf{Temporal sensing} \cite{li2020spatio}, though often embedded implicitly within other modalities, plays a distinct role in embodied perception. Precise timestamping of multimodal inputs, synchronization across sensor streams, and recognition of temporal sequences enable agents to perceive duration, detect causality, and anticipate changes. Some systems integrate dedicated real-time clocks or temporal encoders to maintain synchronized, temporally ordered data streams. Others apply biologically inspired temporal encoding mechanisms-such as spike trains in event cameras \cite{gallego2020event,zhang2024frequency} or phase-coded neural representations, to process time as an explicit perceptual variable. These mechanisms support behavior coordination, sequence learning, and internal simulation of motion and interaction.

Taken together, these sensing modalities collectively establish a neuroscience-inspired multimodal hierarchical sensing architecture that mirrors the layered processing observed in the human brain. This architecture is organized into three functional tiers. At the low-level feature layer, each sensor independently extracts modality-specific spatial and temporal cues: monocular cameras capture high-resolution textures and edges; stereo \cite{hartley1992stereo} and depth sensors \cite{li2019survey} yield local 3D structure; event cameras \cite{gallego2020event} detect rapid brightness transitions; IMUs \cite{ahmad2013reviews} report inertial states; tactile arrays \cite{donlon2018gelslim} register force distributions; and gas sensors \cite{sankaran2012biology} measure chemical gradients. These raw yet information-rich signals form the foundation of the sensory system. At the mid-level alignment layer, multi-sensor data streams are temporally synchronized and spatially calibrated into a unified coordinate system. This layer addresses differences in sampling rate, frame delay, field of view, and physical alignment. It enables coherent fusion of heterogeneous modalities---for instance, aligning LiDAR geometry with RGB contours \cite{di2023photometric}, or co-registering event streams with ToF depth-thus ensuring spatial and temporal consistency across sensing channels. At the high-level fusion and attention layer, context-aware mechanisms integrate and selectively emphasize salient features across modalities to produce semantically meaningful percepts. Inspired by top-down modulation in the brain, this layer may prioritize dynamic objects in vision, amplify relevant sounds in noisy environments, or suppress redundant tactile input during stable grasps. Multimodal attention strategies \cite{nagrani2021attention} allow embodied agents to reweight sensor contributions based on task goals and environmental uncertainty, enabling efficient resource allocation and robust real-time operation.

\begin{figure*}[t!]
\centering
	\includegraphics[width=0.9\linewidth]{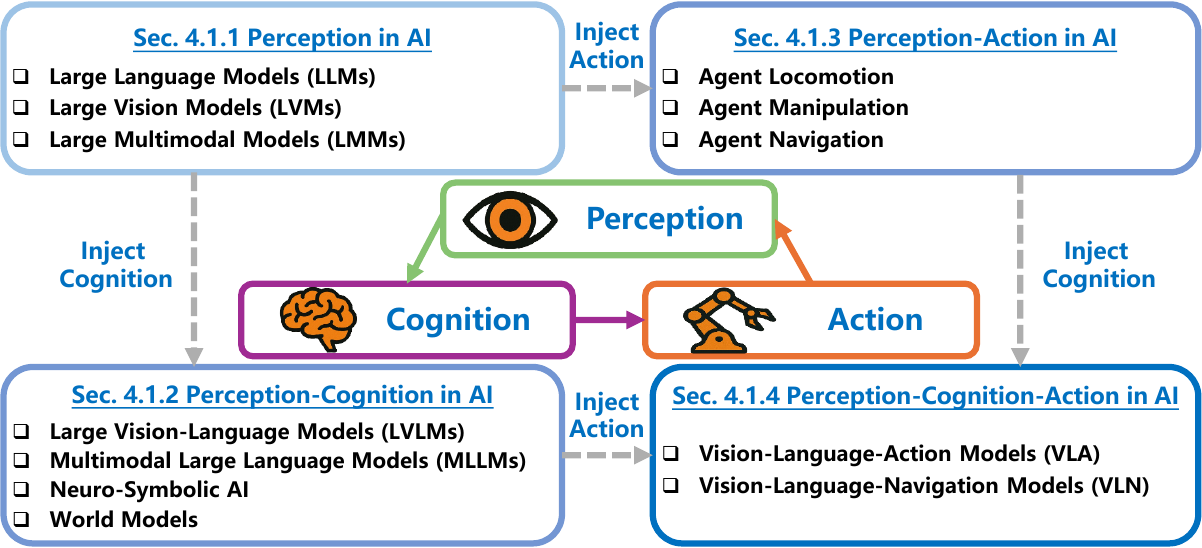}
    \vspace{-1em}
	\caption{\textbf{A schematic overview of the perception-cognition-action framework for embodied agents.} The three components are addressed either in isolation, in integrated pairs, or as a complete loop. The cognition-action pathway is not discussed, as it cannot occur independently of perception. The rightward direction indicates the integration of action, while the downward direction represents the integration of cognition into the process.}
	\label{Fig_4.1_overview}
    \vspace{-1em}
\end{figure*}

\subsubsection{Active Sensing and Adaptive Calibration}\label{Active Sensing and Adaptive Calibration}

To operate robustly in dynamic environments, the Neural Brain of an embodied agent must transcend passive data collection and evolve into an active, adaptive perceptual system. Central to this transformation are two biologically inspired principles: active sensing strategies \cite{bajcsy1988active} and adaptive sensing calibration \cite{aller2022audiovisual}. Together, they endow agents with the ability to modulate how information is acquired and to maintain consistent perceptual fidelity in the face of environmental uncertainty, sensor drift, and system aging.

\textbf{Active sensing} refers to the agent's capacity to dynamically reconfigure its sensing behavior based on task demands, environmental feedback, and internal uncertainty estimates \cite{ren2023robot}. Rather than treating sensors as static data providers, the Neural Brain treats them as controllable, goal-driven interfaces with the world. Inspired by biological systems that utilize saccadic eye movements and selective auditory focus, active sensing in embodied agents involves mechanisms such as gaze control \cite{zaraki2014designing}, viewpoint planning \cite{ren2023robot}, and adaptive sampling \cite{kemna2017multi}. These actions can be optimized through reinforcement learning \cite{ding2023learning} or attention \cite{li2021attention} strategies that prioritize high-value information and suppress irrelevant or redundant input, thereby reducing computational overhead and enhancing real-time decision-making.

For example, monocular and stereo vision systems mounted on mobile bases or pan-tilt actuators can perform active viewpoint selection-redirecting gaze toward salient targets \cite{chen2022learning}, avoiding occlusions, or systematically scanning unexplored regions. Omnidirectional cameras, though mechanically static, support attention-guided digital transformations such as panoramic cropping or region-of-interest magnification \cite{ai2022deep}. Depth cameras such as ToF and structured light systems adjust infrared projection parameters-including modulation frequency, exposure time, and gain-based on scene reflectivity or lighting conditions \cite{adam2016bayesian}. Event cameras, inherently selective by design, fire only in response to brightness changes; some systems further modulate this selectivity through programmable thresholds that adapt to motion scale or illumination \cite{graca2024scidvs}. LiDAR and radar systems exemplify active sensing at a structural level-that is, their capability for active perception is not limited to algorithmic control, but is inherently realized through hardware design. These systems actively emit signals, such as laser pulses or radio waves, to interrogate and sense their surroundings \cite{tran2023adaptive}.

\textbf{Adaptive calibration}, on the other hand, ensures that the sensory data remains stable and accurate as internal and external conditions evolve \cite{preiss2018simultaneous}. This self-correcting capability is crucial in embodied contexts, where physical wear, environmental noise, and operational drift are inevitable. Drawing on the biological principle of adaptive gain control, the Neural Brain implements real-time recalibration routines that update sensor parameters.

In visual systems, this includes online adjustment of exposure, lens distortion compensation, white balance correction, and photometric normalization \cite{luo2022hybrid}. Stereo and depth cameras regularly refine intrinsic and extrinsic calibration using visual landmarks or loop closures, maintaining geometric consistency under temperature shifts, vibration, or mechanical misalignment \cite{liao2023deep}. Tactile arrays, especially those based on optical or elastomeric deformation, adjust baseline pressure maps and response thresholds in response to contact history, ambient conditions, or sensor fatigue \cite{chen2023polymer}. Olfactory sensors, particularly metal-oxide and polymer-based sensor, rely on signal normalization, baseline drift compensation, and adaptive feature scaling to preserve detection accuracy across long-term deployments or volatile environments \cite{chen2024self}. Proprioceptive modalities such as IMUs also benefit from dynamic bias estimation and fusion. IMUs, for instance, are continuously re-aligned with visual or GNSS data to correct for gyroscopic drift and magnetic interference \cite{cao2022gvins}.

Ultimately, the interplay between active sensing and adaptive calibration transforms perception from a static, feedforward process into a closed-loop, context-sensitive, and self-stabilizing system. By selectively directing attention, dynamically tuning sensing behavior, and maintaining internal coherence, the Neural Brain achieves perception that is not only accurate, but behaviorally relevant and resilient. These mechanisms are essential for enabling embodied agents to operate autonomously in the real world.

\subsection{Remarks and Discussions}\label{sec3.2}

% \subsubsection{Current Gaps and Challenges}

Despite substantial progress in sensor hardware, algorithmic fusion, and embodied system integration, several fundamental challenges remain unresolved.

\textbf{(1) Limited Multimodal Fusion Scope.}
Most embodied systems rely on a narrow subset of sensory combinations-most commonly RGB-D vision or vision-inertial odometry \cite{zheng2024fast,wisth2022vilens}. While these pipelines are effective in constrained domains, they do not generalize to the breadth of sensory diversity exhibited in biological organisms. Integrating tactile, auditory, olfactory, and proprioceptive inputs into a single, coherent perception stack remains rare. This limits cross-modal redundancy and inhibits robust sensing in environments with degraded or ambiguous cues.

\textbf{(2) Underdeveloped Active Sensing Behaviors.}
Although mechanical actuation of cameras and limited field-of-view optimization are increasingly used \cite{tilmon2021fast}, current methods fall short of the adaptive, high-frequency sensing strategies observed in biological systems (e.g., saccadic vision, head tilting, whisker movement, or beam steering). Existing solutions are often hardcoded or constrained to a few modalities and lack task-dependent reconfiguration based on attention or environmental salience. Furthermore, non-visual sensors such as LiDAR \cite{placed2023survey}, radar \cite{harlow2024new}, and olfaction \cite{karakaya2020electronic} are seldom dynamically adjusted or selectively enhanced or suppressed at runtime.

\textbf{(3) Inadequate Real-Time Calibration.}
Sensor calibration is typically handled offline or infrequently online \cite{qiu2020real}. However, real-world deployments involve constant fluctuations in temperature, lighting, mechanical stress, or electromagnetic interference-all of which degrade sensor fidelity over time. Existing methods often fail to adapt to such changes, especially in long-duration or mobile settings. Few systems implement continuous, closed-loop self-calibration schemes akin to the gain adaptation and feedback-driven homeostasis observed in human perception.

\section{Neural Brain Perception-Cognition-Action}\label{section_Neural_Brain_Perception_Cognition_Action}

% \par This section explains how human intelligence is structured as a perception-cognition-action loop, where perception drives cognition, cognition informs action, and action modifies perception. The human brain operates as an adaptive feedback system, continuously refining perception, cognition, and action based on real-time interactions with the environment.

This section focuses on the perception-cognition-action loop, a core organizing principle of intelligent behavior. In the human brain, perception, cognition, and action are tightly coupled through predictive, memory-driven feedback mechanisms that enable adaptive interaction with the environment. Inspired by this, the Neural Brain framework seeks to achieve similar closed-loop functionality in embodied agents. We begin by examining foundational neuroscience insights, then review how current AI systems handle perception, cognition, and action. Finally, we summarize existing limitations for building more robust and human-like perception-cognition-action systems.

% \subsection{Human Perception-Cognition-Action: Insights from Neuroscience (A simple description is enough)}

% \par Human intelligence operates as a continuous and dynamic loop of perception, cognition, and action, which is central to how we interact with and adapt to the world.

% \subsection{Embodied Agent Perception-Cognition-Action}

% \par This section reviews existing AI models that integrate perception, cognition, and action, evaluating their strengths, limitations, and relevance for Neural Brains. AI models are progressing toward integrated Perception-Cognition-Action systems, but gaps remain.

% \subsubsection{Perception in AI Models}

% \par This section examines AI models capable of perception, particularly in embodied agents. 

% \par \textbf{(a) Large Language Models (LLMs):} Strong at textual interpretation but lack real-world perception.

% \par \textbf{(b) Large Vision Models (LVMs):} Capable of image recognition but struggle with real-time sensory adaptation.

% \par \textbf{(c) multimodal Large Models (MLMs):} Integrate language and vision, approaching multi-sensory AI.v

% \subsubsection{Perception-Cognition in AI Models}

% \par This section analyzes AI models that attempt to bridge perception with cognition.

% \par \textbf{(a) Models like Transformer-based architectures (GPT, CLIP, Flamingo, etc.) integrate multimodal understanding, but still lack real-time decision-making.} 

% \par \textbf{(b) Hybrid models incorporating reasoning mechanisms show promise for cognitive enhancement in Neural Brains.}

\noindent \textbf{Insights of Human Perception-Cognition-Action from Neuroscience:}
% \par Human intelligence operates as a continuous and dynamic loop of perception, cognition, and action, which is central to how we interact with and adapt to the world.
The complete insights as indicated in Sec.~\ref{2.1.2} emphasize that human perception, cognition, and action are interconnected through a tightly coupled neural loop that supports adaptive intelligence. From a neuroscience perspective, perception is governed by predictive coding mechanisms, where the brain actively anticipates incoming sensory signals and corrects discrepancies using memory-based inference. Cognition builds upon these perceptual processes through internal simulation, causal reasoning, and experience-driven learning, with the hippocampus and related structures supporting memory consolidation and replay. Reinforcement signals, particularly those modulated by dopaminergic pathways, help refine decision-making and guide future actions. Action itself is not a simple motor response but a predictive and continuously adjusted process, shaped by internal models and sensory feedback. Together, these mechanisms create a self-organizing system that enables humans to learn from experience, simulate alternatives, and act purposefully in uncertain and dynamic environments.

%~\cite{huang2001planning, biswal2021development}, balance control~\cite{juang2013design, lin2008development}

\begin{figure*}[t!]
\centering
	\includegraphics[width=0.9\textwidth]{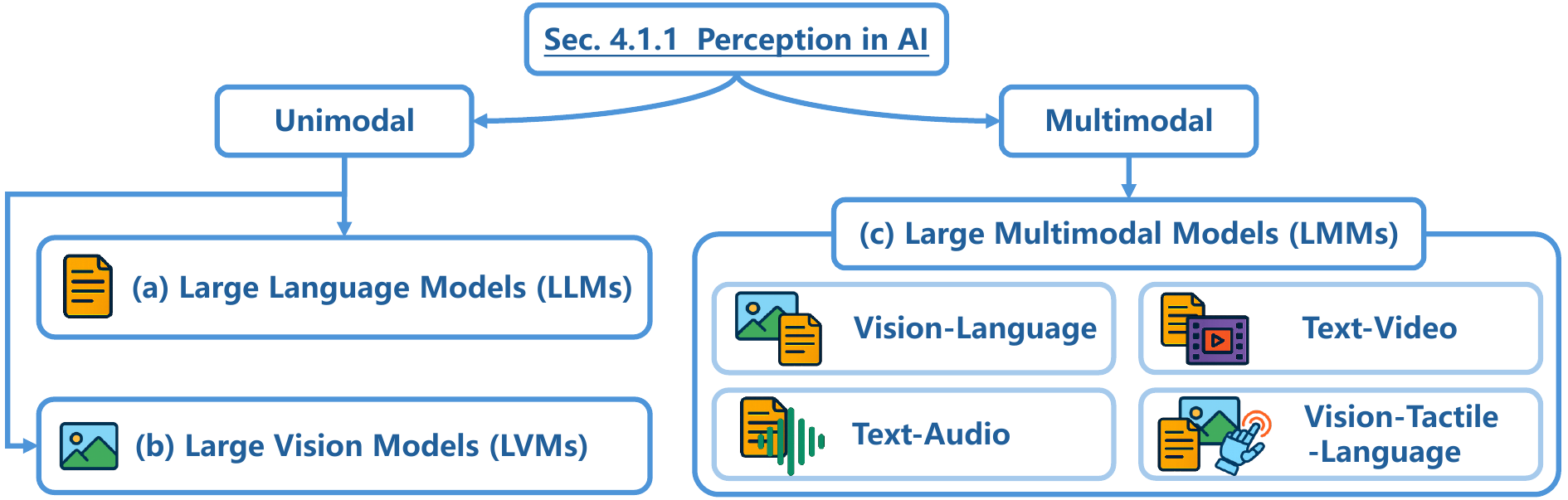}
    \vspace{-1em}
	\caption{General framework of perception in AI models, illustrating the processing of diverse input modalities through unimodal or multimodal paradigms using large foundational models. Unimodal large models primarily include (a) Large Language Models (LLMs) and (b) Large Vision Models (LVMs), while (c) Large Multimodal Models (LMMs) extend perception across multiple modalities, with vision, language, or both often serving as central components, typically combined with additional modalities such as video, audio, or tactile input.}
	\label{Fig_4.1.1}
    \vspace{-1em}
\end{figure*}

\subsection{Embodied Agent Perception-Cognition-Action}

\par This section reviews existing AI models that integrate perception, cognition, and action, evaluating their strengths, limitations, and relevance for Neural Brains. AI models are progressing toward integrated Perception-Cognition-Action systems, but gaps remain. The structure of the section is illustrated in Fig. \ref{Fig_4.1_overview}.

\subsubsection{Perception in AI}
\label{Perception in AI Models}

% \par This section examines AI models capable of perception, particularly in embodied agents. 

\par Perception is the process by which raw sensory data is analyzed and structured into internal representations, enabling further downstream processing~\cite{paolo2024embodiedai}. Advancements in machine learning have led to sophisticated methods for learning feature representations and embedding spaces, significantly improving the transformation of raw data into meaningful information. This section focuses exclusively on approaches that address the perception process with minimal or no complex reasoning. In particular, it emphasizes vision, language, and multimodalities. A general framework on perception in AI models is illustrated in Fig. \ref{Fig_4.1.1}. Some representative works are summarized in Table \ref{tab:perception_models}.

\par \textbf{(a) Large Language Models}

LLMs form a core component of linguistic perception in embodied agents. Trained on large-scale text corpora using unsupervised or self-supervised learning, these models enable systems to interpret and generate human language. LLMs provide deep contextual embeddings that facilitate high-level linguistic understanding, serving as the language-processing backbone in multimodal embodied architectures~\cite{chen2025exploringembodiedmultimodallarge}. However, their capabilities are largely restricted to pattern recognition; they often lack grounding in sensorimotor experience and do not inherently support symbolic reasoning or goal-directed cognition.

Early models such as BERT~\cite{devlin2019bertpretrainingdeepbidirectional} introduced bidirectional Transformer architectures using masked language modeling and next-sentence prediction to learn contextual embeddings. XLNet~\cite{yang2020xlnetgeneralizedautoregressivepretraining} extended this with a permutation-based objective, capturing bidirectional context without predefined masks. T5~\cite{raffel2023exploringlimitstransferlearning} unified diverse NLP tasks into a text-to-text format, while UL2~\cite{tay2023ul2unifyinglanguagelearning} combined generative and discriminative objectives for more flexible understanding.

Autoregressive models such as the GPT series~\cite{radford2018improving, radford2019language, brown2020languagemodelsfewshotlearners} have further advanced generative capabilities. GPT-3, in particular, demonstrated few-shot and zero-shot generalization by scaling model size and training on vast text corpora. Despite impressive fluency and versatility, these models operate through statistical inference and lack perceptual grounding or causal modeling-key components of human-like perception.

More recent architectures like LLaMA and LLaMA-2~\cite{touvron2023llamaopenefficientfoundation, touvron2023llama2openfoundation} offer efficient, open-source alternatives with broad applicability across language tasks. However, like other LLMs, they remain confined to surface-level linguistic cues, with no direct integration of multimodal perception or interaction with embodied environments. Thus, while LLMs are effective in linguistic comprehension, their role in perception remains isolated from physical context and grounded cognition.

\begin{table*}[t!]
\caption{Representative models for perception (Sec. \ref{Perception in AI Models}) in embodied agents.}
\vspace{-1em}
\renewcommand{\arraystretch}{1.5}
\centering
\resizebox{\linewidth}{!}{
\begin{tabular}{ccccc}
\toprule[2pt]
\multicolumn{2}{c}{\textbf{Category/Method}} & \textbf{Year} & \textbf{Details} & \textbf{Highlights} \\
\midrule

% LLMs block
\multirow{7}{*}{\rotatebox{90}{\scriptsize \shortstack{\textbf{(a) Large Language Models}}}}
& BERT~\cite{devlin2019bertpretrainingdeepbidirectional} & 2018 & Masked Language Model & Pretrained with bidirectional context learning \\ 
& XLNet~\cite{yang2020xlnetgeneralizedautoregressivepretraining} & 2020 & Autoregressive Language Model & Permutation-based factorization improves context modeling \\ 
& T5~\cite{raffel2023exploringlimitstransferlearning} & 2020 & Text-to-Text Framework & Unified approach to NLP via seq2seq pretraining \\ 
& UL2~\cite{tay2023ul2unifyinglanguagelearning} & 2023 & Multi-objective Pretraining & Combines prefix, infill, and MLM training \\ 
& GPT-3~\cite{brown2020languagemodelsfewshotlearners} & 2020 & Few-shot Learner & Supports in-context learning across tasks \\ 
& LLaMA~\cite{touvron2023llamaopenefficientfoundation} & 2023 & Open Foundation Model & Optimized for open-weight low-resource training \\ 
& LLaMA-2~\cite{touvron2023llama2openfoundation} & 2023 & Instruction-Tuned LLM & Adds safety tuning and chat capabilities \\ 
\noalign{\global\arrayrulewidth=1.5pt}
\hline
\noalign{\global\arrayrulewidth=0.4pt}

% VLMs block
\multirow{10}{*}{\rotatebox{90}{\scriptsize \shortstack{\textbf{(b) Large Vision Models}}}}
& DETR~\cite{carion2020end} & 2020 & Transformer-based Object Detection & Reformulates detection as direct set prediction \\ 
& Deformable DETR~\cite{zhu2021deformable} & 2021 & Efficient Transformer Detector & Adds multi-scale deformable attention for faster convergence \\ 
& DINO~\cite{caron2021emergingpropertiesselfsupervisedvision} & 2021 & Object Detection Pretraining & Learns semantic features without labels \\ 
& DINOv2~\cite{oquab2024dinov2learningrobustvisual} & 2024 & General-Purpose Detection Backbone & Scalable SSL model with robust representations \\ 
& SAM~\cite{kirillov2023segment} & 2023 & Universal Segmentation Model & Zero-shot segmentation using prompts \\ 
& SAM2~\cite{ravi2024sam2segmentimages} & 2024 & Improved Segmentation Architecture & Refines predictions with better efficiency \\ 
& SEEM~\cite{zou2023segment} & 2023 & Foundation Model for Segmentation & Supports diverse segmentation tasks with referring input \\ 
& Point-BERT~\cite{yu2022point} & 2022 & 3D Object Recognition & Learns discrete point cloud tokens \\ 
& PointNeXt~\cite{qian2022pointnext} & 2022 & 3D Point Cloud Backbone & Residual architecture for spatial reasoning \\ 
& BEVFormer~\cite{li2022bevformer} & 2022 & Bird's-Eye View Perception & 3D scene understanding from cameras \\
\noalign{\global\arrayrulewidth=1.5pt}
\hline
\noalign{\global\arrayrulewidth=0.4pt}

% LMMs wrapper
\multirow{27}{*}{\rotatebox{90}{\scriptsize \shortstack{\textbf{(c) Large Multimodal Models}}}}

% Vision-Language
& \multicolumn{4}{c}{\textbf{(1) Vision-Language}} \\ \cline{2-5}
& VL-T5~\cite{cho2021unifying} & 2021 & Multimodal Text-Generation Encoder-Decoder & Unifies VL tasks via conditional text decoding \\
& ViLT~\cite{kim2021vilt} & 2021 & Transformer with Lightweight Patch Projection & Utilizes ViT to replace CNNs and region features \\
& CLIP~\cite{radford2021learningtransferablevisualmodels} & 2021 & Contrastive Vision-Language Model & Aligns images and texts via dual-encoder training \\ 
& ALIGN~\cite{jia2021scalingvisualvisionlanguagerepresentation} & 2021 & Web-scale Vision-Language Training & Uses large-scale noisy alt-text data \\ 
& Perception Encoder~\cite{bolya2025perception} & 2025 & Multitask Visual Encoder & Adaptable to both spatial and linguistic tasks \\ 
& Grounding DINO~\cite{liu2024groundingdino} & 2024 & Phrase Detection Model & Open-vocabulary detector grounded in text \\ 
& Grounded-SAM~\cite{ren2024groundedsam} & 2024 & Unified VL Segmentation & Integrates SAM with textual input \\ \cline{2-5}  

% Text-Audio
& \multicolumn{4}{c}{\textbf{(2) Text-Audio}} \\ \cline{2-5}
& wav2vec2~\cite{baevski2020wav2vec} & 2020 & SSL Speech Representation & Learns from raw waveform using contrastive loss \\ 
& HuBERT~\cite{hsu2021hubert} & 2021 & Cluster-Based SSL Speech Model & Learns representations by predicting hidden states \\ 
& Whisper~\cite{radford2023whisper} & 2023 & Multilingual ASR & Robust transcriptions across domains and languages \\ 
& AudioCLIP~\cite{guzhov2022audioclip} & 2022 & Audio-Vision-Text Embedding & Maps audio to shared semantic space \\ 
& VATT~\cite{akbari2021vatt} & 2021 & Multimodal Transformer & Unified model for video/audio/text embeddings \\ 
& AudioGen~\cite{kreuk2023audiogen} & 2023 & Text-to-Audio Generation & Synthesizes realistic soundscapes from prompts \\ \cline{2-5}  

% Text-Video
& \multicolumn{4}{c}{\textbf{(3) Text-Video}} \\ \cline{2-5}
& VideoCLIP~\cite{xu2021videoclip} & 2021 & Video-Language Alignment & Learns joint representation for clip-text retrieval \\ 
& Frozen~\cite{bain2021frozenintime} & 2021 & Pretrained VL Model & Performs video QA and captioning \\ 
& All-in-One~\cite{wang2023allinone} & 2023 & Unified Video-Language Framework & Supports multi-tasking across VL benchmarks \\ 
& Sora~\cite{videoworldsimulators2024} & 2024 & Video Simulation from Text & Generates complex dynamics in video worlds \\ 
& Make-A-Video~\cite{singer2022makeavideo} & 2022 & Video Synthesis from Language & Produces short animations from text prompts \\ 
& NUWA~\cite{wu2022nuwa} & 2022 & Visual Story Generation & Unified model for image and video generation \\ 
& Phenaki~\cite{villegas2022phenaki} & 2022 & Long Video Generation & Synthesizes consistent video from long descriptions \\ \cline{2-5} 

% Vision-Tactile-Language
& \multicolumn{4}{c}{\textbf{(4) Vision-Tactile-Language}} \\ \cline{2-5} 
& MViTac~\cite{dave2024multimodal} & 2024 & Tactile-Visual Perception & Aligns tactile sensors with RGB views \\ 
& SSVTP~\cite{kerr2022self} & 2022 & SSL for Touch & Learns touch-visual representations without labels \\ 
& UniTouch~\cite{yang2024binding} & 2024 & Unified Touch Model & Maps vision/touch to shared embedding \\ 
& TouchNeRF~\cite{zhong2023touchnerf} & 2023 & Touch-Based NeRF Rendering & Infers object surfaces from tactile inputs \\ 
& Touch and Go~\cite{yang2022touchgo} & 2022 & Joint Touch-Vision Pretraining & Self-supervised alignment of modalities \\ 
& TVL~\cite{fu2024touch} & 2024 & Multimodal Dataset & GPT-4V-annotated touch-vision-language corpus \\ 
\bottomrule[2pt]

\end{tabular}
}
\vspace{-1em}
\label{tab:perception_models}
\end{table*}

\vspace{1em}
\par \textbf{(b) Large Vision Models (LVMs)}

Large Vision Models (LVMs) underpin the visual perception capabilities of embodied agents by enabling spatial and semantic understanding of high-dimensional visual input. These models are used in tasks such as detection, segmentation, and 3D scene reconstruction, though their operations remain strictly feedforward and task-specific, without incorporating interactive or contextual reasoning.

Backbones like ResNet~\cite{he2015deepresiduallearningimage}, Vision Transformer (ViT)~\cite{dosovitskiy2021imageworth16x16words}, and Swin Transformer~\cite{liu2021swintransformerhierarchicalvision} are foundational to many vision models. ResNet introduced deep residual learning for robust convolutional feature extraction. ViT adapted transformer architectures for vision, using patch-based attention to capture global dependencies. Swin Transformer improved efficiency and scalability by introducing hierarchical and window-based attention, making it suitable for dense prediction tasks commonly found in embodied perception.

In object detection, DETR~\cite{carion2020end} reformulated detection as a set prediction problem using transformers, eliminating the need for post-processing heuristics. Deformable DETR~\cite{zhu2021deformable} improved spatial attention and convergence speed. DINO~\cite{caron2021emergingpropertiesselfsupervisedvision} and DINOv2~\cite{oquab2024dinov2learningrobustvisual}, built upon DETR-like architecture, incorporate self-supervised objectives to learn strong and transferable visual representations. DINOv2 in particular demonstrates robustness in open-world object recognition and is widely used in perception pipelines for embodied agents.

Segmentation models such as SAM~\cite{kirillov2023segment} offer zero-shot, prompt-based segmentation across diverse visual inputs. SAM2~\cite{ravi2024sam2segmentimages} extends this to the video domain with temporal memory modules, enabling consistent instance segmentation across frames. SEEM~\cite{zou2023segment} unifies multiple segmentation paradigms such as semantic, instance, and referring segmentation into a single vision model with dense supervision, enhancing the agent's ability to parse open and dynamic scenes.

For 3D object and scene perception, models like Point-BERT~\cite{yu2022point} and PointNeXt~\cite{qian2022pointnext} operate directly on point cloud data, enabling fine-grained recognition in 3D environments. BEVFormer~\cite{li2022bevformer} transforms multi-view 2D input into a bird's-eye-view representation using transformers, providing agents with spatial reasoning capabilities essential for mapping, navigation, and object localization in embodied contexts.

\par \textbf{(c) Large Multimodal Models (LMMs)}

\par \textbf{Vision-Language:} Early models like LXMERT~\cite{tan2019lxmert} and ViLBERT~\cite{lu2019vilbert} used dual-stream transformers with cross-attention to align language and region-based visual features extracted by object detectors. VL-BERT~\cite{su2019vl} improved efficiency by fusing visual and textual tokens in a single-stream transformer. UNITER~\cite{chen2020uniter} enhanced alignment with refined training objectives, while OSCAR~\cite{li2020oscar} introduced object tags as semantic anchors to guide grounding.

VL-T5~\cite{cho2021unifying} unified tasks as text generation using a sequence-to-sequence model, and ViLT~\cite{kim2021vilt} leverages the ViT architecture, omitting region-based object features and enabling more streamlined, end-to-end training. CLIP~\cite{radford2021learningtransferablevisualmodels} and ALIGN~\cite{jia2021scalingvisualvisionlanguagerepresentation} marked a major shift by using contrastive dual encoders trained on web-scale image-text pairs, enabling strong zero-shot generalization without task-specific tuning or region supervision.

Notably, the Perception Encoder (PE)~\cite{bolya2025perception} advances this contrastive approach by showing that key visual features for diverse downstream tasks often emerge in intermediate layers rather than the final output. To leverage these, it introduces alignment strategies suited for both language-driven and spatial prediction tasks. Trained with large-scale contrastive learning and curated multimodal data, PE achieves strong performance in zero-shot classification, retrieval, and dense prediction across image and video domains, highlighting the scalability of contrastive vision-language training for general-purpose perception.

\par Grounding DINO~\cite{liu2024groundingdino} and Grounded-SAM~\cite{ren2024groundedsam} extend vision-language modeling by enabling fine-grained spatial grounding of natural language. Grounding DINO formulates phrase grounding as open-set object detection, leveraging transformer-based object detectors conditioned on language prompts to localize arbitrary entities in an image. Grounded-SAM integrates this with the promptable segmentation ability of SAM, enabling agents to both detect and segment objects based on textual input. Such models support embodied agents in tasks that require language-driven perception, such as locating and manipulating objects described in natural instructions.

\par \textbf{Text-Audio:} Foundational models bridging audio and language have enabled robust auditory grounding in embodied agents. wav2vec2~\cite{baevski2020wav2vec} introduced a self-supervised framework for learning speech representations directly from raw waveforms, dramatically improving recognition with minimal supervision. HuBERT~\cite{hsu2021hubert} extended this by predicting masked audio segments based on hidden-unit targets, improving contextual understanding. Whisper~\cite{radford2023whisper}, a large-scale multitask model trained on diverse multilingual audio-text pairs, achieves strong performance in transcription, translation, and spoken language detection. AudioCLIP~\cite{guzhov2022audioclip} expands the CLIP architecture to jointly embed audio, image, and text, enabling cross-modal retrieval and zero-shot classification that links environmental sound with visual and linguistic cues. VATT~\cite{akbari2021vatt} further generalizes this by training a unified transformer across video, audio, and text modalities, supporting broad perceptual grounding. Complementing perception, AudioGen~\cite{kreuk2023audiogen} focuses on generative modeling, synthesizing plausible environmental audio from textual prompts. Collectively, these models support both the recognition and generation of auditory experiences, forming the basis for audio-language integration in multimodal embodied systems.

\par \textbf{Text-Video:} Text-video models bridge language and dynamic visual understanding, enabling embodied agents to comprehend temporally rich visual content through natural language. These models align textual descriptions with sequences of video frames, supporting tasks such as video question answering, action recognition, and instruction following. Architectures like VideoCLIP~\cite{xu2021videoclip}, Frozen~\cite{bain2021frozenintime}, and All-in-One~\cite{wang2023allinone} utilize contrastive learning or multimodal transformers to embed text and video into a shared semantic space, enabling zero-shot transfer to novel video-language tasks. Diffusion-based architectures like Sora~\cite{videoworldsimulators2024} and Make-A-Video~\cite{singer2022makeavideo} leverage transformer-enhanced latent diffusion to preserve temporal consistency and motion coherence across extended video sequences, while autoregressive frameworks such as NUWA~\cite{wu2022nuwa} and Phenaki~\cite{villegas2022phenaki} adopt hierarchical representations to handle variable-length, scene-dense perception. These text-video fusion approaches enable fine-grained alignment between language and dynamic visual scenes, making them particularly effective for understanding procedural tasks and interpreting complex, time-varying environments.

\par \textbf{Vision-Tactile-Language:} Integrating vision and tactile sensing is essential for embodied agents to interpret fine-grained physical interactions, particularly in unstructured environments. MViTac~\cite{dave2024multimodal} introduces a contrastive learning framework that fuses vision and touch using intra- and inter-modality objectives, improving material classification and grasp prediction. SSVTP~\cite{kerr2022self} presents a self-supervised method that aligns spatially paired visual and tactile data to learn rotation-invariant embeddings, enabling zero-shot performance on localization and edge-following tasks. UniTouch~\cite{yang2024binding} extends this by aligning tactile signals with vision, language, and audio embeddings using sensor-specific tokens, allowing generalization across multiple sensors and tasks. To mitigate the high cost of tactile data collection, Zhong \emph{et al.}~\cite{zhong2023touchnerf} propose a NeRF and cGAN-based framework that synthesizes tactile images from RGB-D views, supporting cross-sensor transfer. Complementing these methods, Touch and Go~\cite{yang2022touchgo} offers a diverse, human-collected dataset of egocentric vision-touch recordings for cross-modal learning, while the Touch-Vision-Language dataset~\cite{fu2024touch} introduces a large-scale collection of vision-touch pairs with both human and GPT-4V-generated language annotations, used to train a tactile encoder aligned with vision-language models for open-vocabulary classification and touch-conditioned text generation.

\subsubsection{Perception-Cognition in AI}
\label{Perception-Cognition in AI Models}

\begin{figure*}[t!]
\centering
	\includegraphics[width=0.9\textwidth]{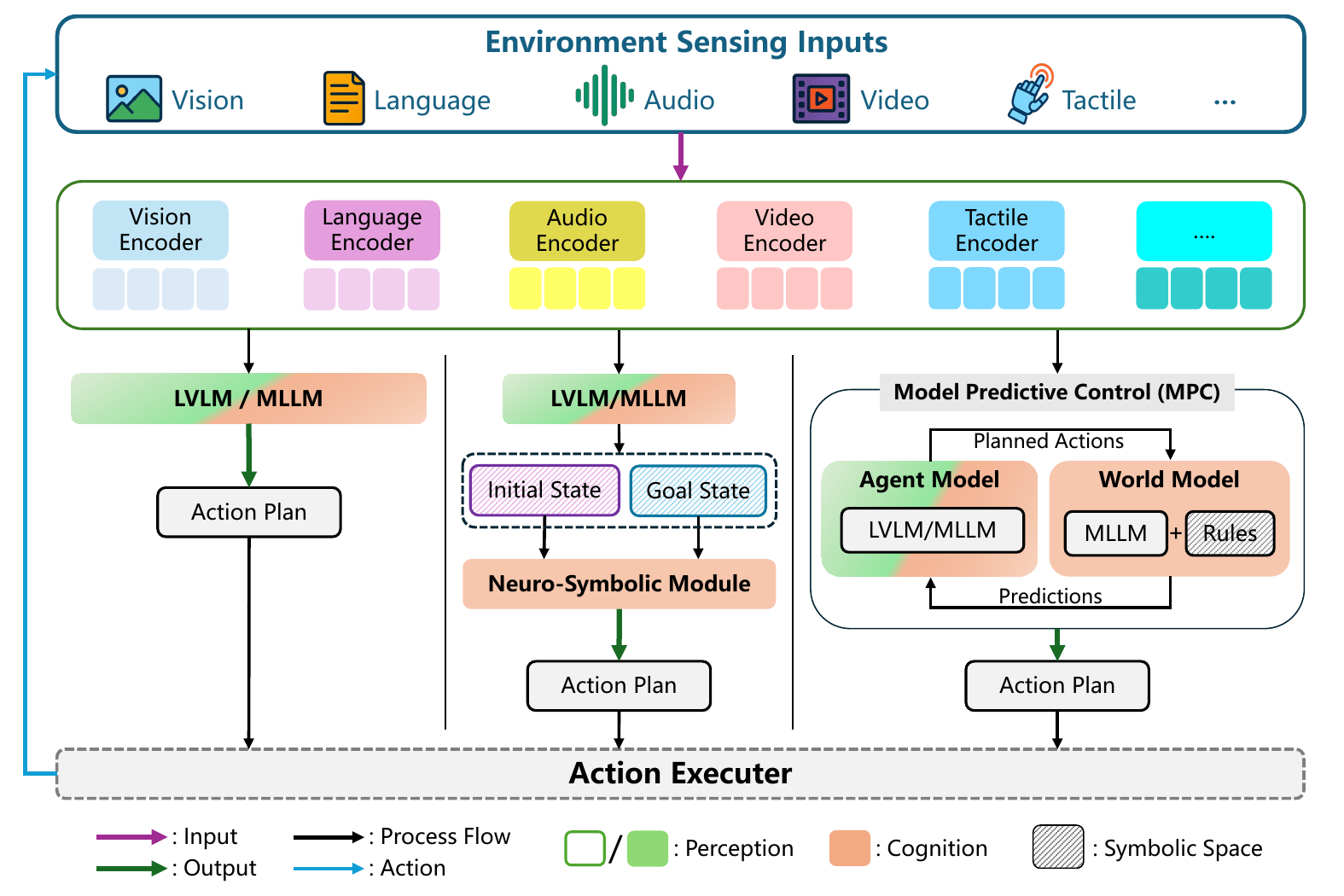}
    \vspace{-1em}
	\caption{General framework of perception-cognition in AI models with Large VLMs. Multimodal LLMs, Neuro-Symbolic Module, and World Models (adapted from~\cite{chen2024pcabenchevaluatingmultimodallarge, chia2024can, zhou2024walle}). The framework illustrates how each model contributes to the perception-cognition process, which receives either unimodal or multimodal inputs from the environment and outputs an action plan. This action plan can take the form of external actions that directly affect the environment, or internal actions, such as planning (producing a sequence of future steps) or decision-making (selecting the best option among available alternatives)~\cite{liu2025advanceschallengesfoundationagents}.}
	\label{Fig_4.1.2}
    \vspace{-1em}
\end{figure*}

This section examines AI models that integrate perception and cognition, primarily through multimodal processing, attention mechanisms, and language-guided reasoning. Attention involves selectively focusing on specific information while filtering out distractions, enabling efficient allocation of cognitive resources~\cite{zhang2019cognitivefunctionsbrainperception}. Multimodal inputs ensure system robustness, allowing functionality even with the loss of one component, as sensory systems can educate each other without external instruction. Furthermore, language plays a pivotal role in cognitive development by providing a shared symbolic system that facilitates higher-level and more abstract distinctions. Incorporating language into AI models enables them to interpret and reason about complex information, enhancing their cognitive capabilities~\cite{smith2005development}. A general framework on different approaches for perception-cognition in AI models is illustrated in Fig. \ref{Fig_4.1.2}. Some representative works are shown in Table \ref{tab:perception_cognition_models}.

\par \textbf{(a) Large Vision-Language Models (LVLMs)} 

LVLMs play a critical role in advancing embodied AI by enabling agents to perceive, interpret, and act within complex environments through the integration of visual and textual modalities. By jointly processing images and language, these models allow embodied agents to identify visual elements, understand contextual cues, and respond appropriately to linguistic instructions. This multimodal fusion supports a wide range of capabilities -- from object recognition and spatial understanding to instruction following and goal-oriented planning. Furthermore, LVLMs are pivotal in enabling cross-modal inference and real-time decision-making, empowering agents to adapt their behavior dynamically as environmental conditions change. This mirrors aspects of human perception-cognition, where sensory input is grounded in language and contextualized through reasoning to support goal-directed behavior.

Several LVLMs adopt distinct but complementary strategies to bridge visual and textual modalities. For example, BLIP~\cite{li2022blipbootstrappinglanguageimagepretraining} uses a two-way self-supervised learning approach with a guided pretraining strategy that enhances fine-grained vision-language understanding. In contrast, Flamingo~\cite{alayrac2022flamingovisuallanguagemodel} focuses on few-shot learning by leveraging interleaved sequences of images and text, enabling robust performance in multimodal tasks with minimal supervision. While BLIP emphasizes perceptual grounding and semantic detail extraction, Flamingo is akin to human rapid generalization, where individuals can understand and act on new visual-linguistic scenarios from just a few examples. VisualBERT~\cite{li2019visualbertsimpleperformantbaseline} incorporates visual features directly into the BERT architecture by embedding image region representations alongside textual tokens, supporting semantic alignment during joint encoding, which reflects how humans integrate visual scenes with language to form coherent mental representations. CoCa~\cite{yu2022cocacontrastivecaptionersimagetext}, on the other hand, merges contrastive and generative learning, allowing the model to both discriminate and generate across modalities.

Despite impressive capabilities, current LVLMs still fall short in comparison to human perception and cognition as they typically rely on training data and struggle to adapt in real-time or learn from continuous experiences. On top of that, while human cognition is highly flexible -- integrating memory with current perception to make decisions, and adjust strategies based on feedback -- LVLMs often struggle with long-term context and flexible problem-solving in unfamiliar scenarios. These gaps highlight the challenge of replicating the full spectrum of human-like adaptive learning and decision-making.

% \begin{table*}[t!]
% \caption{Representative models for perception-cognition (Sec. \ref{Perception-Cognition in AI Models}) in embodied agents.}
% \vspace{-1em}
% \renewcommand{\arraystretch}{1.25}
% \centering
% \begin{tabular}{cccc}
% \toprule[2pt]
% LVLMs & MLLMs & Neuro-Symbolic AI & World Models \\
% \hline
% BLIP~\cite{li2022blipbootstrappinglanguageimagepretraining} & GPT-4V~\cite{yang2023dawnlmmspreliminaryexplorations}, GPT-4o~\cite{gpt-4o} & Can-Do~\cite{chia2024can} & WALL-E~\cite{zhou2024walle} \\
% Flamingo~\cite{alayrac2022flamingovisuallanguagemodel} & Gemini~\cite{geminiteam2024geminifamilyhighlycapable} & NeuroGround~\cite{chia2024can} & DECKARD~\cite{nottingham2023deckard} \\
% VisualBERT~\cite{li2019visualbertsimpleperformantbaseline} & LLaVA~\cite{liu2024improvedbaselinesvisualinstruction} & JARVIS~\cite{zheng2022jarvis} & GLIMO~\cite{liu2024glimo} \\
% CoCa~\cite{yu2022cocacontrastivecaptionersimagetext} & Qwen~\cite{bai2023qwentechnicalreport}, InternVL~\cite{chen2025expandingperformanceboundariesopensource} & NESYC~\cite{choi2025nesyc}, AUKAI~\cite{wang2025understanding} & GenRL~\cite{mazzaglia2024genrl}, AeroVerse~\cite{yao2024aeroverse} \\
% \bottomrule[2pt]
% \end{tabular}
% \label{tab:perception_cognition_models}
% \vspace{-1em}
% \end{table*}

%%%%%%%%%%%%%%

\par \textbf{(b) Multimodal Large Language Models (MLLMs)}

MLLMs are extensions of LLMs that integrate and process multiple types of data, such as text, images, audio, and sometimes video. While LLMs excel in handling textual data, MLLMs extend this capability by incorporating other modalities, enabling them to perform tasks that require cross-modal reasoning. In some cases, MLLMs may overlap with VLMs, which specifically focus on combining text and visual data, but MLLMs are broader in scope, handling more diverse types of data. By enabling the processing and integration of diverse sensory inputs, MLLMs facilitate advanced perception-cognition integration. 

GPT-4~\cite{achiam2023gpt}, along with its multimodal variants, GPT-4V~\cite{yang2023dawnlmmspreliminaryexplorations} and GPT-4o~\cite{gpt-4o}, represent a significant evolution in the integration of language and visual processing. GPT-4 excels at tasks like text generation and visual question answering, while GPT-4V enhances these capabilities by incorporating visual input, making it proficient in tasks such as image captioning and complex reasoning. GPT-4o further refines this integration, improving its ability to manage dynamic environments and interact in real-time with both textual and visual information, making it particularly effective for more complex, multimodal interactions.

Gemini~\cite{geminiteam2024geminifamilyhighlycapable} combines large-scale transformers with vision models, enabling sophisticated cross-modal reasoning and the interpretation of complex visual content such as graphs and real-world scenes. LLaVA~\cite{liu2024improvedbaselinesvisualinstruction} integrates vision transformers with language models, excelling in tasks like image captioning and visual question answering by facilitating seamless communication between text and visual data. Similarly, Qwen~\cite{bai2023qwentechnicalreport} combines vision encoders with language models to handle tasks such as object detection and multimodal reasoning, proving effective in real-world applications. InternVL~\cite{chen2025expandingperformanceboundariesopensource} introduces a novel framework that focuses on optimizing cross-modal retrieval and training, further advancing the capabilities of vision-language models in handling diverse tasks with greater efficiency.

Multiple studies have assessed and benchmarked MLLMs for embodied agents from different perspectives~\cite{yang2025embodiedbenchcomprehensivebenchmarkingmultimodal, chen2024pcabenchevaluatingmultimodallarge, cheng2025embodiedevalevaluatemultimodalllms, ma2025deepperceptionadvancingr1likecognitive, fu2024mmesurveycomprehensivesurveyevaluation}. In vision-driven embodied tasks, EmbodiedBench~\cite{yang2025embodiedbenchcomprehensivebenchmarkingmultimodal} finds that current MLLMs perform well on high-level tasks but struggle with low-level tasks, particularly manipulation. It also shows that low-level tasks rely much more heavily on vision compared to high-level tasks. PCA-Bench~\cite{chen2024pcabenchevaluatingmultimodallarge} evaluates MLLMs within the perception-cognition-action chain, revealing that GPT-4V achieves the highest zero-shot scores across all three stages -- perception, cognition, and action -- outperforming current open-source models. On top of that, the proposed Embodied Instruction Evolution (EIE) method significantly boosts the general decision-making abilities of MLLMs by improving their integrated performance across these dimensions. EmbodiedEval~\cite{cheng2025embodiedevalevaluatemultimodalllms} examines MLLMs as embodied agents across a range of tasks, including Attribute Question Answering (AttrQA), Spatial Question Answering (SpatialQA), Navigation, Object Interaction, and Social Interaction. Results show that MLLMs still perform poorly on embodied tasks and exhibit inconsistent performance across different tasks. Additionally, MLLMs struggle with long-horizon tasks, as their performance declines notably with an increasing number of required steps and subgoals.

Although MLLMs have shown promising results for embodied agents, several limitations remain that need to be addressed, including hallucination in grounding, lack of spatial reasoning, wrong planning~\cite{cheng2025embodiedevalevaluatemultimodalllms}, or incorporating more modalities~\cite{fu2024mmesurveycomprehensivesurveyevaluation}.

\begin{table*}[t!]
\caption{Representative models for perception-cognition (Sec. \ref{Perception-Cognition in AI Models}) in embodied agents.}
\vspace{-1em}
\renewcommand{\arraystretch}{1.6}
\centering
\resizebox{1\linewidth}{!}{
\begin{tabular}{ccccc}
\toprule[2pt]
\multicolumn{2}{c}{\textbf{Category/Method}} & \textbf{Year} & \textbf{Details} & \textbf{Highlights} \\
\midrule

\multirow{4}{*}{\rotatebox{90}{\scriptsize \shortstack{\textbf{(a) Large VLMs}}}}
& BLIP~\cite{li2022blipbootstrappinglanguageimagepretraining} & 2022 & Vision-Language Pre-training & Bootstraps captions for unified vision-language tasks \\ 
& Flamingo~\cite{alayrac2022flamingovisuallanguagemodel} & 2022 & Few-Shot Visual Language Model & Enables few-shot learning across vision-language tasks \\ 
& VisualBERT~\cite{li2019visualbertsimpleperformantbaseline} & 2019 & Vision-Language Model & Integrates visual and textual information for joint understanding \\ 
& CoCa~\cite{yu2022cocacontrastivecaptionersimagetext} & 2022 & Contrastive Captioner & Combines contrastive learning with image-text captioning \\ \midrule

\multirow{6}{*}{\rotatebox{90}{\scriptsize \shortstack{\textbf{(b) Multimodal LLMs}}}}
& GPT-4V~\cite{yang2023dawnlmmspreliminaryexplorations} & 2023 & Multimodal Language Model & Extends GPT-4 with vision capabilities for image understanding \\ 
& GPT-4o~\cite{gpt-4o} & 2024 & Omni Multimodal Model & Processes text, audio, and images with unified architecture \\ 
& Gemini~\cite{geminiteam2024geminifamilyhighlycapable} & 2024 & Multimodal AI Model & Integrates multiple modalities for advanced reasoning \\ 
& LLaVA~\cite{liu2024improvedbaselinesvisualinstruction} & 2024 & Visual Instruction Tuning & Aligns vision-language models with human instructions \\ 
& Qwen~\cite{bai2023qwentechnicalreport} & 2023 & Language Model Series & Offers versatile models for various language tasks \\ 
& InternVL~\cite{chen2025expandingperformanceboundariesopensource} & 2024 & Vision-Language Foundation Model & Scales vision models and aligns them with LLMs for multimodal tasks \\ \midrule

\multirow{5}{*}{\rotatebox{90}{\scriptsize \shortstack{\textbf{(c) Neuro-Symbolic AI}}}}
& Can-Do~\cite{chia2024can} & 2024 & Neuro-Symbolic Framework & Benchmarks embodied planning with neuro-symbolic methods \\ 
& NeuroGround~\cite{chia2024can} & 2024 & Grounded Planning Framework & Grounds plans in perceived environment states \\ 
& JARVIS~\cite{zheng2022jarvis} & 2022 & Neuro-Symbolic Agent & Integrates neural and symbolic reasoning for embodied tasks \\ 
& NESYC~\cite{choi2025nesyc} & 2025 & Neuro-Symbolic Learning & Combines neural networks with symbolic reasoning for learning \\ 
& AUKAI~\cite{wang2025understanding} & 2025 & Unified Knowledge-Action Intelligence & Integrates perception, memory, and decision-making in embodied cognition \\ \midrule

\multirow{5}{*}{\rotatebox{90}{\scriptsize \shortstack{\textbf{(d) World Models}}}}
& WALL-E~\cite{zhou2024walle} & 2024 & Embodied World Model & Constructs world models for embodied agents \\ 
& DECKARD~\cite{nottingham2023deckard} & 2023 & LLM-Guided Agent & Utilizes LLMs for abstract world model generation \\ 
& GLIMO~\cite{liu2024glimo} & 2024 & Generative World Model & Learns generative models for planning and control \\ 
& GenRL~\cite{mazzaglia2024genrl} & 2024 & Generalist Reinforcement Learning & Applies generalist models to reinforcement learning tasks \\ 
& AeroVerse~\cite{yao2024aeroverse} & 2024 & UAV-Agent Benchmark Suite & Provides datasets and benchmarks for aerospace embodied world models \\
\bottomrule[2pt]
\end{tabular}
}
\vspace{-1em}
\label{tab:perception_cognition_models}
\end{table*}

\par \textbf{(c) Neuro-Symbolic AI}

Neuro-symbolic AI integrates neural networks with symbolic reasoning to bridge low-level perception and high-level cognition. In embodied AI, this synergy enables agents to process raw sensory inputs such as vision, audio, and language; while reasoning about their environment through structured, interpretable representations. Neural components excel at learning from high-dimensional data and recognizing patterns, while symbolic systems provide the ability to generalize, reason logically, and plan over time. This fusion enhances the agent's ability to perceive, understand, and interact with complex, dynamic environments in a more robust and explainable manner.

NeuroGround~\cite{chia2024can} shows that advanced models like GPT-4V face limitations in visual perception, understanding, and reasoning capabilities. To address this, it proposes a neuro-symbolic framework that grounds model-generated plans in environmental states and enriches them using symbolic engines, ensuring contextual alignment and improved planning accuracy. JARVIS~\cite{zheng2022jarvis}, tailored for conversational embodied agents, combines LLMs for language understanding with symbolic commonsense reasoning for planning and execution, demonstrating strong generalization and few-shot performance in real-world tasks. PrimeNet~\cite{primenet}, a framework for commonsense knowledge representation and reasoning based on conceptual primitives, uses LLMs to build hierarchical concept representations which have significant applications in diverse domains, e.g., conceptual metaphor understanding and cognitive analysis.
NESYC~\cite{choi2025nesyc} focuses on continual learning by leveraging LLMs and symbolic tools to generalize actionable knowledge across domains, introducing contrastive generality improvement and memory-based monitoring to iteratively refine knowledge and detect action errors. AUKAI~\cite{wang2025understanding} presents a unified multi-scale feedback loop, where neural modules extract perceptual features and symbolic modules guide reasoning and decision-making, enhancing both interpretability and adaptability.

\par \textbf{(d) World Models}

\par A world model refers to a model capable of predicting and simulating dynamic environmental changes, enabling agents to perform counterfactual reasoning---the ability to infer outcomes of previously unencountered scenarios~\cite{mai2024efficient}. Within embodied agents, the notion of world models has evolved into a central paradigm for integrating perception, memory, and planning---core components of brain-inspired cognition. Modern approaches increasingly view world models not merely as predictive simulators, but as internalized, multimodal constructs that support abstraction, imagination, and counterfactual reasoning. By internally aligning linguistic, perceptual, and sensorimotor information, these models act as cognitive scaffolds that facilitate flexible and adaptive behavior in novel contexts.

\par Liu \emph{et al.}~\cite{liu2024surveywm} identify world models and large multimodal models as foundational elements in achieving generalist embodied intelligence. Their analysis outlines core architectural layers---perception, interaction, control, and adaptation---and positions world models as key mechanisms for temporal and cross-modal reasoning. They also highlight persistent integration gaps between symbolic reasoning and physical embodiment, motivating hybrid approaches grounded in both neural representations and structured abstraction. A broader perspective is offered in Zhu \emph{et al.}~\cite{zhu2024sora}, which surveys general-purpose world models across domains such as video generation and autonomous driving, framing them as early foundations for open-ended, physically grounded reasoning. 

There has been approaches advance these ideas by embedding LLMs within world modeling pipelines. WALL-E~\cite{zhou2024walle} enhances LLM-based reasoning with a compact set of learned symbolic rules, improving model alignment with embodied dynamics and enabling sample-efficient planning. In a neuro-symbolic vein, the DECKARD agent~\cite{nottingham2023deckard} constructs abstract world models through LLM-driven subgoal decomposition, iteratively refined by embodied feedback---a structure reminiscent of cognitive predictive loops.

Other approaches address the challenge of grounding through simulated or proxy environments. GLIMO~\cite{liu2024glimo} uses LLM-guided synthetic data from simulators to support temporally coherent planning, while GenRL~\cite{mazzaglia2024genrl} fuses foundation vision-language models with generative dynamics for multi-task generalization. Meanwhile, AeroVerse~\cite{yao2024aeroverse} broadens the scope of world modeling to aerial robotics with a benchmark suite for UAV agents in spatial reasoning and autonomous planning.

\subsubsection{Perception-Action in AI}
\label{Perception-Action in AI Models}

\par
\begin{figure*}[t!]
\centering
	\includegraphics[width=1.\textwidth]{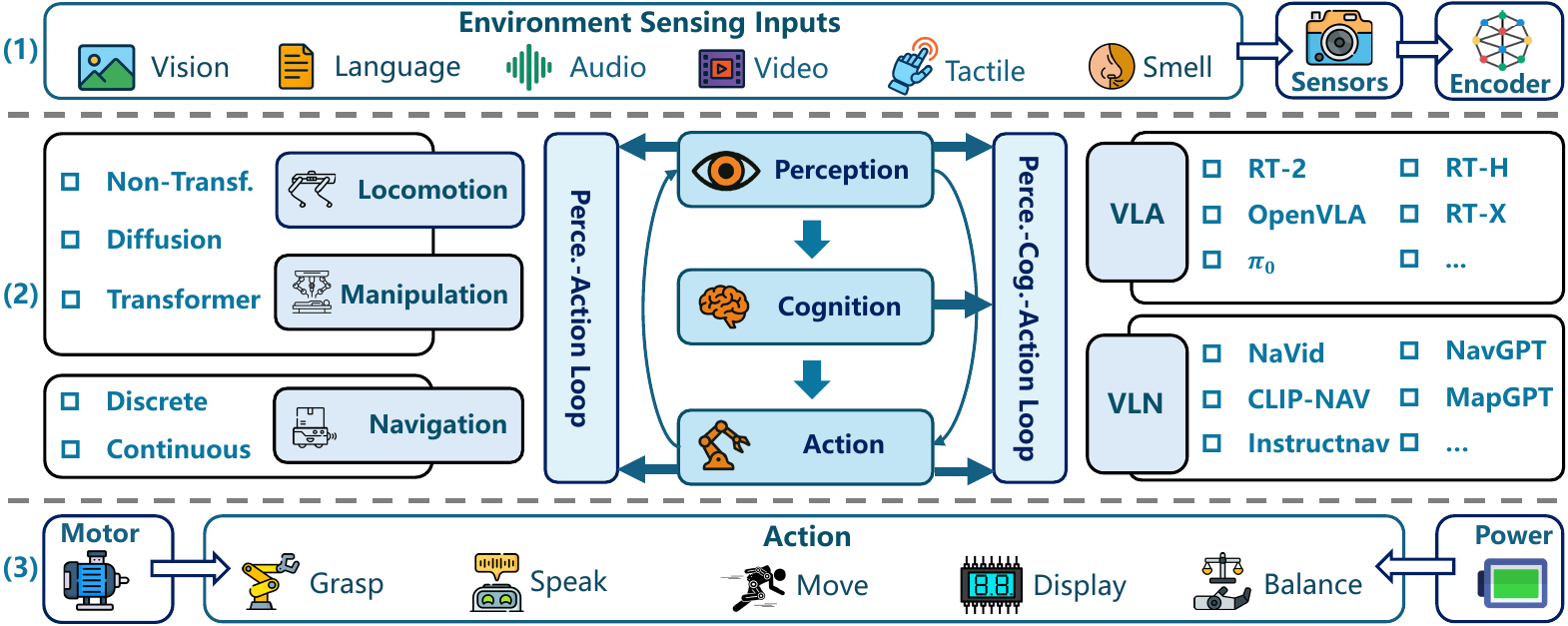}
  \vspace{-2em}
	\caption{The diagram outlines the action mechanisms of embodied AI through two loops: the Perception-Action loop (left)~\cite{zeng2021transporter, shridhar2022cliport, jang2022bc, zhang2024vision, long2024instructnav} and the Perception-Cognition-Action loop (right)~\cite{belkhale2024rt, vuong2023open, kim2024openvla, zhang2024vision, an2023bevbert, zhou2024navgpt}. In both loops, agents perceive the environment using sensors and encoders (Part 1). The output of the model is then used to control the motor and power systems, thereby dictating the agent's actions (Part 3). Both loops contain the VLA or VLN models, which differ in their use of primary cognitive capabilities (\textit{e.g.}, incorporating LLM~\cite{zhou2024navgpt, an2023bevbert} or VLM~\cite{belkhale2024rt, kim2024openvla}) (Part 2). The VLA model is designed for tasks such as locomotion and manipulation~\cite{zeng2021transporter, shridhar2022cliport, zhang2024vision}, while the VLN model is tailored for robot navigation tasks~\cite{zhou2024navgpt, dou2022foam}.}
	\label{Fig_4.1.4}
  \vspace{-1em}
\end{figure*}

 While perception and cognition lay the foundation for an agent's understanding of the world, it is action that serves as the crucial medium for interaction with the physical environment. The Perception-Action Integration~\cite{ridderinkhof2014neurocognitive, fuster2004upper, cassimatis2004integrating, liutwo} describes a continuous loop where the AI model makes decisions and takes actions directly after receiving sensory information without complex thinking. Each decision and action subsequently alters the environment and prompts new perceptions, thereby initiating the next cycle. It focuses more on instant response rather than deliberate reasoning or extensive analysis. For humans, Perception-Action Integration is reflected in many places, \textit{e.g.}, in the reflexive withdrawal of a hand from a hot surface~\cite{andrew2001conditioned}, the instinctive grasping of an object perceived within reach, or the immediate reaction to a sudden loud noise. This section will focus on the AI models that integrate perception with action, \textit{i.e.}, robotics, embodied AI, and real-world interaction, including locomotion, manipulation, and navigation as shown in Fig.~\ref{Fig_4.1.4}. Some representative methods are summarized in Table~\ref{tab:perception_cognition_action_models}.

\par \textbf{(a) Agent Locomotion}
\par Agent Locomotion focuses on the motion control of agents, including walking~\cite{huang2001planning, biswal2021development}, balance control~\cite{juang2013design, lin2008development}, etc. Among them, walking algorithms for robots are one of the most basic and vital parts. Reinforcement learning~\cite{kaelbling1996reinforcement, kober2013reinforcement,sheins} has emerged as a powerful tool for developing walking algorithms for both bipedal and quadrupedal robots. It enables robots to learn effective locomotion strategies by interacting with their environment and receiving feedback through rewards or penalties. Compared to traditional methods, it achieves greater adaptability to diverse terrains and the ability to learn complex behaviors without extensive prior knowledge. Keeping balance is another key problem for agent locomotion. The traditional solutions initially maintained balance by slowly adjusting their movements to maintain a stable center of gravity~\cite{vukobratovic2004zero}. Still, this type of method is not only slow but also unstable and cannot cope with complex environments. To overcome these limitations, recent research has introduced advanced techniques such as Model Predictive Control (MPC)~\cite{schwenzer2021review} and Reinforcement Learning to achieve more efficient dynamic balance control. MPC utilizes predictive models to optimize control strategies by forecasting future system states. It has been successfully applied to humanoid and quadruped robots for gait generation and optimization. Reinforcement Learning has also demonstrated substantial potential in robotic balance control. By interacting with the environment, it enables robots to learn complex dynamic behaviors and adapt to varying environmental conditions. Moreover, the integration of visual perception into locomotion control has advanced significantly. Vision-based locomotion allows robots to anticipate terrain changes and adjust their gait accordingly, achieving more proactive and robust movement~\cite{yang2023neural}. Combining visual and proprioceptive feedback fosters a more complete perception-action loop, enabling agents to adapt to complex and dynamic real-world environments in real time. As locomotion remains foundational to embodied AI, future research continues to explore how richer sensory input, more efficient learning algorithms, and better simulation-to-reality transfer can push the boundaries of autonomous mobility.

\begin{table*}[t!]
\caption{Representative models for perception-action (Sec.~\ref{Perception-Action in AI Models}) and perception-action (Sec.~\ref{Perception-Cognition-Action in AI Models}) in embodied agents.}
\vspace{-1em}
\renewcommand{\arraystretch}{1.6}
\centering
\resizebox{1\linewidth}{!}{
\begin{tabular}{cccccc}
\toprule[2pt]
\multicolumn{3}{c}{\textbf{Category/Method}} & \textbf{Year} & \textbf{Details} & \textbf{Highlights} \\
\midrule

\multirow{23}{*}{\rotatebox{90}{\scriptsize \textbf{Perception-Action in AI}}}
& \multirow{13}{*}{\rotatebox{90}{\scriptsize (a) VLA Models}} 
& MCIL~\cite{lynch2020language} & 2020 & Improves imitation learning & Allows free-form natural language instructions \\

&& Transporter Network~\cite{zeng2021transporter} & 2021 & Visual rearrangement model & Infers spatial displacements from visual inputs \\

&& CLIPort~\cite{shridhar2022cliport} & 2022 & two-stream architecture for manipulation & integrates CLIP with the precision of Transporter Networks \\

&& BC-Z~\cite{jang2022bc} & 2022 & Imitation learning system & Allows generalization by zero-shot or few-shot learning. \\

&& RT-1~\cite{brohan2022rt} & 2022 & A unified framework & Integrates perception and actions into an unified framework \\

&& Q-Transformer~\cite{chebotar2023q} & 2023 & Scalable offline RL & Auto-regressive Q-learning with Transformers\\

&& ACT~\cite{zhao2023learning} & 2023 & Low-cost bimanual control & Low-cost hardware, fine manipulation \\

&& RVT~\cite{goyal2023rvt, goyal2024rvt} & 2023 & Multi-view 3D manipulation & Scalable, accurate, virtual view rendering \\

&& Diffusion Policy~\cite{chi2023diffusion} & 2023 & Diffusion-based control & Action diffusion, multi-modal, high-dimensional \\

&& MBA~\cite{su2024motion} & 2024 & Object motion guided control & Predicts object motion, enhances robotic manipulation \\

&& RDT-1B~\cite{liu2024rdt} & 2024 & Diffusion-based bimanual control & 1.2B parameters, multi-robot pretraining \\

&& DP3~\cite{ze20243d} & 2024 & 3D visual imitation learning & Generalizable, efficient, few demonstrations \\

&& ReKep~\cite{huang2024rekep} & 2024 & Keypoint-based robotic control & Semantic keypoints, spatio-temporal reasoning \\

&& PIVOT~\cite{nasiriany2024pivot} & 2024 & Iterative visual optimization & Zero-shot control, visual refinement \\

\cline{3-6}

&\multirow{10}{*}{\rotatebox{90}{\scriptsize (a) VLN Models}}& VLN-Graph~\cite{anderson2018vision} & 2018 & 3D semantic navigation & 3D semantic maps, instruction-aware paths \\

&& Speaker-Follower~\cite{fried2018speaker} & 2018 & Speaker-follower navigation & Data augmentation, pragmatic reasoning \\

&& Sim-to-Real~\cite{anderson2021sim} & 2021 & Sim-to-real VLN transfer & Subgoal modeling, domain randomization \\

&& WPN~\cite{krantz2021waypoint} & 2021 & Instruction-guided waypoints & Hierarchical planning, continuous navigation \\

&& HAMT~\cite{chen2021history} & 2021 & History-aware navigation & Incorporates long-term history \\

&& Waypoints Predictor~\cite{hong2022bridging} & 2022 & Discrete-Continuous VLN & Predict waypoints, discrete-to-continuous \\

&& Talk2nav~\cite{vasudevan2021talk2nav} & 2022 & Dual attention navigation & Dual attention, spatial memory \\

&& CM2~\cite{georgakis2022cross}& 2022 & Cross-modal semantic mapping & Semantic maps, cross-modal attention \\

&& Bevbert~\cite{an2023bevbert} & 2023 & Topo-metric map pretraining & Hybrid maps, spatial reasoning \\

\midrule

\multirow{11}{*}{\rotatebox{90}{\scriptsize \textbf{Perception-Cognition-Action in AI}}}
& \multirow{6}{*}{\rotatebox{90}{\scriptsize (a) VLA Models}} 
& RT-2~\cite{brohan2023rt} & 2023 & Web-scale Visual Language Model & Web-scale VLMs boost robot generalization \\
&& OpenVLA~\cite{kim2024openvla} & 2024 & Open-source VLA model & Outperforms RT-2-X with 7$\times$ fewer parameters \\
&& RT-X~\cite{vuong2023open} & 2024 & Open X-Embodiment datasets and RT-X models & Global collaboration, diverse robot skills \\
&& RT-H~\cite{belkhale2024rt} & 2024 & Action hierarchies via language & Uses language motions for hierarchical robot control \\
&& $\pi_0$~\cite{black2024pi_0} & 2024 & VLA flow model & Uses flow matching for dexterous robot control \\
&& $\pi_{0.5}$ & 2025 & Co-trained VLA model & Generalizes to new homes via diverse data co-training \\

\cline{3-6}

& \multirow{5}{*}{\rotatebox{90}{\scriptsize (b) VLN Models}} 
& CLIP-NAV~\cite{dorbala2022clip} & 2022 & Zero-shot VLN using CLIP & Zero-shot navigation, CLIP grounding, instruction breakdown \\
&& NaVid~\cite{zhang2024navid} & 2024 & Video-based VLN planning & Map-free, real-time, human-like navigation \\
&& NavGPT~\cite{zhou2024navgpt,zhou2024navgpt2} & 2024 & LLM-powered navigation reasoning & Zero-shot planning, vision-language alignment \\
&& Instructnav~\cite{long2024instructnav} & 2024 & Zero-shot instruction navigation & Unified planning, no prior training needed \\
&& MapGPT~\cite{chen2024mapgpt} & 2024 & Map-guided GPT navigation & Global planning, zero-shot, instruction-adaptive \\

\bottomrule[2pt]
\end{tabular}
}
\vspace{-1em}
\label{tab:perception_cognition_action_models}
\end{table*}

\par \textbf{(b) Agent Manipulation}
\par Agent Manipulation~\cite{ehsani2021manipulathor} focuses on the interaction between the agent and objects in the environment, which requires skills such as precise force control, object perception, and grasping strategies~\cite{saxena2008robotic}, involving complex mechanical models and precise control. In recent years, AI models have been widely used in Agent Manipulation to enhance the agent's ability to interact with objects effectively. For example, Vision-Language-Action (VLA) Models, which are currently a highly popular and effective AI technology, can identify and locate objects in the environment while receiving language commands, helping the agent understand the location, shape, and state of objects. The recent work by Ma \emph{et al.}~\cite{ma2024survey} defines a VLA model as any system capable of processing multimodal inputs such as images, language, and LiDAR to generate robot actions and accomplish a wide range of robotic tasks. A typical VLA model generally comprises encoders for perceiving the current state, such as visual, auditory, or LiDAR inputs using architectures like CNNs, Transformers, or VLMs; a language encoder for processing task instructions; and a decoder for generating action commands. 

Transformer-based policies have become prevalent in many current VLA models. However, prior to their emergence, numerous alternative approaches were developed to implement VLA functionalities~\cite{zeng2021transporter, shridhar2022cliport, jang2022bc, lynch2020language, mees2022matters, mees2023grounding, du2023learning}. For instance, Transporter Network~\cite{zeng2021transporter} introduced a model architecture that rearranges deep features to infer spatial displacements from visual input for robotic manipulation. This model learns to focus on a local region and predict its target spatial displacement through deep feature template matching, thereby parameterizing robot actions. Building upon this, CLIPort~\cite{shridhar2022cliport} proposed a two-stream architecture for robotic manipulation that integrates the semantic understanding of CLIP with the spatial precision of Transporter Networks. It employs a semantic stream conditioned on language features from CLIP~\cite{guzhov2022audioclip} and a spatial stream for precise manipulation, enabling the agent to perform a variety of language-specified tasks. BC-Z~\cite{jang2022bc} presented an interactive imitation learning system that allows a robot to generalize to novel tasks with zero-shot or few-shot learning. This system combines large-scale data collection with shared autonomy and flexible task conditioning using language or human videos, enabling the robot to perform unseen manipulation tasks and demonstrating significant generalization capabilities. Furthermore, MCIL~\cite{lynch2020language} enables robots to follow free-form natural language instructions by incorporating unstructured and unlabeled demonstration data into imitation learning. This approach addresses the challenges of high annotation costs and limited scalability in conventional imitation learning by reducing the need for labeled data. To address the challenges of environment diversity and reward specification in constructing general-purpose agents, UniPi~\cite{du2023learning} proposed a text-conditioned video generation approach, formulating sequential decision making as a planning problem in the video space. Despite offering innovative insights into embodied intelligence, non-transformer-based approaches exhibit certain limitations, \textit{e.g.}, 
limited generalization capabilities when confronted with novel tasks and facing challenges in seamlessly integrating semantic understanding with spatial precision. 

In light of the limitations inherent in non-transformer-based approaches, transformer-based models have emerged as a promising alternative in VLA research~\cite{li2023vision, zhao2023learning, reed2022generalist, gu2023rt, chebotar2023q, lynch2023interactive, bharadhwaj2024roboagent, guhur2023instruction, brohan2022rt}. A classic example is RT-1~\cite{brohan2022rt}, which integrates visual inputs, natural language instructions, and robotic actions into a unified framework. Unlike traditional methods that often rely on manually designed features or single-task learning, RT-1 leverages end-to-end learning and multimodal fusion to achieve strong generalization and robustness. It tokenizes high-dimensional inputs and outputs, enabling efficient real-time control and the ability to absorb heterogeneous data from different sources. Building on this foundation, Q-Transformer~\cite{chebotar2023q} further enhances the capability by incorporating offline reinforcement learning, allowing the model to improve upon demonstrated behaviors through temporal difference learning and effectively utilizing both high-quality demonstrations and sub-optimal data for more robust policy learning. Similarly, ACT~\cite{zhao2023learning} leverages action chunking and temporal ensembling to mitigate compounding errors in imitation learning, enabling the robot to learn precise, closed-loop behaviors from limited human demonstrations. While these advancements have significantly improved the capabilities of robotic learning systems, they primarily operate in 2D spaces, which are inherently limited in capturing the full complexity of real-world environments. Transitioning from 2D to 3D~\cite{shridhar2023perceiver, gervet2023act3d, liu2024volumetric, goyal2023rvt, goyal2024rvt}, PERACT~\cite{shridhar2023perceiver} addresses these limitations by utilizing a voxelized 3D observation and action space. This 3D formulation provides a strong structural prior for efficient 6-DoF behavior cloning with Transformers, allowing for better integration of multi-view observations, learning robust action-centric representations, and enabling data augmentation in 6-DoF. RVT-1~\cite{goyal2023rvt} and RVT-2~\cite{goyal2024rvt} also leverage the power of Transformers to enhance robotic manipulation tasks in 3D space. RVT-1 introduces a novel multi-view representation for encoding the scene, which significantly improves training speed, inference speed, and task performance compared to voxel-based methods like PerAct. RVT-2 further builds on RVT-1 by incorporating architectural improvements such as a multi-stage inference pipeline for more precise end-effector pose prediction and system-level optimizations for faster training and inference. In addition, several approaches have incorporated point-based methods to represent the physical world better, \textit{e.g.}, ReKep~\cite{huang2024rekep}, RoboPoint~\cite{yuan2024robopoint}, and PIVOT~\cite{nasiriany2024pivot}.

Since the remarkable capabilities of diffusion models in modeling complex multimodal distributions and generating high-dimensional sequences with high precision, diffusion-based VLA has emerged to leverage these strengths for robust and versatile robotic control~\cite{chi2023diffusion, reuss2024multimodal, liu2024rdt, team2024octo, ha2023scaling, su2024motion}. Diffusion Policy~\cite{chi2023diffusion}, a pioneering approach in this domain, represents robot actions as a conditional denoising diffusion process, enabling the policy to iteratively refine noisy action samples into precise and contextually appropriate actions through learned gradient fields. This formulation not only allows for stable training and efficient inference but also significantly enhances the policy's ability to handle complex real-world robotic manipulation tasks, as evidenced by its superior performance across various benchmarks and real-world evaluations. MBA~\cite{su2024motion}, a novel module for robotic manipulation, employs two cascaded diffusion processes to first predict future object pose motion sequences~\cite{liu2024surveypose} and then generate robot actions conditioned on these predicted motions. This predictive approach enables the robot to anticipate object dynamics and plan actions accordingly, significantly enhancing the robustness and kinematic consistency of the policy's observation-to-action mapping. RDT-1B~\cite{liu2024rdt} extends the diffusion-based framework by introducing a unified action space that enables cross-robot pretraining, significantly enhancing generalization. It also incorporates a scalable Transformer backbone with specialized modifications to handle the unique challenges of robotic data, such as nonlinearity and high-frequency changes. These advancements allow RDT-1B to achieve state-of-the-art performance in complex bimanual manipulation tasks. Like the evolution of transformer-based approaches, diffusion-based policies have also been extended from 2D to 3D, exemplified by 3D Diffuser Actor~\cite{ke20243d} and DP3~\cite{ze20243d}, to better accommodate the complexities of real-world environments. 

ACT~\cite{zhao2023learning} leverages action chunking and temporal ensembling to mitigate compounding errors in imitation learning, enabling the robot to learn precise, closed-loop behaviors from limited human demonstrations.

\par \textbf{(c) Agent Navigation}

\par Agent Navigation focuses on the autonomous movement of an agent within an environment to reach a specified destination. This process involves a combination of perception, planning, obstacle avoidance, and motion control, making it a crucial area in robotics and artificial intelligence. Simultaneous Localization and Mapping (SLAM)~\cite{durrant2006simultaneous, bailey2006simultaneous} is vital for both indoor and outdoor robotic positioning in traditional methods. It allows robots to construct a map of an unknown environment while simultaneously keeping track of their location within it. SLAM algorithms integrate data from sensors like LIDAR and cameras to create detailed maps and navigate efficiently. Traditional SLAM methods, such as EKF-SLAM~\cite{bailey2006consistency} and Graph SLAM~\cite{grisetti2010tutorial}, rely on probabilistic techniques and hand-engineered features, offering robust performance in structured environments. However, they can be computationally intensive and require significant tuning. In contrast, SLAM based on AI models~\cite{macario2022comprehensive} leverages deep learning models to automatically extract features and optimize navigation strategies, enhancing robustness and efficiency in complex and dynamic environments. Vision-Language Navigation (VLN)~\cite{zhang2024vision, long2024instructnav, an2023bevbert, vasudevan2021talk2nav, li2021improving, xia2020multi, zhou2024navgpt, dou2022foam, lin2021scene} is another key approach in AI navigation: it is a specialized subset of VLA models, focusing specifically on navigation tasks. In VLN, an embodied agent is guided by natural language instructions to traverse complex, often unfamiliar, environments, necessitating the integration of visual perception, linguistic comprehension, and sequential decision-making. Since its inception with the Room-to-Room (R2R) dataset~\cite{anderson2018vision}, VLN research has evolved from employing sequence-to-sequence models with attention mechanisms to leveraging large-scale pre-trained vision-language models, enhancing the agent's ability to generalize across diverse scenarios.

Early research in VLN~\cite{anderson2018vision,gu2022vision, chaplot2020object, krantz2020beyond, ku2020room, fried2018speaker,anderson2019chasing, hong2021vln, hong2022bridging} primarily focused on integrating visual information with language instructions to accomplish basic navigation tasks. These studies often relied on symbolic representations of the environment, such as maps or path graphs, to align instructions with predefined routes. For instance, VLN-Graph~\cite{anderson2018vision} proposed a navigation model based on path graphs, aligning language instructions with nodes on the graph to facilitate navigation. While straightforward, this approach faced limitations in real-world applications, including oversimplified environmental representations and restricted comprehension of complex instructions. Building on this, Speaker-Follower~\cite{fried2018speaker} introduced a speaker-follower framework, where the speaker generates navigation instructions and the follower executes these instructions by aligning them with visual observations in a discrete environment.

Toward more realistic and intuitive modeling, navigation in continuous environments, such as the Room-to-Room in Continuous Environments (R2R-CE) and Room-Across-Room in Continuous Environments (RxR-CE) benchmarks, has received growing attention. These tasks aim to narrow the gap between traditional discrete navigation settings and real-world scenarios by enabling agents to operate freely within continuous 3D spaces. For instance, Anderson \emph{et al.}~\cite{anderson2021sim} introduce a Sim-to-Real transfer framework for vision-and-language navigation, which translates high-level actions learned in simulation into navigable waypoints for physical robots. By employing domain randomization to address visual discrepancies between simulated and real-world environments, their method achieves a 46.8\% success rate in an office environment when provided with an occupancy map and navigation graph. Similarly, Waypoint Prediction Network (WPN)~\cite{krantz2021waypoint} utilizes natural language instructions and panoramic visual input to predict navigational waypoints, incorporating cross-modal attention to align visual and textual modalities. Trained via reinforcement learning, WPN improves both navigation success and execution efficiency. More recently, Hong \emph{et al.}~\cite{hong2022bridging} propose a Candidate Waypoints Predictor to explicitly bridge the gap between discrete and continuous VLN tasks. Their approach employs a lightweight Transformer to infer candidate waypoints from RGB-D inputs, allowing agents to reason over navigable spaces in realistic settings and enhancing cross-environment generalization as well as instruction grounding.

\subsubsection{Perception-Cognition-Action in AI}
\label{Perception-Cognition-Action in AI Models}

While perception allows an agent to see and hear, and action enables it to move and interact, it is cognition that truly differentiates between a simple reactive system and a truly intelligent one. Cognition is the biggest gap between humans and current agents. However, recent advancements in LLMs~\cite{brown2020languagemodelsfewshotlearners, achiam2023gpt, guo2025deepseek} and VLMs~\cite{CLIPose, zhou2022learning} have shown promising potential to bridge this gap. LLMs excel in processing and generating natural language, enabling agents to understand and generate human-like text, while VLMs combine visual perception with language understanding, allowing agents to interpret images and videos in addition to text. Nevertheless, these advancements are essentially a combination of Perception and Cognition, and they mostly operate in a simulated or controlled environment, rather than interacting with the real physical world. Some representative works are shown in Table~\ref{tab:perception_cognition_action_models}.

\par \textbf{(a) Cognition in VLA Models}

To take advantage of the strong perception capabilities of LLM and VLM, some new VLA models have begun to use pre-trained LLM and VLM models~\cite{belkhale2024rt, vuong2023open, kim2024openvla, brohan2023rt, black2024pi_0} to process visual and language information and generate corresponding actions. For example, RT-2~\cite{brohan2023rt} leverages large pre-trained VLMs such as PaLM-E~\cite{driess2023palm} and PaLI-X~\cite{chen2023pali}. It represents robot actions as text tokens and incorporates these tokens directly into the training set of the VLM, allowing the model to be fine-tuned on both web-scale vision-language data and robotic trajectory data. This approach enables RT-2 to exhibit strong generalization capabilities and emergent reasoning abilities in robotic control tasks. OpenVLA~\cite{kim2024openvla}, is constructed upon the Prismatic-7B VLM framework. Employing a dual-visual encoder composed of pre-trained DINOv2~\cite{DINOv2} and SigLIP, it effectively handles visual detection, localization, and semantic analysis. Through fine-tuning on 970,000 robot operation trajectories sourced from the open-xembodiment dataset~\cite{o2024open}, OpenVLA gains the capacity to generate actions in response to visual and language cues, paralleling the functionality of RT-2. $\pi_0$~\cite{black2024pi_0}, as one of the most recent advancements in VLA models, integrates pre-trained Vision-Language Models (VLMs) with a flow matching architecture for continuous action generation, facilitating high-frequency and dexterous control. Its multi-stage training strategy, combining diverse pre-training with high-quality post-training data, enhances adaptability to complex real-world tasks.

\par \textbf{(b) Cognition in VLN Models}

Building upon early VLN methods that relied on symbolic representations and discrete environments, recent advancements have introduced neural networks capable of handling more complex, real-world scenarios~\cite{chen2021history, georgakis2022cross, vasudevan2021talk2nav, li2021improving, xia2020multi, zhou2024navgpt, dou2022foam, lin2021scene, li2023improving}. For example, HAMT~\cite{chen2021history}, which employs a hierarchical vision transformer to encode panoramic observations, capturing spatial and temporal contexts. This approach enables the agent to remember visited locations and actions, leading to improved navigation performance across various VLN tasks. CM2~\cite{georgakis2022cross} introduces a method where agents predict top-down semantic maps from egocentric views, aligning these with language instructions through cross-modal attention mechanisms, enabling effective path planning by grounding language in explicit spatial representations. Although these methods exhibit stronger semantic understanding and generalization capabilities compared to earlier solutions, the backbones they rely on still fall short compared to LLMs and VLMs. Therefore, recent studies in VLN have begun to utilize LLMs or VLMs directly as agents to carry out navigation tasks~\cite{zhang2024vision, an2023bevbert, zhou2024navgpt,zhou2024navgpt2, chen2024mapgpt,long2024discuss,zhan2024mc,long2024instructnav}. NaVid~\cite{zhang2024navid}, in particular, stands out by utilizing a video-based approach that relies solely on RGB inputs, eliminating the need for depth sensors, maps, or odometry data. This simplification not only reduces computational overhead but also enhances generalization to unseen environments, making it more adaptable to real-world scenarios compared to methods that depend on more complex and resource-intensive inputs. To enhance the agent's navigation capabilities, Instructnav~\cite{long2024instructnav} introduces Dynamic Chain-of-Navigation (DCoN) and Multi-sourced Value Maps. DCoN unifies various navigation instructions into a standard planning paradigm, dynamically updating the next action and landmarks based on observed objects at each decision step. This allows Instructnav to align semantic labels, efficiently explore unseen environments, and conduct commonsense reasoning for different types of navigation instructions.

Though some AI approaches partially achieve the Perception-Cognition-Action loop, none fully replicate the human system. As described in~\ref{2.1.2}, the degree of cognition of the agent is divided into four stages from shallow to deep: memory-based, reasoning-based, creation-based, and hypothesis-based. However, at the current stage, including VLAs and VLNs, the development of all agents has only reached the first stage. That is, extensive training on large-scale datasets enables the model to memorize certain patterns for processing perceptual data. This initial stage of cognition relies heavily on pattern recognition and rote learning, which limits the agent's ability to generalize beyond the training data. Future advancements need to focus on enhancing the reasoning and creative capabilities of agents to achieve deeper levels of cognition and more robust performance in diverse environments. Such progress is crucial for the development of more intelligent and adaptive agents that can handle complex and dynamic real-world tasks.

\subsection{Remarks and Discussions}\label{4.2}

% \par This section discusses the current limitations in perception-cognition-action systems and suggests future research directions to close the loop.

% \subsubsection{Challenges in Creating a Full Perception-Cognition-Action Loop}

This section discusses the limitations of existing technologies in the Perception-Cognition-Action Loop, including Multimodal Input Integration, Model Generalization, and Real-World Adaptation.

\par \textbf{(1) Multimodal Input Integration} 

\textbf{Representation and Alignment:} Effectively representing and harmonizing diverse modalities, including visual, auditory, and textual data, continues to be an intricate challenge~\cite{baltruvsaitis2018multimodal}. Each modality has its own unique statistical characteristics and temporal behaviors. These differences pose significant hurdles to creating unified representations that can enable seamless integration. \textbf{Conflict Resolution:} Multimodal systems frequently grapple with conflicting information from disparate sensory inputs~\cite{dritsas2025multimodal}. They concurrently face issues of inconsistent spatial and temporal resolutions. Resolving such conflicts and ensuring consistency between perception and cognition are crucial for reliable decision-making and action execution. \textbf{Real-Time Processing Constraints:} Implementing real-time processing capabilities is vital for applications requiring immediate responses. Balancing the computational demands of multimodal integration with the necessity for low-latency processing remains an ongoing challenge.

\par \textbf{(2) Model Generalization} 

\textbf{Domain Shifts:} Variations in data distributions between training and testing environments can lead to significant declines in model performance~\cite{stacke2020measuring}. For example, differences in sensors or environmental conditions can affect the accuracy of medical image segmentation or autonomous driving systems. \textbf{Out-of-Distribution Generalization:} While data augmentation techniques can improve model robustness and generalization, identifying the optimal strategy for a given task remains an open problem~\cite{yang2024generalized}. Overfitting to augmented data is also a concern, as it can introduce irrelevant features that reduce performance on unseen target domains. \textbf{Data Scarcity and Imbalance:} The scarcity of high-quality, labeled multimodal datasets poses a significant challenge in creating a full Perception-Cognition-Action loop. Existing datasets often suffer from imbalance across modalities, which severely limits the models' ability to learn comprehensive representations and hinders their generalization to diverse real-world scenarios.

\par \textbf{(3) Real-World Adaptation} 

\textbf{Noisy Environments:} Adapting multimodal systems to noisy environments is a critical challenge. In scenarios where ambient noise interferes with speech recognition, alternative modalities like gestures or gaze must be prioritized to maintain seamless interaction. Dynamic modality switching and real-time environmental awareness are essential strategies for handling such conditions. \textbf{Adaptive Interaction:} Systems must adapt dynamically to user preferences and contexts, tailoring modalities like voice or gaze to specific tasks or environments. \textbf{Feedback Mechanisms:} Effective feedback mechanisms are essential for reducing ambiguities in multimodal inputs and ensuring smooth interaction. Adaptive systems should refine input recognition dynamically to enhance adaptability.

\section{Neural Brain Memory Storage and Update}\label{section_Neural_Brain_Knowledge_Memory_Storage_and_Update}

\begin{figure*}[t!]
\centering
	\includegraphics[width=\textwidth]{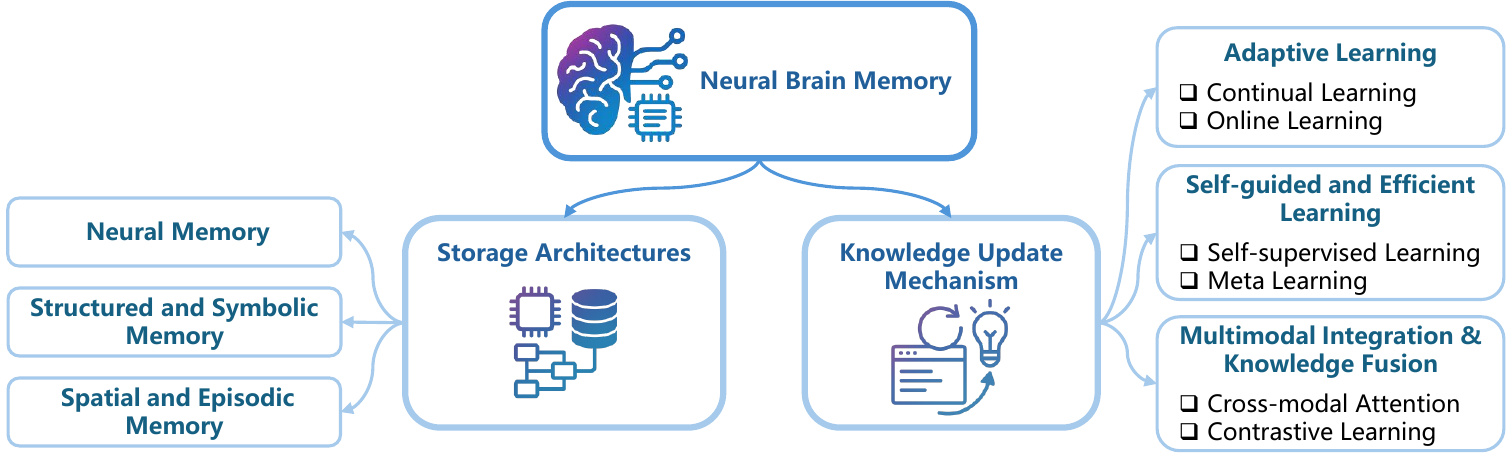}
    \vspace{-2em}
	\caption{A schematic overview of Neural Brain memory processing, divided into two parts: the Memory Storage Architecture (left), which illustrates the different structures and representations through which memory is stored; and the Memory Update Mechanism (right), which describes the methods and learning paradigms by which memories are modified, reinforced, or replaced.}
	\label{Fig_5}
    \vspace{-1em}
\end{figure*}

In this section, we explore how memory storage and knowledge update mechanisms are designed for the Neural Brain of embodied agents, drawing inspiration from cognitive and neuroscience. We first investigate representative memory architectures in embodied agents and then present mechanisms for efficient and adaptive knowledge update. Finally, we conclude with a discussion on major challenges that remain unsolved, especially in achieving human-level memory efficiency, flexibility, and multimodal integration.

\noindent \textbf{Insights of Human Knowledge Storage and Update from Neuroscience:}
As illustrated in Sec. \ref{2.1.3}, knowledge storage and update in the human brain are guided by coordinated neural mechanisms that ensure both stability and adaptability. Information initially enters through short-term and working memory systems, where it is temporarily held and manipulated. Through repetition and rehearsal, select content is transferred to long-term memory via hippocampal encoding. The hippocampus operates as a hashing-like structure, efficiently associating new inputs with existing memory networks. Consolidation is supported by memory replay, which reactivates and strengthens previously encoded sequences. Learning occurs through synaptic modification, where neural connections are adjusted based on experience. Neuronal competition plays a key role, as more excitable neurons are preferentially integrated into memory traces, resulting in selective and efficient encoding. Memory resolution is task-dependent, with higher precision allocated to contextually important details. Synaptic weakening facilitates forgetting, allowing the system to remove outdated information and maintain relevance. Together, these processes form a dynamic and efficient framework for lifelong memory management.

% \begin{figure}[h!]
% \centering
% 	\includegraphics[scale=0.42]{Thai/figures/Fig5.pdf}
%     \vspace{-1em}
% 	\caption{A schematic overview of Neural Brain memory storage and update.}
% 	\label{Fig_5}
%     \vspace{-1em}
% \end{figure}

\subsection{Embodied Agent Knowledge Storage and Update}

A core capability of the human brain is its ability to store richly structured, context-dependent memories and flexibly recall them. Embodied agents operating in complex and dynamic environments require similarly powerful memory systems. This section first reviews memory architectures, including neural, structured and symbolic, and spatial-episodic forms. It then examines how stored knowledge is updated over time to support continual learning, adaptation, and multimodal integration. The overview structure of the section is illustrated in Fig. \ref{Fig_5}.

\subsubsection{Memory Architectures for Embodied Agents}\label{Memory Architectures for Embodied Agents}

\par Effective memory is essential for embodied agents to operate in complex, dynamic environments where information must be retained, organized, and retrieved across time. Inspired by principles from cognitive neuroscience, memory systems in embodied AI are designed to support persistent representations, contextual reasoning, and flexible adaptation. This section explores three key categories of memory mechanisms: neural memory systems that emulate working and episodic recall, structured and symbolic memory for semantic representation and reasoning, and spatial-episodic memory that captures the agent's situated experience across space and time. Some representative works are shown in Table \ref{tab:stacked_memory}.

\begin{table*}[t!]
\caption{Representative knowledge storage methods (Sec. \ref{Memory Architectures for Embodied Agents}) for embodied agent. SEM is the abbreviation for Spatial and Episodic Memory.}
\vspace{-1em}
\renewcommand{\arraystretch}{1.5}
\centering
\resizebox{1\linewidth}{!}{
\begin{tabular}{ccccc}
\toprule[2pt]
\multicolumn{2}{c}{\textbf{Category/Method}} & \textbf{Year} & \textbf{Details} & \textbf{Highlights} \\
\midrule

\multirow{8}{*}{\rotatebox{90}{\scriptsize \textbf{(a) Neural Memory Systems}}}
& NTM \cite{graves2014neural} & 2014 & Neural Turing Machine & Introduced differentiable memory for neural networks \\ 
& DAM \cite{park2020distributed} & 2020 & Distributed Associative Memory & Proposed a memory model with distributed representation \\ 
& Embodied VideoAgent \cite{fan2024embodied} & 2024 & Embodied Agent & Developed an agent for long-form video understanding \\ 
& ESM \cite{lenton2021egospheric} & 2021 & Egospheric Spatial Memory & Encodes spatial memory in an ego-centric frame \\ 
& Mem2Ego \cite{zhang2025mem2ego} & 2025 & Memory-to-Ego Mapping & Maps memory representations to ego-centric views \\ 
& KARMA \cite{wang2024karma} & 2024 & Knowledge-Aware Memory & Integrates knowledge graphs into memory systems \\ 
& MINDSTORES \cite{chari2025mindstores} & 2025 & Memory-Informed Decision Making & Enhances decision-making with memory modules \\ 
& Skip-SCAR \cite{liu2024skip} & 2024 & Embodied Visual Navigation & Optimizes navigation with hardware-friendly design \\ \midrule

\multirow{12}{*}{\rotatebox{90}{\scriptsize \textbf{(b) Structured and Symbolic Memory}}}
& DAMCS \cite{Yang2025HierarchicalKG} & 2025 & Hierarchical Knowledge Graphs & Combines distributed and symbolic memory structures \\ 
& Scene-MMKG \cite{Song2024SceneMMKG} & 2024 & Scene Multi-Modal Knowledge Graph & Integrates multi-modal data into scene understanding \\ 
& AKGVP \cite{akgvp2024} & 2024 & Knowledge Graph Visual Perception & Aligns knowledge graphs with visual perception \\ 
& AdaptBot \cite{Singh2025AdaptBot} & 2025 & Task Decomposition & Uses LLMs and knowledge graphs for task refinement \\ 
& SafetyEKG \cite{Qi2024SafetyEKG} & 2024 & Safety Knowledge Graph & Incorporates safety constraints into embodied agents \\ 
& ESGNN \cite{esgnn2024} & 2024 & Embodied Scene Graph Neural Network & Models scene graphs for embodied reasoning \\ 
& SGRec3D \cite{koch2023sgrec3d} & 2023 & 3D Scene Graph & Self-supervised learning for 3D scene graphs \\ 
& 3DGraphLLM \cite{3dgraphllm2024} & 2024 & 3D Graph Language Model & Integrates 3D graphs with language models \\ 
& SGM \cite{kurenkov2023modeling} & 2023 & Scene Graph Memory & Models memory for scene graph understanding \\ 
& EmbodiedRAG \cite{xie2024embodied} & 2024 & Retrieval-Augmented Generation & Applies RAG to embodied agents \\ 
& EmbodiedVSR \cite{zhang2025embodiedvsr} & 2025 & Visual Scene Reasoning & Enhances scene reasoning in embodied settings \\ 
& Structure-CLIP \cite{huang2023structure} & 2023 & Structured Contrastive Learning & Combines structure with contrastive learning \\ \midrule

\multirow{4}{*}{\rotatebox{90}{\scriptsize \shortstack{\textbf{(c) SEM}}}}
& SAT \cite{cho2024spatially} & 2024 & Spatial Attention Transformer & Applies spatial attention in episodic memory \\ 
& 3D-Mem \cite{yang2024dmem} & 2024 & 3D Memory Mapping & Constructs 3D memory for spatial understanding \\ 
& STMA \cite{lei2025stma} & 2025 & Spatio-Temporal Memory Architecture & Integrates spatial and temporal memory modules \\ 
& SALI \cite{pan2024planning} & 2024 & Spatial-Aware Learning Interface & Enhances planning with spatial awareness \\
\bottomrule[2pt]

\end{tabular}
}
\vspace{-1em}
\label{tab:stacked_memory}
\end{table*}

\par \textbf{(a) Neural Memory Systems:}

Memory-Augmented Neural Networks (MANNs) are a class of neural architectures designed to simulate the function of working memory and episodic recall in biological agents. Inspired by the brain's interplay between the prefrontal cortex, responsible for task control and planning, and the hippocampus, responsible for memory retrieval and binding of spatial-temporal experiences, MANNs incorporate a persistent memory store that can be selectively accessed during task execution. These capabilities are crucial for embodied agents interacting in dynamic and partially observable environments.

The Neural Turing Machine (NTM) \cite{graves2014neural} is the typical example of a MANN architecture. It integrates a recurrent controller with an external, differentiable memory matrix and content-based addressing mechanisms. This design allows the network to store input sequences and retrieve them later for structured reasoning and sequential prediction. However, NTMs often suffer from memory interference and limited long-range consistency. To address these challenges, models such as the Distributed Associative Memory (DAM) \cite{park2020distributed} extend the architecture by introducing multiple parallel memory banks and a memory-refreshing loss that mimics synaptic consolidation observed in biological systems. This allows agents to maintain and reinforce relational patterns across time, enabling more stable learning in long-horizon tasks.

In embodied contexts, MANNs have been integrated into systems that model scene memory from egocentric sensorimotor experience. The Embodied VideoAgent \cite{fan2024embodied} combines vision, depth, and pose information to build a persistent representation of the environment that updates as the agent interacts with objects and navigates through space. Similarly, Mem2Ego \cite{zhang2025mem2ego} introduces a framework that integrates egocentric visual inputs with a global memory module to guide long-horizon decision-making. The system dynamically retrieves relevant cues from memory and aligns them with current sensory input, enabling more effective spatial reasoning in complex, partially observed environments. These approaches are conceptually aligned with the constructive episodic simulation hypothesis, where agents simulate future decisions by recombining stored experiences.

More recent architectures emphasize hybrid memory systems that separate fast-changing task-related memory from slow-updating world knowledge. For example, KARMA \cite{wang2024karma} introduces a dual-memory model, where long-term memory encodes scene graphs capturing 3D semantic understanding, and short-term memory tracks recent object state changes. This design mirrors cognitive theories of complementary learning systems, in which episodic memory supports rapid contextual adaptation, while semantic memory enables generalization and abstraction. Another example, Egospheric Spatial Memory (ESM) \cite{lenton2021egospheric}, uses egocentric geometric encoding to provide location-sensitive memory in robotic visuomotor tasks, enhancing both spatial reasoning and integration with learned control policies.

The integration of symbolic reasoning into memory-augmented architectures is also gaining traction. MINDSTORES \cite{chari2025mindstores} encodes experience in (state, task, plan, outcome) tuples stored as dense embeddings. These memory records are queried by a large language model to generate updated plans and actions. Such a structure reflects mental model formation, where human cognition relies on episodic traces organized semantically for flexible reasoning. Finally, optimization frameworks such as Skip-SCAR \cite{liu2024skip} address the computational burden of memory-heavy architectures by introducing efficient inference techniques for visual navigation, demonstrating that memory augmentation can remain scalable for real-world deployment.

% \begin{table*}[t!]
% \caption{Representative knowledge storage methods (Sec. \ref{Memory Architectures for Embodied Agents}) for embodied agent.}
% \vspace{-1em}
% \renewcommand{\arraystretch}{1.25}
% \centering
% \resizebox{1\linewidth}{!}{
% \label{tab:stacked_memory}
% \begin{tabular}{ccc}
% \toprule[2pt]
% \multicolumn{3}{c}{\textbf{Memory Storage Architectures}} \\
% \hline
% Neural Memory Systems & Structured \& Symbolic Memory & Spatial \& Episodic Memory \\
% \hline
% NTM \cite{graves2014neural}, DAM \cite{park2020distributed} & DAMCS \cite{Yang2025HierarchicalKG}, Scene-MMKG \cite{Song2024SceneMMKG}, AKGVP \cite{akgvp2024} & SAT \cite{cho2024spatially} \\
% Embodied VideoAgent \cite{fan2024embodied}, ESM \cite{lenton2021egospheric} & AdaptBot \cite{Singh2025AdaptBot}, SafetyEKG \cite{Qi2024SafetyEKG}, ESGNN \cite{esgnn2024} & 3D-Mem \cite{yang2024dmem} \\
% Mem2Ego \cite{zhang2025mem2ego}, KARMA \cite{wang2024karma} & SGRec3D \cite{koch2023sgrec3d}, 3DGraphLLM \cite{3dgraphllm2024}, SGM \cite{kurenkov2023modeling} & STMA \cite{lei2025stma} \\
% MINDSTORES \cite{chari2025mindstores}, Skip-SCAR \cite{liu2024skip} & EmbodiedRAG \cite{xie2024embodied}, EmbodiedVSR \cite{zhang2025embodiedvsr}, Structure-CLIP \cite{huang2023structure} & SALI \cite{pan2024planning} \\
% \bottomrule[2pt]
% \end{tabular}
% }
% \vspace{-1em}
% \end{table*}

%%%%%%%%%%%%%%%%%%%%%%%%%%

\par \textbf{(b) Structured and Symbolic Memory:}

Structured and symbolic memory systems offer embodied agents a compact and compositional means of storing and reasoning over semantic information. These systems typically take the form of graph-based structures-such as knowledge graphs and scene graphs-that represent entities and their relationships. Rather than opposing paradigms, scene graphs can be viewed as a spatially grounded subclass of knowledge graphs. Together, they enable agents to integrate perception, commonsense reasoning, and task-relevant knowledge into a coherent memory representation.

Knowledge graphs (KGs) encode structured semantic knowledge as relational triples, allowing embodied agents to represent concepts, associations, and procedural experience in a symbolic form. In partially observable or uncertain environments, KGs support memory-driven reasoning and long-term planning. Kim \emph{et al.} \cite{Kim2024Leveraging} maintains task-relevant histories using KG-based structures to improve decision-making in memory-constrained settings. DAMCS \cite{Yang2025HierarchicalKG} support decentralized multi-agent collaboration by organizing information in modular, adaptive hierarchies that evolve through interaction and context-specific updates.

Multimodal extensions of KGs align symbolic reasoning with perception and action. Scene-MMKG \cite{Song2024SceneMMKG} builds knowledge graphs from visual inputs, grounding symbolic entities in spatial and semantic cues. AKGVP \cite{akgvp2024} enhances object-goal navigation by aligning KG representations with real-time perception. For goal refinement and safety, AdaptBot \cite{Singh2025AdaptBot} integrates structured memory with human input to decompose complex tasks, while Qi \emph{et al.} \cite{Qi2024SafetyEKG} incorporates commonsense constraints into embodied decision-making.

Scene graphs (SGs) specialize this structure to perceptual domains, capturing grounded relations like spatial proximity or support, e.g. (\textit{bottle, on, table}). Numerous works have demonstrated their utility for both visual understanding and spatial reasoning. ESGNN \cite{esgnn2024} introduces equivariant architectures for robust 3D scene graph prediction from point clouds. SGRec3D \cite{koch2023sgrec3d} enables self-supervised graph learning by reconstructing 3D scenes from graph-structured bottlenecks, while 3DGraphLLM \cite{3dgraphllm2024} conditions large language models on learned scene graphs for downstream multimodal tasks. These representations can also serve as memory modules, Kurenkov \emph{et al.} \cite{kurenkov2023modeling} propose Scene Graph Memory to maintain dynamic and partially observed graph states, and EmbodiedRAG \cite{xie2024embodied} uses retrieved scene subgraphs to augment large language model planning in navigation tasks.

Unified memory interfaces increasingly incorporate both knowledge graphs and scene graphs. EmbodiedVSR \cite{zhang2025embodiedvsr} performs dynamic reasoning over scene graphs using physics-aware chain-of-thought models, and the Structure-CLIP model \cite{huang2023structure} leverages structured negative samples informed by scene graphs to enhance multimodal representation learning. By encoding entities, spatial layouts, and conceptual relations, structured memory architectures provide agents with the grounding and flexibility necessary for long-horizon decision-making, generalization, and interpretable behavior.

\par \textbf{(c) Spatial and Episodic Memory:}

Spatial memory provides a persistent map-like representation of an agent's surroundings, supporting tasks such as navigation, object search, and spatial reasoning. Episodic memory complements this by retaining temporally-structured experiences-what was encountered, when, and where-thus enabling agents to make informed decisions based on past trajectories and events. These two forms of memory, inspired by the hippocampal-entorhinal system in biological organisms, are being unified in various architectural innovations.

One representative line of work is the Spatially-Aware Transformer (SAT) \cite{cho2024spatially}, which integrates spatial embeddings directly into attention-based models to construct place-centric episodic memories. Unlike standard sequence models, SAT allows embodied agents to attend to memory not just temporally, but also spatially, by encoding environmental location into transformer keys and queries. This facilitates long-horizon planning and memory-efficient action selection in embodied tasks.

To further structure experience in 3D space, 3D-Mem \cite{yang2024dmem} introduces a scene-centric episodic memory that maintains a compact spatial representation using multi-view RGB inputs. This architecture supports frontier-based exploration by encoding where the agent has been and what parts of the environment remain unknown, leveraging memory to guide curiosity-driven behavior.

Another unified approach is the Spatio-Temporal Memory Agent (STMA) \cite{lei2025stma}, which explicitly combines episodic memory modules with topological and semantic maps. STMA encodes both environmental dynamics and the agent's own experience trajectory into a structured memory, supporting temporally-aware and space-grounded policy learning. This co-evolution of spatial and episodic signals allows agents to operate over extended time horizons while maintaining context-awareness and adaptability.

On top of that, SALI \cite{pan2024planning} introduces a hybrid memory system where agents store real and imagined episodes in a shared architecture. This supports creative reasoning about unseen parts of the environment and allows the agent to simulate possible outcomes before acting-mirroring human mental time travel in problem-solving.

% \par Knowledge Graphs: Graph-based structures for storing and retrieving multimodal knowledge (e.g., OpenAI's CLIP for vision-language alignment).

% \par Continual Learning: Techniques like Elastic Weight Consolidation (EWC) and Generative Replay mitigate catastrophic forgetting in lifelong learning scenarios.

\subsubsection{Knowledge Update Mechanisms}\label{Knowledge Update Mechanisms}

\par Embodied agents must not only store and access knowledge but also continuously update it as they interact with dynamic environments. This requires mechanisms that support incremental learning, efficient adaptation, and integration of new experiences without compromising past knowledge. This section reviews core approaches to knowledge update in embodied systems, including adaptive learning over time, self-guided learning from limited supervision, and multimodal knowledge fusion that aligns diverse inputs into coherent memory representations. Some representative methods are summarized in Table \ref{tab:stacked_update}.

% \begin{table*}[t!]
% \caption{Representative knowledge update methods (Sec. \ref{Knowledge Update Mechanisms}) for embodied agent.}
% \vspace{-1em}
% \renewcommand{\arraystretch}{1.25}
% \centering
% \resizebox{0.85\linewidth}{!}{
% \label{tab:stacked_update}
% \begin{tabular}{ccc}
% \toprule[2pt]
% \multicolumn{3}{c}{\textbf{Knowledge Update Mechanisms}} \\
% \hline
% Adaptive Learning Over Time & Self-Guided and Efficient Learning & Multimodal Integration \\
% \hline
% NESYC \cite{choi2025nesyc}, Voyager \cite{wang2023voyager} & MetaMAE \cite{jang2023modality} & Meta-Transformer \cite{zhang2023meta} \\
% AccuACL \cite{park2025active}, AFEC \cite{wang2021afec} & BMSSL \cite{wang2023bmssl} & MM-ReAct \cite{yang2023mmreact} \\
% Mendez \emph{et al.} \cite{mendez2023embodied}, CAMA \cite{kim2024online} & DRESS \cite{cui2025dress}, M3AE \cite{geng2022multimodalmaskedautoencoderslearn} & UniCL \cite{li2024unicl}, UniTOUCH \cite{yang2024binding} \\
% FSTTA \cite{gao2023fast}, OpenPAL \cite{zhai2023building} & ReMA \cite{wan2025rema}, SSML-AC \cite{he2024selfsupervisedmetalearningalllayerdnnbased} & Uni-Perceiver v2 \cite{li2022uniperceiverv2generalistmodel}, Perceiver AR \cite{hawthorne2022general} \\
% \bottomrule[2pt]
% \end{tabular}
% }
% \vspace{-1em}
% \end{table*}

%%%%%%%%%%%%%%%%

\begin{table*}[t!]
\caption{Representative knowledge update methods (Sec. \ref{Knowledge Update Mechanisms}) for embodied agents.}
\vspace{-1em}
\renewcommand{\arraystretch}{1.5}
\centering
\resizebox{1\linewidth}{!}{
\begin{tabular}{ccccc}
\toprule[2pt]
\textbf{Category} & \textbf{Method} & \textbf{Year} & \textbf{Details} & \textbf{Highlights} \\
\midrule

\multirow{8}{*}{\rotatebox{90}{\scriptsize \shortstack{\textbf{(a) Adaptive Learning} \\ \textbf{Over Time}}}}
& NESYC \cite{choi2025nesyc} & 2025 & Neuro-symbolic Continual Learning & Combines LLMs and symbolic tools for continual learning in open domains \\ 
& Voyager \cite{wang2023voyager} & 2023 & LLM-powered Embodied Agent & Enables autonomous skill acquisition in Minecraft without human intervention \\ 
& AccuACL \cite{park2025active} & 2025 & Active Continual Learning & Enhances learning efficiency through active data selection strategies \\ 
& AFEC \cite{wang2021afec} & 2021 & Active Forgetting Mechanism & Mitigates negative transfer by actively forgetting detrimental knowledge \\ 
& Mendez \emph{et al.} \cite{mendez2023embodied} & 2023 & Lifelong Task and Motion Planning & Integrates lifelong learning with task and motion planning for robots \\ 
& CAMA \cite{kim2024online} & 2024 & Online Adaptation Framework & Facilitates real-time adaptation in dynamic environments \\ 
& FSTTA \cite{gao2023fast} & 2023 & Fast Self-Tuning Task Adaptation & Accelerates adaptation to new tasks with minimal data \\ 
& OpenPAL \cite{zhai2023building} & 2023 & Language-Policy Co-training & Aligns language models with policy learning for open-ended tasks \\ \midrule

\multirow{6}{*}{\rotatebox{90}{\scriptsize \shortstack{\textbf{(b) Self-Guided and} \\ \textbf{Efficient Learning}}}}
& MetaMAE \cite{jang2023modality} & 2023 & Modality-Agnostic SSL & Employs meta-learned masked autoencoders for diverse modalities \\ 
& BMSSL \cite{wang2023bmssl} & 2023 & Bootstrapped Meta SSL & Simulates human learning processes for improved generalization \\ 
& DRESS \cite{cui2025dress} & 2025 & Disentangled Representation Learning & Enables fast adaptation through disentangled self-supervised tasks \\ 
& M3AE \cite{geng2022multimodalmaskedautoencoderslearn} & 2022 & Multimodal Masked Autoencoder & Learns transferable representations across vision and language \\ 
& ReMA \cite{wan2025rema} & 2025 & Multi-Agent Meta-Thinking & Enhances LLM reasoning via multi-agent reinforcement learning \\ 
& SSML-AC \cite{he2024selfsupervisedmetalearningalllayerdnnbased} & 2024 & Self-Supervised Meta-Learning & Adapts deep neural networks for all-layer adaptive control \\ \midrule

\multirow{6}{*}{\rotatebox{90}{\scriptsize \shortstack{\textbf{(c) Multimodal} \\ \textbf{Integration}}}}
& Meta-Transformer \cite{zhang2023meta} & 2023 & Unified Multimodal Framework & Processes 12 modalities with a shared transformer backbone \\ 
& MM-ReAct \cite{yang2023mmreact} & 2023 & Multimodal Reasoning Agent & Integrates vision and language for interactive decision-making \\ 
& UniCL \cite{li2024unicl} & 2024 & Universal Contrastive Learning & Aligns diverse modalities through contrastive objectives \\ 
& UniTOUCH \cite{yang2024binding} & 2024 & Multimodal Binding Mechanism & Binds multimodal inputs for coherent representation learning \\ 
& Uni-Perceiver v2 \cite{li2022uniperceiverv2generalistmodel} & 2022 & Generalist Model & Unifies perception across modalities with a single model \\ 
& Perceiver AR \cite{hawthorne2022general} & 2022 & Autoregressive Perceiver & Models sequential data across modalities with autoregressive decoding \\
\bottomrule[2pt]

\end{tabular}
}
\vspace{-1em}
\label{tab:stacked_update}
\end{table*}

\par \textbf{(a) Adaptive Learning Over Time:} 

As embodied agents operate in evolving environments, they must update their internal models without retraining from scratch. This demands memory and learning systems that support adaptive learning over time, particularly through mechanisms such as online learning and continual learning. These paradigms aim to overcome the problem of catastrophic forgetting while enabling agents to incrementally integrate new experiences, goals, and knowledge structures.

Continual learning approaches often involve parameter isolation, memory replay, or knowledge consolidation strategies. In embodied contexts, NESYC \cite{choi2025nesyc} proposes a neuro-symbolic framework that combines structured commonsense priors with real-time perception, allowing continual learning across tasks in open-ended domains. Complementary to this, Voyager \cite{wang2023voyager} enables embodied agents to autonomously acquire and organize skills over time through LLM-guided exploration and a self-expanding codebase. AccuACL \cite{park2025active} introduces an active sampling strategy based on Fisher information to prioritize informative data, supporting balanced knowledge acquisition and retention. Furthermore, AFEC \cite{wang2021afec} draws inspiration from biological neural mechanisms, employing a synaptic expansion-convergence process to actively forget outdated knowledge that may hinder new learning, thereby enhancing adaptability in dynamic environments. These systems leverage both symbolic and neural memory for lifelong adaptability. Beyond perceptual adaptation, continual learning also supports modular task planning and knowledge transfer. Lifelong planning systems such as the memory-augmented agent proposed in Mendez \emph{et al.} \cite{mendez2023embodied} leverage structured memory to decompose novel goals into reusable subtasks.

Online learning complements continual learning by enabling agents to make parameter updates on a per-instance basis, which is particularly effective in interactive scenarios such as embodied instruction following. Confidence-weighted update mechanisms like CAMA \cite{kim2024online} facilitate task-free online adaptation by controlling memory plasticity without requiring task boundaries. Replay-free online updates have also gained attention as a way to reduce memory overhead in continual adaptation. FSTTA \cite{gao2023fast} proposes a fast-slow test-time adaptation method for vision-and-language navigation that avoids replay buffers by combining stable and transient gradient pathways. Similarly, adaptive control strategies in agents trained for diverse instruction-following tasks demonstrate that performance can be improved through online optimization, without storing past episodes \cite{zhai2023building}. These approaches highlight the feasibility of closed-loop learning in embodied agents, without the need for explicit memory replay.

\par \textbf{(b) Self-Guided and Efficient Learning:}

Efficient learning mechanisms are crucial for agents operating in dynamic environments, where labeled data may be scarce or unavailable. Self-supervised learning (SSL) and meta-learning have emerged as key approaches to address this challenge.

SSL enables agents to extract meaningful representations from unlabeled data by leveraging inherent data structures. It has also been extended to be modality-agnostic, allowing learning across diverse data types. For instance, MetaMAE \cite{jang2023modality} interprets masked autoencoding as a meta-learning task, enhancing reconstruction through gradient-based adaptation and task contrastive learning. Similarly, Bootstrapped Meta Self-Supervised Learning (BMSSL) \cite{wang2023bmssl} combines SSL and meta-learning in a bi-level optimization framework, enabling models to self-improve by generating their own learning signals.

Meta-learning or learning to learn, equips agents with the ability to adapt quickly to new tasks by leveraging prior experience. DRESS \cite{cui2025dress} introduces a task-agnostic approach that utilizes disentangled representations to create self-supervised tasks, facilitating rapid adaptation in few-shot learning scenarios. In the realm of large language models, ReMA \cite{wan2025rema} employs multi-agent reinforcement learning to instill meta-thinking capabilities, decoupling reasoning processes into hierarchical agents for improved generalization and robustness.

Integrating SSL and meta-learning has shown promise in various applications, including adaptive control in robotics. For example, a self-supervised meta-learning framework has been proposed for all-layer deep neural network-based adaptive control, enabling robots to adapt in real-time to dynamic environments without explicit supervision \cite{he2024selfsupervisedmetalearningalllayerdnnbased}. These combined approaches allow agents to autonomously acquire useful priors, accelerate adaptation, and learn flexibly from sparse feedback, which are crucial for long-term autonomy in open-ended environments.

\par \textbf{(c) Multimodal Integration and Knowledge Fusion:}  

To operate effectively in sensory-rich environments, agents must convert heterogeneous inputs into unified representations. Multimodal fusion mechanisms perform this integration at the memory level, aligning and encoding diverse modalities into structured embeddings that support storage, retrieval, and cross-modal reasoning.

A central mechanism for multimodal fusion is the use of cross-modal attention, which enables dynamic weighting and alignment of features across input streams. Architectures such as Meta-Transformer \cite{zhang2023meta} and MM-ReAct \cite{yang2023mmreact} use shared token spaces and unified attention layers to blend features across modalities into cohesive representations. These fused embeddings serve as persistent memory elements that encapsulate contextual and cross-modal correlations, supporting policy learning and decision making.

Contrastive learning provides an alternative and complementary strategy to attention-based fusion. It encourages alignment by pulling representations from semantically similar cross-modal inputs closer in latent space. Models like M3AE \cite{geng2022multimodalmaskedautoencoderslearn} and UniCL \cite{li2024unicl} use contrastive objectives to bind different modalities into a shared representation format, thereby structuring memory around cross-modal consistency. These embeddings improve generalization, facilitate retrieval across modalities, and make memory representations robust to missing or noisy inputs.

Memory-centric fusion also benefits from token unification and modality alignment. UniTOUCH \cite{yang2024binding} and Uni-Perceiver v2 \cite{li2022uniperceiverv2generalistmodel} project modality-specific inputs into task-agnostic encodings, allowing multimodal episodes to be aggregated and reused across tasks. These unified embeddings enable agents to maintain persistent knowledge structures that are accessible regardless of the original sensory channel.

Latent space generalization is further supported in frameworks such as Perceiver AR \cite{hawthorne2022general}, which uses a learned latent memory buffer to integrate and predict across modalities over time. The model incrementally builds a fused state representation from streamed multi-sensory input, demonstrating how multimodal memory can be both scalable and sequentially updated.

\subsection{Remarks and Discussions}\label{5.2}

Although memory systems for embodied agents have seen substantial advances, several fundamental challenges remain. These include maintaining a balance between stability and plasticity in continual learning, addressing the fragmentation between different modalities, managing the computational demands of large-scale memory storage and retrieval, and improving the interpretability of memory-guided behavior in complex, real-world settings.

% \subsubsection{Challenges in Knowledge Storage and Update}

\par \textbf{(1) Stability-Plasticity Dilemma:} 
Embodied agents operating in open-ended environments must adapt rapidly to novel experiences without compromising previously acquired knowledge. This balance between plasticity (for adaptation) and stability (for retention) remains a core challenge. While memory consolidation strategies, such as memory replay and parameter isolation, have shown success in static task sequences, they often struggle when the agent must learn continuously from evolving sensory data and shifting goals. Zheng \textit{et al.} \cite{zheng2025lifelong} highlights that in practice, lifelong learning frameworks either forget older skills or become too rigid to acquire new ones, particularly in long-horizon interactions where task boundaries are implicit and experience is non-stationary. Moreover, the cost of maintaining this balance grows as the memory space increases, making retention mechanisms computationally and architecturally expensive.

\par \textbf{(2) Fragmentation of Memory Modalities:} 
Neural, symbolic, and spatial-episodic memory systems often play complementary roles in embodied reasoning, supporting functions such as perception, semantic understanding, and spatial-temporal recall. However, in many current implementations, these components operate in relative isolation, lacking mechanisms for consistent alignment or mutual referencing. For example, a symbolic knowledge graph might encode task-relevant goals or object relations, while spatial memory maintains location-based context, yet the connection between these modalities is frequently implicit or task-specific. This partial integration can limit the agent's ability to perform unified reasoning across perceptual and abstract domains, especially in environments requiring generalization. Although hybrid memory architectures are increasingly explored, standardized frameworks for harmonizing these representations remain limited \cite{liu2024meia, liu2024surveywm}. This may result in inconsistent behavior, where agents lean heavily on perceptual cues or encounter difficulty translating high-level plans into grounded actions. Cross-modal memory alignment, unified embedding spaces, and layered memory interaction are still emerging.

\par \textbf{(3) Scalability and Computational Bottlenecks:} 
As embodied agents accumulate larger amounts of structured and unstructured information over time, maintaining efficiency becomes increasingly difficult. High-fidelity spatial maps, dense episodic logs, and rich semantic annotations impose high memory storage and retrieval costs. This challenge is particularly significant in real-time settings, where agents must repeatedly query and update memory to respond to dynamic environments. While strategies like attention-based indexing and prioritized replay are designed to reduce unnecessary computation, their effectiveness is often limited by the growing complexity of memory contents and the frequency of access required. As the volume of stored data grows, the effort required to identify and retrieve relevant information also increases, which may impact decision-making speed and reduce responsiveness in latency-sensitive scenarios. 

\par \textbf{(4) Interpretability of Memory-Guided Decision Making:}
Although modern memory systems enable powerful generalization, their inner workings remain largely opaque. In particular, end-to-end models that combine large-scale memory stores with transformer-based policies make it difficult to trace how prior experiences influence current decisions. This lack of transparency becomes problematic in applications requiring explainability, such as human-robot collaboration, where users must understand and anticipate the agent's reasoning. Symbolic approaches partially address this through interpretable graph structures or plan traces, but these solutions rarely scale or integrate smoothly with perceptual learning systems.

\section{Neural Brain Hardware and Software}\label{section_Neural_Brain_Hardware_and_Software}

This section investigates the hardware and software infrastructure underlying the Neural Brain of embodied agents. Drawing from neuroscience, we note that human intelligence arises from the deep integration of biological hardware and software. Similarly, embodied AI systems must couple physical computing with dynamic learning processes. We begin by examining neuromorphic hardware technologies, then discuss the software frameworks enabling embodied cognition, and finally address key mechanisms for energy-efficient learning at the edge. Finally, some related remarks and discussions are given.

%%%%% Tianxiang

% \subsection{Human Hardware \& Software: Insights from Neuroscience}
\noindent \textbf{Insights of Human Hardware \& Software from Neuroscience:}
The complete insights as indicated in Sec. \ref{2.1.4}. In brief, in the human brain, hardware and software work together seamlessly to enable intelligent behavior~\cite{laydevant2024hardware}.
Neuroscience reveals that this integration is not merely metaphorical, but fundamentally embedded in human brain's biological architecture~\cite{brette2022brains}. 
The "hardware" refers to the brain's physical infrastructure, including neurons, synaptic networks and specialized brain regions they form~\cite{galakhova2022evolution}. This biological architecture has been shaped through evolution to support complex perception-cognition-action workflows~\cite{barrett2012hierarchical}. 
In contrast, the "software" refers to the brain's dynamic and distributed patterns of neural activity, which coordinate basic functions such as reflexes, sensory processing, autonomic regulation, and basic motor control through continuous interactions among neural populations~\cite{pulvermuller2018neural}. These low-level processes form the foundation for more advanced mechanisms, including multimodal sensing, perception-cognition-action loops, and memory systems shaped by neuroplasticity~\cite{gupta2023neuroprosthetics}.
The brain's hardware and software are deeply intertwined and continuously co-evolve. Neural activity reshapes structural connectivity, while the anatomical architecture constrains and guides neural dynamics.
% Understanding human brain through this integrated lens not only advances our knowledge of biological intelligence, but also provides valuable insights for developing neural architectures in embodied agents-systems that need to perceive, learn, and act within complex physical world. 

\begin{figure*}[t!]
	\centering
	\includegraphics[width=\textwidth]{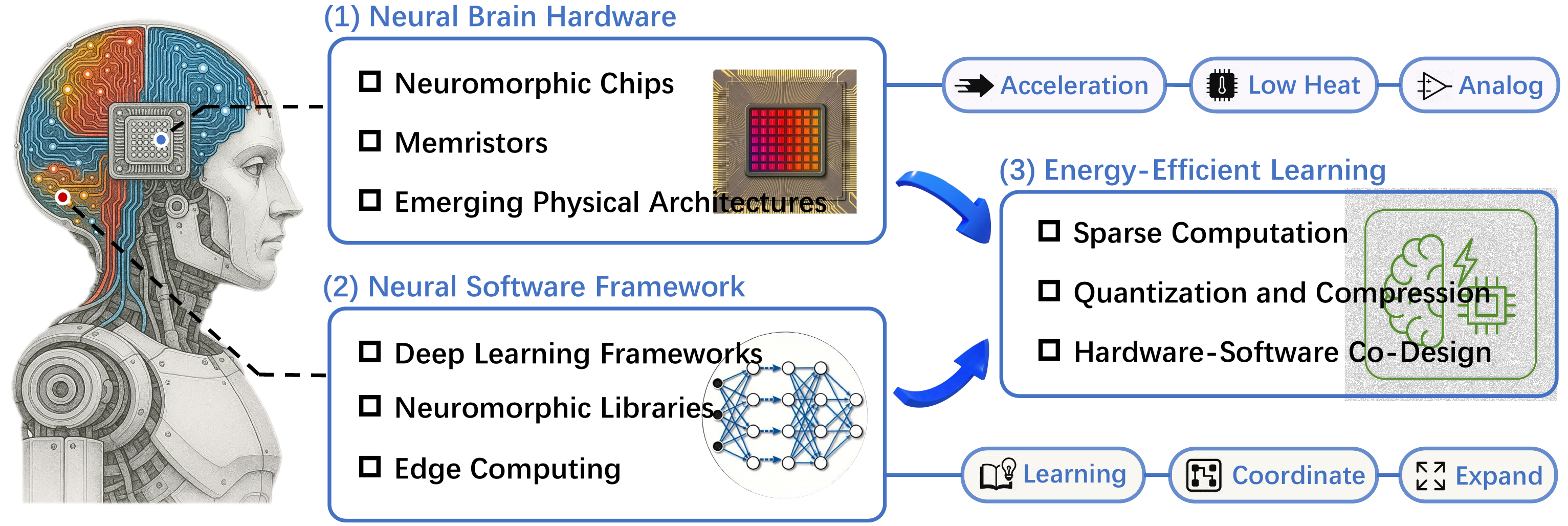}
	\vspace{-2em}
	\caption{The hardware and software frameworks in Neural Brain for embodied humanoid agents. The hardware system consists of three key components: neuromorphic chips, memristors and emerging physical architectures. The software framework cover deep learning frameworks, neuromorphic libraries and edge computing. To implement the energy-efficient learning on edge platforms, the Neural Brain is further optimized with sparse computation, quantization and compression, and hardware-software co-design.}
	\label{chap6_figure}
	\vspace{-1em}
\end{figure*}

\subsection{Embodied Agent Hardware \& Software}

To support robust and adaptive behavior in the real world, embodied agents require tight integration between hardware capabilities and software intelligence. Inspired by the co-evolution of neural dynamics and biological structure in the brain, this section reviews the design of hardware-software systems for Neural Brain architectures. An overview of the section's organization is provided in Fig. \ref{chap6_figure}.

\subsubsection{Neuromorphic Hardware}\label{Neuromorphic Hardware}

\par Neuromorphic hardware refers to a class of physical systems inspired by both the structural organization and the dynamic processing mechanisms of biological nervous systems.
These mechanisms encompass not only the anatomical structure of neurons and synapses, but also the temporal dynamics of neural activity~\cite{bohnstingl2019neuromorphic}, such as spike-based signaling, synaptic plasticity, and evolving circuit-level states~\cite{walter2015neuromorphic}. Some representative methods are summarized in Table \ref{tab:NeuromorphicHardware}.
% The scope of neuromorphic hardware extends beyond computational devices like spiking neural network chips and memristive arrays, to include systems that interface directly with biological tissue, such as the brain-machine interface and programmable biohybrid platform. 
% This diversity reflects a growing recognition that embodied intelligence depends not only on neuromorphic computation, but also on deep and continuous interaction between ego-motion and the external environment.

\par \textbf{(a) Neuromorphic Chips:}
The neuromorphic chips are specialized energy-efficient computing platforms designed to emulate the structure and dynamics of biological neural systems. 
Unlike traditional von Neumann architectures that separate memory and processing, neuromorphic chips integrate storage, computation and communication to enable real-time, energy-efficient information processing~\cite{schuman2022opportunities}. 
Their design is guided by core principles observed in the brain, such as massive parallelism, event-driven operation, and asynchronous processing, aiming to replicate the adaptability and efficiency of biological intelligence~\cite{shahsavari2023advancements}.
The architectural foundation of neuromorphic chips is built on synapse-neuron cores that combine local memory, spike-based computation, and asynchronous communication. 
These units are linked via on-chip networks (NoC), supporting scalable and modular low-latency communication~\cite{carrillo2012scalable}. 
To enhance system performance, advanced designs incorporate wireless NoC, multi-chip integration, and event-based routing, increasing bandwidth and timing control. 
Some architectures further integrate sensing and computation, co-locating dynamic vision input and spike-based inference on the same asynchronous substrate to achieve fast, power-efficient interaction with the environment~\cite{rathi2023exploring}.
Neuromorphic chips are typically designed to support SNNs, which encode information through discrete temporal spikes rather than continuous activations. 
This spike-based coding fits naturally with the event-driven architecture, allowing precise and sparse neural computation. 
Local learning mechanisms, such as spike-timing-dependent plasticity (STDP), support real-time on-chip adaptation without centralized training. 
% Many chips also provide support for conventional models like convolutional and recurrent networks, enhancing versatility across applications.

With event-driven activation, asynchronous processing, and sparse neural activity, neuromorphic chips achieve low idle power and fast response under strict resource constraints~\cite{mehonic2024roadmap}. 
The co-localization of memory and computation reduces data movement and enables high throughput, making these systems ideal for embodied intelligence in power-limited, real-time settings.
Representative platforms showcase diverse strategies: Intel's Loihi~\cite{davies2018loihi} offers programmable neurons and embedded plasticity; BrainChip's Akida integrates convolutional and recurrent structures within an event-based framework; BrainScaleS-2~\cite{pehle2022brainscales} uses a mixed-signal design for accelerated emulation of neural dynamics. 
SynSense's Speck~\cite{richter2023speck} combines dynamic vision sensing and spike-based inference in a fully asynchronous chip, suitable for mobile applications with sub-milliwatt power consumption. 
Other platforms, such as Innatera's T1 and GrAI Matter's NeuronFlow, adopt analog-digital hybrid schemes for near-sensor intelligence, while ODIN, DynapCNN, and ROLLS explore STDP learning, memristive integration, and spike-based deep networks~\cite{frenkel20180}. 
% These advances underline the growing maturity and potential of neuromorphic chips in supporting real-world, low-power cognitive systems.

% \begin{table*}[t!]
% \caption{Representative neuromorphic hardware (Sec. \ref{Neuromorphic Hardware}) for embodied agent.}
% \vspace{-1em}
% \renewcommand{\arraystretch}{1.25}
% \centering
% \resizebox{0.78\linewidth}{!}{
% \label{tab:Hardware}
% \begin{tabular}{ccc}
% \toprule[2pt]
% \multicolumn{3}{c}{\textbf{Neuromorphic Hardware}} \\
% \hline
% Neuromorphic Chips & Memristors & Emerging Physical Architectures \\
% \hline

% Loihi~\cite{davies2018loihi}, BrainScaleS-2~\cite{pehle2022brainscales} &  ODIN \cite{shivanandamurthy2021odin}, NeuroSim \cite{chen2018neurosim} & LightOn  \cite{brossollet2021lighton} \\

% Akida~\cite{ivanov2022neuromorphic}, Speck~\cite{richter2023speck} & PCM \cite{nandakumar2018phase}, PUMA \cite{ankit2019puma} & PIC \cite{ashtiani2022chip}, QRC \cite{martinez2023quantum} \\

% T1 \cite{vogginger2024neuromorphic}, NeuronFlow \cite{moreira2020neuronflow} & Loihi 2 \cite{orchard2021efficient}, Prime \cite{chi2016prime} & TTN-VQC \cite{qi2023theoretical} \\

% \bottomrule[2pt]
% \end{tabular}
% }
% \vspace{-1em}
% \end{table*}

\begin{table*}[t!]
\caption{Representative neuromorphic hardware (Sec. \ref{Neuromorphic Hardware}) for embodied agent. EP is the abbreviation for Emerging Physical.}
\vspace{-1em}
\renewcommand{\arraystretch}{1.5}
\centering
\resizebox{1\linewidth}{!}{
\label{tab:NeuromorphicHardware}
\begin{tabular}{cccc}
\toprule[2pt]
\textbf{Category} & \textbf{Platform} & \textbf{Institution (Year)} & \textbf{Key Features \& Strengths} \\
\midrule

\multirow{6}{*}{\rotatebox{90}{\scriptsize \shortstack{\textbf{(a) Neuromorphic} \\ \textbf{Chips}}}} & Loihi~\cite{davies2018loihi}        & Intel (2018)         & On-chip learning, event-driven SNN, Lava ecosystem \\
& NeuronFlow~\cite{moreira2020neuronflow}   & GrAI Matter Labs (2020) & Local adaptive learning, energy-efficient motor control \\
& Loihi 2~\cite{orchard2021efficient}      & Intel (2021)        & Finer-grained spiking, flexible plasticity, improved scalability \\
& BrainScaleS-2~\cite{pehle2022brainscales} & Heidelberg U. (2022) & Analog neuron circuits, accelerated SNN emulation, hybrid learning \\
& Akida~\cite{ivanov2022neuromorphic}        & BrainChip (2022)    & Ultra-low power edge inference, CNN-to-SNN support \\
& Speck~\cite{richter2023speck}        & SynSense (2023)     & Visual event-based SNN, compact mobile vision, micro-robotics integration \\

\midrule
\multirow{5}{*}{\rotatebox{90}{\scriptsize \shortstack{\textbf{(b) Memristor} \\ \textbf{-Based Systems}}}}  & PRIME~\cite{chi2016prime}        & Chinese Academy of Sciences (2016) & In-memory computing with RRAM crossbar arrays \\
& NeuroSim~\cite{chen2018neurosim}     & Stanford (2018)     & Circuit-level simulation for memristor-CNN/SNN co-design \\
& PCM~\cite{nandakumar2018phase}       & IBM Research (2018) & Multi-level conductance, phase-change material, analog MAC operation \\
& PUMA~\cite{ankit2019puma}         & UT Austin (2019)     & Programmable analog-digital hybrid RRAM accelerator \\
& ODIN~\cite{shivanandamurthy2021odin}         & CEA-Leti (2021)     & Spiking RRAM, online Hebbian/STDP learning \\

\midrule
\multirow{4}{*}{\rotatebox{90}{\scriptsize \shortstack{\textbf{(c) EP} \\ \textbf{Architectures}}}}  & LightOn OPU~\cite{brossollet2021lighton}  & LightOn (2021)      & Optical random projections, real-time NLP inference \\
& Photonic IC (PIC)~\cite{ashtiani2022chip} & Princeton/MIT (2022) & Silicon photonics for MAC ops at GHz scale \\
& QRC~\cite{martinez2023quantum}  & RIKEN (2023)         & Quantum chaos for time-series learning and control \\
& TTN-VQC~\cite{qi2023theoretical}      & Tsinghua U. (2023)   & Variational quantum circuits for neuromorphic logic \\

\bottomrule[2pt]
\end{tabular}
}
\vspace{-1em}
\end{table*}

\par \textbf{(b) Memristor-Based Systems:}
The memristors are nanoscale resistive switching devices that change their conductance in response to electrical stimuli, making them highly suitable for emulating synaptic plasticity in neuromorphic systems~\cite{kwon2022synaptic}. As a core enabler of in-memory computing, memristors co-locate data storage and analog computation, thereby reducing the energy and latency costs associated with traditional von Neumann memory hierarchies~\cite{mehonic2020memristors}.
Unlike conventional memory technologies such as SRAM and DRAM, memristors exhibit non-volatility, high integration density, and analog tunability. Their operation is based on ion or vacancy migration in active materials, which modulates device resistance and enables multi-level conductance states~\cite{du2021synaptic}. This behavior is analogous to biological synaptic strength modulation and supports essential neural primitives such as analog multiply-accumulate (MAC) operations and spike-based learning.

Memristor crossbar arrays implement vector-matrix multiplication directly in the memory plane, enabling massively parallel and energy-efficient neural computation~\cite{xiao2023review}. This architecture supports local, unsupervised learning mechanisms such as spike-timing-dependent plasticity and Hebbian learning. 
% Among various device types, resistive RAM (RRAM), phase-change RAM (PRAM), and magnetic RAM (MRAM) are particularly promising. RRAM offers simple structure and analog behavior; PRAM supports multilevel storage via phase transitions; MRAM combines speed with endurance using magnetic tunnel junctions.
In practical neuromorphic platforms, memristors serve both as dense, non-volatile weight storage and as analog computing elements for spiking and non-spiking models~\cite{du2021synaptic}. ODIN employs embedded RRAM for spike-driven local learning; DynapCNN integrates memristive elements into convolutional accelerators. Hybrid architectures such as Intel's Loihi 2 are exploring memristor integration to enhance on-chip plasticity and energy efficiency.
Although challenges such as device variability, limited endurance, and conductance precision remain, continued advances in materials, fabrication techniques, and peripheral circuit design are gradually improving the reliability and scalability of memristive systems~\cite{park2022experimental}. These devices provide a biologically inspired and hardware-efficient foundation for future neuromorphic architectures, especially in edge applications that require compactness, low power, and adaptability.

\par \textbf{(c) Emerging Physical Architectures:}
Beyond traditional electronic substrates, neuromorphic computing is increasingly exploring alternative physical platforms that offer novel trade-offs in latency, scalability, and energy efficiency. Two promising directions are photonic neuromorphics~\cite{prucnal2017neuromorphic} and quantum neuromorphics~\cite{ajagekar2022quantum}, which aim to exploit the unique properties of light and quantum mechanics for neural computation.

Photonic neuromorphic systems leverage the inherent parallelism and high-speed nature of light to achieve ultra-low-latency processing with minimal thermal dissipation~\cite{shastri2021photonics}. Optical components such as microring resonators, Mach-Zehnder interferometers, and photonic waveguides have been utilized to construct neurons and synapses operating at gigahertz speeds~\cite{goi2020perspective}. These architectures can perform essential neural operations such as matrix-vector multiplication by encoding information in the intensity or phase of light, achieving significant acceleration in applications like image recognition and high-frequency sensor processing. 
% Moreover, recent research has demonstrated on-chip photonic networks that support spiking dynamics and event-based computation, making them compatible with SNN models.
Quantum neuromorphic computing represents a more speculative but potentially transformative approach~\cite{sangwan2020neuromorphic}. By exploiting quantum properties such as superposition and entanglement, quantum neural networks can theoretically encode and process exponentially large state spaces. This capability is particularly relevant for optimization, probabilistic inference, and reinforcement learning tasks that involve high-dimensional search spaces~\cite{pershin2011neuromorphic}. 
% Recent developments have introduced variational quantum circuits inspired by neural architectures, aiming to implement learning algorithms directly within quantum hardware. 
While large-scale, practical quantum neuromorphic systems are still in their infancy, ongoing advances in quantum coherence, error correction, and qubit scalability are steadily improving their feasibility.

Both photonic and quantum neuromorphic systems remain largely in the experimental stage, but they demonstrate the potential of alternative physical substrates to expand the capabilities of embodied intelligence~\cite{ferreira2017progress,hoffmann2022quantum}. Their inherent properties, such as high processing speed, large bandwidth, and scalability, make them promising options for future systems that require real-time operation in complex, data-intensive environments while maintaining strict energy efficiency. 

\begin{table*}[t!]
\caption{Representative software frameworks (Sec. \ref{Software Frameworks for Neural Brains}) for embodied agents.}
\vspace{-1em}
\renewcommand{\arraystretch}{1.5}
\centering
\resizebox{1\linewidth}{!}{
\label{tab:SoftwareFrameworks}
\begin{tabular}{cccc}
\toprule[2pt]
\textbf{Category} & \textbf{Framework} & \textbf{Institution (Year)} & \textbf{Key Features \& Strengths} \\
\midrule
\multirow{9}{*}{\rotatebox{90}{\scriptsize \shortstack{\textbf{(a) Deep Learning Frameworks}}}}  & R2R~\cite{fried2018speaker}         & Stanford (2018) & Vision-language navigation benchmark for instruction-following \\
& CLIP~\cite{guzhov2022audioclip}       & OpenAI (2021) & Vision-language model, zero-shot image-text alignment \\
& VLN-BERT~\cite{hong2021vln}     & CUHK (2021) & Transformer-based model for navigation with natural language \\
& BLIP-2~\cite{li2023blip}       & Salesforce (2023) & Vision-language reasoning with pre-trained image encoders \\
& GPT-4~\cite{achiam2023gpt}     & OpenAI (2023) & Multimodal language model, capable of reasoning and generation \\
& PaLM~\cite{anil2023palm}       & Google (2023) & Scalable language model, few-shot and multilingual capabilities \\
& Qwen~\cite{bai2023qwentechnicalreport}     & Alibaba DAMO (2023) & Open-source Chinese-centric LLM, alignment-focused \\
& RT-2~\cite{brohan2023rt}         & Google DeepMind (2023) & Language-conditioned robot policy via vision-language model \\
& Deepseek~\cite{liu2024deepseek} & DeepSeek AI (2024) & Multimodal foundation model, supports code and language tasks \\

\midrule
\multirow{5}{*}{\rotatebox{90}{\scriptsize \shortstack{\textbf{(b) Neuromorphic} \\ \textbf{Libraries}}}}  &  NEST~\cite{dupuy1990nest}        & INCF (1990-) & Scalable simulator for large spiking neural networks \\
& BindsNET~\cite{hazan2018bindsnet} & Columbia University (2018) & Spiking network simulation with reinforcement learning support \\
& Brian2~\cite{stimberg2019brian}  & Neuroinformatics UK (2019) & Intuitive simulator for prototyping SNNs in Python \\
& SpiNNaker~\cite{furber2020spinnaker} & University of Manchester (2020) & Distributed SNN execution on custom neuromorphic hardware \\
& Norse~\cite{pehle2021norse}      & TU Graz (2021) & SNN computation library built on PyTorch \\

\midrule
\multirow{5}{*}{\rotatebox{90}{\scriptsize \shortstack{\textbf{(c) Edge Computing} \\ \textbf{Frameworks}}}}  &  TensorRT~\cite{davoodi2019tensorrt}        & NVIDIA (2019) & Optimized inference engine for low-latency deep learning on GPUs \\
& ONNX Runtime~\cite{jin2020compiling}       & Microsoft (2020) & Interoperable runtime for accelerated inference across platforms \\
& Tengine~\cite{benoit2020impact}            & OPEN AI Lab (2020) & Lightweight inference engine for mobile and IoT devices \\
& PyTorch Mobile~\cite{luo2020comparison}    & Meta AI (2020) & Optimized runtime for deploying PyTorch models to edge hardware \\
& TensorFlow Lite~\cite{david2021tensorflow} & Google (2021) & Lightweight deployment framework for embedded and mobile devices \\

\bottomrule[2pt]
\end{tabular}
}
\vspace{-1em}
\end{table*}

\subsubsection{Software Frameworks for Neural Brains}\label{Software Frameworks for Neural Brains}

\par The implementation of Neural Brain architectures demands a comprehensive software stack that supports the full lifecycle of intelligent systems. From training and simulation to real-time deployment and adaptive interaction, these frameworks play a central role in translating neural principles into practical, embodied intelligence~\cite{zhang2020system,mehonic2022brain}. 
Unlike traditional applications where computation is centralized and static, Neural Brain systems may operate in dynamic, uncertain environments and continuously integrate multimodal sensory input, learn from interaction, and produce behavior in real time. 
A successful framework should offer flexibility across hardware, scalability for increasing task complexity, and responsiveness to environmental change, while also facilitating modular design for perception, decision-making, and control~\cite{remmelzwaal2021brain}. In this context, software is not a passive layer but a central enabler of embodied cognition, responsible for coordinating computation across tightly coupled physical and neural domains. Some representative works are summarized in Table \ref{tab:SoftwareFrameworks}.

% \par Deep Learning Frameworks: PyTorch and TensorFlow for training large-scale models.
 % Model Training and Inference
\par \textbf{(a) Deep Learning Frameworks: }

Deep learning frameworks form the foundation of modern artificial neural network development, enabling efficient model design, training, and deployment at scale~\cite{richards2019deep}. These frameworks abstract low-level operations and provide modular components that facilitate rapid experimentation and integration across diverse applications~\cite{nguyen2019machine}. In the context of Neural Brain systems and embodied agents, they support a wide range of functionalities including perception, decision-making, motor control, and high-level reasoning.

PyTorch and TensorFlow are the two most prominent platforms for implementing artificial neural networks~\cite{novac2022analysis}. PyTorch emphasizes dynamic computation graphs and a Pythonic interface that is well suited for research and rapid prototyping. TensorFlow, with its static graph model and optimized runtime, excels in scalable training and production deployment. Both frameworks support GPU and TPU acceleration, automatic differentiation, and a variety of network architectures including CNNs, RNNs, and transformers~\cite{alomar2024rnns}.
These frameworks are not limited to supervised learning. They offer extensive capabilities for unsupervised and self-supervised learning, such as contrastive learning, autoencoders, and generative models~\cite{chen2022semi}. In reinforcement learning contexts, extensions like TensorFlow Agents, Keras-RL, and TorchRL provide modular implementations of popular algorithms including DQN, PPO, and A3C~\cite{akalin2021reinforcement}. These tools enable agents to learn from interaction and adapt to complex environments, making them especially relevant for embodied applications~\cite{schmidhuber2015deep}.

Recent advances in large-scale foundation models have further extended the capabilities of deep learning frameworks, establishing critical links between perception, language, and action in embodied systems~\cite{wang2023voyager}. LLMs such as GPT, PaLM, and LLaMA are developed and fine-tuned using deep learning libraries to generate human-like text, reason over symbolic information, and support high-level cognitive functions in embodied agents~\cite{huang2022inner}. These models can be grounded in sensorimotor experiences through fine-tuning on interaction data or through in-context learning with multimodal prompts.
 VLMs, including CLIP, Flamingo, and BLIP, enable agents to align visual perception with linguistic understanding~\cite{li2023blip}. Trained on large-scale image-text pairs, these models allow embodied systems to interpret visual scenes, follow instructions, and perform zero-shot generalization across tasks. In practical settings, VLMs are integrated with robotic frameworks for applications such as visual navigation, object retrieval, and human-robot interaction.

Further integrating vision and action, Vision-Language Navigation (VLN) models and Visual Language Actuators (VLAs) support end-to-end pipelines where agents follow natural language commands to navigate or manipulate their environment~\cite{ma2024survey}. 
% These models are typically trained using datasets that simulate real-world environments and multi-step instructions, and they rely on deep learning frameworks for multimodal data handling, memory encoding, and policy optimization.
Another emerging paradigm is the Vision Transformer (ViT) and its descendants such as DiT and BEiT, which provide scalable architectures for unified image, video, and sensor processing~\cite{peebles2023scalable}. 
% These transformer-based models have been adapted for robotics, where they serve as backbones for semantic understanding, embodied policy learning, and cross-modal attention.
Together, these foundation models demonstrate the evolving role of deep learning frameworks not only in training neural networks, but also in integrating high-level cognition with low-level sensorimotor loops, thus bridging the gap between abstract reasoning and embodied intelligence.

Deep learning frameworks are foundational to enabling embodied agents to learn generalizable skills across simulation and real-world environments. Their ability to facilitate the development, training, and transfer of neural models is key to building adaptable systems capable of robust perception and decision-making in complex settings~\cite{de2022next}.
These frameworks integrate seamlessly with high-fidelity simulation platforms such as Habitat, AI2-THOR, and Isaac Gym. Within these virtual environments, agents can be trained at scale under diverse and controllable conditions~\cite{li2022metadrive,makoviychuk2021isaac}. Models learn to perform tasks involving perception-action policies, spatial navigation, semantic understanding, and object manipulation, often in scenarios that would be difficult or costly to replicate in physical environments~\cite{mu2023embodiedgpt}. 
% By leveraging large-scale datasets and domain randomization techniques, agents trained in simulation can acquire transferable skills that apply to unseen real-world situations.
% Moreover, modern deep learning frameworks are designed with modular architectures that support the independent development of perception, reasoning, and control components. These modules can be pre-trained in different environments and later integrated into unified systems, enhancing flexibility and scalability.

In summary, deep learning frameworks provide the computational backbone for building Neural Brain systems that span from abstract reasoning to low-level sensorimotor control. They unify high-level cognitive models such as language and vision transformers with grounded interaction policies required for embodied agents. By enabling scalable learning, generalization across domains, and integration with simulation and hardware, these frameworks serve as a critical bridge between artificial neural computation and the demands of real-world intelligence.

% \par Neuromorphic Libraries: NEST and Brian2 for simulating SNNs.
\par \textbf{(b) Neuromorphic Libraries: }  
Neuromorphic libraries are essential software infrastructures that enable the simulation and prototyping of SNNs, which are biologically inspired computational models mimicking the structure and dynamics of real nervous systems~\cite{yik2025neurobench}. These libraries provide researchers and engineers with the tools necessary to investigate neural computation across multiple scales, ranging from single-neuron activity to the behavior of complex cortical circuits. 
% By supporting the design, implementation, and experimentation of brain-like systems, neuromorphic libraries bridge the gap between neuroscience and artificial intelligence.

Fundamentally, neuromorphic libraries are built upon the principles of event-driven computation. Unlike traditional artificial neural networks, which rely on continuous-valued activations and synchronous updates, spiking models simulate neurons that communicate via discrete spikes~\cite{ivanov2022neuromorphic}. A neuron emits a spike only when its membrane potential surpasses a threshold, resulting in sparse, asynchronous, and temporally accurate activity. This mirrors the way biological neurons operate and supports more energy-efficient and scalable information processing~\cite{sandamirskaya2022neuromorphic}. In these models, time is treated as a first-class variable, and information is encoded through spike timing, firing rate, or complex temporal patterns instead of scalar outputs.

To support such dynamics, neuromorphic libraries offer a range of neuron models including leaky integrate-and-fire (LIF), adaptive exponential integrate-and-fire (AdEx), and conductance-based Hodgkin-Huxley formulations~\cite{teeter2018generalized}. These models vary in biological realism and computational cost, allowing users to tailor simulations to specific use cases. Additionally, the libraries support detailed definitions of synaptic connectivity and plasticity mechanisms such as short-term synaptic dynamics, long-term potentiation and depression, and spike-timing-dependent plasticity (STDP). These features enable the modeling of phenomena such as learning, memory consolidation, synchronization, and emergent behavior in neural populations. 

Key simulation libraries include NEST, Brian2, and SpiNNaker, which support large-scale SNN modeling with customizable neuron models, synaptic dynamics, and learning rules~\cite{kulkarni2021benchmarking}. NEST is optimized for simulating networks with high structural complexity and supports parallel processing, making it suitable for studying cortical microcircuits and network-level brain dynamics. Brian2 offers an intuitive Python interface and is well-suited for rapid model prototyping, especially in academic and research contexts~\cite{stimberg2019brian}. SpiNNaker, a neuromorphic platform developed by the University of Manchester, includes both software and hardware components for simulating real-time spiking activity on specialized multi-core processors~\cite{furber2020spinnaker}.
To bridge the gap between spiking and deep learning paradigms, hybrid libraries such as BindsNET and Norse have emerged~\cite{pehle2021norse}. These tools integrate with PyTorch, enabling researchers to incorporate spiking neuron models and surrogate gradient methods into GPU-accelerated pipelines. 
% This compatibility allows for the training of SNNs using backpropagation-like approaches, facilitating experimentation with biologically plausible alternatives to traditional ANNs.

Neuromorphic libraries not only support efficient simulation but also contribute to the broader objective of constructing Neural Brain architectures for embodied agents~\cite{huynh2022implementing}. By providing biologically grounded models of neurons and synapses, these libraries enable the construction of neural control systems that reflect the structural and functional principles of the human brain. This capability is particularly important for applications that require continuous sensorimotor integration and real-time feedback, such as autonomous navigation, interactive manipulation, and multimodal perception~\cite{wunderlich2019demonstrating}. 
% Spiking neural networks developed using these libraries can form closed-loop perception-action cycles, where dynamic environmental inputs directly influence decision-making and motor output.
The modular design of neuromorphic libraries further supports the integration of diverse sensory channels, including vision, touch, and proprioception. This facilitates adaptive behavior in agents that must respond flexibly to changing environments. By aligning temporal coding, event-driven computation, and synaptic plasticity, these libraries enable neural models that are not only computationally efficient but also structurally compatible with embodied cognition. 
Through this convergence of biological realism and engineering practicality, neuromorphic libraries serve as a critical interface between neuroscience-inspired theory and real-world artificial intelligence systems.

% \par Edge Computing: Federated learning (e.g., NVIDIA Clara) for decentralized, privacy-preserving computation.
\par \textbf{(c) Edge Computing Frameworks: }  
Edge computing frameworks are increasingly vital for deploying intelligent systems in real-world settings where low latency, energy efficiency, and privacy are critical. Unlike centralized cloud-based approaches, edge computing distributes computation closer to the data source, enabling embodied agents to process sensory input, perform inference, and adapt behavior in real time~\cite{villar2023edge}. This decentralized architecture is particularly suited for neuromorphic systems, which often operate in power-constrained and dynamic environments such as mobile robotics, autonomous vehicles, wearable devices, and brain-computer interfaces.

At the software level, edge computing frameworks must manage resource-aware scheduling, model partitioning, and on-device inference across heterogeneous platforms including CPUs, GPUs, FPGAs, and specialized neuromorphic chips~\cite{thomas2009comparison}. Modern toolkits such as TensorFlow Lite, PyTorch Mobile, and ONNX Runtime enable neural models to be compressed, quantized, and deployed on embedded hardware with minimal overhead~\cite{kim2022extending}. These platforms support real-time execution of deep learning models for tasks such as image classification, object tracking, speech recognition, and decision-making in sensor-rich contexts.
For more advanced scenarios, edge intelligence requires adaptive learning mechanisms that allow models to evolve post-deployment. Frameworks such as NVIDIA Jetson and Intel OpenVINO support incremental learning and hardware-aware optimization, while federated learning toolkits such as PySyft and Flower facilitate decentralized training across distributed agents. This paradigm ensures data privacy and reduces bandwidth usage by performing local updates and aggregating global models without centralized data transfer.

In the domain of neuromorphic computing, runtime environments like Intel Lava and Nengo are designed to execute event-driven spiking neural networks directly on low-power hardware~\cite{seekings2024towards}. These frameworks manage time-sensitive updates, asynchronous communication, and local plasticity rules, enabling real-time, biologically inspired computation at the edge. Combined with event-based sensors such as dynamic vision sensors (DVS), these platforms allow for efficient perception-action loops in robotics and embedded agents.

For embodied intelligence, edge computing is essential to bridging the gap between local sensorimotor feedback and global cognitive functions~\cite{krishnasamy2020edge}. By enabling fast, context-aware decision-making and continuous learning in the field, edge frameworks support robust performance in environments where cloud connectivity is limited or unavailable~\cite{ma2022neuromorphic}. The modularity and interoperability of edge platforms also make them ideal for hybrid systems that integrate deep learning with neuromorphic controllers or real-time biological interfaces.
% Beyond its technical utility, edge computing represents a paradigm shift in the design of Neural Brain systems by relocating key cognitive and control functions to the periphery of intelligent agents. This shift allows for real-time adaptation, contextual awareness, and autonomous decision-making at the site of perception and action, which are hallmarks of biological cognition. 
In Neural Brain architectures, especially those combining deep learning and neuromorphic components, edge computing enables the co-location of sensory processing, inference, and motor control within a unified, low-power system. This integration is essential for closing the loop between perception and behavior, supporting continuous learning and dynamic interaction with the environment. 
% As such, edge computing is not merely a deployment strategy, but a foundational element in building embodied, context-sensitive intelligence that mirrors the distributed and decentralized processing found in biological nervous systems.

% \begin{table*}[t!]
% \caption{Representative energy-efficient strategies (Sec. \ref{Energy-efficient Strategies}) at the edge.}
% \vspace{-1em}
% \renewcommand{\arraystretch}{1.25}
% \centering
% \resizebox{0.9\linewidth}{!}{
% \label{tab:Energy-Efficient}
% \begin{tabular}{ccc}
% \toprule[2pt]
% \multicolumn{3}{c}{\textbf{Energy-Efficient Strategies}} \\
% \hline
% Sparse Computation & Quantization and Compression & Hardware-Software Co-Design \\
% \hline
% Srinivas \emph{et al.} \cite{srinivas2017training}, Liu \emph{et al.} \cite{liu2015sparse}, Zambrano \emph{et al.} \cite{zambrano2019sparse} & Li \emph{et al.} \cite{li2022quantization}, Rathi \emph{et al.} \cite{rathi2018stdp} & Eyeriss \cite{chen2016eyeriss}, TVM \cite{chen2018tvm}  \\

% Sbnet \cite{ren2018sbnet}, GLaM \cite{du2022glam}, Switch Transformer \cite{fedus2022switch} & Yang \emph{et al.} \cite{yang2019quantization}, Jacob \emph{et al.} \cite{jacob2018quantization} & TrueNorth \cite{merolla2014million}, DNNBuilder \cite{zhang2018dnnbuilder} \\

% BASE Layers \cite{lewis2021base}, BigBird \cite{zaheer2020big}, SCNN \cite{parashar2017scnn} & Hinton \emph{et al.} \cite{hinton2015distilling}, Liu \emph{et al.} \cite{liu2021efficient}  & SECDA \cite{haris2021secda}, Maeri \cite{kwon2018maeri} \\

% \bottomrule[2pt]
% \end{tabular}
% }
% \vspace{-1em}
% \end{table*}

\begin{table*}[t!]
\caption{Representative energy-efficient strategies (Sec. \ref{Energy-efficient Strategies}) at the edge.}
\vspace{-1em}
\renewcommand{\arraystretch}{1.5}
\centering
\resizebox{1\linewidth}{!}{
\label{tab:EnergyEfficientStrategies}
\begin{tabular}{cccc}
\toprule[2pt]
\textbf{Category}  &  \textbf{Method} & \textbf{Institution (Year)} & \textbf{Key Features \& Strengths} \\
\midrule
\multirow{9}{*}{\rotatebox{90}{\scriptsize \shortstack{\textbf{(a) Sparse Computation}}}}  &  Liu \emph{et al.}~\cite{liu2015sparse} & Intel (2015) & Explored hardware-aware sparsity for energy reduction \\
& Srinivas \emph{et al.}~\cite{srinivas2017training} & IISc Bangalore (2017) & Pruning via parameter prediction during training \\
& SCNN~\cite{parashar2017scnn} & Princeton (2017) & Sparse CNN accelerator exploiting activation and weight sparsity \\
& Zambrano \emph{et al.}~\cite{zambrano2019sparse} & University of Chile (2019) & Sparse SNN training with regularization methods \\
& Sbnet~\cite{ren2018sbnet} & NVIDIA (2018) & Block-wise sparse computation for high-speed CNNs \\
& BigBird~\cite{zaheer2020big} & Google Research (2020) & Long-sequence transformer with sparse attention \\
& GLaM~\cite{du2022glam} & Google (2022) & Scalable MoE architecture using sparse experts \\
& Switch Transformer~\cite{fedus2022switch} & Google Brain (2022) & Sparse routing among transformer blocks for efficiency \\
& BASE Layers~\cite{lewis2021base} & Google Research (2021) & Binary sparse attention for scalable transformer models \\

\midrule
\multirow{6}{*}{\rotatebox{90}{\scriptsize \shortstack{\textbf{(b) Quantization and} \\ \textbf{Compression}}}}  &  Hinton \emph{et al.}~\cite{hinton2015distilling} & University of Toronto (2015) & Introduced knowledge distillation to compress large models \\
& Jacob \emph{et al.}~\cite{jacob2018quantization} & Google (2018) & Proposed quantization-aware training for inference efficiency \\
& Rathi \emph{et al.}~\cite{rathi2018stdp} & IIT Delhi (2018) & STDP-based weight quantization for spiking models \\
& Yang \emph{et al.}~\cite{yang2019quantization} & UCLA (2019) & Post-training quantization for deep networks on edge devices \\
& Li \emph{et al.}~\cite{li2022quantization} & Alibaba DAMO (2022) & Survey on recent advances in neural quantization \\
& Liu \emph{et al.}~\cite{liu2021efficient} & Zhejiang University (2021) & Efficient model compression techniques for embedded AI \\

\midrule
\multirow{6}{*}{\rotatebox{90}{\scriptsize \shortstack{\textbf{(c) Hardware-Software} \\ \textbf{Co-Design}}}}  & TrueNorth~\cite{merolla2014million} & IBM Research (2014) & Early neuromorphic chip supporting event-driven SNNs \\
& Eyeriss~\cite{chen2016eyeriss} & MIT (2016) & Energy-efficient DNN accelerator with data reuse-aware architecture \\
& Maeri~\cite{kwon2018maeri} & UIUC (2018) & Flexible dataflow mapping architecture for DNNs \\
& DNNBuilder~\cite{zhang2018dnnbuilder} & NTU Singapore (2018) & Hardware-software codesign tool for FPGA-based deployment \\
& TVM~\cite{chen2018tvm} & UW + Amazon (2018) & End-to-end deep learning compiler stack for CPU/GPU/edge targets \\
& SECDA~\cite{haris2021secda} & University of Toronto (2021) & Co-design automation for deploying CNNs to custom accelerators \\

\bottomrule[2pt]
\end{tabular}
}
\vspace{-1em}
\end{table*}

\subsubsection{Energy-Efficient Learning at the Edge}\label{Energy-efficient Strategies}

As Neural Brain systems increasingly migrate from controlled lab environments to real-world deployment, the demand for energy-efficient learning at the edge becomes not only a practical concern but a foundational requirement~\cite{daghero2021energy}. Neural architectures inspired by the human brain must operate under constraints similar to those faced by biological systems, including limited energy availability, decentralized processing, and the need for real-time adaptation~\cite{mehlin2023towards}. In embodied agents, which must continuously sense, reason, and act in dynamic environments, energy-efficient learning ensures long-term autonomy, responsiveness, and operational sustainability. Some representative works are shown in Table \ref{tab:EnergyEfficientStrategies}.

% Unlike cloud-centric models that benefit from virtually unlimited resources, Neural Brains embedded in edge devices face stringent limitations in computation, memory, and power. These constraints are especially pronounced in mobile robots, wearable interfaces, implantable devices, and distributed sensor networks where battery life and thermal budgets are critical bottlenecks. Energy-efficient learning enables such agents to locally process sensory data, update models incrementally, and adapt to environmental changes without relying on energy-intensive communication with remote servers.
% Furthermore, integrating energy-aware learning algorithms into Neural Brain systems enhances scalability and robustness. It allows for distributed intelligence across a swarm of agents, each capable of contributing to collective behavior while managing its own energy footprint. This is vital for tasks such as autonomous exploration, environmental monitoring, and cooperative manipulation. By supporting real-time, context-sensitive, and power-conscious computation, energy-efficient learning at the edge bridges the gap between artificial neural computation and the physical demands of embodied intelligence, laying the groundwork for a new generation of sustainable and responsive intelligent systems.

% \par Sparse Computation: Mixture-of-Experts (MoE) models (e.g., Google's GLaM) reduce computational load.
\par \textbf{(a) Sparse Computation:}
Sparse computation is a foundational strategy for achieving energy-efficient learning at the edge, inspired both by biological neural systems and recent advances in artificial intelligence~\cite{bai2024beyond}. In biological brains, neural activity is typically sparse, with only a small subset of neurons firing at any given moment. This sparse coding not only minimizes metabolic energy expenditure but also improves representational efficiency and reduces signal interference. Inspired by these principles, artificial neural architectures have increasingly adopted sparsity as a design feature to reduce computational overhead, lower memory consumption, and increase robustness in real-world environments~\cite{kachris2025survey}.

One of the most advanced implementations of sparse computation is found in Mixture-of-Experts (MoE) models~\cite{liu2024survey}. These architectures consist of a large pool of expert subnetworks, with a learned gating mechanism that activates only a small number of relevant experts for each input. This conditional computation allows the model to maintain high capacity and representational power while significantly lowering the number of operations per inference. MoE models, such as Google's GLaM and Switch Transformer, have shown that it is possible to scale up to billions of parameters without proportionally increasing compute cost, as only a fraction of the network is active at any given time~\cite{zhao2018deep}. This enables high-performance learning and inference with substantially lower energy usage compared to dense architectures.
The advantages of MoE models are particularly valuable in edge-deployed Neural Brain systems for embodied agents. In these scenarios, devices often operate with tight energy budgets and limited hardware resources~\cite{yi2023edgemoe}. Sparse activation allows the agent to dynamically adapt its computational effort to the current sensory context, selectively engaging the parts of its neural model that are most relevant to the task~\cite{li2025uni}. This capability is essential for real-time, context-aware behavior in environments where inputs are heterogeneous, noisy, or multimodal. For example, an autonomous robot may rely on different expert pathways for visual navigation, object recognition, or auditory commands, all orchestrated efficiently within a sparse MoE framework.

% Moveover, sparse computation encompasses a broader set of techniques aimed at minimizing the number of active parameters and operations. Structured pruning removes redundant weights or entire neurons from the model, while dynamic routing directs data through only a subset of layers based on input-dependent criteria. Activation masking disables non-contributing units at runtime, and low-rank factorization decomposes weight matrices to reduce complexity. These techniques have been successfully employed in mobile deep learning applications, compact vision models, and spiking neural network implementations, where energy efficiency and responsiveness are paramount.

Sparse computation is also supported by specialized hardware and software infrastructures~\cite{liu2017unified}. Accelerators such as custom ASICs and neuromorphic processors are designed to exploit data sparsity by skipping inactive operations, reducing memory access, and enhancing throughput~\cite{ren2018sbnet}. Sparse tensor libraries further optimize execution by compressing data representations and aligning computation with available hardware bandwidth. These advances make it feasible to deploy complex Neural Brain models at the edge while maintaining real-time performance and minimizing power consumption.

As a convergence of neuroscience principles and machine learning innovation, sparse computation offers a compelling foundation for building scalable and efficient intelligent systems. By leveraging conditional activation, model compression, and hardware-aware optimization, this approach enables Neural Brain architectures to operate effectively under limited computational and energy budgets. 
% It empowers embodied agents to exhibit adaptive, context-aware behaviors in dynamic environments, ensuring responsiveness and autonomy without relying on resource-heavy infrastructures. In doing so, sparse computation lays the groundwork for energy-efficient Neural Brains that closely mirror the operational strategies of biological systems.

% \par Quantization and Compression: Low-precision arithmetic and model pruning for efficient deployment on edge devices.
\par \textbf{(b) Quantization and Compression:}
Quantization and compression are also the key strategies for enabling energy-efficient deployment of Neural Brain models at the edge. These techniques aim to reduce the size and computational complexity of neural networks while preserving accuracy and functional integrity~\cite{li2022quantization}. By minimizing memory footprint, lowering arithmetic precision, and reducing the number of active parameters, quantization and compression allow intelligent agents to perform real-time inference and continual learning on devices with limited hardware resources and strict power constraints.

Quantization refers to the process of reducing the precision of numerical representations used in neural networks~\cite{rathi2018stdp}. Instead of using 32-bit floating-point numbers, models can be converted to use 16-bit, 8-bit, or even binary formats. This not only reduces the size of the model but also allows faster and more energy-efficient computation, as lower-precision arithmetic requires fewer logic gates and memory accesses~\cite{castagnetti2023trainable}. Quantized models benefit from acceleration on edge-oriented hardware platforms, such as mobile AI processors, FPGAs, and custom ASICs that are optimized for low-bit operations. Techniques like post-training quantization and quantization-aware training further ensure that model accuracy is preserved despite the reduced precision.

Compression encompasses a broader set of model optimization techniques that reduce redundancy and improve computational efficiency~\cite{liu2021efficient}. Pruning is a widely used compression method that removes weights or neurons that contribute little to the model's output. Structured pruning focuses on removing entire filters or layers to create a compact model structure that is easier to optimize and deploy~\cite{blalock2020state}. Other compression strategies include weight sharing, which reduces the number of unique weight values through parameter tying, and knowledge distillation, which trains a smaller "student" network to mimic the output of a larger, pre-trained "teacher" model. These methods collectively reduce the model's memory and compute demands while maintaining generalization capability.

For the Neural Brain systems in embodied agents, quantization and compression enable on-device inference and adaptive behavior without the need for constant offloading to the cloud~\cite{gholami2022survey}. This is crucial for agents operating in latency-sensitive or network-limited environments, such as autonomous vehicles, wearable medical systems, or drones. Quantized and compressed models can also be updated more frequently using edge-compatible training techniques, facilitating continual learning and context-aware adaptation in the field~\cite{reguero2025energy}.
Neuromorphic platforms further benefit from quantization and compression strategies. Event-driven computation in spiking neural networks naturally supports sparse, low-bit communication and storage. Hardware like Intel's Loihi or SynSense's Speck leverages these properties by supporting low-precision synaptic weights and local weight updates, reducing energy consumption and enhancing responsiveness. 
Compression techniques can be adapted to SNNs by pruning unnecessary synaptic connections or quantizing spike rates, thereby reducing energy consumption and improving real-time performance.

By tightly integrating algorithmic optimization with hardware efficiency, quantization and compression play a pivotal role in enabling compact and low-power Neural Brain systems~\cite{kuzmin2023pruning}.
% These techniques empower models to learn and adapt in real time while operating within the strict resource constraints of edge environments. 
Their ability to reduce computational cost and memory usage without compromising performance makes them essential for deploying intelligent agents that must act autonomously in dynamic, real-world settings. As embodied systems continue to scale in complexity and number, quantization and compression will remain foundational tools in realizing sustainable, energy-aware artificial intelligence.

% \par Hardware-Software Co-Design: Algorithm-hardware co-optimization for FPGA/ASIC implementations.
\par \textbf{(c) Hardware-Software Co-Design:}
Hardware-software co-design is an essential paradigm for achieving energy-efficient learning at the edge, involving the joint optimization of algorithmic models and the hardware architectures on which they run~\cite{de1997hardware}. Rather than treating hardware as a passive execution layer, co-design emphasizes a synergistic development process where software algorithms are tailored to hardware constraints and, conversely, hardware is customized to meet the computational and communication patterns of the software~\cite{varshika2023hardware}. This close alignment enables the deployment of intelligent systems that are both highly efficient and responsive in real-world environments.

In edge scenarios where energy budgets, processing throughput, and latency are tightly constrained, traditional one-size-fits-all hardware solutions are often inadequate~\cite{ponzina2023hardware}. Hardware-software co-design addresses this by enabling application-specific customization. For example, field-programmable gate arrays (FPGAs) can be configured to accelerate specific neural network operations, such as convolution, attention, or spike-based integration~\cite{haris2021secda}. When paired with tailored neural models that exploit sparsity, quantization, or temporal dynamics, these platforms can achieve orders of magnitude improvements in energy efficiency and performance.
% Neuromorphic processors such as Intel's Loihi and IBM's TrueNorth exemplify hardware-software co-design in action. These chips are built with architectural features that support the event-driven, parallel, and distributed nature of spiking neural networks. They work in tandem with software ecosystems like Lava or Nengo, which allow researchers to design neural models that map naturally onto the underlying hardware structure. This integration allows for localized plasticity, asynchronous spike communication, and low-power operation, making them suitable for continuous learning and perception-action loops in embodied agents.
Another important aspect of co-design is optimizing memory access patterns and dataflow. In many Neural Brain systems, data movement consumes more energy than computation itself. Co-design approaches seek to minimize this cost through techniques such as near-memory computing, hierarchical memory architectures, and pipelined data paths~\cite{bringmann2021automated}. Algorithms are adapted to these architectures using techniques like loop tiling, layer fusion, and workload partitioning. The result is a system that can maintain high throughput while operating within tight power envelopes.
% Furthermore, co-design facilitates rapid prototyping and deployment of neural models onto diverse edge platforms. Frameworks such as TVM, Xilinx Vitis AI, and TensorRT provide toolchains that automate the mapping of high-level models onto low-level hardware primitives, incorporating performance profiling and power estimation during the design phase. These tools support iterative optimization cycles, enabling developers to balance accuracy, latency, and energy consumption according to application-specific constraints.

For embodied agents operating in unstructured and dynamic environments, hardware-software co-design ensures that Neural Brain systems are not only computationally viable but also behaviorally adaptive and responsive. It enables the seamless integration of perception, learning, and actuation on edge hardware, reducing reliance on remote processing and enhancing autonomy. 

\subsection{Remarks and Discussions}\label{6.2}

% \subsubsection{Challenges in Hardware \& Software Co-Design}

Despite substantial progress in Neural Brain systems, the co-development of hardware and software continues to face several persistent challenges. 

\par \textbf{(1) Scalability:}  
One of the central issues is ensuring that neuromorphic hardware platforms can scale to support increasingly complex cognitive tasks~\cite{maguire2007challenges}. While current systems demonstrate efficiency in small-scale networks, extending these capabilities to large, hierarchical models that span perception, memory, and decision-making remains difficult~\cite{han2025progress}. Scalability challenges also arise in terms of communication bandwidth, memory hierarchy, and heat dissipation, all of which limit the integration of large-scale spiking or hybrid neural models.

\par \textbf{(2) Software Support:}  
Another key challenge lies in the lack of comprehensive software frameworks that can fully utilize the capabilities of emerging neuromorphic and edge-oriented hardware~\cite{ahmed2015review}. Many hardware platforms require highly specialized coding and configuration, which limits accessibility and slows down adoption. Bridging this gap requires the development of high-level programming environments, modular APIs, and compiler toolchains that abstract hardware details while preserving performance and flexibility.

\par \textbf{(3) Real-Time Constraints:}  
Balancing real-time responsiveness with energy efficiency is critical for embodied agents operating at the edge. Neural Brain systems must perform fast and reliable inference under tight latency constraints while maintaining low power consumption~\cite{sanyal2025real}. This trade-off is especially demanding when integrating continuous sensorimotor input, dynamic adaptation, and closed-loop feedback control. Meeting real-time constraints without sacrificing accuracy or adaptability is an open and pressing challenge.

\section{Summary and Future Directions}\label{section_Dissuasion_and_Further_Directions}

This final section summarizes the core insights of the Neural Brain framework and outlines key future directions for building more intelligent, adaptive, and embodied agents. Grounded in neuroscience, the Neural Brain integrates four essential components: multimodal sensing, perception-cognition-action loops, neuroplastic memory systems, and energy-efficient hardware/software co-design. While recent advances across these dimensions are promising, significant challenges remain in achieving human-level autonomy and generalization. We begin with a concise summary of the main takeaways, followed by an exploration of open problems and a forward-looking research roadmap.

\subsection{Summary of Key Insights}

The Neural Brain framework draws upon principles from neuroscience to enhance the cognitive capabilities of embodied agents. This section recaps the major design insights from previous chapters.

\begin{itemize} 
\item{The Neural Brain is a biologically inspired framework comprising four parts: multimodal active sensing, closed-loop perception-cognition-action function, neuroplasticity-driven memory systems, and energy-efficient neuromorphic hardware-software co-design.}

\item{The Neural Brain should integrate sensory inputs hierarchically, align and fuse them through attention for coherent perception, actively adjust sensing via predictive attention and learning, and continuously self-calibrate to maintain stability under noise and environmental changes.}

\item{The Neural Brain needs to refine actions via predictive feedback, separate high-level planning from low-level control, and adapt motor commands in real time for precise, flexible responses.}

\item{The Neural Brain should integrate causal inference, symbolic reasoning, and contextual memory for robust cognition, combine neural and symbolic processing to enhance reasoning under uncertainty, and continuously update internal models via neuroplasticity for long-term adaptability.}

\item{The Neural Brain should predict sensory outcomes and adjust actions to improve efficiency. It needs to align perception and action with task goals, prioritizing relevant cues. Contextual memory retrieval should link past experiences to current tasks, enhancing decision-making in dynamic environments.}

\item{The Neural Brain needs to use a hierarchical memory system with short-term memory for immediate decisions and long-term memory for learning and adaptation. It should consolidate important information, filter redundancy, and dynamically adjust memory based on relevance. An advanced retrieval system should activate relevant knowledge based on contextual cues, enhancing decision-making and adaptability.}

\item{The Neural Brain should use SNNs for efficient, event-driven computation and neuromorphic hardware for scalable, low-power processing. Hardware and software should be co-optimized for real-time inference. This is complemented by a distributed edge computing approach, enabling local processing with cloud synchronization for continuous adaptation.}

\end{itemize} 

% \par each sec one point. Current limitations include gaps in real-time adaptation, reasoning, and energy efficiency.

\subsection{Open Challenges and Future Directions}
Despite the rapid advancements in embodied agents, numerous challenges remain, as discussed in Sec.~\ref{sec3.2}, Sec.~\ref{4.2}, Sec.~\ref{5.2}, and Sec.~\ref{6.2}. Building upon these challenges, this part aims to analyze potential future research directions from the perspective of the four components of the Neural Brain.

\subsubsection{Sensing}
Future sensing research may focus on the following priorities to address the challenges detailed in Sec.~\ref{sec3.2} and build toward a robust Neural Brain for embodied agents.

\textbf{(1) Unified Multimodal Sensing Architectures.}
Future systems should strive for deeper integration of diverse sensory modalities-including underused channels such as olfaction, audio, temperature, and tactile perception-into a unified hierarchical sensing stack. This demands advances in time-synchronized data pipelines, spatial co-registration, and fusion.

\textbf{(2) Learning-Based Active Sensing Policies.}
Instead of fixed sensor behaviors, agents should learn to dynamically allocate sensory attention based on task objectives and environmental feedback. Reinforcement learning, predictive coding, or information-theoretic optimization can enable adaptive modulation of sampling frequency, resolution, field of view, and even sensor modality activation. Cross-modal attention mechanisms and goal-conditioned viewpoint planning-analogous to human gaze shifts or auditory focus-should be expanded to non-visual sensors as well.

\textbf{(3) Continuous Self-Calibration and Robustification.}
Neural Brain frameworks should embed real-time calibration mechanisms that are context-aware and continuously updated. This includes dynamic re-estimation of intrinsic and extrinsic parameters, online bias correction for IMUs, auto-tuning of olfactory sensitivity, and exposure/temperature compensation in vision. Probabilistic modeling, self-supervised error tracking, and confidence-based fusion pipelines can ensure perception remains stable over time. Long-term deployments, outdoor missions, and sensor degradation scenarios will particularly benefit from such resilience.

\par By systematically addressing these three directions, future embodied agents can more closely emulate the flexible, adaptive, and energy-efficient sensing capabilities that characterize biological intelligence. This foundation sets the stage for further enhancements in perception-cognition-action loops, memory storage, and hardware/software co-design within the Neural Brain framework.

\subsubsection{Perception-Cognition-Action}
% \par Perception: Achieving efficient multimodal fusion and active sensing.
% \par Cognition: Advancing counterfactual and causal reasoning.
To advance the development of comprehensive Perception-Cognition-Action systems, future research should focus on the following areas:

\textbf{(1) Unified Multimodal Representations:} Developing architectures capable of learning cohesive representations across diverse modalities, such as visual, auditory, and textual data. This includes creating models that can effectively align and fuse information from different sources to enhance understanding and decision-making processes.

\textbf{2) Adaptive Generalization Mechanisms:} Implementing strategies that enable models to generalize across varying domains and tasks is crucial. This involves designing systems that can adapt to new environments and conditions without extensive retraining, thereby improving robustness and flexibility. 

\textbf{(3) Real-Time Processing and Efficiency:} Enhancing the computational efficiency of multimodal systems to facilitate real-time processing is vital. Research should aim to optimize algorithms and hardware utilization to meet the low-latency requirements of dynamic, real-world applications.

\textbf{(4) Robustness to Environmental Variability:} Ensuring that systems can maintain performance in the face of environmental changes, such as noise or sensor variability, is important. This includes developing methods for dynamic modality weighting and context-aware processing to handle unforeseen challenges.

Addressing these research directions will contribute to the creation of more robust, adaptable, and efficient perception-cognition-action systems capable of operating effectively in diverse and unpredictable environments.

\subsubsection{Memory Storage and Update}
Future research on memory in embodied agents should move toward more flexible, context-aware, and cognitively inspired systems. The following directions highlight areas where further exploration is likely to yield meaningful advances.

\textbf{(1) Adaptive Memory Structuring:} Agents should be able to reorganize and refine memory layouts over time, such as consolidating stable patterns, isolating transient data, or expanding modules for task-specific learning, without relying on fixed architectural constraints.

\textbf{2) Multimodal Representation Alignment:} Cross-modal memory components must go beyond co-existence. Building representations that encode consistent semantic structure across perceptual, spatial, and symbolic modalities is essential for coherent long-term reasoning. 

\textbf{(3) Relevance-Based Forgetting and Compression:} Rather than accumulating memory indiscriminately, agents should employ forgetting and compression strategies that prioritize relevance. By abstracting or discarding low-impact content based on usage, context, or predictive utility, memory systems can remain efficient while preserving essential information for behavior and learning.

\textbf{(4) Constructive Memory Utilization:} Instead of serving solely as a retrieval mechanism, memory should enable agents to reconstruct, adapt, and extend past experiences. Structured memory can be reused to simulate variations, compose generalizations, and drive action in novel conditions, especially when grounded in task-specific structure.

These directions reflect a broader shift toward brain-inspired memory systems that move beyond static, task-bound designs to become self-organizing, semantically integrated, and capable of supporting adaptive, context-sensitive behavior.

\subsubsection{Hardware and Software}
% Developing scalable, energy-efficient architectures.
Looking forward, there are several strategic directions that promise to advance the field of Neural Brain hardware and software:

\par \textbf{(1) Advanced Neuromorphic Hardware:} 
Future generations of neuromorphic chips will need to offer greater scalability, improved robustness, and seamless integration with sensing and actuation components~\cite{yang2020neuromorphic}. Enhancing support for plasticity, temporal coding, and hybrid spiking-dense operations will be essential for bridging biological plausibility and engineering practicality. Research into new materials, 3D integration, and heterogeneous architectures may provide pathways for higher density and more efficient event-driven computation.

\par \textbf{(2) Novel Computing Paradigms:} 
Quantum and photonic computing present exciting opportunities for expanding the computational capabilities of Neural Brain systems. These paradigms offer fundamentally different mechanisms for parallelism, memory encoding, and non-classical logic~\cite{peng2018neuromorphic,markovic2020quantum}. As experimental platforms mature, they may enable high-speed, low-energy implementations of neural models for tasks such as probabilistic reasoning, high-dimensional search, and multimodal inference.

\par \textbf{(3) Deep Co-Design:} 
To maximize efficiency and generalizability, future research must embrace deep co-design strategies that integrate algorithm development, model architecture, and hardware implementation in a unified workflow~\cite{ponzina2022hardware}. This includes joint optimization of neural representations, dataflow scheduling, and memory access across the stack. Deep co-design not only enhances performance but also improves interpretability, adaptability, and system-level robustness. Ultimately, it will be the foundation for scalable, energy-aware embodied intelligence in the real world.

\subsubsection{Conclusion and Future Roadmap}
\par This work explored the design of a brain for embodied agents from a neuroscience perspective, providing a biologically inspired foundation to enhance their adaptability and cognitive capabilities. We identified the key components of a Neural Brain, including sensing, function, memory, and hardware/software, which form the foundation for supporting embodied agents in unstructured environments. Building on this, we proposed a biologically inspired architecture that integrates multimodal active sensing, perception-cognition-action functions, neuroplasticity-driven memory, and neuromorphic hardware/software co-design. Additionally, through a comprehensive review of current AI agents, we critically analyzed the gap between existing systems and human intelligence, offering valuable insights that can guide future developments toward achieving human-level intelligence in real-world scenarios.

\par Based on the current research status and technological development, we argue that the future roadmap for the Neural Brain involves three key stages. In the short term (1-3 years), the focus should be on implementing integrated perception-cognition-action loops in robotics. The mid-term goal (5 years) is to develop brain-inspired memory and reasoning systems that enhance decision-making and adaptability. Over the long term (10+ years), the goal should be to achieve human-level intelligence, with energy-efficient hardware that can ensure scalability and sustainability.

\noindent \textbf{Acknowledgements:} GPT-4o~\cite{gpt-4o} is used as an assistive tool in the preparation of certain biological diagrams.

% \par Short-Term (1-3 years): Implement integrated perception-cognition-action loops in robotics (e.g., Tesla Bot).

% \par Mid-Term (5 years): Develop brain-inspired memory and reasoning systems.

% \par Long-Term (10+ years): Achieve human-level intelligence with energy-efficient hardware.

% \subsection{Ethical and Societal Implications}

% \par Safety: Ensuring reliability in dynamic environments (e.g., ISO 21448 for robotics).

% \par Privacy: Protecting multimodal data in healthcare and assistive robots.

% \par Societal Impact: Addressing labor market disruptions and ethical concerns.

% \subsection{Case Studies}

% \par Medical Robotics: Neural Brain-driven surgical assistants (e.g., Intuitive Surgical's da Vinci).

% \par Autonomous Vehicles: Active perception-cognition-action loops in Waymo's self-driving systems.

% \section{Summary}\label{section_Summary}

% \par Some simple content for summary.

% (Test~\cite{FPFH})

% Can use something like this to put references on a page
% by themselves when using endfloat and the captionsoff option.
\ifCLASSOPTIONcaptionsoff
\newpage
\fi

% <OR> manually copy in the resultant .bbl file
% set second argument of \begin to the number of references
% (used to reserve space for the reference number labels box)
% \footnotesize

% \setbeamerfont{bibliography entry author}{shape=\scshape,size=\tiny}%
% \setbeamerfont{bibliography entry title}{shape=\scshape,size=\tiny}
% \setbeamerfont{bibliography entry journal}{shape=\scshape,size=\tiny}
% \setbeamerfont{bibliography entry note}{shape=\scshape,size=\tiny}
%\bibliographystyle{IEEEtran}
%\bibliography{reference}
\vspace{-1em}
\bibliographystyle{ieeetr.bst}

%\bibliographystyle{IEEEtran}

%\renewcommand*{\bibfont}{\footnotesize}

% \bibliography{reference, Thai/references_thai}
\bibliography{reference}

\begin{thebibliography}{100}

\bibitem{survey1}
M.~Shoaran, U.~Shin, and M.~Shaeri, ``Intelligent and miniaturized neural
  interfaces: An emerging era in neurotechnology,'' {\em arXiv preprint
  arXiv:2405.10780}, 2024.

\bibitem{survey2}
Y.~Liu, X.~Cao, T.~Chen, Y.~Jiang, J.~You, M.~Wu, X.~Wang, M.~Feng, Y.~Jin, and
  J.~Chen, ``A survey of embodied {AI} in healthcare: Techniques, applications,
  and opportunities,'' {\em arXiv preprint arXiv:2501.07468}, 2025.

\bibitem{liu2024surveywm}
Y.~Liu, W.~Chen, Y.~Bai, X.~Liang, G.~Li, W.~Gao, and L.~Lin, ``Aligning cyber
  space with physical world: A comprehensive survey on embodied {AI},'' {\em
  arXiv preprint arXiv:2407.06886}, 2024.

\bibitem{zhang2024survey}
Z.~Zhang, X.~Bo, C.~Ma, R.~Li, X.~Chen, Q.~Dai, J.~Zhu, Z.~Dong, and J.-R. Wen,
  ``A survey on the memory mechanism of large language model based agents,''
  {\em arXiv preprint arXiv:2404.13501}, 2024.

\bibitem{7pillars}
E.~Cambria, R.~Mao, M.~Chen, Z.~Wang, and S.-B. Ho, ``Seven pillars for the
  future of artificial intelligence,'' {\em {IEEE} Intelligent Systems},
  vol.~38, no.~6, pp.~62--69, 2023.

\bibitem{brown2020languagemodelsfewshotlearners}
T.~Brown, B.~Mann, N.~Ryder, M.~Subbiah, J.~D. Kaplan, P.~Dhariwal,
  A.~Neelakantan, P.~Shyam, G.~Sastry, A.~Askell, {\em et~al.}, ``Language
  models are few-shot learners,'' {\em Advances in neural information
  processing systems}, vol.~33, pp.~1877--1901, 2020.

\bibitem{achiam2023gpt}
J.~Achiam, S.~Adler, S.~Agarwal, L.~Ahmad, I.~Akkaya, F.~L. Aleman, D.~Almeida,
  J.~Altenschmidt, S.~Altman, S.~Anadkat, {\em et~al.}, ``{GPT}-4 technical
  report,'' {\em arXiv preprint arXiv:2303.08774}, 2023.

\bibitem{chowdhery2023palm}
A.~Chowdhery, S.~Narang, J.~Devlin, M.~Bosma, G.~Mishra, A.~Roberts, P.~Barham,
  H.~W. Chung, C.~Sutton, S.~Gehrmann, {\em et~al.}, ``Palm: Scaling language
  modeling with pathways,'' {\em Journal of Machine Learning Research},
  vol.~24, no.~240, pp.~1--113, 2023.

\bibitem{raffel2023exploringlimitstransferlearning}
C.~Raffel, N.~Shazeer, A.~Roberts, K.~Lee, S.~Narang, M.~Matena, Y.~Zhou,
  W.~Li, and P.~J. Liu, ``Exploring the limits of transfer learning with a
  unified text-to-text transformer,'' {\em Journal of machine learning
  research}, vol.~21, no.~140, pp.~1--67, 2020.

\bibitem{meta2024llama3}
M.~AI, ``{LLaMA} 3: Open and efficient foundation language models,'' {\em Meta
  AI Blog}, 2024.

\bibitem{NMIschulze2025visual}
L.~M. Schulze~Buschoff, E.~Akata, M.~Bethge, and E.~Schulz, ``Visual cognition
  in multimodal large language models,'' {\em Nature Machine Intelligence},
  pp.~1--11, 2025.

\bibitem{bostonDynamics2023atlas}
B.~Dynamics, ``Boston dynamics' atlas robot: A leap towards human-like mobility
  and flexibility,'' 2023.
\newblock Available at \url{https://www.bostondynamics.com/atlas}.

\bibitem{tesla2023optimus}
Tesla, ``Tesla's optimus robot: Pushing the limits of humanoid robotics,''
  2023.
\newblock Available at \url{https://www.tesla.com/AI}.

\bibitem{unitreeG1}
U.~Robotics, ``Unitree {G1}: The next-generation humanoid robot,'' 2024.
\newblock Available at \url{https://www.unitree.com/g1}.

\bibitem{NMIachterberg2023spatially}
J.~Achterberg, D.~Akarca, D.~Strouse, J.~Duncan, and D.~E. Astle, ``Spatially
  embedded recurrent neural networks reveal widespread links between structural
  and functional neuroscience findings,'' {\em Nature Machine Intelligence},
  vol.~5, no.~12, pp.~1369--1381, 2023.

\bibitem{NMIhan2024lifelike}
L.~Han, Q.~Zhu, J.~Sheng, C.~Zhang, T.~Li, Y.~Zhang, H.~Zhang, Y.~Liu, C.~Zhou,
  R.~Zhao, {\em et~al.}, ``Lifelike agility and play in quadrupedal robots
  using reinforcement learning and generative pre-trained models,'' {\em Nature
  Machine Intelligence}, vol.~6, no.~7, pp.~787--798, 2024.

\bibitem{NMImeng2025preserving}
Y.~Meng, Z.~Bing, X.~Yao, K.~Chen, K.~Huang, Y.~Gao, F.~Sun, and A.~Knoll,
  ``Preserving and combining knowledge in robotic lifelong reinforcement
  learning,'' {\em Nature Machine Intelligence}, pp.~1--14, 2025.

\bibitem{zaadnoordijk2022lessons}
L.~Zaadnoordijk, T.~R. Besold, and R.~Cusack, ``Lessons from infant learning
  for unsupervised machine learning,'' {\em Nature Machine Intelligence},
  vol.~4, no.~6, pp.~510--520, 2022.

\bibitem{NNSmiconi2025neural}
T.~Miconi and K.~Kay, ``Neural mechanisms of relational learning and fast
  knowledge reassembly in plastic neural networks,'' {\em Nature Neuroscience},
  pp.~1--9, 2025.

\bibitem{NMICOGvon2024cognitive}
M.~C. von Ebers and X.-X. Wei, ``Cognitive maps from predictive vision,'' {\em
  Nature Machine Intelligence}, vol.~6, no.~8, pp.~850--851, 2024.

\bibitem{NMICOGgornet2024automated}
J.~Gornet and M.~Thomson, ``Automated construction of cognitive maps with
  visual predictive coding,'' {\em Nature Machine Intelligence}, vol.~6, no.~7,
  pp.~820--833, 2024.

\bibitem{jia2023neuroscience}
H.~Jia, {\em Neuroscience for Artificial Intelligence}.
\newblock Jenny Stanford Publishing, 2023.

\bibitem{xisurvey}
Z.~Xi, W.~Chen, X.~Guo, W.~He, Y.~Ding, B.~Hong, M.~Zhang, J.~Wang, S.~Jin,
  E.~Zhou, {\em et~al.}, ``The rise and potential of large language model based
  agents: A survey,'' {\em Science China Information Sciences}, vol.~68, no.~2,
  p.~121101, 2025.

\bibitem{wangsurvey}
L.~Wang, C.~Ma, X.~Feng, Z.~Zhang, H.~Yang, J.~Zhang, Z.~Chen, J.~Tang,
  X.~Chen, Y.~Lin, {\em et~al.}, ``A survey on large language model based
  autonomous agents,'' {\em Frontiers of Computer Science}, vol.~18, no.~6,
  p.~186345, 2024.

\bibitem{xiesurvey}
J.~Xie, Z.~Chen, R.~Zhang, X.~Wan, and G.~Li, ``Large multimodal agents: A
  survey,'' {\em arXiv preprint arXiv:2402.15116}, 2024.

\bibitem{durantesurvey}
Z.~Durante, Q.~Huang, N.~Wake, R.~Gong, J.~S. Park, B.~Sarkar, R.~Taori,
  Y.~Noda, D.~Terzopoulos, Y.~Choi, {\em et~al.}, ``Agent {AI}: Surveying the
  horizons of multimodal interaction,'' {\em arXiv preprint arXiv:2401.03568},
  2024.

\bibitem{ma2024survey}
Y.~Ma, Z.~Song, Y.~Zhuang, J.~Hao, and I.~King, ``A survey on
  vision-language-action models for embodied {AI},'' {\em arXiv preprint
  arXiv:2405.14093}, 2024.

\bibitem{kim2024openvla}
M.~J. Kim, K.~Pertsch, S.~Karamcheti, T.~Xiao, A.~Balakrishna, S.~Nair,
  R.~Rafailov, E.~Foster, G.~Lam, P.~Sanketi, {\em et~al.}, ``Openvla: An
  open-source vision-language-action model,'' {\em arXiv preprint
  arXiv:2406.09246}, 2024.

\bibitem{huanginner}
W.~Huang, F.~Xia, T.~Xiao, H.~Chan, J.~Liang, P.~Florence, A.~Zeng, J.~Tompson,
  I.~Mordatch, Y.~Chebotar, {\em et~al.}, ``Inner monologue: {Embodied}
  reasoning through planning with language models,'' in {\em CoRL}, 2022.

\bibitem{sarch2025vlm}
G.~Sarch, L.~Jang, M.~Tarr, W.~W. Cohen, K.~Marino, and K.~Fragkiadaki, ``{VLM}
  agents generate their own memories: {Distilling} experience into embodied
  programs of thought,'' {\em NeurIPS}, 2025.

\bibitem{he2025asap}
T.~He, J.~Gao, W.~Xiao, Y.~Zhang, Z.~Wang, J.~Wang, Z.~Luo, G.~He, N.~Sobanbab,
  C.~Pan, {\em et~al.}, ``{ASAP}: {Aligning} simulation and real-world physics
  for learning agile humanoid whole-body skills,'' {\em arXiv preprint
  arXiv:2502.01143}, 2025.

\bibitem{ji2024exbody2}
M.~Ji, X.~Peng, F.~Liu, J.~Li, G.~Yang, X.~Cheng, and X.~Wang, ``Exbody2:
  {Advanced} expressive humanoid whole-body control,'' {\em arXiv preprint
  arXiv:2412.13196}, 2024.

\bibitem{cheng2024expressive}
X.~Cheng, Y.~Ji, J.~Chen, R.~Yang, G.~Yang, and X.~Wang, ``Expressive
  whole-body control for humanoid robots,'' {\em arXiv preprint
  arXiv:2402.16796}, 2024.

\bibitem{peng2021amp}
X.~B. Peng, Z.~Ma, P.~Abbeel, S.~Levine, and A.~Kanazawa, ``{AMP}: Adversarial
  motion priors for stylized physics-based character control,'' {\em ACM
  Transactions on Graphics (ToG)}, 2021.

\bibitem{gu2025humanoid}
Z.~Gu, J.~Li, W.~Shen, W.~Yu, Z.~Xie, S.~McCrory, X.~Cheng, A.~Shamsah,
  R.~Griffin, C.~K. Liu, {\em et~al.}, ``Humanoid locomotion and manipulation:
  Current progress and challenges in control, planning, and learning,'' {\em
  arXiv preprint arXiv:2501.02116}, 2025.

\bibitem{guo2025deepseek}
D.~Guo, D.~Yang, H.~Zhang, J.~Song, R.~Zhang, R.~Xu, Q.~Zhu, S.~Ma, P.~Wang,
  X.~Bi, {\em et~al.}, ``Deepseek-{R1}: {Incentivizing} reasoning capability in
  {LLMs} via reinforcement learning,'' {\em arXiv preprint arXiv:2501.12948},
  2025.

\bibitem{zhao2023learning}
T.~Z. Zhao, V.~Kumar, S.~Levine, and C.~Finn, ``Learning fine-grained bimanual
  manipulation with low-cost hardware,'' {\em arXiv preprint arXiv:2304.13705},
  2023.

\bibitem{zhao2024aloha}
T.~Z. Zhao, J.~Tompson, D.~Driess, P.~Florence, S.~K.~S. Ghasemipour, C.~Finn,
  and A.~Wahid, ``{ALOHA} unleashed: {A} simple recipe for robot dexterity,''
  in {\em CoRL}, 2024.

\bibitem{chi2023diffusion}
C.~Chi, Z.~Xu, S.~Feng, E.~Cousineau, Y.~Du, B.~Burchfiel, R.~Tedrake, and
  S.~Song, ``Diffusion policy: Visuomotor policy learning via action
  diffusion,'' {\em The International Journal of Robotics Research},
  p.~02783649241273668, 2023.

\bibitem{team2024octo}
O.~M. Team, D.~Ghosh, H.~Walke, K.~Pertsch, K.~Black, O.~Mees, S.~Dasari,
  J.~Hejna, T.~Kreiman, C.~Xu, {\em et~al.}, ``Octo: An open-source generalist
  robot policy,'' {\em arXiv preprint arXiv:2405.12213}, 2024.

\bibitem{wang2025scaling}
L.~Wang, X.~Chen, J.~Zhao, and K.~He, ``Scaling proprioceptive-visual learning
  with heterogeneous pre-trained transformers,'' {\em NeurIPS}, vol.~37,
  pp.~124420--124450, 2025.

\bibitem{liu2024rdt}
S.~Liu, L.~Wu, B.~Li, H.~Tan, H.~Chen, Z.~Wang, K.~Xu, H.~Su, and J.~Zhu,
  ``Rdt-1b: a diffusion foundation model for bimanual manipulation,'' {\em
  arXiv preprint arXiv:2410.07864}, 2024.

\bibitem{black2024pi_0}
K.~Black, N.~Brown, D.~Driess, A.~Esmail, M.~Equi, C.~Finn, N.~Fusai, L.~Groom,
  K.~Hausman, B.~Ichter, {\em et~al.}, ``$\pi_0$: A vision-language-action flow
  model for general robot control,'' {\em arXiv preprint arXiv:2410.24164},
  2024.

\bibitem{zhanghirt}
J.~Zhang, Y.~Guo, X.~Chen, Y.-J. Wang, Y.~Hu, C.~Shi, and J.~Chen, ``{HiRT}:
  {Enhancing} robotic control with hierarchical robot transformers,'' in {\em
  CoRL}, 2024.

\bibitem{chen2020soundspaces}
C.~Chen, U.~Jain, C.~Schissler, S.~V.~A. Gari, Z.~Al-Halah, V.~K. Ithapu,
  P.~Robinson, and K.~Grauman, ``Soundspaces: Audio-visual navigation in {3D}
  environments,'' in {\em ECCV}, 2020.

\bibitem{liu2020embodied}
H.~Liu, D.~Guo, F.~Sun, W.~Yang, S.~Furber, and T.~Sun, ``Embodied tactile
  perception and learning,'' {\em Brain Science Advances}, 2020.

\bibitem{severino2024evolution}
G.~J. Severino and A.-S. Barwich, ``Evolution and analysis of respiratory odor
  navigation in embodied agents,'' in {\em ALIFE}, 2024.

\bibitem{zhu2024spa}
H.~Zhu, H.~Yang, Y.~Wang, J.~Yang, L.~Wang, and T.~He, ``{SPA}: {3D}
  spatial-awareness enables effective embodied representation,'' {\em arXiv
  preprint arXiv:2410.08208}, 2024.

\bibitem{yang2024thinking}
J.~Yang, S.~Yang, A.~W. Gupta, R.~Han, L.~Fei-Fei, and S.~Xie, ``Thinking in
  space: {How} multimodal large language models see, remember, and recall
  spaces,'' {\em arXiv preprint arXiv:2412.14171}, 2024.

\bibitem{gupta2021embodied}
A.~Gupta, S.~Savarese, S.~Ganguli, and L.~Fei-Fei, ``Embodied intelligence via
  learning and evolution,'' {\em Nature Communications}, 2021.

\bibitem{zawalski2024robotic}
M.~Zawalski, W.~Chen, K.~Pertsch, O.~Mees, C.~Finn, and S.~Levine, ``Robotic
  control via embodied chain-of-thought reasoning,'' {\em arXiv preprint
  arXiv:2407.08693}, 2024.

\bibitem{li2025embodied}
M.~Li, S.~Zhao, Q.~Wang, K.~Wang, Y.~Zhou, S.~Srivastava, C.~Gokmen, T.~Lee,
  E.~L. Li, R.~Zhang, {\em et~al.}, ``Embodied agent interface: {Benchmarking}
  {LLMs} for embodied decision making,'' {\em NeurIPS}, 2025.

\bibitem{chen2023towards}
L.~Chen, Y.~Zhang, S.~Ren, H.~Zhao, Z.~Cai, Y.~Wang, P.~Wang, T.~Liu, and
  B.~Chang, ``Towards end-to-end embodied decision making via multi-modal large
  language model: {Explorations} with {GPT4}-vision and beyond,'' {\em arXiv
  preprint arXiv:2310.02071}, 2023.

\bibitem{fang2019scene}
K.~Fang, A.~Toshev, L.~Fei-Fei, and S.~Savarese, ``Scene memory transformer for
  embodied agents in long-horizon tasks,'' in {\em CVPR}, 2019.

\bibitem{davies2021advancing}
M.~Davies, A.~Wild, G.~Orchard, Y.~Sandamirskaya, G.~A.~F. Guerra, P.~Joshi,
  P.~Plank, and S.~R. Risbud, ``Advancing neuromorphic computing with loihi:
  {A} survey of results and outlook,'' {\em Proceedings of the IEEE}, vol.~109,
  no.~5, pp.~911--934, 2021.

\bibitem{haessig2018spiking}
G.~Haessig, A.~Cassidy, R.~Alvarez, R.~Benosman, and G.~Orchard, ``Spiking
  optical flow for event-based sensors using {IBM}'s truenorth neurosynaptic
  system,'' {\em IEEE Transactions on Biomedical Circuits and Systems}, 2018.

\bibitem{bhattasali2022neural}
N.~Bhattasali, A.~M. Zador, and T.~Engel, ``Neural circuit architectural priors
  for embodied control,'' {\em NeurIPS}, 2022.

\bibitem{paszke2019pytorch}
A.~Paszke, S.~Gross, F.~Massa, A.~Lerer, J.~Bradbury, G.~Chanan, T.~Killeen,
  Z.~Lin, N.~Gimelshein, L.~Antiga, {\em et~al.}, ``Pytorch: {An} imperative
  style, high-performance deep learning library,'' {\em NeurIPS}, 2019.

\bibitem{abadi2016tensorflow}
M.~Abadi, A.~Agarwal, P.~Barham, E.~Brevdo, Z.~Chen, C.~Citro, G.~S. Corrado,
  A.~Davis, J.~Dean, M.~Devin, {\em et~al.}, ``Tensorflow: {Large}-scale
  machine learning on heterogeneous distributed systems,'' {\em arXiv preprint
  arXiv:1603.04467}, 2016.

\bibitem{liu2025advanceschallengesfoundationagents}
B.~Liu, X.~Li, J.~Zhang, J.~Wang, T.~He, S.~Hong, H.~Liu, S.~Zhang, K.~Song,
  K.~Zhu, {\em et~al.}, ``Advances and challenges in foundation agents: From
  brain-inspired intelligence to evolutionary, collaborative, and safe
  systems,'' {\em arXiv preprint arXiv:2504.01990}, 2025.

\bibitem{maocom}
R.~Mao, G.~Chen, X.~Li, M.~Ge, and E.~Cambria, ``A comparative analysis of
  metaphorical cognition in chatgpt and human minds,'' {\em Cognitive
  Computation}, vol.~17, p.~35, 2025.

\bibitem{hourglass}
Y.~Susanto, A.~Livingstone, B.~C. Ng, and E.~Cambria, ``The {H}ourglass {M}odel
  revisited,'' {\em {IEEE} Intelligent Systems}, vol.~35, no.~5, pp.~96--102,
  2020.

\bibitem{damdes}
A.~Damasio, {\em Descartes' Error: Emotion, Reason, and the Human Brain}.
\newblock New York: Grossett/Putnam, 1994.

\bibitem{liukno}
H.~Liu, R.~Wei, G.~Tu, J.~Lin, D.~Jiang, and E.~Cambria, ``Knowing what and
  why: Causal emotion entailment for emotion recognition in conversations,''
  {\em Expert Systems with Applications}, vol.~274, p.~126924, 2025.

\bibitem{senticnet}
E.~Cambria, X.~Zhang, R.~Mao, M.~Chen, and K.~Kwok, ``{SenticNet} 8: Fusing
  emotion {AI} and commonsense {AI} for interpretable, trustworthy, and
  explainable affective computing,'' in {\em {Proceedings of the International
  Conference on Human-Computer Interaction (HCII)}}, (Washington DC, USA),
  pp.~197--216, 2024.

\bibitem{zhuneu}
L.~Zhu, R.~Mao, E.~Cambria, and B.~J. Jansen, ``Neurosymbolic {AI} for
  personalized sentiment analysis,'' in {\em {Proceedings of the International
  Conference on Human-Computer Interaction (HCII)}}, (Washington DC, USA),
  pp.~269--290, 2024.

\bibitem{affectivebci}
T.~Wang, R.~Mao, S.~Liu, E.~Cambria, and D.~Ming, ``Explainable multi-frequency
  and multi-region fusion model for affective brain-computer interfaces,'' {\em
  Information Fusion}, vol.~118, p.~102971, 2025.

\bibitem{pansel}
V.~Pandelea, E.~Ragusa, P.~Gastaldo, and E.~Cambria, ``Selecting language
  models features via software-hardware co-design,'' in {\em Proceedings of
  {IEEE ICASSP}}, 2023.

\bibitem{tiippana2014mcgurk}
K.~Tiippana, ``What is the mcgurk effect?,'' 2014.

\bibitem{hartley1992stereo}
R.~I. Hartley, R.~Gupta, and T.~Chang, ``Stereo from uncalibrated cameras.,''
  in {\em CVPR}, vol.~92, pp.~761--764, 1992.

\bibitem{eargle2012microphone}
J.~Eargle, {\em The Microphone Book: From mono to stereo to surround-a guide to
  microphone design and application}.
\newblock Routledge, 2012.

\bibitem{dahiya2009tactile}
R.~S. Dahiya, G.~Metta, M.~Valle, and G.~Sandini, ``Tactile sensing-from humans
  to humanoids,'' {\em IEEE TRO}, vol.~26, no.~1, pp.~1--20, 2009.

\bibitem{yang20211d}
B.~Yang, N.~V. Myung, and T.-T. Tran, ``{1D} metal oxide semiconductor
  materials for chemiresistive gas sensors: a review,'' {\em Advanced
  Electronic Materials}, vol.~7, no.~9, p.~2100271, 2021.

\bibitem{lechner2000global}
W.~Lechner and S.~Baumann, ``Global navigation satellite systems,'' {\em
  Computers and Electronics in Agriculture}, vol.~25, no.~1-2, pp.~67--85,
  2000.

\bibitem{zaraki2014designing}
A.~Zaraki, D.~Mazzei, M.~Giuliani, and D.~De~Rossi, ``Designing and evaluating
  a social gaze-control system for a humanoid robot,'' {\em IEEE Transactions
  on Human-Machine Systems}, vol.~44, no.~2, pp.~157--168, 2014.

\bibitem{liao2023deep}
K.~Liao, L.~Nie, S.~Huang, C.~Lin, J.~Zhang, Y.~Zhao, M.~Gabbouj, and D.~Tao,
  ``Deep learning for camera calibration and beyond: A survey,'' {\em arXiv
  preprint arXiv:2303.10559}, 2023.

\bibitem{wasenmuller2017comparison}
O.~Wasenm{\"u}ller and D.~Stricker, ``Comparison of kinect v1 and v2 depth
  images in terms of accuracy and precision,'' in {\em Computer Vision--ACCV
  2016 Workshops: ACCV 2016 International Workshops, Taipei, Taiwan, November
  20-24, 2016, Revised Selected Papers, Part II 13}, pp.~34--45, Springer,
  2017.

\bibitem{yang2022deepear}
Q.~Yang and Y.~Zheng, ``Deepear: Sound localization with binaural
  microphones,'' {\em IEEE Transactions on Mobile Computing}, vol.~23, no.~1,
  pp.~359--375, 2022.

\bibitem{heins2024force}
A.~Heins and A.~P. Schoellig, ``Force push: Robust single-point pushing with
  force feedback,'' {\em IEEE Robotics and Automation Letters}, 2024.

\bibitem{lewis2004comparisons}
N.~S. Lewis, ``Comparisons between mammalian and artificial olfaction based on
  arrays of carbon black-polymer composite vapor detectors,'' {\em Accounts of
  chemical research}, vol.~37, no.~9, pp.~663--672, 2004.

\bibitem{lee2004robust}
S.~Lee and J.-B. Song, ``Robust mobile robot localization using optical flow
  sensors and encoders,'' in {\em ICRA}, vol.~1, pp.~1039--1044, IEEE, 2004.

\bibitem{ren2023robot}
H.~Ren and A.~H. Qureshi, ``Robot active neural sensing and planning in unknown
  cluttered environments,'' {\em IEEE Transactions on Robotics}, vol.~39,
  no.~4, pp.~2738--2750, 2023.

\bibitem{chen2023polymer}
W.~Chen, Y.~Yan, Z.~Zhang, L.~Yang, and J.~Pan, ``Polymer-based self-calibrated
  optical fiber tactile sensor,'' in {\em 2023 IEEE/RSJ International
  Conference on Intelligent Robots and Systems (IROS)}, pp.~10197--10203, IEEE,
  2023.

\bibitem{gallego2020event}
G.~Gallego, T.~Delbr{\"u}ck, G.~Orchard, C.~Bartolozzi, B.~Taba, A.~Censi,
  S.~Leutenegger, A.~J. Davison, J.~Conradt, K.~Daniilidis, {\em et~al.},
  ``Event-based vision: A survey,'' {\em TPAMI}, vol.~44, no.~1, pp.~154--180,
  2020.

\bibitem{elko2008microphone}
G.~W. Elko and J.~Meyer, ``Microphone arrays,'' {\em Springer handbook of
  speech processing}, pp.~1021--1041, 2008.

\bibitem{tenzer2014feel}
Y.~Tenzer, L.~P. Jentoft, and R.~D. Howe, ``The feel of {MEMS} barometers:
  Inexpensive and easily customized tactile array sensors,'' {\em IEEE Robotics
  \& Automation Magazine}, vol.~21, no.~3, pp.~89--95, 2014.

\bibitem{gautschi2002piezoelectric}
G.~Gautschi and G.~Gautschi, {\em Piezoelectric sensors}.
\newblock Springer, 2002.

\bibitem{zhang2018tutorial}
Z.~Zhang and D.~Scaramuzza, ``A tutorial on quantitative trajectory evaluation
  for visual (-inertial) odometry,'' in {\em IROS}, pp.~7244--7251, IEEE, 2018.

\bibitem{kemna2017multi}
S.~Kemna, J.~G. Rogers, C.~Nieto-Granda, S.~Young, and G.~S. Sukhatme,
  ``Multi-robot coordination through dynamic voronoi partitioning for
  informative adaptive sampling in communication-constrained environments,'' in
  {\em 2017 IEEE International Conference on Robotics and Automation (ICRA)},
  pp.~2124--2130, 2017.

\bibitem{chen2024self}
Y.~Chen, M.~Wang, Z.~Chen, W.~Zhao, and Y.~Shi, ``Self-validating sensor
  technology and its application in artificial olfaction: A review,'' {\em
  Measurement}, p.~116025, 2024.

\bibitem{behroozpour2017lidar}
B.~Behroozpour, P.~A. Sandborn, M.~C. Wu, and B.~E. Boser, ``Lidar system
  architectures and circuits,'' {\em IEEE Communications Magazine}, vol.~55,
  no.~10, pp.~135--142, 2017.

\bibitem{shung1996ultrasonic}
K.~K. Shung and M.~Zippuro, ``Ultrasonic transducers and arrays,'' {\em IEEE
  Engineering in Medicine and Biology Magazine}, vol.~15, no.~6, pp.~20--30,
  1996.

\bibitem{donlon2018gelslim}
E.~Donlon, S.~Dong, M.~Liu, J.~Li, E.~Adelson, and A.~Rodriguez, ``Gelslim: A
  high-resolution, compact, robust, and calibrated tactile-sensing finger,'' in
  {\em IEEE/RSJ International Conference on Intelligent Robots and Systems
  (IROS)}, pp.~1927--1934, IEEE, 2018.

\bibitem{liu2020recent}
B.~Liu, J.~Zhuang, and G.~Wei, ``Recent advances in the design of colorimetric
  sensors for environmental monitoring,'' {\em Environmental Science: Nano},
  vol.~7, no.~8, pp.~2195--2213, 2020.

\bibitem{li2020spatio}
J.~Li, Y.~Li, L.~He, J.~Chen, and A.~Plaza, ``Spatio-temporal fusion for remote
  sensing data: An overview and new benchmark,'' {\em Science China Information
  Sciences}, vol.~63, pp.~1--17, 2020.

\bibitem{ai2022deep}
H.~Ai, Z.~Cao, J.~Zhu, H.~Bai, Y.~Chen, and L.~Wang, ``Deep learning for
  omnidirectional vision: A survey and new perspectives,'' {\em arXiv preprint
  arXiv:2205.10468}, 2022.

\bibitem{cao2022gvins}
S.~Cao, X.~Lu, and S.~Shen, ``{GVINS}: Tightly coupled {GNSS}--visual--inertial
  fusion for smooth and consistent state estimation,'' {\em IEEE Transactions
  on Robotics}, vol.~38, no.~4, pp.~2004--2021, 2022.

\bibitem{zou2023object}
Z.~Zou, K.~Chen, Z.~Shi, Y.~Guo, and J.~Ye, ``Object detection in 20 years: A
  survey,'' {\em Proceedings of the IEEE}, vol.~111, no.~3, pp.~257--276, 2023.

\bibitem{yilmaz2006object}
A.~Yilmaz, O.~Javed, and M.~Shah, ``Object tracking: A survey,'' {\em Acm
  computing surveys (CSUR)}, vol.~38, no.~4, pp.~13--es, 2006.

\bibitem{kiechle2018model}
M.~Kiechle, M.~Storath, A.~Weinmann, and M.~Kleinsteuber, ``Model-based
  learning of local image features for unsupervised texture segmentation,''
  {\em IEEE Transactions on Image Processing}, vol.~27, no.~4, pp.~1994--2007,
  2018.

\bibitem{engel2015large}
J.~Engel, J.~St{\"u}ckler, and D.~Cremers, ``Large-scale direct {SLAM} with
  stereo cameras,'' in {\em 2015 IEEE/RSJ international conference on
  intelligent robots and systems (IROS)}, pp.~1935--1942, IEEE, 2015.

\bibitem{hrabar20083d}
S.~Hrabar, ``{3D} path planning and stereo-based obstacle avoidance for
  rotorcraft uavs,'' in {\em IROS}, pp.~807--814, IEEE, 2008.

\bibitem{Guo_navigation}
J.~Guo and F.~Zhong, ``A visual navigation perspective for category-level
  object pose estimation,'' in {\em ECCV}, 2022.

\bibitem{li2019survey}
R.~Li, Z.~Liu, and J.~Tan, ``A survey on {3D} hand pose estimation: Cameras,
  methods, and datasets,'' {\em Pattern Recognition}, vol.~93, pp.~251--272,
  2019.

\bibitem{nayar1997catadioptric}
S.~K. Nayar, ``Catadioptric omnidirectional camera,'' in {\em Proceedings of
  IEEE computer society conference on computer vision and pattern recognition},
  pp.~482--488, IEEE, 1997.

\bibitem{patole2017automotive}
S.~M. Patole, M.~Torlak, D.~Wang, and M.~Ali, ``Automotive radars: A review of
  signal processing techniques,'' {\em IEEE Signal Processing Magazine},
  vol.~34, no.~2, pp.~22--35, 2017.

\bibitem{ort2020autonomous}
T.~Ort, I.~Gilitschenski, and D.~Rus, ``Autonomous navigation in inclement
  weather based on a localizing ground penetrating radar,'' {\em IEEE RAL},
  vol.~5, no.~2, pp.~3267--3274, 2020.

\bibitem{vanarse2016review}
A.~Vanarse, A.~Osseiran, and A.~Rassau, ``A review of current neuromorphic
  approaches for vision, auditory, and olfactory sensors,'' {\em Frontiers in
  neuroscience}, vol.~10, p.~115, 2016.

\bibitem{rubel2002auditory}
E.~W. Rubel and B.~Fritzsch, ``Auditory system development: primary auditory
  neurons and their targets,'' {\em Annual review of neuroscience}, vol.~25,
  no.~1, pp.~51--101, 2002.

\bibitem{bizley2013and}
J.~K. Bizley and Y.~E. Cohen, ``The what, where and how of auditory-object
  perception,'' {\em Nature Reviews Neuroscience}, vol.~14, no.~10,
  pp.~693--707, 2013.

\bibitem{guenther2015role}
F.~H. Guenther and G.~Hickok, ``Role of the auditory system in speech
  production,'' {\em Handbook of clinical neurology}, vol.~129, pp.~161--175,
  2015.

\bibitem{battaglia2003bayesian}
P.~W. Battaglia, R.~A. Jacobs, and R.~N. Aslin, ``Bayesian integration of
  visual and auditory signals for spatial localization,'' {\em Journal of the
  optical society of America A}, vol.~20, no.~7, pp.~1391--1397, 2003.

\bibitem{rodriguez2012assisting}
A.~Rodr{\'\i}guez, J.~J. Yebes, P.~F. Alcantarilla, L.~M. Bergasa,
  J.~Almaz{\'a}n, and A.~Cela, ``Assisting the visually impaired: obstacle
  detection and warning system by acoustic feedback,'' {\em Sensors}, vol.~12,
  no.~12, pp.~17476--17496, 2012.

\bibitem{song2023multi}
Y.~Song, Z.~Li, G.~Li, B.~Wang, M.~Zhu, and P.~Shi, ``Multi-sensory
  visual-auditory fusion of wearable navigation assistance for people with
  impaired vision,'' {\em IEEE Transactions on Automation Science and
  Engineering}, 2023.

\bibitem{tegin2005tactile}
J.~Tegin and J.~Wikander, ``Tactile sensing in intelligent robotic
  manipulation--a review,'' {\em Industrial Robot: An International Journal},
  vol.~32, no.~1, pp.~64--70, 2005.

\bibitem{yu2023mimictouch}
K.~Yu, Y.~Han, Q.~Wang, V.~Saxena, D.~Xu, and Y.~Zhao, ``Mimictouch: Leveraging
  multi-modal human tactile demonstrations for contact-rich manipulation,''
  {\em arXiv preprint arXiv:2310.16917}, 2023.

\bibitem{huang20243d}
B.~Huang, Y.~Wang, X.~Yang, Y.~Luo, and Y.~Li, ``{3D}-{ViTac}: Learning
  fine-grained manipulation with visuo-tactile sensing,'' {\em arXiv preprint
  arXiv:2410.24091}, 2024.

\bibitem{sankaran2012biology}
S.~Sankaran, L.~R. Khot, and S.~Panigrahi, ``Biology and applications of
  olfactory sensing system: A review,'' {\em Sensors and Actuators B:
  Chemical}, vol.~171, pp.~1--17, 2012.

\bibitem{feng2010colorimetric}
L.~Feng, C.~J. Musto, J.~W. Kemling, S.~H. Lim, W.~Zhong, and K.~S. Suslick,
  ``Colorimetric sensor array for determination and identification of toxic
  industrial chemicals,'' {\em Analytical chemistry}, vol.~82, no.~22,
  pp.~9433--9440, 2010.

\bibitem{laska1999olfactory}
M.~Laska, C.~G. Galizia, M.~Giurfa, and R.~Menzel, ``Olfactory discrimination
  ability and odor structure--activity relationships in honeybees,'' {\em
  Chemical senses}, vol.~24, no.~4, pp.~429--438, 1999.

\bibitem{kononenko2001machine}
I.~Kononenko, ``Machine learning for medical diagnosis: history, state of the
  art and perspective,'' {\em Artificial Intelligence in medicine}, vol.~23,
  no.~1, pp.~89--109, 2001.

\bibitem{gomiero2018food}
T.~Gomiero, ``Food quality assessment in organic vs. conventional agricultural
  produce: Findings and issues,'' {\em Applied Soil Ecology}, vol.~123,
  pp.~714--728, 2018.

\bibitem{mpatani2021adsorption}
F.~M. Mpatani, R.~Han, A.~A. Aryee, A.~N. Kani, Z.~Li, and L.~Qu, ``Adsorption
  performance of modified agricultural waste materials for removal of emerging
  micro-contaminant bisphenol {A}: A comprehensive review,'' {\em Science of
  the Total Environment}, vol.~780, p.~146629, 2021.

\bibitem{ahmad2013reviews}
N.~Ahmad, R.~A.~R. Ghazilla, N.~M. Khairi, and V.~Kasi, ``Reviews on various
  inertial measurement unit ({IMU}) sensor applications,'' {\em International
  Journal of Signal Processing Systems}, vol.~1, no.~2, pp.~256--262, 2013.

\bibitem{xu2007gps}
G.~Xu and Y.~Xu, {\em GPS}, vol.~2.
\newblock Springer, 2007.

\bibitem{revnivykh2017glonass}
S.~Revnivykh, A.~Bolkunov, A.~Serdyukov, and O.~Montenbruck, ``Glonass,'' {\em
  Springer Handbook of Global Navigation Satellite Systems}, pp.~219--245,
  2017.

\bibitem{joubert2020developments}
N.~Joubert, T.~G. Reid, and F.~Noble, ``Developments in modern {GNSS} and its
  impact on autonomous vehicle architectures,'' in {\em 2020 IEEE Intelligent
  Vehicles Symposium (IV)}, pp.~2029--2036, IEEE, 2020.

\bibitem{vougioukas2019agricultural}
S.~G. Vougioukas, ``Agricultural robotics,'' {\em Annual review of control,
  robotics, and autonomous systems}, vol.~2, no.~1, pp.~365--392, 2019.

\bibitem{zhang2024frequency}
H.~Zhang, X.~Wang, C.~Xu, X.~Wang, F.~Xu, H.~Yu, L.~Yu, and W.~Yang,
  ``Frequency-adaptive low-latency object detection using events and frames,''
  {\em arXiv preprint arXiv:2412.04149}, 2024.

\bibitem{di2023photometric}
L.~Di~Giammarino, E.~Giacomini, L.~Brizi, O.~Salem, and G.~Grisetti,
  ``Photometric {LiDAR} and {RGB-D} bundle adjustment,'' {\em IEEE Robotics and
  Automation Letters}, vol.~8, no.~7, pp.~4362--4369, 2023.

\bibitem{nagrani2021attention}
A.~Nagrani, S.~Yang, A.~Arnab, A.~Jansen, C.~Schmid, and C.~Sun, ``Attention
  bottlenecks for multimodal fusion,'' {\em Advances in neural information
  processing systems}, vol.~34, pp.~14200--14213, 2021.

\bibitem{bajcsy1988active}
R.~Bajcsy, ``Active perception,'' {\em Proceedings of the IEEE}, vol.~76,
  no.~8, pp.~966--1005, 1988.

\bibitem{aller2022audiovisual}
M.~Aller, A.~Mihalik, and U.~Noppeney, ``Audiovisual adaptation is expressed in
  spatial and decisional codes,'' {\em Nature Communications}, vol.~13, no.~1,
  p.~3924, 2022.

\bibitem{ding2023learning}
W.~Ding, N.~Majcherczyk, M.~Deshpande, X.~Qi, D.~Zhao, R.~Madhivanan, and
  A.~Sen, ``Learning to view: Decision transformers for active object
  detection,'' in {\em 2023 IEEE International Conference on Robotics and
  Automation (ICRA)}, pp.~7140--7146, IEEE, 2023.

\bibitem{li2021attention}
T.~Li, C.~Wang, M.~Q.-H. Meng, and C.~W. de~Silva, ``Attention-driven active
  sensing with hybrid neural network for environmental field mapping,'' {\em
  IEEE Transactions on Automation Science and Engineering}, vol.~19, no.~3,
  pp.~2135--2152, 2021.

\bibitem{chen2022learning}
P.~Chen, D.~Ji, K.~Lin, W.~Hu, W.~Huang, T.~Li, M.~Tan, and C.~Gan, ``Learning
  active camera for multi-object navigation,'' {\em Advances in Neural
  Information Processing Systems}, vol.~35, pp.~28670--28682, 2022.

\bibitem{adam2016bayesian}
A.~Adam, C.~Dann, O.~Yair, S.~Mazor, and S.~Nowozin, ``Bayesian time-of-flight
  for realtime shape, illumination and albedo,'' {\em IEEE transactions on
  pattern analysis and machine intelligence}, vol.~39, no.~5, pp.~851--864,
  2016.

\bibitem{graca2024scidvs}
R.~Graca, S.~Zhou, B.~McReynolds, and T.~Delbruck, ``{SciDVS}: A scientific
  event camera with 1.7\% temporal contrast sensitivity at 0.7 lux,'' in {\em
  2024 IEEE European Solid-State Electronics Research Conference (ESSERC)},
  pp.~205--208, IEEE, 2024.

\bibitem{tran2023adaptive}
D.~M. Tran, N.~Ahlgren, C.~Depcik, and H.~He, ``Adaptive active fusion of
  camera and single-point lidar for depth estimation,'' {\em IEEE Transactions
  on Instrumentation and Measurement}, vol.~72, pp.~1--9, 2023.

\bibitem{preiss2018simultaneous}
J.~A. Preiss, K.~Hausman, G.~S. Sukhatme, and S.~Weiss, ``Simultaneous
  self-calibration and navigation using trajectory optimization,'' {\em The
  International Journal of Robotics Research}, vol.~37, no.~13-14,
  pp.~1573--1594, 2018.

\bibitem{luo2022hybrid}
D.~Luo, Y.~Zhuang, and S.~Wang, ``Hybrid sparse monocular visual odometry with
  online photometric calibration,'' {\em The International Journal of Robotics
  Research}, vol.~41, no.~11-12, pp.~993--1021, 2022.

\bibitem{zheng2024fast}
C.~Zheng, W.~Xu, Z.~Zou, T.~Hua, C.~Yuan, D.~He, B.~Zhou, Z.~Liu, J.~Lin,
  F.~Zhu, {\em et~al.}, ``Fast-livo2: Fast, direct lidar-inertial-visual
  odometry,'' {\em IEEE Transactions on Robotics}, vol.~41, pp.~326--346, 2025.

\bibitem{wisth2022vilens}
D.~Wisth, M.~Camurri, and M.~Fallon, ``Vilens: Visual, inertial, lidar, and leg
  odometry for all-terrain legged robots,'' {\em IEEE Transactions on
  Robotics}, vol.~39, no.~1, pp.~309--326, 2022.

\bibitem{tilmon2021fast}
B.~Tilmon, E.~Jain, S.~Ferrari, and S.~Koppal, ``Fast foveating cameras for
  dense adaptive resolution,'' {\em IEEE Transactions on Pattern Analysis and
  Machine Intelligence}, vol.~44, no.~9, pp.~4867--4878, 2021.

\bibitem{placed2023survey}
J.~A. Placed, J.~Strader, H.~Carrillo, N.~Atanasov, V.~Indelman, L.~Carlone,
  and J.~A. Castellanos, ``A survey on active simultaneous localization and
  mapping: State of the art and new frontiers,'' {\em IEEE Transactions on
  Robotics}, vol.~39, no.~3, pp.~1686--1705, 2023.

\bibitem{harlow2024new}
K.~Harlow, H.~Jang, T.~D. Barfoot, A.~Kim, and C.~Heckman, ``A new wave in
  robotics: Survey on recent mmwave radar applications in robotics,'' {\em IEEE
  Transactions on Robotics}, vol.~40, pp.~4544--4560, 2024.

\bibitem{karakaya2020electronic}
D.~Karakaya, O.~Ulucan, and M.~Turkan, ``Electronic nose and its applications:
  A survey,'' {\em International journal of Automation and Computing}, vol.~17,
  no.~2, pp.~179--209, 2020.

\bibitem{qiu2020real}
K.~Qiu, T.~Qin, J.~Pan, S.~Liu, and S.~Shen, ``Real-time temporal and
  rotational calibration of heterogeneous sensors using motion correlation
  analysis,'' {\em IEEE Transactions on Robotics}, vol.~37, no.~2,
  pp.~587--602, 2020.

\bibitem{paolo2024embodiedai}
G.~Paolo, J.~Gonzalez-Billandon, and B.~K{\'e}gl, ``A call for embodied ai,''
  {\em arXiv preprint arXiv:2402.03824}, 2024.

\bibitem{chen2025exploringembodiedmultimodallarge}
S.~Chen, Z.~Wu, K.~Zhang, C.~Li, B.~Zhang, F.~Ma, F.~R. Yu, and Q.~Li,
  ``Exploring embodied multimodal large models: Development, datasets, and
  future directions,'' {\em arXiv preprint arXiv:2502.15336}, 2025.

\bibitem{devlin2019bertpretrainingdeepbidirectional}
J.~Devlin, M.-W. Chang, K.~Lee, and K.~Toutanova, ``Bert: Pre-training of deep
  bidirectional transformers for language understanding,'' in {\em Proceedings
  of the 2019 conference of the North American chapter of the association for
  computational linguistics: human language technologies, volume 1 (long and
  short papers)}, pp.~4171--4186, 2019.

\bibitem{yang2020xlnetgeneralizedautoregressivepretraining}
Z.~Yang, Z.~Dai, Y.~Yang, J.~Carbonell, R.~R. Salakhutdinov, and Q.~V. Le,
  ``Xlnet: Generalized autoregressive pretraining for language understanding,''
  {\em Advances in neural information processing systems}, vol.~32, 2019.

\bibitem{tay2023ul2unifyinglanguagelearning}
Y.~Tay, M.~Dehghani, V.~Q. Tran, X.~Garcia, J.~Wei, X.~Wang, H.~W. Chung,
  S.~Shakeri, D.~Bahri, T.~Schuster, {\em et~al.}, ``Ul2: Unifying language
  learning paradigms,'' {\em arXiv preprint arXiv:2205.05131}, 2022.

\bibitem{radford2018improving}
A.~Radford, K.~Narasimhan, T.~Salimans, I.~Sutskever, {\em et~al.}, ``Improving
  language understanding by generative pre-training,'' 2018.

\bibitem{radford2019language}
A.~Radford, J.~Wu, R.~Child, D.~Luan, D.~Amodei, and I.~Sutskever, ``Language
  models are unsupervised multitask learners,'' 2019.

\bibitem{touvron2023llamaopenefficientfoundation}
H.~Touvron, T.~Lavril, G.~Izacard, X.~Martinet, M.-A. Lachaux, T.~Lacroix,
  B.~Rozi{\`e}re, N.~Goyal, E.~Hambro, F.~Azhar, {\em et~al.}, ``Llama: Open
  and efficient foundation language models,'' {\em arXiv preprint
  arXiv:2302.13971}, 2023.

\bibitem{touvron2023llama2openfoundation}
H.~Touvron, L.~Martin, K.~Stone, P.~Albert, A.~Almahairi, Y.~Babaei,
  N.~Bashlykov, S.~Batra, P.~Bhargava, S.~Bhosale, {\em et~al.}, ``Llama 2:
  Open foundation and fine-tuned chat models,'' {\em arXiv preprint
  arXiv:2307.09288}, 2023.

\bibitem{carion2020end}
N.~Carion, F.~Massa, G.~Synnaeve, N.~Usunier, A.~Kirillov, and S.~Zagoruyko,
  ``End-to-end object detection with transformers,'' in {\em European
  Conference on Computer Vision (ECCV)}, 2020.

\bibitem{zhu2021deformable}
X.~Zhu, W.~Su, L.~Lu, B.~Li, X.~Wang, and J.~Dai, ``Deformable {DETR}:
  Deformable transformers for end-to-end object detection,'' in {\em
  International Conference on Learning Representations (ICLR)}, 2021.

\bibitem{caron2021emergingpropertiesselfsupervisedvision}
M.~Caron, H.~Touvron, I.~Misra, H.~J{\'e}gou, J.~Mairal, P.~Bojanowski, and
  A.~Joulin, ``Emerging properties in self-supervised vision transformers,'' in
  {\em Proceedings of the IEEE/CVF international conference on computer
  vision}, pp.~9650--9660, 2021.

\bibitem{oquab2024dinov2learningrobustvisual}
M.~Oquab, T.~Darcet, T.~Moutakanni, H.~Vo, M.~Szafraniec, V.~Khalidov,
  P.~Fernandez, D.~Haziza, F.~Massa, A.~El-Nouby, {\em et~al.}, ``Dinov2:
  Learning robust visual features without supervision,'' {\em arXiv preprint
  arXiv:2304.07193}, 2023.

\bibitem{kirillov2023segment}
A.~Kirillov, E.~Mintun, N.~Ravi, H.~Mao, C.~Rolland, L.~Gustafson, T.~Xiao,
  S.~Whitehead, A.~C. Berg, W.-Y. Lo, {\em et~al.}, ``Segment anything,'' in
  {\em Proceedings of the IEEE/CVF international conference on computer
  vision}, pp.~4015--4026, 2023.

\bibitem{ravi2024sam2segmentimages}
N.~Ravi, V.~Gabeur, Y.-T. Hu, R.~Hu, C.~Ryali, T.~Ma, H.~Khedr, R.~R{\"a}dle,
  C.~Rolland, L.~Gustafson, {\em et~al.}, ``Sam 2: Segment anything in images
  and videos,'' {\em arXiv preprint arXiv:2408.00714}, 2024.

\bibitem{zou2023segment}
X.~Zou, J.~Yang, H.~Zhang, F.~Li, L.~Li, J.~Wang, L.~Wang, J.~Gao, and Y.~J.
  Lee, ``Segment everything everywhere all at once,'' {\em Advances in neural
  information processing systems}, vol.~36, pp.~19769--19782, 2023.

\bibitem{yu2022point}
X.~Yu, L.~Tang, Y.~Rao, T.~Huang, J.~Zhou, and J.~Lu, ``Point-{BERT}:
  Pre-training {3D} point cloud transformers with masked point modeling,'' in
  {\em IEEE/CVF Conference on Computer Vision and Pattern Recognition (CVPR)},
  2022.

\bibitem{qian2022pointnext}
G.~Qian, Y.~Li, H.~Peng, J.~Mai, H.~A. A.~K. Hammoud, M.~Elhoseiny, and
  B.~Ghanem, ``{PointNeXt}: Revisiting pointnet++ with improved training and
  scaling strategies,'' in {\em Advances in Neural Information Processing
  Systems (NeurIPS)}, 2022.

\bibitem{li2022bevformer}
Z.~Li, W.~Wang, H.~Li, E.~Xie, C.~Sima, T.~Lu, Q.~Yu, and J.~Dai,
  ``{BEVFormer}: Learning bird's-eye-view representation from multi-camera
  images via spatiotemporal transformers,'' in {\em European Conference on
  Computer Vision (ECCV)}, 2022.

\bibitem{cho2021unifying}
J.~Cho, J.~Lei, H.~Tan, and M.~Bansal, ``Unifying vision-and-language tasks via
  text generation,'' in {\em International Conference on Machine Learning},
  pp.~1931--1942, PMLR, 2021.

\bibitem{kim2021vilt}
W.~Kim, B.~Son, and I.~Kim, ``Vilt: Vision-and-language transformer without
  convolution or region supervision,'' in {\em International conference on
  machine learning}, pp.~5583--5594, PMLR, 2021.

\bibitem{radford2021learningtransferablevisualmodels}
A.~Radford, J.~W. Kim, C.~Hallacy, A.~Ramesh, G.~Goh, S.~Agarwal, G.~Sastry,
  A.~Askell, P.~Mishkin, J.~Clark, {\em et~al.}, ``Learning transferable visual
  models from natural language supervision,'' in {\em International conference
  on machine learning}, pp.~8748--8763, PmLR, 2021.

\bibitem{jia2021scalingvisualvisionlanguagerepresentation}
C.~Jia, Y.~Yang, Y.~Xia, Y.-T. Chen, Z.~Parekh, H.~Pham, Q.~Le, Y.-H. Sung,
  Z.~Li, and T.~Duerig, ``Scaling up visual and vision-language representation
  learning with noisy text supervision,'' in {\em International conference on
  machine learning}, pp.~4904--4916, PMLR, 2021.

\bibitem{bolya2025perception}
D.~Bolya, P.-Y. Huang, P.~Sun, J.~H. Cho, A.~Madotto, C.~Wei, T.~Ma, J.~Zhi,
  J.~Rajasegaran, H.~Rasheed, {\em et~al.}, ``Perception encoder: The best
  visual embeddings are not at the output of the network,'' {\em arXiv preprint
  arXiv:2504.13181}, 2025.

\bibitem{liu2024groundingdino}
S.~Liu, Z.~Zeng, T.~Ren, F.~Li, H.~Zhang, J.~Yang, Q.~Jiang, C.~Li, J.~Yang,
  H.~Su, {\em et~al.}, ``Grounding dino: Marrying dino with grounded
  pre-training for open-set object detection,'' in {\em European Conference on
  Computer Vision}, pp.~38--55, Springer, 2024.

\bibitem{ren2024groundedsam}
T.~Ren, S.~Liu, A.~Zeng, J.~Lin, K.~Li, H.~Cao, J.~Chen, X.~Huang, Y.~Chen,
  F.~Yan, {\em et~al.}, ``Grounded sam: Assembling open-world models for
  diverse visual tasks,'' {\em arXiv preprint arXiv:2401.14159}, 2024.

\bibitem{baevski2020wav2vec}
A.~Baevski, Y.~Zhou, A.~Mohamed, and M.~Auli, ``wav2vec 2.0: A framework for
  self-supervised learning of speech representations,'' {\em Advances in Neural
  Information Processing Systems}, vol.~33, pp.~12449--12460, 2020.

\bibitem{hsu2021hubert}
W.-N. Hsu, B.~Bolte, Y.-H.~H. Tsai, K.~Lakhotia, R.~Salakhutdinov, and
  A.~Mohamed, ``{HuBERT}: Self-supervised speech representation learning by
  masked prediction of hidden units,'' {\em IEEE/ACM Transactions on Audio,
  Speech, and Language Processing}, vol.~29, pp.~3451--3460, 2021.

\bibitem{radford2023whisper}
A.~Radford, J.~W. Gao, C.~Roberts, {\em et~al.}, ``Robust speech recognition
  via large-scale weak supervision,'' {\em arXiv preprint arXiv:2212.04356},
  2023.

\bibitem{guzhov2022audioclip}
A.~Guzhov, S.~Shokhrankhani, B.~Elizalde, F.-R. St{\"o}ter, and T.~Kuhn,
  ``{AudioCLIP}: Extending {CLIP} to image, text and audio,'' in {\em ICASSP
  2022--IEEE International Conference on Acoustics, Speech and Signal
  Processing (ICASSP)}, pp.~976--980, IEEE, 2022.

\bibitem{akbari2021vatt}
H.~Akbari, L.~Yuan, R.~Qian, C.-H. Chuang, S.-F. Chang, and B.~Gong, ``{VATT}:
  Transformers for multimodal self-supervised learning from raw video, audio
  and text,'' {\em arXiv preprint arXiv:2104.11178}, 2021.

\bibitem{kreuk2023audiogen}
F.~Kreuk, C.-Y. Yeh, A.~Polyak, D.~Lavi, O.~Litany, J.~Ren, T.~Shacham,
  A.~Globerson, and L.~Wolf, ``{AudioGen}: Textually guided audio generation,''
  {\em arXiv preprint arXiv:2209.15352}, 2023.

\bibitem{xu2021videoclip}
H.~Xu, G.~Ghosh, P.-Y. Huang, D.~Okhonko, A.~Aghajanyan, F.~Metze,
  L.~Zettlemoyer, and C.~Feichtenhofer, ``Videoclip: Contrastive pre-training
  for zero-shot video-text understanding,'' {\em arXiv preprint
  arXiv:2109.14084}, 2021.

\bibitem{bain2021frozenintime}
M.~Bain, A.~Nagrani, G.~Varol, and A.~Zisserman, ``Frozen in time: A joint
  video and image encoder for end-to-end retrieval,'' in {\em Proceedings of
  the IEEE/CVF international conference on computer vision}, pp.~1728--1738,
  2021.

\bibitem{wang2023allinone}
J.~Wang, Y.~Ge, R.~Yan, Y.~Ge, K.~Q. Lin, S.~Tsutsui, X.~Lin, G.~Cai, J.~Wu,
  Y.~Shan, {\em et~al.}, ``All in one: Exploring unified video-language
  pre-training,'' in {\em Proceedings of the IEEE/CVF Conference on Computer
  Vision and Pattern Recognition}, pp.~6598--6608, 2023.

\bibitem{videoworldsimulators2024}
T.~Brooks, B.~Peebles, C.~Holmes, W.~DePue, Y.~Guo, L.~Jing, D.~Schnurr,
  J.~Taylor, T.~Luhman, E.~Luhman, {\em et~al.}, ``Video generation models as
  world simulators,'' 2024.

\bibitem{singer2022makeavideo}
U.~Singer, A.~Polyak, T.~Hayes, X.~Yin, J.~An, S.~Zhang, Q.~Hu, H.~Yang,
  O.~Ashual, O.~Gafni, {\em et~al.}, ``Make-a-video: Text-to-video generation
  without text-video data,'' {\em arXiv preprint arXiv:2209.14792}, 2022.

\bibitem{wu2022nuwa}
C.~Wu, J.~Liang, L.~Ji, F.~Yang, Y.~Fang, D.~Jiang, and N.~Duan, ``N{\"u}wa:
  Visual synthesis pre-training for neural visual world creation,'' in {\em
  European conference on computer vision}, pp.~720--736, Springer, 2022.

\bibitem{villegas2022phenaki}
R.~Villegas, M.~Babaeizadeh, P.-J. Kindermans, H.~Moraldo, H.~Zhang, M.~T.
  Saffar, S.~Castro, J.~Kunze, and D.~Erhan, ``Phenaki: Variable length video
  generation from open domain textual description,'' {\em arXiv preprint
  arXiv:2210.02399}, 2022.

\bibitem{dave2024multimodal}
V.~Dave, F.~Lygerakis, and E.~Rueckert, ``Multimodal visual-tactile
  representation learning through self-supervised contrastive pre-training,''
  in {\em 2024 IEEE International Conference on Robotics and Automation
  (ICRA)}, pp.~8013--8020, IEEE, 2024.

\bibitem{kerr2022self}
J.~Kerr, H.~Huang, A.~Wilcox, R.~Hoque, J.~Ichnowski, R.~Calandra, and
  K.~Goldberg, ``Self-supervised visuo-tactile pretraining to locate and follow
  garment features,'' {\em arXiv preprint arXiv:2209.13042}, 2022.

\bibitem{yang2024binding}
F.~Yang, C.~Feng, Z.~Chen, H.~Park, D.~Wang, Y.~Dou, Z.~Zeng, X.~Chen,
  R.~Gangopadhyay, A.~Owens, {\em et~al.}, ``Binding touch to everything:
  Learning unified multimodal tactile representations,'' in {\em Proceedings of
  the IEEE/CVF Conference on Computer Vision and Pattern Recognition},
  pp.~26340--26353, 2024.

\bibitem{zhong2023touchnerf}
S.~Zhong, A.~Albini, O.~P. Jones, P.~Maiolino, and I.~Posner, ``Touching a
  nerf: Leveraging neural radiance fields for tactile sensory data
  generation,'' in {\em Conference on Robot Learning}, pp.~1618--1628, PMLR,
  2023.

\bibitem{yang2022touchgo}
F.~Yang, C.~Ma, J.~Zhang, J.~Zhu, W.~Yuan, and A.~Owens, ``Touch and go:
  Learning from human-collected vision and touch,'' {\em arXiv preprint
  arXiv:2211.12498}, 2022.

\bibitem{fu2024touch}
L.~Fu, G.~Datta, H.~Huang, W.~C.-H. Panitch, J.~Drake, J.~Ortiz, M.~Mukadam,
  M.~Lambeta, R.~Calandra, and K.~Goldberg, ``A touch, vision, and language
  dataset for multimodal alignment,'' {\em arXiv preprint arXiv:2402.13232},
  2024.

\bibitem{he2015deepresiduallearningimage}
K.~He, X.~Zhang, S.~Ren, and J.~Sun, ``Deep residual learning for image
  recognition,'' in {\em Proceedings of the IEEE conference on computer vision
  and pattern recognition}, pp.~770--778, 2016.

\bibitem{dosovitskiy2021imageworth16x16words}
A.~Dosovitskiy, L.~Beyer, A.~Kolesnikov, D.~Weissenborn, X.~Zhai,
  T.~Unterthiner, M.~Dehghani, M.~Minderer, G.~Heigold, S.~Gelly, {\em et~al.},
  ``An image is worth 16x16 words: Transformers for image recognition at
  scale,'' {\em arXiv preprint arXiv:2010.11929}, 2020.

\bibitem{liu2021swintransformerhierarchicalvision}
Z.~Liu, Y.~Lin, Y.~Cao, H.~Hu, Y.~Wei, Z.~Zhang, S.~Lin, and B.~Guo, ``Swin
  transformer: Hierarchical vision transformer using shifted windows,'' in {\em
  Proceedings of the IEEE/CVF international conference on computer vision},
  pp.~10012--10022, 2021.

\bibitem{tan2019lxmert}
H.~Tan and M.~Bansal, ``Lxmert: Learning cross-modality encoder representations
  from transformers,'' {\em arXiv preprint arXiv:1908.07490}, 2019.

\bibitem{lu2019vilbert}
J.~Lu, D.~Batra, D.~Parikh, and S.~Lee, ``Vilbert: Pretraining task-agnostic
  visiolinguistic representations for vision-and-language tasks,'' {\em
  Advances in neural information processing systems}, vol.~32, 2019.

\bibitem{su2019vl}
W.~Su, X.~Zhu, Y.~Cao, B.~Li, L.~Lu, F.~Wei, and J.~Dai, ``Vl-bert:
  Pre-training of generic visual-linguistic representations,'' {\em arXiv
  preprint arXiv:1908.08530}, 2019.

\bibitem{chen2020uniter}
Y.-C. Chen, L.~Li, L.~Yu, A.~El~Kholy, F.~Ahmed, Z.~Gan, Y.~Cheng, and J.~Liu,
  ``Uniter: Universal image-text representation learning,'' in {\em European
  conference on computer vision}, pp.~104--120, Springer, 2020.

\bibitem{li2020oscar}
X.~Li, X.~Yin, C.~Li, P.~Zhang, X.~Hu, L.~Zhang, L.~Wang, H.~Hu, L.~Dong,
  F.~Wei, {\em et~al.}, ``Oscar: Object-semantics aligned pre-training for
  vision-language tasks,'' in {\em Computer Vision--ECCV 2020: 16th European
  Conference, Glasgow, UK, August 23--28, 2020, Proceedings, Part XXX 16},
  pp.~121--137, Springer, 2020.

\bibitem{zhang2019cognitivefunctionsbrainperception}
J.~Zhang, ``Cognitive functions of the brain: perception, attention and
  memory,'' {\em arXiv preprint arXiv:1907.02863}, 2019.

\bibitem{smith2005development}
L.~Smith and M.~Gasser, ``The development of embodied cognition: Six lessons
  from babies,'' {\em Artificial life}, vol.~11, no.~1-2, pp.~13--29, 2005.

\bibitem{chen2024pcabenchevaluatingmultimodallarge}
L.~Chen, Y.~Zhang, S.~Ren, H.~Zhao, Z.~Cai, Y.~Wang, P.~Wang, X.~Meng, T.~Liu,
  and B.~Chang, ``Pca-bench: Evaluating multimodal large language models in
  perception-cognition-action chain,'' {\em arXiv preprint arXiv:2402.15527},
  2024.

\bibitem{chia2024can}
Y.~K. Chia, Q.~Sun, L.~Bing, and S.~Poria, ``{Can-Do}! {A} dataset and
  neuro-symbolic grounded framework for embodied planning with large multimodal
  models,'' {\em arXiv preprint arXiv:2409.14277}, 2024.

\bibitem{zhou2024walle}
S.~Zhou, T.~Zhou, Y.~Yang, G.~Long, D.~Ye, J.~Jiang, and C.~Zhang, ``{WALL-E}:
  World alignment by rule learning improves world model-based llm agents,''
  {\em arXiv preprint arXiv:2410.07484}, 2024.

\bibitem{li2022blipbootstrappinglanguageimagepretraining}
J.~Li, D.~Li, C.~Xiong, and S.~Hoi, ``Blip: Bootstrapping language-image
  pre-training for unified vision-language understanding and generation,'' in
  {\em International conference on machine learning}, pp.~12888--12900, PMLR,
  2022.

\bibitem{alayrac2022flamingovisuallanguagemodel}
J.-B. Alayrac, J.~Donahue, P.~Luc, A.~Miech, I.~Barr, Y.~Hasson, K.~Lenc,
  A.~Mensch, K.~Millican, M.~Reynolds, {\em et~al.}, ``Flamingo: a visual
  language model for few-shot learning,'' {\em Advances in neural information
  processing systems}, vol.~35, pp.~23716--23736, 2022.

\bibitem{li2019visualbertsimpleperformantbaseline}
L.~H. Li, M.~Yatskar, D.~Yin, C.-J. Hsieh, and K.-W. Chang, ``Visualbert: A
  simple and performant baseline for vision and language,'' {\em arXiv preprint
  arXiv:1908.03557}, 2019.

\bibitem{yu2022cocacontrastivecaptionersimagetext}
J.~Yu, Z.~Wang, V.~Vasudevan, L.~Yeung, M.~Seyedhosseini, and Y.~Wu, ``Coca:
  Contrastive captioners are image-text foundation models,'' {\em arXiv
  preprint arXiv:2205.01917}, 2022.

\bibitem{yang2023dawnlmmspreliminaryexplorations}
Z.~Yang, L.~Li, K.~Lin, J.~Wang, C.-C. Lin, Z.~Liu, and L.~Wang, ``The dawn of
  lmms: Preliminary explorations with gpt-4v (ision),'' {\em arXiv preprint
  arXiv:2309.17421}, vol.~9, no.~1, p.~1, 2023.

\bibitem{gpt-4o}
A.~Hurst, A.~Lerer, A.~P. Goucher, A.~Perelman, A.~Ramesh, A.~Clark, A.~Ostrow,
  A.~Welihinda, A.~Hayes, A.~Radford, {\em et~al.}, ``{GPT}-4o system card,''
  {\em arXiv preprint arXiv:2410.21276}, 2024.

\bibitem{geminiteam2024geminifamilyhighlycapable}
G.~Team, R.~Anil, S.~Borgeaud, J.-B. Alayrac, J.~Yu, R.~Soricut, J.~Schalkwyk,
  A.~M. Dai, A.~Hauth, K.~Millican, {\em et~al.}, ``Gemini: a family of highly
  capable multimodal models,'' {\em arXiv preprint arXiv:2312.11805}, 2023.

\bibitem{liu2024improvedbaselinesvisualinstruction}
H.~Liu, C.~Li, Y.~Li, and Y.~J. Lee, ``Improved baselines with visual
  instruction tuning,'' in {\em Proceedings of the IEEE/CVF Conference on
  Computer Vision and Pattern Recognition}, pp.~26296--26306, 2024.

\bibitem{bai2023qwentechnicalreport}
J.~Bai, S.~Bai, Y.~Chu, Z.~Cui, K.~Dang, X.~Deng, Y.~Fan, W.~Ge, Y.~Han,
  F.~Huang, {\em et~al.}, ``Qwen technical report,'' {\em arXiv preprint
  arXiv:2309.16609}, 2023.

\bibitem{chen2025expandingperformanceboundariesopensource}
Z.~Chen, W.~Wang, Y.~Cao, Y.~Liu, Z.~Gao, E.~Cui, J.~Zhu, S.~Ye, H.~Tian,
  Z.~Liu, {\em et~al.}, ``Expanding performance boundaries of open-source
  multimodal models with model, data, and test-time scaling,'' {\em arXiv
  preprint arXiv:2412.05271}, 2024.

\bibitem{yang2025embodiedbenchcomprehensivebenchmarkingmultimodal}
R.~Yang, H.~Chen, J.~Zhang, M.~Zhao, C.~Qian, K.~Wang, Q.~Wang, T.~V.
  Koripella, M.~Movahedi, M.~Li, {\em et~al.}, ``Embodiedbench: Comprehensive
  benchmarking multi-modal large language models for vision-driven embodied
  agents,'' {\em arXiv preprint arXiv:2502.09560}, 2025.

\bibitem{cheng2025embodiedevalevaluatemultimodalllms}
Z.~Cheng, Y.~Tu, R.~Li, S.~Dai, J.~Hu, S.~Hu, J.~Li, Y.~Shi, T.~Yu, W.~Chen,
  {\em et~al.}, ``Embodiedeval: Evaluate multimodal llms as embodied agents,''
  {\em arXiv preprint arXiv:2501.11858}, 2025.

\bibitem{ma2025deepperceptionadvancingr1likecognitive}
X.~Ma, Z.~Ding, Z.~Luo, C.~Chen, Z.~Guo, D.~F. Wong, X.~Feng, and M.~Sun,
  ``Deepperception: Advancing r1-like cognitive visual perception in mllms for
  knowledge-intensive visual grounding,'' {\em arXiv preprint
  arXiv:2503.12797}, 2025.

\bibitem{fu2024mmesurveycomprehensivesurveyevaluation}
C.~Fu, Y.-F. Zhang, S.~Yin, B.~Li, X.~Fang, S.~Zhao, H.~Duan, X.~Sun, Z.~Liu,
  L.~Wang, {\em et~al.}, ``Mme-survey: A comprehensive survey on evaluation of
  multimodal llms,'' {\em arXiv preprint arXiv:2411.15296}, 2024.

\bibitem{zheng2022jarvis}
K.~Zheng, K.~Zhou, J.~Gu, Y.~Fan, J.~Wang, Z.~Di, X.~He, and X.~E. Wang,
  ``Jarvis: A neuro-symbolic commonsense reasoning framework for conversational
  embodied agents,'' {\em arXiv preprint arXiv:2208.13266}, 2022.

\bibitem{choi2025nesyc}
W.~Choi, J.~Park, S.~Ahn, D.~Lee, and H.~Woo, ``{NeSyC}: A neuro-symbolic
  continual learner for complex embodied tasks in open domains,'' {\em arXiv
  preprint arXiv:2503.00870}, 2025.

\bibitem{wang2025understanding}
M.~Wang, ``From understanding the world to intervening in it: A unified
  multi-scale framework for embodied cognition,'' {\em arXiv preprint
  arXiv:2503.00727}, 2025.

\bibitem{nottingham2023deckard}
K.~Nottingham, P.~Ammanabrolu, A.~Suhr, Y.~Choi, H.~Hajishirzi, S.~Singh, and
  R.~Fox, ``Do embodied agents dream of pixelated sheep: Embodied decision
  making using language guided world modelling,'' {\em arXiv preprint
  arXiv:2301.12050}, 2023.

\bibitem{liu2024glimo}
H.~Liu and J.~Zhao, ``Grounding large language models in embodied environment
  with imperfect world models,'' {\em arXiv preprint arXiv:2410.02742}, 2024.

\bibitem{mazzaglia2024genrl}
P.~Mazzaglia, T.~Verbelen, B.~Dhoedt, A.~Courville, and S.~Rajeswar, ``{GenRL}:
  Multimodal-foundation world models for generalization in embodied agents,''
  {\em arXiv preprint arXiv:2406.18043}, 2024.

\bibitem{yao2024aeroverse}
F.~Yao, Y.~Yue, Y.~Liu, X.~Sun, and K.~Fu, ``{AeroVerse}: {UAV}-agent benchmark
  suite for simulating, pre-training, finetuning, and evaluating aerospace
  embodied world models,'' {\em arXiv preprint arXiv:2408.15511}, 2024.

\bibitem{primenet}
Q.~Liu, S.~Han, Y.~Li, E.~Cambria, and K.~Kwok, ``{PrimeNet}: A framework for
  commonsense knowledge representation and reasoning based on conceptual
  primitives,'' {\em Cognitive Computation}, vol.~16, no.~6, pp.~3429--3456,
  2024.

\bibitem{mai2024efficient}
X.~Mai, Z.~Tao, J.~Lin, H.~Wang, Y.~Chang, Y.~Kang, Y.~Wang, and W.~Zhang,
  ``From efficient multimodal models to world models: A survey,'' {\em arXiv
  preprint arXiv:2407.00118}, 2024.

\bibitem{zhu2024sora}
Z.~Zhu, X.~Wang, W.~Zhao, C.~Min, N.~Deng, M.~Dou, Y.~Wang, B.~Shi, K.~Wang,
  C.~Zhang, {\em et~al.}, ``Is {Sora} a world simulator? a comprehensive survey
  on general world models and beyond,'' {\em arXiv preprint arXiv:2405.03520},
  2024.

\bibitem{ridderinkhof2014neurocognitive}
K.~R. Ridderinkhof, ``Neurocognitive mechanisms of perception--action
  coordination: A review and theoretical integration,'' {\em Neuroscience \&
  Biobehavioral Reviews}, vol.~46, pp.~3--29, 2014.

\bibitem{fuster2004upper}
J.~M. Fuster, ``Upper processing stages of the perception--action cycle,'' {\em
  Trends in cognitive sciences}, vol.~8, no.~4, pp.~143--145, 2004.

\bibitem{cassimatis2004integrating}
N.~L. Cassimatis, J.~G. Trafton, M.~D. Bugajska, and A.~C. Schultz,
  ``Integrating cognition, perception and action through mental simulation in
  robots,'' {\em Robotics and Autonomous Systems}, vol.~49, no.~1-2,
  pp.~13--23, 2004.

\bibitem{liutwo}
L.~Liu, D.~Zhang, S.~Li, G.~Zhou, and E.~Cambria, ``Two heads are better than
  one: Zero-shot cognitive reasoning via multi-llm knowledge fusion,'' in {\em
  Proceedings of {CIKM}}, pp.~1462--1472, 2024.

\bibitem{andrew2001conditioned}
A.~Salter, {\em Conditioned reflex therapy}.
\newblock Wellness Institute, Inc., 2001.

\bibitem{huang2001planning}
Q.~Huang, K.~Yokoi, S.~Kajita, K.~Kaneko, H.~Arai, N.~Koyachi, and K.~Tanie,
  ``Planning walking patterns for a biped robot,'' {\em IEEE TRO}, vol.~17,
  no.~3, pp.~280--289, 2001.

\bibitem{biswal2021development}
P.~Biswal and P.~K. Mohanty, ``Development of quadruped walking robots: A
  review,'' {\em Ain Shams Engineering Journal}, vol.~12, no.~2,
  pp.~2017--2031, 2021.

\bibitem{juang2013design}
H.-S. Juang and K.-Y. Lum, ``Design and control of a two-wheel self-balancing
  robot using the arduino microcontroller board,'' in {\em 2013 10th IEEE
  International Conference on Control and Automation (ICCA)}, pp.~634--639,
  IEEE, 2013.

\bibitem{lin2008development}
S.-C. Lin and C.-C. Tsai, ``Development of a self-balancing human
  transportation vehicle for the teaching of feedback control,'' {\em IEEE
  Transactions on Education}, vol.~52, no.~1, pp.~157--168, 2008.

\bibitem{kaelbling1996reinforcement}
L.~P. Kaelbling, M.~L. Littman, and A.~W. Moore, ``Reinforcement learning: A
  survey,'' {\em Journal of artificial intelligence research}, vol.~4,
  pp.~237--285, 1996.

\bibitem{kober2013reinforcement}
J.~Kober, J.~A. Bagnell, and J.~Peters, ``Reinforcement learning in robotics: A
  survey,'' {\em The International Journal of Robotics Research}, vol.~32,
  no.~11, pp.~1238--1274, 2013.

\bibitem{sheins}
T.~Shen, E.~Cambria, J.~Wang, Y.~Cai, and X.~Zhang, ``Insight at the right
  spot: Provide decisive subgraph information to graph llm with reinforcement
  learning,'' {\em Information Fusion}, vol.~117, p.~102860, 2025.

\bibitem{vukobratovic2004zero}
M.~Vukobratovi{\'c} and B.~Borovac, ``Zero-moment point-thirty five years of
  its life,'' {\em International journal of humanoid robotics}, vol.~1, no.~01,
  pp.~157--173, 2004.

\bibitem{schwenzer2021review}
M.~Schwenzer, M.~Ay, T.~Bergs, and D.~Abel, ``Review on model predictive
  control: An engineering perspective,'' {\em The International Journal of
  Advanced Manufacturing Technology}, vol.~117, no.~5, pp.~1327--1349, 2021.

\bibitem{yang2023neural}
R.~Yang, G.~Yang, and X.~Wang, ``Neural volumetric memory for visual locomotion
  control,'' in {\em Proceedings of the IEEE/CVF conference on computer vision
  and pattern recognition}, pp.~1430--1440, 2023.

\bibitem{zeng2021transporter}
A.~Zeng, P.~Florence, J.~Tompson, S.~Welker, J.~Chien, M.~Attarian,
  T.~Armstrong, I.~Krasin, D.~Duong, V.~Sindhwani, {\em et~al.}, ``Transporter
  networks: Rearranging the visual world for robotic manipulation,'' in {\em
  Conference on Robot Learning}, pp.~726--747, PMLR, 2021.

\bibitem{shridhar2022cliport}
M.~Shridhar, L.~Manuelli, and D.~Fox, ``Cliport: What and where pathways for
  robotic manipulation,'' in {\em Conference on robot learning}, pp.~894--906,
  PMLR, 2022.

\bibitem{jang2022bc}
E.~Jang, A.~Irpan, M.~Khansari, D.~Kappler, F.~Ebert, C.~Lynch, S.~Levine, and
  C.~Finn, ``Bc-z: Zero-shot task generalization with robotic imitation
  learning,'' in {\em Conference on Robot Learning}, pp.~991--1002, PMLR, 2022.

\bibitem{zhang2024vision}
Y.~Zhang, Z.~Ma, J.~Li, Y.~Qiao, Z.~Wang, J.~Chai, Q.~Wu, M.~Bansal, and
  P.~Kordjamshidi, ``Vision-and-language navigation today and tomorrow: A
  survey in the era of foundation models,'' {\em arXiv preprint
  arXiv:2407.07035}, 2024.

\bibitem{long2024instructnav}
Y.~Long, W.~Cai, H.~Wang, G.~Zhan, and H.~Dong, ``Instructnav: Zero-shot system
  for generic instruction navigation in unexplored environment,'' {\em arXiv
  preprint arXiv:2406.04882}, 2024.

\bibitem{belkhale2024rt}
S.~Belkhale, T.~Ding, T.~Xiao, P.~Sermanet, Q.~Vuong, J.~Tompson, Y.~Chebotar,
  D.~Dwibedi, and D.~Sadigh, ``Rt-h: Action hierarchies using language,'' {\em
  arXiv preprint arXiv:2403.01823}, 2024.

\bibitem{vuong2023open}
Q.~Vuong, S.~Levine, H.~R. Walke, K.~Pertsch, A.~Singh, R.~Doshi, C.~Xu,
  J.~Luo, L.~Tan, D.~Shah, {\em et~al.}, ``Open x-embodiment: Robotic learning
  datasets and rt-x models,'' in {\em Towards Generalist Robots: Learning
  Paradigms for Scalable Skill Acquisition@ CoRL2023}, 2023.

\bibitem{an2023bevbert}
D.~An, Y.~Qi, Y.~Li, Y.~Huang, L.~Wang, T.~Tan, and J.~Shao, ``Bevbert:
  Multimodal map pre-training for language-guided navigation,'' {\em arXiv
  preprint arXiv:2212.04385}, 2022.

\bibitem{zhou2024navgpt}
G.~Zhou, Y.~Hong, and Q.~Wu, ``Navgpt: Explicit reasoning in
  vision-and-language navigation with large language models,'' in {\em AAAI},
  vol.~38, pp.~7641--7649, 2024.

\bibitem{dou2022foam}
Z.-Y. Dou and N.~Peng, ``Foam: A follower-aware speaker model for
  vision-and-language navigation,'' {\em arXiv preprint arXiv:2206.04294},
  2022.

\bibitem{lynch2020language}
C.~Lynch and P.~Sermanet, ``Language conditioned imitation learning over
  unstructured data,'' {\em arXiv preprint arXiv:2005.07648}, 2020.

\bibitem{brohan2022rt}
A.~Brohan, N.~Brown, J.~Carbajal, Y.~Chebotar, J.~Dabis, C.~Finn,
  K.~Gopalakrishnan, K.~Hausman, A.~Herzog, J.~Hsu, {\em et~al.}, ``Rt-1:
  Robotics transformer for real-world control at scale,'' {\em arXiv preprint
  arXiv:2212.06817}, 2022.

\bibitem{chebotar2023q}
Y.~Chebotar, Q.~Vuong, K.~Hausman, F.~Xia, Y.~Lu, A.~Irpan, A.~Kumar, T.~Yu,
  A.~Herzog, K.~Pertsch, {\em et~al.}, ``Q-transformer: Scalable offline
  reinforcement learning via autoregressive {Q}-functions,'' in {\em Conference
  on Robot Learning}, pp.~3909--3928, PMLR, 2023.

\bibitem{goyal2023rvt}
A.~Goyal, J.~Xu, Y.~Guo, V.~Blukis, Y.-W. Chao, and D.~Fox, ``Rvt: Robotic view
  transformer for {3D} object manipulation,'' in {\em Conference on Robot
  Learning}, pp.~694--710, PMLR, 2023.

\bibitem{goyal2024rvt}
A.~Goyal, V.~Blukis, J.~Xu, Y.~Guo, Y.-W. Chao, and D.~Fox, ``Rvt-2: Learning
  precise manipulation from few demonstrations,'' {\em arXiv preprint
  arXiv:2406.08545}, 2024.

\bibitem{su2024motion}
Y.~Su, X.~Zhan, H.~Fang, Y.-L. Li, C.~Lu, and L.~Yang, ``Motion before action:
  Diffusing object motion as manipulation condition,'' {\em arXiv preprint
  arXiv:2411.09658}, 2024.

\bibitem{ze20243d}
Y.~Ze, G.~Zhang, K.~Zhang, C.~Hu, M.~Wang, and H.~Xu, ``{3D} diffusion policy:
  Generalizable visuomotor policy learning via simple {3D} representations,''
  {\em arXiv preprint arXiv:2403.03954}, 2024.

\bibitem{huang2024rekep}
W.~Huang, C.~Wang, Y.~Li, R.~Zhang, and L.~Fei-Fei, ``Rekep: Spatio-temporal
  reasoning of relational keypoint constraints for robotic manipulation,'' {\em
  arXiv preprint arXiv:2409.01652}, 2024.

\bibitem{nasiriany2024pivot}
S.~Nasiriany, F.~Xia, W.~Yu, T.~Xiao, J.~Liang, I.~Dasgupta, A.~Xie, D.~Driess,
  A.~Wahid, Z.~Xu, {\em et~al.}, ``Pivot: Iterative visual prompting elicits
  actionable knowledge for {VLMs},'' {\em arXiv preprint arXiv:2402.07872},
  2024.

\bibitem{anderson2018vision}
P.~Anderson, Q.~Wu, D.~Teney, J.~Bruce, M.~Johnson, N.~S{\"u}nderhauf, I.~Reid,
  S.~Gould, and A.~Van Den~Hengel, ``Vision-and-language navigation:
  Interpreting visually-grounded navigation instructions in real
  environments,'' in {\em Proceedings of the IEEE conference on computer vision
  and pattern recognition}, pp.~3674--3683, 2018.

\bibitem{fried2018speaker}
D.~Fried, R.~Hu, V.~Cirik, A.~Rohrbach, J.~Andreas, L.-P. Morency,
  T.~Berg-Kirkpatrick, K.~Saenko, D.~Klein, and T.~Darrell, ``Speaker-follower
  models for vision-and-language navigation,'' {\em Advances in neural
  information processing systems}, vol.~31, 2018.

\bibitem{anderson2021sim}
P.~Anderson, A.~Shrivastava, J.~Truong, A.~Majumdar, D.~Parikh, D.~Batra, and
  S.~Lee, ``Sim-to-real transfer for vision-and-language navigation,'' in {\em
  Conference on Robot Learning}, pp.~671--681, PMLR, 2021.

\bibitem{krantz2021waypoint}
J.~Krantz, A.~Gokaslan, D.~Batra, S.~Lee, and O.~Maksymets, ``Waypoint models
  for instruction-guided navigation in continuous environments,'' in {\em
  ICCV}, pp.~15162--15171, 2021.

\bibitem{chen2021history}
S.~Chen, P.-L. Guhur, C.~Schmid, and I.~Laptev, ``History aware multimodal
  transformer for vision-and-language navigation,'' {\em NeurIPS}, vol.~34,
  pp.~5834--5847, 2021.

\bibitem{hong2022bridging}
Y.~Hong, Z.~Wang, Q.~Wu, and S.~Gould, ``Bridging the gap between learning in
  discrete and continuous environments for vision-and-language navigation,'' in
  {\em CVPR}, pp.~15439--15449, 2022.

\bibitem{vasudevan2021talk2nav}
A.~B. Vasudevan, D.~Dai, and L.~Van~Gool, ``Talk2nav: Long-range
  vision-and-language navigation with dual attention and spatial memory,'' {\em
  International Journal of Computer Vision}, vol.~129, pp.~246--266, 2021.

\bibitem{georgakis2022cross}
G.~Georgakis, K.~Schmeckpeper, K.~Wanchoo, S.~Dan, E.~Miltsakaki, D.~Roth, and
  K.~Daniilidis, ``Cross-modal map learning for vision and language
  navigation,'' in {\em CVPR}, pp.~15460--15470, 2022.

\bibitem{brohan2023rt}
A.~Brohan, N.~Brown, J.~Carbajal, Y.~Chebotar, X.~Chen, K.~Choromanski,
  T.~Ding, D.~Driess, A.~Dubey, C.~Finn, {\em et~al.}, ``Rt-2:
  Vision-language-action models transfer web knowledge to robotic control,''
  {\em arXiv preprint arXiv:2307.15818}, 2023.

\bibitem{dorbala2022clip}
V.~S. Dorbala, G.~Sigurdsson, R.~Piramuthu, J.~Thomason, and G.~S. Sukhatme,
  ``{CLIP-Nav}: Using clip for zero-shot vision-and-language navigation,'' {\em
  arXiv preprint arXiv:2211.16649}, 2022.

\bibitem{zhang2024navid}
J.~Zhang, K.~Wang, R.~Xu, G.~Zhou, Y.~Hong, X.~Fang, Q.~Wu, Z.~Zhang, and
  H.~Wang, ``Navid: Video-based vlm plans the next step for vision-and-language
  navigation,'' in {\em Robotics: Science and Systems}, 2024.

\bibitem{zhou2024navgpt2}
G.~Zhou, Y.~Hong, Z.~Wang, X.~E. Wang, and Q.~Wu, ``Navgpt-2: Unleashing
  navigational reasoning capability for large vision-language models,'' in {\em
  European Conference on Computer Vision}, pp.~260--278, Springer, 2024.

\bibitem{chen2024mapgpt}
J.~Chen, B.~Lin, R.~Xu, Z.~Chai, X.~Liang, and K.-Y.~K. Wong, ``Mapgpt:
  Map-guided prompting for unified vision-and-language navigation,'' {\em arXiv
  e-prints}, pp.~arXiv--2401, 2024.

\bibitem{ehsani2021manipulathor}
K.~Ehsani, W.~Han, A.~Herrasti, E.~VanderBilt, L.~Weihs, E.~Kolve, A.~Kembhavi,
  and R.~Mottaghi, ``Manipulathor: A framework for visual object
  manipulation,'' in {\em CVPR}, pp.~4497--4506, 2021.

\bibitem{saxena2008robotic}
A.~Saxena, J.~Driemeyer, and A.~Y. Ng, ``Robotic grasping of novel objects
  using vision,'' {\em The International Journal of Robotics Research},
  vol.~27, no.~2, pp.~157--173, 2008.

\bibitem{mees2022matters}
O.~Mees, L.~Hermann, and W.~Burgard, ``What matters in language conditioned
  robotic imitation learning over unstructured data,'' {\em IEEE Robotics and
  Automation Letters}, vol.~7, no.~4, pp.~11205--11212, 2022.

\bibitem{mees2023grounding}
O.~Mees, J.~Borja-Diaz, and W.~Burgard, ``Grounding language with visual
  affordances over unstructured data,'' in {\em ICRA}, pp.~11576--11582, IEEE,
  2023.

\bibitem{du2023learning}
Y.~Du, S.~Yang, B.~Dai, H.~Dai, O.~Nachum, J.~Tenenbaum, D.~Schuurmans, and
  P.~Abbeel, ``Learning universal policies via text-guided video generation,''
  {\em Advances in neural information processing systems}, vol.~36,
  pp.~9156--9172, 2023.

\bibitem{li2023vision}
X.~Li, M.~Liu, H.~Zhang, C.~Yu, J.~Xu, H.~Wu, C.~Cheang, Y.~Jing, W.~Zhang,
  H.~Liu, {\em et~al.}, ``Vision-language foundation models as effective robot
  imitators,'' {\em arXiv preprint arXiv:2311.01378}, 2023.

\bibitem{reed2022generalist}
S.~Reed, K.~Zolna, E.~Parisotto, S.~G. Colmenarejo, A.~Novikov, G.~Barth-Maron,
  M.~Gimenez, Y.~Sulsky, J.~Kay, J.~T. Springenberg, {\em et~al.}, ``A
  generalist agent,'' {\em arXiv preprint arXiv:2205.06175}, 2022.

\bibitem{gu2023rt}
J.~Gu, S.~Kirmani, P.~Wohlhart, Y.~Lu, M.~G. Arenas, K.~Rao, W.~Yu, C.~Fu,
  K.~Gopalakrishnan, Z.~Xu, {\em et~al.}, ``Rt-trajectory: Robotic task
  generalization via hindsight trajectory sketches,'' {\em arXiv preprint
  arXiv:2311.01977}, 2023.

\bibitem{lynch2023interactive}
C.~Lynch, A.~Wahid, J.~Tompson, T.~Ding, J.~Betker, R.~Baruch, T.~Armstrong,
  and P.~Florence, ``Interactive language: Talking to robots in real time,''
  {\em IEEE Robotics and Automation Letters}, 2023.

\bibitem{bharadhwaj2024roboagent}
H.~Bharadhwaj, J.~Vakil, M.~Sharma, A.~Gupta, S.~Tulsiani, and V.~Kumar,
  ``Roboagent: Generalization and efficiency in robot manipulation via semantic
  augmentations and action chunking,'' in {\em 2024 IEEE International
  Conference on Robotics and Automation (ICRA)}, pp.~4788--4795, IEEE, 2024.

\bibitem{guhur2023instruction}
P.-L. Guhur, S.~Chen, R.~G. Pinel, M.~Tapaswi, I.~Laptev, and C.~Schmid,
  ``Instruction-driven history-aware policies for robotic manipulations,'' in
  {\em Conference on Robot Learning}, pp.~175--187, PMLR, 2023.

\bibitem{shridhar2023perceiver}
M.~Shridhar, L.~Manuelli, and D.~Fox, ``Perceiver-actor: A multi-task
  transformer for robotic manipulation,'' in {\em Conference on Robot
  Learning}, pp.~785--799, PMLR, 2023.

\bibitem{gervet2023act3d}
T.~Gervet, Z.~Xian, N.~Gkanatsios, and K.~Fragkiadaki, ``Act3d: {3D} feature
  field transformers for multi-task robotic manipulation,'' {\em arXiv preprint
  arXiv:2306.17817}, 2023.

\bibitem{liu2024volumetric}
R.~Liu, W.~Wang, and Y.~Yang, ``Volumetric environment representation for
  vision-language navigation,'' in {\em Proceedings of the IEEE/CVF Conference
  on Computer Vision and Pattern Recognition}, pp.~16317--16328, 2024.

\bibitem{yuan2024robopoint}
W.~Yuan, J.~Duan, V.~Blukis, W.~Pumacay, R.~Krishna, A.~Murali, A.~Mousavian,
  and D.~Fox, ``Robopoint: A vision-language model for spatial affordance
  prediction for robotics,'' {\em arXiv preprint arXiv:2406.10721}, 2024.

\bibitem{reuss2024multimodal}
M.~Reuss, {\"O}.~E. Ya{\u{g}}murlu, F.~Wenzel, and R.~Lioutikov, ``Multimodal
  diffusion transformer: Learning versatile behavior from multimodal goals,''
  {\em arXiv preprint arXiv:2407.05996}, 2024.

\bibitem{ha2023scaling}
H.~Ha, P.~Florence, and S.~Song, ``Scaling up and distilling down:
  Language-guided robot skill acquisition,'' in {\em Conference on Robot
  Learning}, pp.~3766--3777, PMLR, 2023.

\bibitem{liu2024surveypose}
J.~Liu, W.~Sun, H.~Yang, Z.~Zeng, C.~Liu, J.~Zheng, X.~Liu, H.~Rahmani,
  N.~Sebe, and A.~Mian, ``Deep learning-based object pose estimation: A
  comprehensive survey,'' {\em arXiv preprint arXiv:2405.07801}, 2024.

\bibitem{ke20243d}
T.-W. Ke, N.~Gkanatsios, and K.~Fragkiadaki, ``{3D} diffuser actor: Policy
  diffusion with {3D} scene representations,'' {\em arXiv preprint
  arXiv:2402.10885}, 2024.

\bibitem{durrant2006simultaneous}
H.~Durrant-Whyte and T.~Bailey, ``Simultaneous localization and mapping: part
  {I},'' {\em IEEE robotics \& automation magazine}, vol.~13, no.~2,
  pp.~99--110, 2006.

\bibitem{bailey2006simultaneous}
T.~Bailey and H.~Durrant-Whyte, ``Simultaneous localization and mapping (slam):
  Part {II},'' {\em IEEE robotics \& automation magazine}, vol.~13, no.~3,
  pp.~108--117, 2006.

\bibitem{bailey2006consistency}
T.~Bailey, J.~Nieto, J.~Guivant, M.~Stevens, and E.~Nebot, ``Consistency of the
  {EKF-SLAM} algorithm,'' in {\em IROS}, pp.~3562--3568, IEEE, 2006.

\bibitem{grisetti2010tutorial}
G.~Grisetti, R.~K{\"u}mmerle, C.~Stachniss, and W.~Burgard, ``A tutorial on
  graph-based {SLAM},'' {\em IEEE Intelligent Transportation Systems Magazine},
  vol.~2, no.~4, pp.~31--43, 2010.

\bibitem{macario2022comprehensive}
A.~Macario~Barros, M.~Michel, Y.~Moline, G.~Corre, and F.~Carrel, ``A
  comprehensive survey of visual {SLAM} algorithms,'' {\em Robotics}, vol.~11,
  no.~1, p.~24, 2022.

\bibitem{li2021improving}
J.~Li, H.~Tan, and M.~Bansal, ``Improving cross-modal alignment in vision
  language navigation via syntactic information,'' {\em arXiv preprint
  arXiv:2104.09580}, 2021.

\bibitem{xia2020multi}
Q.~Xia, X.~Li, C.~Li, Y.~Bisk, Z.~Sui, J.~Gao, Y.~Choi, and N.~A. Smith,
  ``Multi-view learning for vision-and-language navigation,'' {\em arXiv
  preprint arXiv:2003.00857}, 2020.

\bibitem{lin2021scene}
X.~Lin, G.~Li, and Y.~Yu, ``Scene-intuitive agent for remote embodied visual
  grounding,'' in {\em CVPR}, pp.~7036--7045, 2021.

\bibitem{gu2022vision}
J.~Gu, E.~Stefani, Q.~Wu, J.~Thomason, and X.~E. Wang, ``Vision-and-language
  navigation: A survey of tasks, methods, and future directions,'' {\em arXiv
  preprint arXiv:2203.12667}, 2022.

\bibitem{chaplot2020object}
D.~S. Chaplot, D.~P. Gandhi, A.~Gupta, and R.~R. Salakhutdinov, ``Object goal
  navigation using goal-oriented semantic exploration,'' {\em Advances in
  Neural Information Processing Systems}, vol.~33, pp.~4247--4258, 2020.

\bibitem{krantz2020beyond}
J.~Krantz, E.~Wijmans, A.~Majumdar, D.~Batra, and S.~Lee, ``Beyond the
  nav-graph: Vision-and-language navigation in continuous environments,'' in
  {\em Computer Vision--ECCV 2020: 16th European Conference, Glasgow, UK,
  August 23--28, 2020, Proceedings, Part XXVIII 16}, pp.~104--120, Springer,
  2020.

\bibitem{ku2020room}
A.~Ku, P.~Anderson, R.~Patel, E.~Ie, and J.~Baldridge, ``Room-across-room:
  Multilingual vision-and-language navigation with dense spatiotemporal
  grounding,'' {\em arXiv preprint arXiv:2010.07954}, 2020.

\bibitem{anderson2019chasing}
P.~Anderson, A.~Shrivastava, D.~Parikh, D.~Batra, and S.~Lee, ``Chasing ghosts:
  Instruction following as bayesian state tracking,'' {\em Advances in neural
  information processing systems}, vol.~32, 2019.

\bibitem{hong2021vln}
Y.~Hong, Q.~Wu, Y.~Qi, C.~Rodriguez-Opazo, and S.~Gould, ``{VLN BERT}: A
  recurrent vision-and-language bert for navigation,'' in {\em Proceedings of
  the IEEE/CVF conference on Computer Vision and Pattern Recognition},
  pp.~1643--1653, 2021.

\bibitem{CLIPose}
X.~Lin and M.~Zhu, ``Clipose: Category-level object pose estimation with
  pre-trained vision-language knowledge,'' {\em arXiv preprint
  arXiv:2402.15726}, 2024.

\bibitem{zhou2022learning}
K.~Zhou, J.~Yang, C.~C. Loy, and Z.~Liu, ``Learning to prompt for
  vision-language models,'' {\em International Journal of Computer Vision},
  vol.~130, no.~9, pp.~2337--2348, 2022.

\bibitem{driess2023palm}
D.~Driess, F.~Xia, M.~Sajjadi, C.~Lynch, A.~Chowdhery, B.~Ichter, A.~Wahid,
  J.~Tompson, Q.~Vuong, T.~Yu, {\em et~al.}, ``{PaLM-E}: An embodied multimodal
  language model,'' {\em arXiv preprint arXiv:2303.03378}, 2023.

\bibitem{chen2023pali}
X.~Chen, J.~Djolonga, P.~Padlewski, B.~Mustafa, S.~Changpinyo, J.~Wu, C.~R.
  Ruiz, S.~Goodman, X.~Wang, Y.~Tay, {\em et~al.}, ``Pali-x: On scaling up a
  multilingual vision and language model,'' {\em arXiv preprint
  arXiv:2305.18565}, 2023.

\bibitem{DINOv2}
M.~Oquab and T.~Darcet, ``Dinov2: Learning robust visual features without
  supervision,'' {\em arXiv preprint arXiv:2304.07193}, 2023.

\bibitem{o2024open}
A.~O'Neill, A.~Rehman, A.~Maddukuri, A.~Gupta, A.~Padalkar, A.~Lee, A.~Pooley,
  A.~Gupta, A.~Mandlekar, A.~Jain, {\em et~al.}, ``Open x-embodiment: Robotic
  learning datasets and rt-x models: Open x-embodiment collaboration$^{0}$,''
  in {\em ICRA}, pp.~6892--6903, IEEE, 2024.

\bibitem{li2023improving}
J.~Li and M.~Bansal, ``Improving vision-and-language navigation by generating
  future-view image semantics,'' in {\em CVPR}, pp.~10803--10812, 2023.

\bibitem{long2024discuss}
Y.~Long, X.~Li, W.~Cai, and H.~Dong, ``Discuss before moving: Visual language
  navigation via multi-expert discussions,'' in {\em ICRA}, pp.~17380--17387,
  IEEE, 2024.

\bibitem{zhan2024mc}
Z.~Zhan, L.~Yu, S.~Yu, and G.~Tan, ``Mc-{GPT}: Empowering vision-and-language
  navigation with memory map and reasoning chains,'' {\em arXiv preprint
  arXiv:2405.10620}, 2024.

\bibitem{baltruvsaitis2018multimodal}
T.~Baltru{\v{s}}aitis, C.~Ahuja, and L.-P. Morency, ``Multimodal machine
  learning: A survey and taxonomy,'' {\em IEEE transactions on pattern analysis
  and machine intelligence}, vol.~41, no.~2, pp.~423--443, 2018.

\bibitem{dritsas2025multimodal}
E.~Dritsas, M.~Trigka, C.~Troussas, and P.~Mylonas, ``Multimodal interaction,
  interfaces, and communication: A survey,'' {\em Multimodal Technologies and
  Interaction}, vol.~9, no.~1, p.~6, 2025.

\bibitem{stacke2020measuring}
K.~Stacke, G.~Eilertsen, J.~Unger, and C.~Lundstr{\"o}m, ``Measuring domain
  shift for deep learning in histopathology,'' {\em IEEE journal of biomedical
  and health informatics}, vol.~25, no.~2, pp.~325--336, 2020.

\bibitem{yang2024generalized}
J.~Yang, K.~Zhou, Y.~Li, and Z.~Liu, ``Generalized out-of-distribution
  detection: A survey,'' {\em International Journal of Computer Vision},
  vol.~132, no.~12, pp.~5635--5662, 2024.

\bibitem{graves2014neural}
A.~Graves, G.~Wayne, and I.~Danihelka, ``Neural turing machines,'' {\em arXiv
  preprint arXiv:1410.5401}, 2014.

\bibitem{park2020distributed}
T.~Park, I.~Choi, and M.~Lee, ``Distributed associative memory network with
  memory refreshing loss,'' {\em arXiv preprint arXiv:2007.10637}, 2020.

\bibitem{fan2024embodied}
Y.~Fan, X.~Ma, R.~Su, J.~Guo, R.~Wu, X.~Chen, and Q.~Li, ``Embodied videoagent:
  Persistent memory from egocentric videos and embodied sensors enables dynamic
  scene understanding,'' {\em arXiv preprint arXiv:2501.00358}, 2024.

\bibitem{lenton2021egospheric}
D.~Lenton, S.~James, R.~Clark, and A.~J. Davison, ``End-to-end egospheric
  spatial memory,'' {\em arXiv preprint arXiv:2102.07764}, 2021.

\bibitem{zhang2025mem2ego}
L.~Zhang, Y.~Liu, Z.~Zhang, M.~Aghaei, Y.~Hu, H.~Gu, M.~A. Alomrani, D.~G.~A.
  Bravo, R.~Karimi, A.~Hamidizadeh, {\em et~al.}, ``Mem2ego: Empowering
  vision-language models with global-to-ego memory for long-horizon embodied
  navigation,'' {\em arXiv preprint arXiv:2502.14254}, 2025.

\bibitem{wang2024karma}
Z.~Wang, B.~Yu, J.~Zhao, W.~Sun, S.~Hou, S.~Liang, X.~Hu, Y.~Han, and Y.~Gan,
  ``{KARMA}: Augmenting embodied {AI} agents with long-and-short term memory
  systems,'' {\em arXiv preprint arXiv:2409.14908}, 2024.

\bibitem{chari2025mindstores}
A.~Chari, S.~Reddy, A.~Tiwari, R.~Lian, and B.~Zhou, ``{MINDSTORES}:
  Memory-informed neural decision synthesis for task-oriented reinforcement in
  embodied systems,'' {\em arXiv preprint arXiv:2501.19318}, 2025.

\bibitem{liu2024skip}
Y.~Liu, Y.~Cao, and J.~Zhang, ``{Skip-SCAR}: Hardware-friendly high-quality
  embodied visual navigation,'' {\em arXiv preprint arXiv:2405.14154}, 2024.

\bibitem{Yang2025HierarchicalKG}
H.~e.~a. Yang, ``{LLM}-powered decentralized generative agents with adaptive
  hierarchical knowledge graph for cooperative planning,'' {\em arXiv preprint
  arXiv:2502.05453}, 2025.

\bibitem{Song2024SceneMMKG}
Y.~e.~a. Song, ``Scene-driven multimodal knowledge graph construction for
  embodied {AI},'' {\em arXiv preprint arXiv:2311.03783}, 2024.

\bibitem{akgvp2024}
L.~Zhao and et~al., ``Aligning knowledge graph with visual perception for
  object-goal navigation,'' {\em arXiv preprint arXiv:2402.18892}, 2024.

\bibitem{Singh2025AdaptBot}
S.~e.~a. Singh, ``{AdaptBot}: Combining {LLM} with knowledge graphs and human
  input for generic-to-specific task decomposition and knowledge refinement,''
  {\em arXiv preprint arXiv:2502.02067}, 2025.

\bibitem{Qi2024SafetyEKG}
Y.~e.~a. Qi, ``Safety control of service robots with {LLMs} and embodied
  knowledge graphs,'' {\em arXiv preprint arXiv:2405.17846}, 2024.

\bibitem{esgnn2024}
Y.~Zhang and et~al., ``{ESGNN}: Towards equivariant scene graph neural network
  for {3D} scene understanding,'' {\em arXiv preprint arXiv:2407.00609}, 2024.

\bibitem{koch2023sgrec3d}
S.~Koch and et~al., ``{SGRec3D}: Self-supervised {3D} scene graph learning via
  object-level scene reconstruction,'' {\em arXiv preprint arXiv:2309.15702},
  2023.

\bibitem{3dgraphllm2024}
Z.~Wang and et~al., ``{3DGraphLLM}: Combining semantic graphs and large
  language models for {3D} vision-language tasks,'' {\em arXiv preprint
  arXiv:2412.18450}, 2024.

\bibitem{kurenkov2023modeling}
A.~Kurenkov and et~al., ``Modeling dynamic environments with scene graph
  memory,'' {\em arXiv preprint arXiv:2305.17537}, 2023.

\bibitem{xie2024embodied}
Q.~Xie, S.~Y. Min, P.~Ji, Y.~Yang, T.~Zhang, K.~Xu, A.~Bajaj, R.~Salakhutdinov,
  M.~Johnson-Roberson, and Y.~Bisk, ``Embodied-{RAG}: General non-parametric
  embodied memory for retrieval and generation,'' {\em arXiv preprint
  arXiv:2409.18313}, 2024.

\bibitem{zhang2025embodiedvsr}
Y.~Zhang and et~al., ``{EmbodiedVSR}: Dynamic scene graph-guided
  chain-of-thought reasoning for visual spatial tasks,'' {\em arXiv preprint
  arXiv:2503.11089}, 2025.

\bibitem{huang2023structure}
Y.~Huang and et~al., ``Structure-{CLIP}: Towards scene graph knowledge to
  enhance multi-modal structured representations,'' {\em arXiv preprint
  arXiv:2305.06152}, 2023.

\bibitem{cho2024spatially}
J.~Cho, J.~Yoon, and S.~Ahn, ``Spatially-aware transformer for embodied
  agents,'' {\em arXiv preprint arXiv:2402.15160}, 2024.

\bibitem{yang2024dmem}
Y.~Yang, H.~Yang, J.~Zhou, P.~Chen, H.~Zhang, Y.~Du, and C.~Gan, ``{3D}-mem:
  {3D} scene memory for embodied exploration and reasoning,'' {\em arXiv
  preprint arXiv:2411.17735}, 2024.

\bibitem{lei2025stma}
M.~Lei, Y.~Zhao, G.~Wang, Z.~Mai, S.~Cui, Y.~Han, and J.~Ren, ``{STMA}: A
  spatio-temporal memory agent for long-horizon embodied task planning,'' {\em
  arXiv preprint arXiv:2502.10177}, 2025.

\bibitem{pan2024planning}
Y.~Pan, Y.~Xu, Z.~Liu, and H.~Wang, ``Planning from imagination: Episodic
  simulation and episodic memory for vision-and-language navigation,'' {\em
  arXiv preprint arXiv:2412.01857}, 2024.

\bibitem{Kim2024Leveraging}
T.~e.~a. Kim, ``Leveraging knowledge graph-based human-like memory systems to
  solve partially observable markov decision processes,'' {\em arXiv preprint
  arXiv:2408.05861}, 2024.

\bibitem{wang2023voyager}
G.~Wang, Y.~Xie, Y.~Jiang, A.~Mandlekar, C.~Xiao, Y.~Zhu, L.~Fan, and
  A.~Anandkumar, ``Voyager: An open-ended embodied agent with large language
  models,'' {\em arXiv preprint arXiv:2305.16291}, 2023.

\bibitem{park2025active}
J.~Park, D.~Park, and J.-G. Lee, ``Active learning for continual learning:
  Keeping the past alive in the present,'' {\em arXiv preprint
  arXiv:2501.14278}, 2025.

\bibitem{wang2021afec}
L.~Wang, M.~Zhang, Z.~Jia, Q.~Li, C.~Bao, K.~Ma, J.~Zhu, and Y.~Zhong, ``Afec:
  Active forgetting of negative transfer in continual learning,'' {\em Advances
  in Neural Information Processing Systems}, vol.~34, pp.~22379--22391, 2021.

\bibitem{mendez2023embodied}
J.~Mendez-Mendez, L.~P. Kaelbling, and T.~Lozano-Perez, ``Embodied lifelong
  learning for task and motion planning,'' {\em arXiv preprint
  arXiv:2307.06870}, 2023.

\bibitem{kim2024online}
B.~Kim, M.~Seo, and J.~Choi, ``Online continual learning for interactive
  instruction following agents,'' {\em arXiv preprint arXiv:2403.07548}, 2024.

\bibitem{gao2023fast}
J.~Gao, X.~Yao, and C.~Xu, ``Fast-slow test-time adaptation for online
  vision-and-language navigation,'' {\em arXiv preprint arXiv:2311.13209},
  2023.

\bibitem{zhai2023building}
S.~Zhai, J.~Wang, T.~Zhang, F.~Huang, Q.~Zhang, M.~Zhou, J.~Hou, Y.~Qiao, and
  Y.~Liu, ``Building open-ended embodied agent via language-policy
  bidirectional adaptation,'' {\em arXiv preprint arXiv:2401.00006}, 2023.

\bibitem{jang2023modality}
H.~Jang, J.~Tack, D.~Choi, J.~Jeong, and J.~Shin, ``Modality-agnostic
  self-supervised learning with meta-learned masked auto-encoder,'' {\em arXiv
  preprint arXiv:2310.16318}, 2023.

\bibitem{wang2023bmssl}
J.~Wang, Z.~Song, W.~Qiang, and C.~Zheng, ``Unleash model potential:
  Bootstrapped meta self-supervised learning,'' {\em arXiv preprint
  arXiv:2308.14267}, 2023.

\bibitem{cui2025dress}
W.~Cui, T.~Wu, J.~C. Cresswell, Y.~Sui, and K.~Golestan, ``{DRESS}:
  Disentangled representation-based self-supervised meta-learning for diverse
  tasks,'' {\em arXiv preprint arXiv:2503.09679}, 2025.

\bibitem{geng2022multimodalmaskedautoencoderslearn}
X.~Geng, H.~Liu, L.~Lee, D.~Schuurmans, S.~Levine, and P.~Abbeel, ``Multimodal
  masked autoencoders learn transferable representations,'' {\em arXiv preprint
  arXiv:2205.14204}, 2022.

\bibitem{wan2025rema}
Z.~Wan, Y.~Li, Y.~Song, H.~Wang, L.~Yang, M.~Schmidt, J.~Wang, W.~Zhang, S.~Hu,
  and Y.~Wen, ``{ReMA}: Learning to meta-think for {LLMs} with multi-agent
  reinforcement learning,'' {\em arXiv preprint arXiv:2503.09501}, 2025.

\bibitem{he2024selfsupervisedmetalearningalllayerdnnbased}
G.~He, Y.~Choudhary, and G.~Shi, ``Self-supervised meta-learning for all-layer
  dnn-based adaptive control with stability guarantees,'' {\em arXiv preprint
  arXiv:2410.07575}, 2024.

\bibitem{zhang2023meta}
Y.~Zhang, K.~Gong, K.~Zhang, H.~Li, Y.~Qiao, W.~Ouyang, and X.~Yue,
  ``Meta-transformer: A unified framework for multimodal learning,'' {\em arXiv
  preprint arXiv:2307.10802}, 2023.

\bibitem{yang2023mmreact}
Z.~Yang, H.~Hu, H.~Li, Y.~Wu, Y.~Zhang, Y.~Zhu, Y.~Xu, J.~Wu, and L.~Fei-Fei,
  ``{MM-ReAct}: Prompting chatgpt for multimodal reasoning and action,'' {\em
  arXiv preprint arXiv:2303.11381}, 2023.

\bibitem{li2024unicl}
J.~Li, J.~Peng, H.~Li, and L.~Chen, ``Unicl: A universal contrastive learning
  framework for large time series models,'' {\em arXiv preprint
  arXiv:2405.10597}, 2024.

\bibitem{li2022uniperceiverv2generalistmodel}
H.~Li, J.~Zhu, X.~Jiang, X.~Zhu, H.~Li, C.~Yuan, X.~Wang, Y.~Qiao, X.~Wang,
  W.~Wang, {\em et~al.}, ``Uni-perceiver v2: A generalist model for large-scale
  vision and vision-language tasks,'' in {\em Proceedings of the IEEE/CVF
  Conference on Computer Vision and Pattern Recognition}, pp.~2691--2700, 2023.

\bibitem{hawthorne2022general}
C.~Hawthorne, A.~Jaegle, C.~Cangea, S.~Borgeaud, C.~Nash, M.~Malinowski,
  S.~Dieleman, O.~Vinyals, M.~Botvinick, I.~Simon, {\em et~al.},
  ``General-purpose, long-context autoregressive modeling with perceiver ar,''
  in {\em International Conference on Machine Learning}, pp.~8535--8558, PMLR,
  2022.

\bibitem{zheng2025lifelong}
J.~Zheng, C.~Shi, X.~Cai, Q.~Li, D.~Zhang, C.~Li, D.~Yu, and Q.~Ma, ``Lifelong
  learning of large language model based agents: A roadmap,'' {\em arXiv
  preprint arXiv:2501.07278}, 2025.

\bibitem{liu2024meia}
Y.~Liu, X.~Song, K.~Jiang, W.~Chen, J.~Luo, G.~Li, and L.~Lin, ``Meia:
  Multimodal embodied perception and interaction in unknown environments,''
  {\em arXiv preprint arXiv:2402.00290}, 2024.

\bibitem{laydevant2024hardware}
J.~Laydevant, L.~G. Wright, T.~Wang, and P.~L. McMahon, ``The hardware is the
  software,'' {\em Neuron}, vol.~112, no.~2, pp.~180--183, 2024.

\bibitem{brette2022brains}
R.~Brette, ``Brains as computers: Metaphor, analogy, theory or fact?,'' {\em
  Frontiers in Ecology and Evolution}, vol.~10, p.~878729, 2022.

\bibitem{galakhova2022evolution}
A.~A. Galakhova, S.~Hunt, R.~Wilbers, D.~B. Heyer, C.~P. de~Kock, H.~D.
  Mansvelder, and N.~A. Goriounova, ``Evolution of cortical neurons supporting
  human cognition,'' {\em Trends in cognitive sciences}, vol.~26, no.~11,
  pp.~909--922, 2022.

\bibitem{barrett2012hierarchical}
H.~C. Barrett, ``A hierarchical model of the evolution of human brain
  specializations,'' {\em Proceedings of the national Academy of Sciences},
  vol.~109, no.~supplement\_1, pp.~10733--10740, 2012.

\bibitem{pulvermuller2018neural}
F.~Pulverm{\"u}ller, ``Neural reuse of action perception circuits for language,
  concepts and communication,'' {\em Progress in neurobiology}, vol.~160,
  pp.~1--44, 2018.

\bibitem{gupta2023neuroprosthetics}
A.~Gupta, N.~Vardalakis, and F.~B. Wagner, ``Neuroprosthetics: from
  sensorimotor to cognitive disorders,'' {\em Communications biology}, vol.~6,
  no.~1, p.~14, 2023.

\bibitem{bohnstingl2019neuromorphic}
T.~Bohnstingl, F.~Scherr, C.~Pehle, K.~Meier, and W.~Maass, ``Neuromorphic
  hardware learns to learn,'' {\em Frontiers in neuroscience}, vol.~13, p.~483,
  2019.

\bibitem{walter2015neuromorphic}
F.~Walter, F.~R{\"o}hrbein, and A.~Knoll, ``Neuromorphic implementations of
  neurobiological learning algorithms for spiking neural networks,'' {\em
  Neural Networks}, vol.~72, pp.~152--167, 2015.

\bibitem{schuman2022opportunities}
C.~D. Schuman, S.~R. Kulkarni, M.~Parsa, J.~P. Mitchell, P.~Date, and B.~Kay,
  ``Opportunities for neuromorphic computing algorithms and applications,''
  {\em Nature Computational Science}, vol.~2, no.~1, pp.~10--19, 2022.

\bibitem{shahsavari2023advancements}
M.~Shahsavari, D.~Thomas, M.~van Gerven, A.~Brown, and W.~Luk, ``Advancements
  in spiking neural network communication and synchronization techniques for
  event-driven neuromorphic systems,'' {\em Array}, vol.~20, p.~100323, 2023.

\bibitem{carrillo2012scalable}
S.~Carrillo, J.~Harkin, L.~J. McDaid, F.~Morgan, S.~Pande, S.~Cawley, and
  B.~McGinley, ``Scalable hierarchical network-on-chip architecture for spiking
  neural network hardware implementations,'' {\em IEEE Transactions on Parallel
  and Distributed Systems}, vol.~24, no.~12, pp.~2451--2461, 2012.

\bibitem{rathi2023exploring}
N.~Rathi, I.~Chakraborty, A.~Kosta, A.~Sengupta, A.~Ankit, P.~Panda, and
  K.~Roy, ``Exploring neuromorphic computing based on spiking neural networks:
  Algorithms to hardware,'' {\em ACM Computing Surveys}, vol.~55, no.~12,
  pp.~1--49, 2023.

\bibitem{mehonic2024roadmap}
A.~Mehonic, D.~Ielmini, K.~Roy, O.~Mutlu, S.~Kvatinsky, T.~Serrano-Gotarredona,
  B.~Linares-Barranco, S.~Spiga, S.~Savel'ev, A.~G. Balanov, {\em et~al.},
  ``Roadmap to neuromorphic computing with emerging technologies,'' {\em APL
  Materials}, vol.~12, no.~10, 2024.

\bibitem{davies2018loihi}
M.~Davies, N.~Srinivasa, T.-H. Lin, G.~Chinya, Y.~Cao, S.~H. Choday, G.~Dimou,
  P.~Joshi, N.~Imam, S.~Jain, {\em et~al.}, ``Loihi: A neuromorphic manycore
  processor with on-chip learning,'' {\em Ieee Micro}, vol.~38, no.~1,
  pp.~82--99, 2018.

\bibitem{pehle2022brainscales}
C.~Pehle, S.~Billaudelle, B.~Cramer, J.~Kaiser, K.~Schreiber, Y.~Stradmann,
  J.~Weis, A.~Leibfried, E.~M{\"u}ller, and J.~Schemmel, ``The {BrainScaleS}-2
  accelerated neuromorphic system with hybrid plasticity,'' {\em Frontiers in
  Neuroscience}, vol.~16, p.~795876, 2022.

\bibitem{richter2023speck}
O.~Richter, Y.~Xing, M.~De~Marchi, C.~Nielsen, M.~Katsimpris, R.~Cattaneo,
  Y.~Ren, Y.~Hu, Q.~Liu, S.~Sheik, {\em et~al.}, ``Speck: A smart event-based
  vision sensor with a low latency 327k neuron convolutional neuronal network
  processing pipeline,'' {\em arXiv preprint arXiv:2304.06793}, 2023.

\bibitem{frenkel20180}
C.~Frenkel, M.~Lefebvre, J.-D. Legat, and D.~Bol, ``A 0.086-mm $^2$ 12.7-pj/sop
  64k-synapse 256-neuron online-learning digital spiking neuromorphic processor
  in 28-nm {CMOS},'' {\em IEEE transactions on biomedical circuits and
  systems}, vol.~13, no.~1, pp.~145--158, 2018.

\bibitem{moreira2020neuronflow}
O.~Moreira, A.~Yousefzadeh, F.~Chersi, A.~Kapoor, R.-J. Zwartenkot, P.~Qiao,
  G.~Cinserin, M.~A. Khoei, M.~Lindwer, and J.~Tapson, ``Neuronflow: A hybrid
  neuromorphic--dataflow processor architecture for {AI} workloads,'' in {\em
  2020 2nd IEEE International Conference on Artificial Intelligence Circuits
  and Systems (AICAS)}, pp.~01--05, IEEE, 2020.

\bibitem{orchard2021efficient}
G.~Orchard, E.~P. Frady, D.~B.~D. Rubin, S.~Sanborn, S.~B. Shrestha, F.~T.
  Sommer, and M.~Davies, ``Efficient neuromorphic signal processing with loihi
  2,'' in {\em 2021 IEEE Workshop on Signal Processing Systems (SiPS)},
  pp.~254--259, IEEE, 2021.

\bibitem{ivanov2022neuromorphic}
D.~Ivanov, A.~Chezhegov, M.~Kiselev, A.~Grunin, and D.~Larionov, ``Neuromorphic
  artificial intelligence systems,'' {\em Frontiers in Neuroscience}, vol.~16,
  p.~959626, 2022.

\bibitem{chi2016prime}
P.~Chi, S.~Li, C.~Xu, T.~Zhang, J.~Zhao, Y.~Liu, Y.~Wang, and Y.~Xie, ``Prime:
  A novel processing-in-memory architecture for neural network computation in
  reram-based main memory,'' {\em ACM SIGARCH Computer Architecture News},
  vol.~44, no.~3, pp.~27--39, 2016.

\bibitem{chen2018neurosim}
P.-Y. Chen, X.~Peng, and S.~Yu, ``Neurosim: A circuit-level macro model for
  benchmarking neuro-inspired architectures in online learning,'' {\em IEEE
  Transactions on Computer-Aided Design of Integrated Circuits and Systems},
  vol.~37, no.~12, pp.~3067--3080, 2018.

\bibitem{nandakumar2018phase}
S.~Nandakumar, M.~Le~Gallo, I.~Boybat, B.~Rajendran, A.~Sebastian, and
  E.~Eleftheriou, ``A phase-change memory model for neuromorphic computing,''
  {\em Journal of Applied Physics}, vol.~124, no.~15, 2018.

\bibitem{ankit2019puma}
A.~Ankit, I.~E. Hajj, S.~R. Chalamalasetti, G.~Ndu, M.~Foltin, R.~S. Williams,
  P.~Faraboschi, W.-m.~W. Hwu, J.~P. Strachan, K.~Roy, {\em et~al.}, ``{PUMA}:
  A programmable ultra-efficient memristor-based accelerator for machine
  learning inference,'' in {\em Proceedings of the twenty-fourth international
  conference on architectural support for programming languages and operating
  systems}, pp.~715--731, 2019.

\bibitem{shivanandamurthy2021odin}
S.~M. Shivanandamurthy, I.~Thakkar, S.~A. Salehi, {\em et~al.}, ``{ODIN}: A
  bit-parallel stochastic arithmetic based accelerator for in-situ neural
  network processing in phase change {RAM},'' {\em arXiv preprint
  arXiv:2103.03953}, 2021.

\bibitem{brossollet2021lighton}
C.~Brossollet, A.~Cappelli, I.~Carron, C.~Chaintoutis, A.~Chatelain, L.~Daudet,
  S.~Gigan, D.~Hesslow, F.~Krzakala, J.~Launay, {\em et~al.}, ``Lighton optical
  processing unit: Scaling-up {AI} and {HPC} with a non von neumann
  co-processor,'' {\em arXiv preprint arXiv:2107.11814}, 2021.

\bibitem{ashtiani2022chip}
F.~Ashtiani, A.~J. Geers, and F.~Aflatouni, ``An on-chip photonic deep neural
  network for image classification,'' {\em Nature}, vol.~606, no.~7914,
  pp.~501--506, 2022.

\bibitem{martinez2023quantum}
R.~Mart{\'\i}nez-Pe{\~n}a and J.-P. Ortega, ``Quantum reservoir computing in
  finite dimensions,'' {\em Physical Review E}, vol.~107, no.~3, p.~035306,
  2023.

\bibitem{qi2023theoretical}
J.~Qi, C.-H.~H. Yang, P.-Y. Chen, and M.-H. Hsieh, ``Theoretical error
  performance analysis for variational quantum circuit based functional
  regression,'' {\em npj Quantum Information}, vol.~9, no.~1, p.~4, 2023.

\bibitem{kwon2022synaptic}
O.~Kwon, Y.~Lee, M.~Kang, and S.~Kim, ``Synaptic plasticity features and
  neuromorphic system simulation in aln-based memristor devices,'' {\em Journal
  of Alloys and Compounds}, vol.~911, p.~164870, 2022.

\bibitem{mehonic2020memristors}
A.~Mehonic, A.~Sebastian, B.~Rajendran, O.~Simeone, E.~Vasilaki, and A.~J.
  Kenyon, ``Memristors-from in-memory computing, deep learning acceleration,
  and spiking neural networks to the future of neuromorphic and bio-inspired
  computing,'' {\em Advanced Intelligent Systems}, vol.~2, no.~11, p.~2000085,
  2020.

\bibitem{du2021synaptic}
N.~Du, X.~Zhao, Z.~Chen, B.~Choubey, M.~Di~Ventra, I.~Skorupa, D.~B{\"u}rger,
  and H.~Schmidt, ``Synaptic plasticity in memristive artificial synapses and
  their robustness against noisy inputs,'' {\em Frontiers in neuroscience},
  vol.~15, p.~660894, 2021.

\bibitem{xiao2023review}
Y.~Xiao, B.~Jiang, Z.~Zhang, S.~Ke, Y.~Jin, X.~Wen, and C.~Ye, ``A review of
  memristor: material and structure design, device performance, applications
  and prospects,'' {\em Science and Technology of Advanced Materials}, vol.~24,
  no.~1, p.~2162323, 2023.

\bibitem{park2022experimental}
S.-O. Park, H.~Jeong, J.~Park, J.~Bae, and S.~Choi, ``Experimental
  demonstration of highly reliable dynamic memristor for artificial neuron and
  neuromorphic computing,'' {\em Nature Communications}, vol.~13, no.~1,
  p.~2888, 2022.

\bibitem{prucnal2017neuromorphic}
P.~R. Prucnal and B.~J. Shastri, {\em Neuromorphic photonics}.
\newblock CRC press, 2017.

\bibitem{ajagekar2022quantum}
A.~Ajagekar and F.~You, ``Quantum computing and quantum artificial intelligence
  for renewable and sustainable energy: A emerging prospect towards climate
  neutrality,'' {\em Renewable and Sustainable Energy Reviews}, vol.~165,
  p.~112493, 2022.

\bibitem{shastri2021photonics}
B.~J. Shastri, A.~N. Tait, T.~Ferreira~de Lima, W.~H. Pernice, H.~Bhaskaran,
  C.~D. Wright, and P.~R. Prucnal, ``Photonics for artificial intelligence and
  neuromorphic computing,'' {\em Nature Photonics}, vol.~15, no.~2,
  pp.~102--114, 2021.

\bibitem{goi2020perspective}
E.~Goi, Q.~Zhang, X.~Chen, H.~Luan, and M.~Gu, ``Perspective on photonic
  memristive neuromorphic computing,'' {\em PhotoniX}, vol.~1, pp.~1--26, 2020.

\bibitem{sangwan2020neuromorphic}
V.~K. Sangwan and M.~C. Hersam, ``Neuromorphic nanoelectronic materials,'' {\em
  Nature nanotechnology}, vol.~15, no.~7, pp.~517--528, 2020.

\bibitem{pershin2011neuromorphic}
Y.~V. Pershin and M.~Di~Ventra, ``Neuromorphic, digital, and quantum
  computation with memory circuit elements,'' {\em Proceedings of the IEEE},
  vol.~100, no.~6, pp.~2071--2080, 2011.

\bibitem{ferreira2017progress}
T.~Ferreira De~Lima, B.~J. Shastri, A.~N. Tait, M.~A. Nahmias, and P.~R.
  Prucnal, ``Progress in neuromorphic photonics,'' {\em Nanophotonics}, vol.~6,
  no.~3, pp.~577--599, 2017.

\bibitem{hoffmann2022quantum}
A.~Hoffmann, S.~Ramanathan, J.~Grollier, A.~D. Kent, M.~J. Rozenberg, I.~K.
  Schuller, O.~G. Shpyrko, R.~C. Dynes, Y.~Fainman, A.~Frano, {\em et~al.},
  ``Quantum materials for energy-efficient neuromorphic computing:
  Opportunities and challenges,'' {\em APL Materials}, vol.~10, no.~7, 2022.

\bibitem{li2023blip}
J.~Li, D.~Li, S.~Savarese, and S.~Hoi, ``Blip-2: Bootstrapping language-image
  pre-training with frozen image encoders and large language models,'' in {\em
  International conference on machine learning}, pp.~19730--19742, PMLR, 2023.

\bibitem{anil2023palm}
R.~Anil, A.~M. Dai, O.~Firat, M.~Johnson, D.~Lepikhin, A.~Passos, S.~Shakeri,
  E.~Taropa, P.~Bailey, Z.~Chen, {\em et~al.}, ``Palm 2 technical report,''
  {\em arXiv preprint arXiv:2305.10403}, 2023.

\bibitem{liu2024deepseek}
A.~Liu, B.~Feng, B.~Xue, B.~Wang, B.~Wu, C.~Lu, C.~Zhao, C.~Deng, C.~Zhang,
  C.~Ruan, {\em et~al.}, ``Deepseek-{V3} technical report,'' {\em arXiv
  preprint arXiv:2412.19437}, 2024.

\bibitem{dupuy1990nest}
A.~Dupuy, J.~Schwartz, Y.~Yemini, and D.~Bacon, ``{NEST}: A network simulation
  and prototyping testbed,'' {\em Communications of the ACM}, vol.~33, no.~10,
  pp.~63--74, 1990.

\bibitem{hazan2018bindsnet}
H.~Hazan, D.~J. Saunders, H.~Khan, D.~Patel, D.~T. Sanghavi, H.~T. Siegelmann,
  and R.~Kozma, ``Bindsnet: A machine learning-oriented spiking neural networks
  library in python,'' {\em Frontiers in neuroinformatics}, vol.~12, p.~89,
  2018.

\bibitem{stimberg2019brian}
M.~Stimberg, R.~Brette, and D.~F. Goodman, ``Brian 2, an intuitive and
  efficient neural simulator,'' {\em elife}, vol.~8, p.~e47314, 2019.

\bibitem{furber2020spinnaker}
S.~Furber and P.~Bogdan, {\em Spinnaker-a spiking neural network architecture}.
\newblock Now publishers, 2020.

\bibitem{pehle2021norse}
C.-G. Pehle and J.~Egholm~Pedersen, ``Norse-a deep learning library for spiking
  neural networks,'' {\em Zenodo}, 2021.

\bibitem{davoodi2019tensorrt}
P.~Davoodi, C.~Gwon, G.~Lai, and T.~Morris, ``Tensorrt inference with
  tensorflow,'' in {\em GPU Technology Conference}, 2019.

\bibitem{jin2020compiling}
T.~Jin, G.-T. Bercea, T.~D. Le, T.~Chen, G.~Su, H.~Imai, Y.~Negishi, A.~Leu,
  K.~O'Brien, K.~Kawachiya, {\em et~al.}, ``Compiling onnx neural network
  models using mlir,'' {\em arXiv preprint arXiv:2008.08272}, 2020.

\bibitem{benoit2020impact}
T.~Benoit-Cattin, D.~Velasco-Montero, and J.~Fern{\'a}ndez-Berni, ``Impact of
  thermal throttling on long-term visual inference in a {CPU}-based edge
  device,'' {\em Electronics}, vol.~9, no.~12, p.~2106, 2020.

\bibitem{luo2020comparison}
C.~Luo, X.~He, J.~Zhan, L.~Wang, W.~Gao, and J.~Dai, ``Comparison and
  benchmarking of {AI} models and frameworks on mobile devices,'' {\em arXiv
  preprint arXiv:2005.05085}, 2020.

\bibitem{david2021tensorflow}
R.~David, J.~Duke, A.~Jain, V.~Janapa~Reddi, N.~Jeffries, J.~Li, N.~Kreeger,
  I.~Nappier, M.~Natraj, T.~Wang, {\em et~al.}, ``Tensorflow lite micro:
  Embedded machine learning for tinyml systems,'' {\em Proceedings of Machine
  Learning and Systems}, vol.~3, pp.~800--811, 2021.

\bibitem{zhang2020system}
Y.~Zhang, P.~Qu, Y.~Ji, W.~Zhang, G.~Gao, G.~Wang, S.~Song, G.~Li, W.~Chen,
  W.~Zheng, {\em et~al.}, ``A system hierarchy for brain-inspired computing,''
  {\em Nature}, vol.~586, no.~7829, pp.~378--384, 2020.

\bibitem{mehonic2022brain}
A.~Mehonic and A.~J. Kenyon, ``Brain-inspired computing needs a master plan,''
  {\em Nature}, vol.~604, no.~7905, pp.~255--260, 2022.

\bibitem{remmelzwaal2021brain}
L.~A. Remmelzwaal, A.~K. Mishra, and G.~F. Ellis, ``Brain-inspired distributed
  cognitive architecture,'' {\em Cognitive Systems Research}, vol.~66,
  pp.~13--20, 2021.

\bibitem{richards2019deep}
B.~A. Richards, T.~P. Lillicrap, P.~Beaudoin, Y.~Bengio, R.~Bogacz,
  A.~Christensen, C.~Clopath, R.~P. Costa, A.~de~Berker, S.~Ganguli, {\em
  et~al.}, ``A deep learning framework for neuroscience,'' {\em Nature
  neuroscience}, vol.~22, no.~11, pp.~1761--1770, 2019.

\bibitem{nguyen2019machine}
G.~Nguyen, S.~Dlugolinsky, M.~Bob{\'a}k, V.~Tran, {\'A}.~L{\'o}pez~Garc{\'\i}a,
  I.~Heredia, P.~Mal{\'\i}k, and L.~Hluch{\`y}, ``Machine learning and deep
  learning frameworks and libraries for large-scale data mining: A survey,''
  {\em Artificial Intelligence Review}, vol.~52, pp.~77--124, 2019.

\bibitem{novac2022analysis}
O.-C. Novac, M.~C. Chirodea, C.~M. Novac, N.~Bizon, M.~Oproescu, O.~P. Stan,
  and C.~E. Gordan, ``Analysis of the application efficiency of {TensorFlow}
  and pytorch in convolutional neural network,'' {\em Sensors}, vol.~22,
  no.~22, p.~8872, 2022.

\bibitem{alomar2024rnns}
K.~Alomar, H.~I. Aysel, and X.~Cai, ``{RNNs}, {CNNs} and transformers in human
  action recognition: A survey and a hybrid model,'' {\em arXiv preprint
  arXiv:2407.06162}, 2024.

\bibitem{chen2022semi}
Y.~Chen, M.~Mancini, X.~Zhu, and Z.~Akata, ``Semi-supervised and unsupervised
  deep visual learning: A survey,'' {\em IEEE transactions on pattern analysis
  and machine intelligence}, vol.~46, no.~3, pp.~1327--1347, 2022.

\bibitem{akalin2021reinforcement}
N.~Akalin and A.~Loutfi, ``Reinforcement learning approaches in social
  robotics,'' {\em Sensors}, vol.~21, no.~4, p.~1292, 2021.

\bibitem{schmidhuber2015deep}
J.~Schmidhuber, ``Deep learning in neural networks: An overview,'' {\em Neural
  networks}, vol.~61, pp.~85--117, 2015.

\bibitem{huang2022inner}
W.~Huang, F.~Xia, T.~Xiao, H.~Chan, J.~Liang, P.~Florence, A.~Zeng, J.~Tompson,
  I.~Mordatch, Y.~Chebotar, {\em et~al.}, ``Inner monologue: Embodied reasoning
  through planning with language models,'' {\em arXiv preprint
  arXiv:2207.05608}, 2022.

\bibitem{peebles2023scalable}
W.~Peebles and S.~Xie, ``Scalable diffusion models with transformers,'' in {\em
  Proceedings of the IEEE/CVF international conference on computer vision},
  pp.~4195--4205, 2023.

\bibitem{de2022next}
C.~M. De~Melo, A.~Torralba, L.~Guibas, J.~DiCarlo, R.~Chellappa, and
  J.~Hodgins, ``Next-generation deep learning based on simulators and synthetic
  data,'' {\em Trends in cognitive sciences}, vol.~26, no.~2, pp.~174--187,
  2022.

\bibitem{li2022metadrive}
Q.~Li, Z.~Peng, L.~Feng, Q.~Zhang, Z.~Xue, and B.~Zhou, ``Metadrive: Composing
  diverse driving scenarios for generalizable reinforcement learning,'' {\em
  IEEE transactions on pattern analysis and machine intelligence}, vol.~45,
  no.~3, pp.~3461--3475, 2022.

\bibitem{makoviychuk2021isaac}
V.~Makoviychuk, L.~Wawrzyniak, Y.~Guo, M.~Lu, K.~Storey, M.~Macklin,
  D.~Hoeller, N.~Rudin, A.~Allshire, A.~Handa, {\em et~al.}, ``Isaac {Gym}:
  High performance gpu-based physics simulation for robot learning,'' {\em
  arXiv preprint arXiv:2108.10470}, 2021.

\bibitem{mu2023embodiedgpt}
Y.~Mu, Q.~Zhang, M.~Hu, W.~Wang, M.~Ding, J.~Jin, B.~Wang, J.~Dai, Y.~Qiao, and
  P.~Luo, ``Embodiedgpt: Vision-language pre-training via embodied chain of
  thought,'' {\em Advances in Neural Information Processing Systems}, vol.~36,
  pp.~25081--25094, 2023.

\bibitem{yik2025neurobench}
J.~Yik, K.~Van~den Berghe, D.~den Blanken, Y.~Bouhadjar, M.~Fabre, P.~Hueber,
  W.~Ke, M.~A. Khoei, D.~Kleyko, N.~Pacik-Nelson, {\em et~al.}, ``The
  neurobench framework for benchmarking neuromorphic computing algorithms and
  systems,'' {\em Nature Communications}, vol.~16, no.~1, p.~1545, 2025.

\bibitem{sandamirskaya2022neuromorphic}
Y.~Sandamirskaya, M.~Kaboli, J.~Conradt, and T.~Celikel, ``Neuromorphic
  computing hardware and neural architectures for robotics,'' {\em Science
  Robotics}, vol.~7, no.~67, p.~eabl8419, 2022.

\bibitem{teeter2018generalized}
C.~Teeter, R.~Iyer, V.~Menon, N.~Gouwens, D.~Feng, J.~Berg, A.~Szafer, N.~Cain,
  H.~Zeng, M.~Hawrylycz, {\em et~al.}, ``Generalized leaky integrate-and-fire
  models classify multiple neuron types,'' {\em Nature communications}, vol.~9,
  no.~1, p.~709, 2018.

\bibitem{kulkarni2021benchmarking}
S.~R. Kulkarni, M.~Parsa, J.~P. Mitchell, and C.~D. Schuman, ``Benchmarking the
  performance of neuromorphic and spiking neural network simulators,'' {\em
  Neurocomputing}, vol.~447, pp.~145--160, 2021.

\bibitem{huynh2022implementing}
P.~K. Huynh, M.~L. Varshika, A.~Paul, M.~Isik, A.~Balaji, and A.~Das,
  ``Implementing spiking neural networks on neuromorphic architectures: A
  review,'' {\em arXiv preprint arXiv:2202.08897}, 2022.

\bibitem{wunderlich2019demonstrating}
T.~Wunderlich, A.~F. Kungl, E.~M{\"u}ller, A.~Hartel, Y.~Stradmann, S.~A.
  Aamir, A.~Gr{\"u}bl, A.~Heimbrecht, K.~Schreiber, D.~St{\"o}ckel, {\em
  et~al.}, ``Demonstrating advantages of neuromorphic computation: A pilot
  study,'' {\em Frontiers in neuroscience}, vol.~13, p.~260, 2019.

\bibitem{villar2023edge}
E.~Villar-Rodriguez, M.~A. P{\'e}rez, A.~I. Torre-Bastida, C.~R. Senderos, and
  J.~L{\'o}pez-de Armentia, ``Edge intelligence secure frameworks: Current
  state and future challenges,'' {\em Computers \& Security}, vol.~130,
  p.~103278, 2023.

\bibitem{thomas2009comparison}
D.~B. Thomas, L.~Howes, and W.~Luk, ``A comparison of {CPUs}, {GPUs}, {FPGAs},
  and massively parallel processor arrays for random number generation,'' in
  {\em Proceedings of the ACM/SIGDA international symposium on Field
  programmable gate arrays}, pp.~63--72, 2009.

\bibitem{kim2022extending}
S.~Y. Kim, J.~Lee, C.~H. Kim, W.~J. Lee, and S.~W. Kim, ``Extending the onnx
  runtime framework for the processing-in-memory execution,'' in {\em 2022
  International Conference on Electronics, Information, and Communication
  (ICEIC)}, pp.~1--4, IEEE, 2022.

\bibitem{seekings2024towards}
J.~Seekings, P.~Chandarana, M.~Ardakani, M.~Mohammadi, and R.~Zand, ``Towards
  efficient deployment of hybrid snns on neuromorphic and edge {AI} hardware,''
  in {\em 2024 International Conference on Neuromorphic Systems (ICONS)},
  pp.~71--77, IEEE, 2024.

\bibitem{krishnasamy2020edge}
E.~Krishnasamy, S.~Varrette, and M.~Mucciardi, ``Edge computing: An overview of
  framework and applications,'' {\em PRACE Technical Report}, 2020.

\bibitem{ma2022neuromorphic}
S.~Ma, J.~Pei, W.~Zhang, G.~Wang, D.~Feng, F.~Yu, C.~Song, H.~Qu, C.~Ma, M.~Lu,
  {\em et~al.}, ``Neuromorphic computing chip with spatiotemporal elasticity
  for multi-intelligent-tasking robots,'' {\em Science Robotics}, vol.~7,
  no.~67, p.~eabk2948, 2022.

\bibitem{liu2015sparse}
B.~Liu, M.~Wang, H.~Foroosh, M.~Tappen, and M.~Pensky, ``Sparse convolutional
  neural networks,'' in {\em Proceedings of the IEEE conference on computer
  vision and pattern recognition}, pp.~806--814, 2015.

\bibitem{srinivas2017training}
S.~Srinivas, A.~Subramanya, and R.~Venkatesh~Babu, ``Training sparse neural
  networks,'' in {\em Proceedings of the IEEE conference on computer vision and
  pattern recognition workshops}, pp.~138--145, 2017.

\bibitem{parashar2017scnn}
A.~Parashar, M.~Rhu, A.~Mukkara, A.~Puglielli, R.~Venkatesan, B.~Khailany,
  J.~Emer, S.~W. Keckler, and W.~J. Dally, ``{SCNN}: An accelerator for
  compressed-sparse convolutional neural networks,'' {\em ACM SIGARCH computer
  architecture news}, vol.~45, no.~2, pp.~27--40, 2017.

\bibitem{zambrano2019sparse}
D.~Zambrano, R.~Nusselder, H.~S. Scholte, and S.~M. Boht{\'e}, ``Sparse
  computation in adaptive spiking neural networks,'' {\em Frontiers in
  neuroscience}, vol.~12, p.~987, 2019.

\bibitem{ren2018sbnet}
M.~Ren, A.~Pokrovsky, B.~Yang, and R.~Urtasun, ``Sbnet: Sparse blocks network
  for fast inference,'' in {\em Proceedings of the IEEE conference on Computer
  Vision and Pattern Recognition}, pp.~8711--8720, 2018.

\bibitem{zaheer2020big}
M.~Zaheer, G.~Guruganesh, K.~A. Dubey, J.~Ainslie, C.~Alberti, S.~Ontanon,
  P.~Pham, A.~Ravula, Q.~Wang, L.~Yang, {\em et~al.}, ``Big bird: Transformers
  for longer sequences,'' {\em Advances in neural information processing
  systems}, vol.~33, pp.~17283--17297, 2020.

\bibitem{du2022glam}
N.~Du, Y.~Huang, A.~M. Dai, S.~Tong, D.~Lepikhin, Y.~Xu, M.~Krikun, Y.~Zhou,
  A.~W. Yu, O.~Firat, {\em et~al.}, ``Glam: Efficient scaling of language
  models with mixture-of-experts,'' in {\em International conference on machine
  learning}, pp.~5547--5569, PMLR, 2022.

\bibitem{fedus2022switch}
W.~Fedus, B.~Zoph, and N.~Shazeer, ``Switch transformers: Scaling to trillion
  parameter models with simple and efficient sparsity,'' {\em Journal of
  Machine Learning Research}, vol.~23, no.~120, pp.~1--39, 2022.

\bibitem{lewis2021base}
M.~Lewis, S.~Bhosale, T.~Dettmers, N.~Goyal, and L.~Zettlemoyer, ``Base layers:
  Simplifying training of large, sparse models,'' in {\em International
  Conference on Machine Learning}, pp.~6265--6274, PMLR, 2021.

\bibitem{hinton2015distilling}
G.~Hinton, O.~Vinyals, and J.~Dean, ``Distilling the knowledge in a neural
  network,'' {\em arXiv preprint arXiv:1503.02531}, 2015.

\bibitem{jacob2018quantization}
B.~Jacob, S.~Kligys, B.~Chen, M.~Zhu, M.~Tang, A.~Howard, H.~Adam, and
  D.~Kalenichenko, ``Quantization and training of neural networks for efficient
  integer-arithmetic-only inference,'' in {\em Proceedings of the IEEE
  conference on computer vision and pattern recognition}, pp.~2704--2713, 2018.

\bibitem{rathi2018stdp}
N.~Rathi, P.~Panda, and K.~Roy, ``{STDP}-based pruning of connections and
  weight quantization in spiking neural networks for energy-efficient
  recognition,'' {\em IEEE Transactions on Computer-Aided Design of Integrated
  Circuits and Systems}, vol.~38, no.~4, pp.~668--677, 2018.

\bibitem{yang2019quantization}
J.~Yang, X.~Shen, J.~Xing, X.~Tian, H.~Li, B.~Deng, J.~Huang, and X.-s. Hua,
  ``Quantization networks,'' in {\em Proceedings of the IEEE/CVF conference on
  computer vision and pattern recognition}, pp.~7308--7316, 2019.

\bibitem{li2022quantization}
C.~Li, L.~Ma, and S.~Furber, ``Quantization framework for fast spiking neural
  networks,'' {\em Frontiers in Neuroscience}, vol.~16, p.~918793, 2022.

\bibitem{liu2021efficient}
S.~Liu, D.~S. Ha, F.~Shen, and Y.~Yi, ``Efficient neural networks for edge
  devices,'' {\em Computers \& Electrical Engineering}, vol.~92, p.~107121,
  2021.

\bibitem{merolla2014million}
P.~A. Merolla, J.~V. Arthur, R.~Alvarez-Icaza, A.~S. Cassidy, J.~Sawada,
  F.~Akopyan, B.~L. Jackson, N.~Imam, C.~Guo, Y.~Nakamura, {\em et~al.}, ``A
  million spiking-neuron integrated circuit with a scalable communication
  network and interface,'' {\em Science}, vol.~345, no.~6197, pp.~668--673,
  2014.

\bibitem{chen2016eyeriss}
Y.-H. Chen, J.~Emer, and V.~Sze, ``Eyeriss: A spatial architecture for
  energy-efficient dataflow for convolutional neural networks,'' {\em ACM
  SIGARCH computer architecture news}, vol.~44, no.~3, pp.~367--379, 2016.

\bibitem{kwon2018maeri}
H.~Kwon, A.~Samajdar, and T.~Krishna, ``Maeri: Enabling flexible dataflow
  mapping over {DNN} accelerators via reconfigurable interconnects,'' {\em ACM
  Sigplan Notices}, vol.~53, no.~2, pp.~461--475, 2018.

\bibitem{zhang2018dnnbuilder}
X.~Zhang, J.~Wang, C.~Zhu, Y.~Lin, J.~Xiong, W.-m. Hwu, and D.~Chen,
  ``Dnnbuilder: An automated tool for building high-performance dnn hardware
  accelerators for fpgas,'' in {\em 2018 IEEE/ACM International Conference on
  Computer-Aided Design (ICCAD)}, pp.~1--8, IEEE, 2018.

\bibitem{chen2018tvm}
T.~Chen, T.~Moreau, Z.~Jiang, L.~Zheng, E.~Yan, H.~Shen, M.~Cowan, L.~Wang,
  Y.~Hu, L.~Ceze, {\em et~al.}, ``{TVM}: An automated {End-to-End} optimizing
  compiler for deep learning,'' in {\em 13th USENIX Symposium on Operating
  Systems Design and Implementation (OSDI 18)}, pp.~578--594, 2018.

\bibitem{haris2021secda}
J.~Haris, P.~Gibson, J.~Cano, N.~B. Agostini, and D.~Kaeli, ``{SECDA}:
  Efficient hardware/software co-design of {FPGA}-based {DNN} accelerators for
  edge inference,'' in {\em 2021 IEEE 33rd International Symposium on Computer
  Architecture and High Performance Computing (SBAC-PAD)}, pp.~33--43, IEEE,
  2021.

\bibitem{daghero2021energy}
F.~Daghero, D.~J. Pagliari, and M.~Poncino, ``Energy-efficient deep learning
  inference on edge devices,'' in {\em Advances in Computers}, vol.~122,
  pp.~247--301, Elsevier, 2021.

\bibitem{mehlin2023towards}
V.~Mehlin, S.~Schacht, and C.~Lanquillon, ``Towards energy-efficient deep
  learning: An overview of energy-efficient approaches along the deep learning
  lifecycle,'' {\em arXiv preprint arXiv:2303.01980}, 2023.

\bibitem{bai2024beyond}
G.~Bai, Z.~Chai, C.~Ling, S.~Wang, J.~Lu, N.~Zhang, T.~Shi, Z.~Yu, M.~Zhu,
  Y.~Zhang, {\em et~al.}, ``Beyond efficiency: A systematic survey of
  resource-efficient large language models,'' {\em arXiv preprint
  arXiv:2401.00625}, 2024.

\bibitem{kachris2025survey}
C.~Kachris, ``A survey on hardware accelerators for large language models,''
  {\em Applied Sciences}, vol.~15, no.~2, p.~586, 2025.

\bibitem{liu2024survey}
J.~Liu, P.~Tang, W.~Wang, Y.~Ren, X.~Hou, P.-A. Heng, M.~Guo, and C.~Li, ``A
  survey on inference optimization techniques for mixture of experts models,''
  {\em arXiv preprint arXiv:2412.14219}, 2024.

\bibitem{zhao2018deep}
T.~Zhao, Q.~Chen, Z.~Kuang, J.~Yu, W.~Zhang, and J.~Fan, ``Deep mixture of
  diverse experts for large-scale visual recognition,'' {\em IEEE transactions
  on pattern analysis and machine intelligence}, vol.~41, no.~5,
  pp.~1072--1087, 2018.

\bibitem{yi2023edgemoe}
R.~Yi, L.~Guo, S.~Wei, A.~Zhou, S.~Wang, and M.~Xu, ``Edgemoe: Fast on-device
  inference of moe-based large language models,'' {\em arXiv preprint
  arXiv:2308.14352}, 2023.

\bibitem{li2025uni}
Y.~Li, S.~Jiang, B.~Hu, L.~Wang, W.~Zhong, W.~Luo, L.~Ma, and M.~Zhang,
  ``Uni-moe: Scaling unified multimodal {LLMs} with mixture of experts,'' {\em
  IEEE Transactions on Pattern Analysis and Machine Intelligence}, 2025.

\bibitem{liu2017unified}
B.~Liu, C.~Wen, A.~D. Sarwate, and M.~M. Dehnavi, ``A unified optimization
  approach for sparse tensor operations on {GPUs},'' in {\em 2017 IEEE
  international conference on cluster computing (CLUSTER)}, pp.~47--57, IEEE,
  2017.

\bibitem{castagnetti2023trainable}
A.~Castagnetti, A.~Pegatoquet, and B.~Miramond, ``Trainable quantization for
  speedy spiking neural networks,'' {\em Frontiers in Neuroscience}, vol.~17,
  p.~1154241, 2023.

\bibitem{blalock2020state}
D.~Blalock, J.~J. Gonzalez~Ortiz, J.~Frankle, and J.~Guttag, ``What is the
  state of neural network pruning?,'' {\em Proceedings of machine learning and
  systems}, vol.~2, pp.~129--146, 2020.

\bibitem{gholami2022survey}
A.~Gholami, S.~Kim, Z.~Dong, Z.~Yao, M.~W. Mahoney, and K.~Keutzer, ``A survey
  of quantization methods for efficient neural network inference,'' in {\em
  Low-power computer vision}, pp.~291--326, Chapman and Hall/CRC, 2022.

\bibitem{reguero2025energy}
{\'A}.~D. Reguero, S.~Mart{\'\i}nez-Fern{\'a}ndez, and R.~Verdecchia,
  ``Energy-efficient neural network training through runtime layer freezing,
  model quantization, and early stopping,'' {\em Computer Standards \&
  Interfaces}, vol.~92, p.~103906, 2025.

\bibitem{kuzmin2023pruning}
A.~Kuzmin, M.~Nagel, M.~Van~Baalen, A.~Behboodi, and T.~Blankevoort, ``Pruning
  vs quantization: Which is better?,'' {\em Advances in neural information
  processing systems}, vol.~36, pp.~62414--62427, 2023.

\bibitem{de1997hardware}
G.~De~Michell and R.~K. Gupta, ``Hardware/software co-design,'' {\em
  Proceedings of the IEEE}, vol.~85, no.~3, pp.~349--365, 1997.

\bibitem{varshika2023hardware}
M.~L. Varshika, A.~K. Mishra, N.~Kandasamy, and A.~Das, ``Hardware-software
  co-design for on-chip learning in {AI} systems,'' in {\em Proceedings of the
  28th Asia and South Pacific Design Automation Conference}, pp.~624--631,
  2023.

\bibitem{ponzina2023hardware}
F.~Ponzina, {\em Hardware-software co-design methodologies for edge {AI}
  optimization}.
\newblock PhD thesis, EPFL, 2023.

\bibitem{bringmann2021automated}
O.~Bringmann, W.~Ecker, I.~Feldner, A.~Frischknecht, C.~Gerum,
  T.~H{\"a}m{\"a}l{\"a}inen, M.~A. Hanif, M.~J. Klaiber,
  D.~Mueller-Gritschneder, P.~P. Bernardo, {\em et~al.}, ``Automated {HW/SW}
  co-design for edge {AI}: State, challenges and steps ahead,'' in {\em
  Proceedings of the 2021 International Conference on Hardware/Software
  Codesign and System Synthesis}, pp.~11--20, 2021.

\bibitem{maguire2007challenges}
L.~P. Maguire, T.~M. McGinnity, B.~Glackin, A.~Ghani, A.~Belatreche, and
  J.~Harkin, ``Challenges for large-scale implementations of spiking neural
  networks on {FPGAs},'' {\em Neurocomputing}, vol.~71, no.~1-3, pp.~13--29,
  2007.

\bibitem{han2025progress}
W.~Han, T.~Zhang, H.~Xue, X.~Long, L.~Guanting, W.~Zhang, L.~Wang, M.~Li,
  Y.~Zhou, and J.~Chen, ``Progress and challenges in large scale spiking neural
  networks for {AI} and neuroscience,'' {\em Authorea Preprints}, 2025.

\bibitem{ahmed2015review}
M.~R. Ahmed and B.~Sujatha, ``A review on methods, issues and challenges in
  neuromorphic engineering,'' in {\em 2015 International Conference on
  Communications and Signal Processing (ICCSP)}, pp.~0899--0903, IEEE, 2015.

\bibitem{sanyal2025real}
S.~Sanyal, A.~Joshi, A.~Kosta, and K.~Roy, ``Real-time neuromorphic navigation:
  Guiding physical robots with event-based sensing and task-specific
  reconfigurable autonomy stack,'' {\em arXiv preprint arXiv:2503.09636}, 2025.

\bibitem{yang2020neuromorphic}
J.-Q. Yang, R.~Wang, Y.~Ren, J.-Y. Mao, Z.-P. Wang, Y.~Zhou, and S.-T. Han,
  ``Neuromorphic engineering: from biological to spike-based hardware nervous
  systems,'' {\em Advanced Materials}, vol.~32, no.~52, p.~2003610, 2020.

\bibitem{peng2018neuromorphic}
H.-T. Peng, M.~A. Nahmias, T.~F. De~Lima, A.~N. Tait, and B.~J. Shastri,
  ``Neuromorphic photonic integrated circuits,'' {\em IEEE Journal of Selected
  Topics in Quantum Electronics}, vol.~24, no.~6, pp.~1--15, 2018.

\bibitem{markovic2020quantum}
D.~Markovi{\'c} and J.~Grollier, ``Quantum neuromorphic computing,'' {\em
  Applied physics letters}, vol.~117, no.~15, 2020.

\bibitem{ponzina2022hardware}
F.~Ponzina, S.~Machetti, M.~Rios, B.~W. Denkinger, A.~Levisse, G.~Ansaloni,
  M.~Pe{\'o}n-Quir{\'o}s, and D.~Atienza, ``A hardware/software co-design
  vision for deep learning at the edge,'' {\em IEEE Micro}, vol.~42, no.~6,
  pp.~48--54, 2022.

\end{thebibliography}

\begin{comment}

\end{comment}

\end{document}